\documentclass[12pt,a4paper]{article}

\usepackage[utf8]{inputenc}
\usepackage[T1]{fontenc}
\usepackage{lmodern}
\usepackage{geometry}
\usepackage{graphicx}
\usepackage{amsmath, amssymb, amsfonts}
\usepackage{mathrsfs}
\usepackage{caption}
\usepackage{subcaption}
\usepackage{float}
\usepackage{booktabs}
\usepackage{multirow}
\usepackage{longtable}
\usepackage[table]{xcolor}
\usepackage{textcomp}
\usepackage{url}
\usepackage{algorithm}
\usepackage{algpseudocode}
\usepackage{siunitx}
\usepackage{acronym}
\usepackage[numbers,sort&compress]{natbib}
\usepackage{authblk}
\usepackage{hyperref}

\definecolor{bestgray}{gray}{0.92}

\makeatletter
\newcommand\myparagraph{%
  \@startsection{paragraph}{4}{0mm}%
    {-\baselineskip}%
    {-0.7\baselineskip}%
    {\normalfont\normalsize\bfseries}}
\makeatother

\title{Similarity-Guided Optimization Strategies for Approximate Network Symmetry Detection}

\author[1,2]{David Hartman\thanks{Corresponding author}}
\author[1,3]{Jaroslav Hlinka}
\author[1]{Dominik Klepl}
\author[1,3]{Anna Pidnebesna}
\author[1]{Petra Vidnerov\'{a}}

\affil[1]{Institute of Computer Science of the Czech Academy of Sciences, Prague, Czech Republic}
\affil[2]{Unicorn University, Prague, Czech Republic}
\affil[3]{National Institute of Mental Health, Klecany, Czech Republic}

\acrodef{SA}{simulated annealing}
\acrodef{GA}{genetic algorithm}
\acrodef{SG}{similarity guidance}
\acrodef{SGSA}{similarity-guided simulated annealing}
\acrodef{SGGA}{similarity-guided genetic algorithm}
\acrodef{QSA}{quadratic symmetry approximator}
\acrodef{BASIN}{bilinear approximate symmetry inference}
\acrodef{SGQSA}{similarity-guided quadratic symmetry approximator}
\acrodef{TS}{tabu search}
\acrodef{SGTS}{similarity-guided tabu search}
\acrodef{PSO}{particle swarm optimization}
\acrodef{SGPSO}{similarity-guided particle swarm optimization}
\acrodef{DE}{differential evolution}
\acrodef{SGDE}{similarity-guided differential evolution}
\acrodef{ER}{Erd\H{o}s--R\'{e}nyi}
\acrodef{BA}{Barab\'{a}si--Albert}
\acrodef{LRM}{lateral random model}
\acrodef{MMM}{modular mirror model}
\acrodef{FP}{fixed point}
\acrodef{SC}{structural connectivity}
\acrodef{AAL}{automated anatomical labeling}
\acrodef{OMST}{orthogonal minimum spanning trees}

\begin{document}

\maketitle

\begin{abstract}
Symmetry in complex networks is a structural property with broad implications for network function, dynamics, and robustness. While classical approaches rely on exact automorphisms and thus remain highly sensitive to small perturbations, recent work of Liu has introduced approximate formulations that relax the strict invariance condition and recast symmetry detection as an optimization problem over permutations. We investigate how node-level similarity information can be used to bias the search toward structurally plausible vertex correspondences, with network centralities serving as one interpretable instance. Building on this perspective, we adapt simulated annealing, genetic algorithms, tabu search, particle swarm optimization, differential evolution and the quadratic symmetry approximator to this problem, add similarity guidance to each, and introduce Bilinear Approximate Symmetry Inference (BASIN), which learns a soft assignment matrix via bilinear embeddings. Using a unified evaluation pipeline, we compare their guided and unguided variants, BASIN and a random baseline on four synthetic graph families, structural brain connectomes, and three empirical networks with known automorphism groups. Node-similarity guidance is decisive where a planted orbit exists that the unguided search misses, and algorithm-dependent elsewhere, on synthetic and empirical networks alike. Overall, tabu search achieves the strongest performance; its guided variant (SGTS) has the lowest median score among the guided variants in all 44 synthetic conditions, while BASIN is the better of the continuous relaxations on rewired lateral graphs. The effect of node-similarity guidance is therefore regime-dependent rather than uniformly positive.
\end{abstract}


\section{Introduction}\label{s:Intro}

Traditionally, symmetry in complex networks has been understood in the strict sense of exact automorphisms: structural permutations of nodes that preserve all adjacency relations of the network. Automorphisms provide a precise language for network structure and have been used to study empirical networks, model reduction, and collective dynamics \cite{macarthur2008symmetry,sanchez2020exploiting,Hart2019topological,Mondal2024symmetry}. Structural symmetry is also relevant to brain networks, although anatomical and network-level asymmetries must be distinguished \cite{delange2016role,kong2018mapping}. Exact automorphisms, however, can disappear after a small number of edge changes; their absence does not rule out an informative near-symmetry~\cite{Liu2020approx,pidnebesna2025computing}.

We study approximate symmetry as a nontrivial permutation of a \emph{single} graph that approximately preserves its adjacency matrix. This retains the vertex-bijection interpretation of an automorphism while allowing edge disagreement \cite{Liu2020approx}. Finding a useful permutation is difficult because the search space grows factorially. Moreover, unlike matching two different graphs, unrestricted self-matching has an exact but uninformative identity solution. The treatment of fixed points is therefore part of the problem specification, not a detail of the optimizer.

Existing approaches include Liu's \ac{SA}, the doubly stochastic relaxation of the \ac{QSA}, and the basic results about the centrality-guided local search \cite{Liu2020approx,pidnebesna2025computing,hartman2025local}. These provide direct points of comparison but do not establish how node-similarity guidance behaves across several discrete and continuous optimizers under a common evaluation protocol. Structural descriptors are natural search cues, but equal descriptors do not imply a symmetry (Section~\ref{s:related_work}). Whether such guidance helps is consequently an empirical question.

We investigate this question by adapting \ac{SA}, \ac{GA}, \ac{TS}, \ac{PSO}, \ac{DE}
and \ac{QSA} to nontrivial approximate self-symmetry, and by comparing guided and unguided versions of each. The \ac{SGGA} combines centrality-informed fitness and mutation; genetic search itself is established in graph correspondence \cite{Myers2001leastcommitment}. \Ac{SGTS} also selects its guidance configuration per instance. In addition, \ac{BASIN} learns node embeddings, forms a row-stochastic soft assignment from bilinear scores, includes terms for structural disagreement and optional column-sum and fixed-point penalties, and projects the result to a permutation. Its components draw on established continuous and learned correspondence ideas; the contribution here is their combination for nontrivial approximate \emph{self}-symmetry.

The evaluation covers four synthetic graph families, including two with a planted correspondence, structural brain connectomes, and three empirical
networks with computed automorphism groups. Guidance is most useful when it exposes a relevant correspondence missed by unguided search, but can be neutral or detrimental in other regimes. \Ac{SGTS} has the lowest median objective value among the guided variants in the reported synthetic conditions, with unguided \ac{TS} never significantly worse outside the \ac{MMM}, while \ac{BASIN} is stronger than \ac{QSA} on most rewired \ac{LRM} conditions. These observations are bounded by the tested graphs and tuning protocol; they are not a claim of universal algorithmic superiority or scalability.

\section{Related Work}\label{s:related_work}

\subsection{Exact symmetry and non-equivalent approximations}

The automorphism group describes exact invariance under vertex relabeling.
Empirical-network studies have found substantial, often local, automorphism
structure, while asymmetry is typical in broad classes of random graphs
\cite{macarthur2008symmetry}, which is also supported by theoretical results~\cite{erdos1963asymmetric}. Automorphisms can also
organize network reductions and synchronized dynamics
\cite{sanchez2020exploiting,Hart2019topological}. These observations motivate
robust extensions, but \emph{approximate symmetry} has more than one
mathematical meaning. Concentric network symmetry measures regularity across
successive neighborhoods \cite{silva2015concentric}; dynamical
quasi-symmetry concerns phase relations rather than an approximately
adjacency-preserving vertex permutation \cite{roselltarrag2021quasi}.
Neither is interchangeable with the objective studied here.

Liu formulated a global, permutation-based measure of adjacency
disagreement, restricting the search to fixed-point-free permutations to
avoid the identity \cite{Liu2020approx}. Pidnebesna et al. retained the
self-matching interpretation but introduced a fixed-point penalty and \ac{QSA},
which relaxes permutations to doubly stochastic matrices and projects a
continuous solution back to a discrete assignment
\cite{pidnebesna2025computing}. Hartman et al. studied centrality-guided
iterative search for the same family of approximate-symmetry questions
\cite{hartman2025local}. These three papers are the closest methodological
predecessors. Their different fixed-point treatments should be considered
when comparing reported symmetry scores and candidate permutations.

\subsection{Graph matching and continuous assignment}

Matching two graphs often optimizes a quadratic objective over permutation
matrices and is closely related to the quadratic assignment problem
\cite{Loiola2007qap,vogelstein2015fast}. Replacing permutations by doubly
stochastic matrices makes continuous optimization possible, but relaxation
and projection need not preserve the discrete optimum. For example, Aflalo
et al. established recovery results for particular spectrally characterized
graph classes and bounded perturbations, whereas Lyzinski et al. showed
important differences between convex and indefinite relaxations under their
random-graph assumptions \cite{Aflalo2015convex,Lyzinski2016risk}.
Vogelstein et al. used a Frank--Wolfe-type procedure with linear-assignment
directions and final projection for graph matching
\cite{vogelstein2015fast}; \ac{QSA} adapts this relaxation strategy to approximate
self-symmetry \cite{pidnebesna2025computing}.

These are methodological connections, not equivalent tasks. Ordinary
two-graph matching seeks an unknown correspondence between distinct inputs;
it does not require excluding the identity of a graph matched to itself.
Theoretical guarantees and empirical rankings for two-graph matching
therefore do not transfer automatically to the nontrivial self-matching
objective. \ac{QSA}'s fixed-point penalty and \ac{BASIN}'s own penalty and
post-projection selection are specifically relevant to that distinction.

\subsection{Structural guidance and discrete search}

Node-level information can narrow a correspondence search. Recursive
neighborhood scores have been used in graph matching
\cite{Zager2008similarity}; more recently, attribute and structural
similarities have been combined in a two-graph matching algorithm
\cite{Candelier2025gasm}. Such work supports using structural evidence to
prioritize candidate pairs, but does not show that similarly scored vertices
are approximately automorphic. Centrality is constant on an exact orbit, but
equal centrality does not imply a shared orbit
\cite{gago2013betweenness,hartman2024connectivity}, and under approximate
symmetry even the constancy holds only approximately.
The more direct evidence for this setting is the reported improvement of
centrality-guided local search on selected instances
\cite{hartman2025local}.

Genetic search has long been applied to inexact matching between different
graphs \cite{Myers2001leastcommitment}, and the quadratic-assignment
literature includes permutation-based evolutionary and local-search
strategies \cite{Loiola2007qap,taillard1991robust}. Our \ac{SGGA} applies this search to the self-symmetry objective, with a
fixed-point bound and centrality terms in its fitness and mutation. The
\ac{TS} adaptation instead uses a permutation neighborhood and
instance-level selection among guidance settings. The \ac{PSO} and
\ac{DE} adaptations use continuous population search followed
by projection to permutations; they are not native permutation algorithms.
These distinct mechanisms justify empirical comparison but do not predict
their ranking on self-matching. Similarity guidance can be useful when a
descriptor separates plausible from implausible mappings; it may contribute
little or misdirect search when the descriptor is degenerate or unrelated to
the target correspondence. The present experiments test, rather than presume,
its value.

\subsection{Learned soft correspondence and \acs{BASIN}}

Soft assignments are also used in differentiable matching. Sinkhorn--Knopp
normalization alternates row and column scaling toward a doubly stochastic
matrix \cite{SinkhornKnopp1967}, and differentiable variants have enabled
learning of latent permutations \cite{Mena2018gumbelsinkhorn}. Learned node
embeddings and affinity scores have been combined with assignment mechanisms
in graph matching between pairs of graphs \cite{Wang2019combinatorial}.
These works are relevant precedents for \ac{BASIN}'s components, not prior
solutions to the present self-symmetry problem.

\ac{BASIN} learns bilinear node scores from the symmetry loss on one graph and
applies a row-wise softmax. It regularizes, but does not enforce, column
sums of one; thus its soft matrix is generally not doubly stochastic and
\ac{BASIN} does not implement Sinkhorn normalization. A discrete permutation is
obtained only after projection, which can change the structural objective.
Nor does \ac{BASIN} inherit recovery guarantees from convex graph-matching
relaxations. Its claim here is a distinct parametrization and empirical
comparison in permutation-based approximate symmetry detection, rather than
novelty of bilinear embeddings or differentiable assignment in general.

\section{Definitions}\label{s:definitions}

This section introduces the formal concepts and mathematical notation underlying approximate symmetry detection in networks. We begin with the definition of exact graph automorphisms and then introduce the relaxed formulation used throughout this work.

\subsection{Networks}\label{ss:networks}

As a network, we understand a loop-less, undirected, unweighted graph, defined by a set of vertices (nodes) $V$ and edges (links) $E$, denoted as $G = (V, E)$. We denote the number of vertices by $|V| = n$, and the number of edges by $|E| = m$. We then define the density of the graph by $\rho(G) = m / {n \choose 2}$.

The graph is represented by its {\em adjacency matrix} \( A \in \{0,1\}^{n \times n} \), where $A_{ij}=1$ if $\{v_i,v_j\}\in E$ and $A_{ij}=0$ otherwise. A {\em graph automorphism} is a permutation \( \pi \) of nodes that preserves adjacency: \( A = PAP^T \), where \( P \) is the corresponding {\em permutation matrix}, i.e. a square binary matrix with exactly one entry of 1 in each row and each column. We denote the set of all permutation matrices by $\mathcal{P}_n$ and an identity matrix as $I_n$. The set of all such adjacency-preserving matrices forms the automorphism group of the graph~\cite{bollobas1998modern}.

Equivalently, an automorphism can be understood as a bijective mapping from $V$ to $V$ preserving both the presence and absence of all edges. Automorphisms can map (permute) some vertices so that their image is identical to themselves. We will call such vertices {\em fixed vertices} of the permutation. Note that for every graph there is always one automorphism called a {\em trivial automorphism}, in which all vertices are fixed, i.e., it does not permute any vertices. We will call a graph symmetric if there is at least one non-trivial automorphism, otherwise it is \emph{asymmetric}.

\subsection{Problem statement}\label{ss:problem}

Exact automorphism-based symmetry is highly sensitive to small structural perturbations: even minor changes in the adjacency structure may eliminate all non-trivial automorphisms. To address this limitation, we adopt the approximate symmetry framework introduced by~\cite{Liu2020approx}, which measures how closely a permutation preserves the adjacency matrix.

Given a permutation matrix $P$, the approximate symmetry error is defined as
\begin{equation}
\epsilon(A, P) = \frac{1}{4} \|A - PAP^T\|_1.
\end{equation}
The factor $\tfrac{1}{4}$ counts each mismatched undirected edge once. Normalizing by the number of ordered vertex pairs gives the size-independent score
\begin{equation}\label{eq:s_value}
\mathcal{S}(A,P) = \frac{\left\| A - PAP^{\mathrm{T}} \right\|_{1}}{n(n-1)} = \frac{4\,\epsilon(A,P)}{n(n-1)},
\end{equation}
which is minimized over $P \in \mathcal{P}_n \setminus \{I_n\}$.

Very local permutations involving only a small number of vertices may yield artificially low symmetry scores. The original formulation in~\cite{Liu2020approx} therefore restricted to permutations with no fixed points, i.e. $\mathrm{tr}(P)=0$. In many empirical networks, however, meaningful approximate symmetries may include partially fixed substructures. Thus, to improve the interpretability of the discovered symmetries, while avoiding near-identity solutions, we allow permutations with fixed points but explicitly penalize their influence through a modified normalization of the approximate symmetry score:
\begin{equation}\label{eq:s_fp_value}
    \mathcal{S}'(A,P) = \frac{\|A-PAP^{T}\|_1}{n(n-1)-F(P)\bigl(F(P)-1\bigr)}, \qquad P \in \mathcal{P}_n \setminus \{I_n\}
\end{equation}
where $F(P)=\mathrm{tr}(P)$ denotes the number of fixed points of permutation $P$. This formulation discourages permutation that leave most vertices unchanged while still allowing partially fixed symmetries to be detected.

This self-referential form also explains why off-the-shelf graph matching and quadratic assignment solvers cannot be applied unmodified. Such solvers align two distinct graphs, and with both arguments equal to $A$ the identity permutation attains zero discrepancy and is a global optimum of their objective, so an unconstrained solver returns it. Approximate symmetry is therefore a matching of a network to itself under an explicit non-identity constraint, and the constraint has to be part of the search rather than a post-hoc filter. With it built in, such solvers do apply: \ac{QSA} and \ac{BASIN} are continuous relaxations of the kind used in graph matching, and \ac{TS} is a quadratic assignment solver (Section~\ref{s:methods}).

\subsection{Node similarity and Centralities}\label{ss:centralities}
The optimization problem defined above involves search over the space of all permutations, whose size grows factorially with the number of vertices. As a result, purely random exploration of candidate permutations is inefficient and fails to prioritize structurally meaningful vertex correspondences. In this work, we therefore introduce the concept of \emph{node similarity guidance}: additional node-level information that biases the optimization toward permutations mapping vertices with comparable structural roles. We use the term \emph{node similarity} to denote any descriptor that assigns comparable value to vertices that are likely to be interchangeable under approximate symmetries. Centrality measures constitute one interpretable and computationally convenient instance of such descriptors~\cite{newman2018networks}.

Centralities quantify different aspects of a vertex's structural role within a network, such as connectivity, influence, or position within communication pathways. Because automorphisms preserve node-rooted structural descriptors, vertices that are symmetric (i.e. can be mapped to each other via automorphism) or nearly symmetric often exhibit similar centrality profiles. Consequently, centralities provide a principled way to construct node-similarity signals that can be incorporated into optimization strategies.
The alternative notions of approximate symmetry discussed in Section~\ref{s:related_work} can correlate only weakly with centralities~\cite{silva2015concentric}. This is not at odds with the above: those notions define symmetry by other node properties, whereas here symmetry is defined by the permutation objective and centralities only bias the search towards vertex correspondences that an automorphism would preserve.

In this work, node centralities provide an interpretable implementation of node-similarity guidance, but they are not the only option. In particular, our proposed \acs{BASIN} strategy (Sec.~\ref{ss:basin}) replaces centrality descriptors with a learnable similarity mechanism. For the explicit (hand-crafted) versions of strategies, we consider several commonly used centrality measures:

\begin{itemize}
    \item \emph{Degree centrality}, capturing local connectivity and forming a basic structural descriptor frequently used in graph characterization~\cite{barabasi1999emergence}.
    \item \emph{Eigenvector centrality}~\cite{bonacich1972technique}, which generalizes degree by incorporating global network structure and is closely related to spectral properties of the graph.
    \item \emph{Bonacich centrality}~\cite{bonacich1972technique}, a parametric extension of eigenvector centrality that accounts for indirect influence through neighboring vertices.
    \item \emph{Betweenness centrality}~\cite{freeman1977set}, measuring how often a vertex lies on shortest paths between other vertices, with connections to symmetry-related structural properties~\cite{gago2013betweenness,ghanbari2023structure,hartman2024connectivity}.
    \item \emph{Clustering coefficient}, describing the density of local neighborhoods and providing structural information~\cite{newman2018networks,watts1998collective}.
\end{itemize}

In this work, centrality measures serve mainly as a clear and interpretable example of node-similarity guidance that can be incorporated into optimization procedures. At the same time, they provide a useful reference point for comparison with more flexible similarity mechanisms, including learnable formulations introduced later. Centralities should therefore be viewed here as one application of the node similarity guidance rather than a restrictive modeling choice.

\section{Methods}\label{s:methods}

This section describes the optimization strategies evaluated for identifying approximate symmetries in networks. We consider both established reference methods and new strategies proposed in this work. Specifically we compare: (i) a random-search baseline, (ii) \ac{SA} following~\cite{Liu2020approx}, (iii) \ac{QSA}~\cite{pidnebesna2025computing}, (iv) \ac{TS} following~\cite{taillard1991robust}, (v) \ac{PSO}~\cite{kennedy1995particle} and (vi) \ac{DE}~\cite{storn1997differential}, (vii) a genetic-algorithm-based search (\ac{GA}), and (viii) \ac{BASIN}, a continuous optimization strategy introduced here. Every strategy but \ac{BASIN} and the baseline is evaluated in an unguided and a similarity-guided variant, prefixed ``SG''. The guided variant of \ac{SA} follows~\cite{hartman2025local}; the guided variants of the other strategies, are introduced here.

All methods optimize the same approximate-symmetry objective but differ in how the permutation space is explored and in how node-similarity information is incorporated. Similarity guidance is introduced explicitly through predefined structural descriptors, that is, centrality measures. \Ac{BASIN} instead implements a data-driven node-similarity mechanism through continuous embeddings. \ac{QSA} represents a state-of-the-art continuous relaxation approach that does not in its original form explicitly incorporate node-similarity guidance. The following subsections describe each strategy in detail.

\subsection{Random Search}\label{ss:random}
To contextualize the performance of the tested optimization strategies, we include a simple brute--force random search baseline. This strategy directly samples the permutation space by repeatedly drawing uniformly random permutations and evaluating their approximate symmetry score $S'(A,P)$.

For a given graph with adjacency matrix $A$, we generate $10000$ independent random permutations $P^{(1)},\,P^{(2)},\,\dots,\,P^{(10\,000)}$, compute $\mathcal{S}'(A,P^{(k)})$ for every sampled permutation, and retain the lowest-scoring one as the solution produced by the method.

\subsection{Simulated Annealing}\label{ss:sim_anneal}
\cite{Liu2020approx} address the factorial growth of $\mathcal{P}_n$ with a \ac{SA} strategy that searches over permutations by iteratively proposing local modifications and accepting them according to a temperature-controlled criterion.

Briefly, at iteration $t$, a candidate permutation $P'$ is generated from the current permutation $P$ and accepted if it improves the objective, i.e. $\epsilon(A,P') \le \epsilon(A,P)$. Otherwise, the move is accepted with probability
\[
\exp\!\left(-\frac{\epsilon(A,P')-\epsilon(A,P)}{T(t)}\right)
\]
where $T(t)$ is the temperature. The temperature follows a logarithmic cooling schedule,
\[
T(t) = \frac{c}{\log(t + d)},
\]
where $t$ denotes the iteration index and $c,d>0$ are parameters controlling the initial temperature and the rate of cooling. Higher temperatures allow the algorithm to accept worsening moves with non-negligible probability, facilitating exploration of permutation space, while decreasing temperatures progressively bias the search toward locally improving moves. Here $\epsilon$ is $\mathcal{S}'(A,P)$ (Eq.~\ref{eq:s_fp_value}), and moves that would raise $F$ above $\lfloor 0.25n \rfloor$ are refused.

Transitions between permutations are realized by swapping the images of two vertices. Specifically, the move is generated by a random transposition that exchanges the images of two vertices $a,b\in V$ under the current permutation $\pi$, replacing the pairs $(a,\pi(a))$ and $(b,\pi(b))$ with $(a,\pi(b))$ and $(b,\pi(a))$, where $a,b \in V$.

In our implementation, we retain the \ac{SA} framework of~\cite{Liu2020approx}, but replace random transpositions with an \emph{informed move proposal} that favors swaps between vertices with similar structural roles. This modification is motivated by the fact that graph automorphisms preserve node centralities. Consequently, mapping vertices to images with similar centrality values is expected to reduce edge disagreements and guide the search toward structurally meaningful permutations. Similarity is measured by node centralities, yielding a guided proposal distribution over candidate transpositions, defined next.

\myparagraph{Vertex similarity structures}
We represent the pairwise similarity of vertices under a chosen centrality $\Gamma$ by an $n \times n$ matrix $M$ whose rows and columns correspond to the vertices. We first define a \emph{difference matrix} $\Delta$ with entries $d_{ij} = |\Gamma(i) - \Gamma(j)|$, and convert differences into similarities by setting $m_{ij} = (d_{ij} + \beta)^{-1}$, where $\beta > 0$ is a regularization parameter preventing division by zero and controlling the strength of the bias. Larger $\beta$ values yield a flatter $M$ and thus weaker guidance. Rather than selecting a single centrality, a guided strategy uses either a weighted blend
\begin{equation}\label{eq:similarity}
M = \sum_k w_k \frac{M_k}{\overline{M_k}}
\end{equation}
of the matrices $M_k$ built from degree, eigenvector, betweenness and clustering, each rescaled to unit mean, with weights $w_k \ge 0$, or guides each restart by a single $M_k$, the four taken in turn; which of the two forms a strategy uses is one of its hyperparameters.

\myparagraph{Guided move function}
Given the current permutation $\pi$, we select a vertex $a$ uniformly at random and consider swaps of its image $\pi(a)$ with the images $\pi(b)$ of the other vertices $b \in V$. Let $m_{ij}$ denote the vertex similarity as defined previously. We associate the current permutation with a global similarity score $\sum_i m_{i\pi(i)}$. Since swapping the images of $a$ and $b$ affects only the terms corresponding to these two vertices, the resulting change in this global similarity score is $\Delta M_b = m_{a\pi(b)} + m_{b\pi(a)} - m_{a\pi(a)} - m_{b\pi(b)}$.

These values are transformed into non-negative weights via $\Delta E_b = \max(\Delta M_b, \phi)$, where $\phi > 0$ is a probability-smoothing parameter. As with the division parameter $\beta$, larger values of $\phi$ flatten the relative differences between candidates. After normalization, the resulting weights define the probability distribution over candidate vertices $b$, from which the swap partner is sampled. The selected transposition is then evaluated using the standard \ac{SA} acceptance rule.

\myparagraph{Hyperparameters}
The tunable hyperparameter of \ac{SA} is the temperature scale $c$; \ac{SGSA} adds $\beta$, $\phi$ and the weights $w_k$ of Eq.~\ref{eq:similarity}.

\subsection{Quadratic symmetry approximator}\label{ss:qsa}
This strategy is described in~\cite{pidnebesna2025computing}. The central idea of this strategy is to relax the discrete search over permutation matrices into a continuous optimization problem over doubly stochastic matrices. This relaxation is motivated by the close relationship between approximate symmetry detection and the quadratic assignment problem~\cite{pidnebesna2025computing}. Under this formulation, the optimization problem can be written as
$$
\min_{P\in \mathcal{D}_n} -\text{tr}\!\left(A P A^{\mathrm{T}}P^{\mathrm{T}} - \mathrm{diag}(c)P\right),
$$
where the $\mathcal{D}_n$ denotes the Birkhoff polytope of $n\times n$ doubly stochastic matrices and the vector $c$ encodes penalties on fixed points.

Since the solution of this optimization lies in the polytope $\mathcal{D}_n$, the resulting matrix is generally doubly stochastic rather than a valid permutation matrix. A discrete permutation is therefore recovered in a post-processing step by selecting a permutation matrix that is close to the relaxed solution. In practice, this is achieved by solving a linear assignment problem~\cite{pidnebesna2025computing}.

The continuous part of the optimization is carried out using a first-order algorithm based on the Frank-Wolfe method, in which the linear assignment problem can be efficiently exploited to calculate the search direction, following ideas introduced in~\cite{vogelstein2015fast}.

\myparagraph{Similarity guidance}\ac{SGQSA} adds a linear reward to the relaxed objective, $f(P) - \lambda_{\mathrm{c}}\,s\,\langle M/\overline{M}, P\rangle$, with $M$ the similarity matrix defined in Eq.~\ref{eq:similarity} and $s$ the mean absolute gradient at the initial point, which puts the term on the scale of the gradient at every graph size. The objective stays quadratic plus linear, so the Frank--Wolfe line search remains exact, and $\lambda_{\mathrm{c}}=0$ recovers \ac{QSA}.

\myparagraph{Hyperparameters}
The tunable hyperparameters of \ac{QSA} are the fixed-point penalty $c$ and the convergence tolerance of the Frank--Wolfe iterations; \ac{SGQSA} adds $\lambda_{\mathrm{c}}$, $\beta$, the form of $M$ and, for a blend, the weights $w_k$.

\subsection{Tabu Search}\label{ss:tabu}
\ac{TS}~\cite{glover1989tabu} searches the discrete space of permutations through local modifications of the current permutation. At every iteration it moves to the best solution in the neighborhood of the current one, even when it is worse, and forbids the reversal of recent moves for a number of iterations, which keeps the search from falling back into a local minimum it has just left. We follow the robust tabu search of~\cite{taillard1991robust} for the quadratic assignment problem.

As in \ac{SA}, the neighborhood consists of the $\binom{n}{2}$ transpositions that exchange the images of two vertices, but \ac{TS} evaluates all of them rather than sampling one. The change in $\|A - PAP^{\mathrm{T}}\|_1$ of each transposition is read from a table updated in $O(n)$ per accepted move, so an iteration costs $O(n^2)$. A move exchanging the images of $a$ and $b$ is then logged as tabu for a tenure whose length is drawn uniformly from $[t_{\mathrm{lo}}n,\,t_{\mathrm{hi}}n]$ iterations. A tabu move is still accepted if it improves on the best solution found so far.

Our formulation departs from~\cite{taillard1991robust} in three respects. Both the flow and the distance matrix are $A$, so an unconstrained search converges to the identity (Section~\ref{ss:problem}); we require $\mathrm{tr}(P) \le n-2$ throughout, which is exactly $P \neq I_n$. Moves are ranked by $\mathcal{S}'(A,P)$ (Eq.~\ref{eq:s_fp_value}) rather than by the unnormalized discrepancy $\|A-PAP^{T}\|_1$. Its denominator depends only on the number of fixed points $F(P)$, which a transposition can change only at the two vertices it exchanges, so $\mathcal{S}'(A,P)$ of every candidate move follows in $O(1)$ from its tabulated change in discrepancy, and the scan remains $O(n^2)$. The long-term frequency memory is omitted.

\myparagraph{Similarity-guided moves}
Similarity guidance changes which transposition is chosen at each iteration, in one of two ways: by restricting the neighborhood to smaller candidate list or by biasing the ranking of moves. With $M$ the similarity matrix of Eq.~\ref{eq:similarity}, let $\Delta M_{uv} = m_{u\pi(v)} + m_{v\pi(u)} - m_{u\pi(u)} - m_{v\pi(v)}$ be the change in global similarity from transposing the images of $u$ and $v$. The \emph{candidate list} restricts the scan to the fraction $q \in (0,1]$ of transpositions with the largest $\Delta M_{uv}$, recomputed every iteration. The \emph{bias} keeps the full scan and ranks moves on $\mathcal{S}'(A,P') - \lambda\,\hat{\sigma}\,z_{uv}$, with $z_{uv}$ the standardized $\Delta M_{uv}$ and $\hat{\sigma}$ the standard deviation of $\mathcal{S}'$ over the admissible moves, so $\lambda$ is a shift in units of that deviation. Whether a tabu move improves on the best solution is judged on the unbiased $\mathcal{S}'(A,P')$, so the bias cannot release a move from the tabu list. With $q=1$ and $\lambda=0$ the search reduces to plain \ac{TS}.

\myparagraph{Similarity-guided tabu search}
\Ac{SGTS} selects its guidance per instance. Four configurations --- plain \ac{TS}, two bias strengths $\lambda$ and one candidate fraction $q$ --- are the arms of a multi-armed bandit, and one pull is one restart from a fresh random permutation. Restarts share no state, so each arm's scores are independent draws from a fixed distribution, as UCB1~\cite{auer2002finite} assumes. After one pull of each arm, restart $t$ goes to the arm maximizing
\[
\bar{r}_a + \kappa \sqrt{\frac{2\ln t}{n_a}},
\]
with $n_a$ the pulls of arm $a$ so far, $\bar{r}_a$ its mean reward and $\kappa$ an exploration constant. A restart returning $\mathcal{S}'=s$ is rewarded $(s_{\max}-s)/(s_{\max}-s_{\min})$, clipped to $[0,1]$, with the range taken from the opening round on the same instance; a tied round gives $0.5$.

The product of restarts and iterations per restart equals that of unguided \ac{TS}, so the selector pays for its exploration out of the same budget. Only the allocation of restarts is adaptive; each restart is the \ac{TS} above with the arm's $q$ and $\lambda$.

\myparagraph{Hyperparameters}
The tunable hyperparameters of \ac{TS} are the tenure bounds $t_{\mathrm{lo}}$ and $t_{\mathrm{hi}}$; \ac{SGTS} adds $\beta$, the weights $w_k$, the two bias strengths $\lambda$ and the candidate fraction $q$ of its arms, and $\kappa$.

\subsection{Particle swarm optimization}\label{ss:pso}
\ac{PSO}~\cite{kennedy1995particle} is defined on a real vector space, so
applying it here requires a continuous relaxation of $\mathcal{P}_n$. A
candidate is a matrix $X \in [0,1]^{n \times n}$, read as a score for assigning
vertex $i$ the image $j$, committed to a permutation by solving the linear
assignment problem exactly,
\[
P = \arg\max_{P \in \mathcal{P}_n} \langle P, X\rangle,
\]
by the Hungarian algorithm~\cite{kuhn1955hungarian}, and scored by
$\mathcal{S}'(A,P)$ on the feasible set $\mathrm{tr}(P) \le n-2$ used throughout.
Unlike \ac{QSA}, no constraint ties $X$ to the Birkhoff polytope: the rounding,
not the parametrization, is what enforces that the answer is a permutation.

The swarm holds $k$ particles, each carrying a position $X_i$ and a velocity
$V_i$ of the same shape, both initialized uniformly. At iteration $t$ the
velocity is updated in the global-best form,
\[
V_i \leftarrow w(t)\,V_i + c_1 R_1 \odot (X_i^{\ast} - X_i)
              + c_2 R_2 \odot (X^{\ast} - X_i),
\]
where $X_i^{\ast}$ is the particle's own best position, $X^{\ast}$ the swarm's,
$\odot$ the element-wise product, and $R_1, R_2$ are matrices of independent
$U(0,1)$ draws, so every coordinate is perturbed independently. The inertia
weight is annealed linearly over the budget,
$w(t) = w_0 + (w_1 - w_0)\,t/(t_{\max}-1)$, following~\cite{shi1998modified};
velocities are clamped to $[-v_{\max}, v_{\max}]$ and positions clipped back to
$[0,1]$ after each update. The search stops after a fixed number of iterations
without an improvement of the swarm best, counted per iteration rather than per
particle.

\myparagraph{Similarity guidance}
\ac{SGPSO} adds a fourth velocity term, $c_g R_3 \odot (M - X_i)$, pulling each
particle toward assignments between vertices of similar centrality, with $M$ the
similarity matrix of Eq.~\ref{eq:similarity} rescaled to $[0,1]$ so that its
pull is commensurate with the other terms. Its diagonal is set to a large
\emph{difference}, so the guidance does not reward mapping a vertex to itself.
Setting $c_g = 0$ recovers plain \ac{PSO} exactly.

\myparagraph{Hyperparameters}
The tunable hyperparameters of \ac{PSO} are the swarm size $k$, the inertia weights $w_0$ and $w_1$, the acceleration coefficients $c_1$ and $c_2$ and the velocity bound $v_{\max}$; \ac{SGPSO} adds $c_g$, $\beta$ and the weights $w_k$.

\subsection{Differential evolution}\label{ss:de}
\ac{DE}~\cite{storn1997differential} searches the same relaxation as \ac{PSO}
(Section~\ref{ss:pso}) --- the same encoding, the same rounding and the same
feasible set --- and differs only in how candidates are generated: from
differences between population members rather than by accumulating velocity. The
comparison between the two is therefore a comparison of search operators, not of
encoding of the problem.

The population holds $k$ matrices. For each target $X_i$, three further members
$r_1, r_2, r_3$ are drawn without replacement and distinct from $i$, and a
mutant is formed as
\[
U = X_{r_1} + \zeta\,(X_{r_2} - X_{r_3}),
\]
with scale factor $\zeta$, and its entries are clipped to $[0,1]$. Binomial crossover then takes each coordinate from $U$ with
probability $\mathit{CR}$ and from $X_i$ otherwise, with one coordinate forced to
come from $U$ so that the trial is never a copy of the target. The trial
replaces the target when it scores at least as well; the weak inequality lets the
population drift across plateaus of $\mathcal{S}'$. This is the DE/rand/1/bin scheme of~\cite{storn1997differential}.

\myparagraph{Similarity guidance}
\ac{SGDE} guides the mutant with the same matrix $M$ that guides \ac{SGPSO},
adding $c_g (M - X_{r_1})$ before the clipping to $[0,1]$. The term is deterministic, with
no random coefficient. As with \ac{SGPSO}, $c_g = 0$ recovers plain \ac{DE} exactly.

\myparagraph{Hyperparameters}
The tunable hyperparameters of \ac{DE} are the population size $k$, the scale factor $\zeta$ and the crossover probability $\mathit{CR}$; \ac{SGDE} adds $c_g$, $\beta$ and the weights $w_k$.

\subsection{Genetic Algorithms}\label{ss:ga}

To search for approximate symmetries directly in the discrete space of permutations, we adapt a \acf{GA}-based strategy. \Acp{GA}, see~\cite{Holland1975, goldberg1989genetic}, are population-based optimization methods that iteratively improve a set of candidate solutions through selection, combination and random variation. In our setting, each solution represents a permutation of the graph vertices, and its quality ({\em fitness}) is assessed by both the mismatch in adjacency structure and the similarity between vertices and their images under the permutation, as quantified by centrality measures.

The algorithm maintains parallel populations ($k=10$), each containing $N=90$ individuals (candidate permutations). Populations evolve over discrete steps (``generations") using an elitist update scheme: a subset of the best-performing individuals remains unchanged between generations, while the rest of the population is replaced by ``offspring" candidates.

\myparagraph{Fitness evaluation}
For each individual permutation $P$, to define its fitness we evaluate three quantities: (i) the approximate symmetry score induced by $P$, (ii) the number of fixed points, and (iii) the mismatch of node-level centrality profiles.

The approximate symmetry induced by a permutation $P$ is quantified by the unnormalized
mismatch $D(P)=\lVert A-PAP^{T}\rVert_1$, related to the normalized symmetry score of Eq.~(\ref{eq:s_value}) by $\mathcal{S}(A,P)=D(P)/\bigl(n(n-1)\bigr)$. The number of fixed points is given by $F_{\mathrm{fp}}(P)=\sum_{i=1}^n P_{ii}$, where $n$ is the number of vertices of the graph. Permutations with more than $\lfloor 0.25n \rfloor$ fixed points are rejected to avoid trivial (near-identity) solutions.

Centrality mismatch is computed by comparing centrality values only for not-fixed vertices:
\[
C(P)=\frac{n}{3\bigl(n - F_{\mathrm{fp}}(P)\bigr)}
\sum_{\substack{i=1 \\ \pi(i)\neq i}}^n
\Big(
\Delta\Gamma_{deg}(i,P)+\Delta\Gamma_{bet}(i,P)+\Delta\Gamma_{bon}(i,P)
\Big),
\]
where $\Delta\Gamma_{\ast}(i,P)$ denotes the absolute difference between the degree, betweenness and Bonacich centrality value of vertex $i$ and those of its image $\pi(i)$ under the permutation $P$.

The composite fitness value is then defined as
\begin{equation}\label{eq:ga_fitness}
f(P)=\frac{n-1}{2n}\,D(P)+w\,C(P),
\end{equation}
which is minimized by the algorithm, with $w$ a weight; the unguided \ac{GA} minimizes $D(P)$ alone and mutates only by random swaps. In addition to $f(P)$, the unnormalised mismatch $D(P)$ and the number of fixed points $F_{\mathrm{fp}}(P)$ are retained for further selection control.

\myparagraph{Offspring generation}
New individuals are generated by selecting parents and applying recombination and mutation. Parent selection is performed using tournament selection: $k_{\mathrm{t}}$ individuals are sampled randomly, and the best one is chosen by minimizing $f(P)$; ties are broken by preferring lower $D(P)$ and then fewer fixed points.

Given the parents, offspring are produced via crossover with probability $p_{\mathrm{co}}$, using cyclic crossover (probability $0.7$) or order crossover (probability $0.3$). If crossover is not applied, the parents are copied without a change. Mutation is applied independently to each offspring with probability $p_\mathrm{mt}$: with probability $p_{\mathrm{r}}$ a random swap of two vertices is performed, otherwise a centrality-guided swap favoring vertices with similar centrality profiles is used. The mutation rate is adapted as $p_{\mathrm{mt}}=1-\delta_p$, where the population diversity $\delta_p$ is defined as the ratio of distinct individuals in the population: two individuals are considered distinct if their  $f(P)$ values differ or, if tied, if their permutation matrices differ. As a result, when the population collapses to many copies of the same or very similar solution, diversity drops and mutation pressure increases, renewing exploration of the permutation space.

\myparagraph{Global control}
With all populations having progressed to next generation, one step of the algorithm is completed, elite individuals are exchanged across populations, and the global best candidate is identified as one with lowest $D(P)$ value with ties broken by preferring lower $f(P)$ and then fewer fixed points $F_{\mathrm{fp}}(P)$. Elitism is implemented by retaining $N_{\mathrm{e}}$ of best-performing individuals unchanged into subsequent generations. The elite groups are exchanged across all populations, thus allowing globally strong solutions to propagate quickly. If the global best solution does not improve, a patience counter is increased. When patience counter is not multiple of $10$, populations are mixed cyclically exchanging $N/3$ individuals between populations. When patience counter reaches a multiple of $10$, a restart is performed in which all populations are reinitialized with only the accumulated elite set preserved (i.e. the $N_{\mathrm{e}}$ globally best individuals). The search is terminated if patience counter reaches $50$, or if a perfect solution is found, i.e. $D(P)=0$.

\myparagraph{Hyperparameters}
The tunable hyperparameters of \ac{GA} are $p_{\mathrm{co}}$, the initial value of $p_{\mathrm{mt}}$, $k_{\mathrm{t}}$ and $N_{\mathrm{e}}$; \ac{SGGA} adds $w$ and $p_{\mathrm{r}}$.

\subsection{Bilinear Approximate Symmetry Inference}\label{ss:basin}
\Ac{BASIN} is a novel continuous optimization strategy that we propose for detecting approximate graph symmetries by relaxing the discrete permutation search to a differentiable problem. Like \ac{QSA}~\cite{pidnebesna2025computing}, and like continuous formulations of graph matching and of the quadratic assignment problem more generally (e.g. Frank--Wolfe approaches~\cite{vogelstein2015fast}), it replaces the combinatorial search over permutations with optimization over soft assignment matrices, which is subsequently discretized into a permutation. It differs in parametrization: where \ac{QSA} optimizes a doubly stochastic matrix directly, \ac{BASIN} parametrizes vertex correspondences through latent embeddings learned as part of the objective, so node similarity is data-driven rather than given by predefined centrality descriptors. The objective combines this bilinear similarity model with regularization terms penalizing deviations from approximate double stochasticity and near-identity solutions with many fixed points.

\myparagraph{Model formulation}
Given an adjacency matrix $A \in \mathbb{R}^{n \times n}$, \ac{BASIN} introduces two learnable embedding matrices $S,T \in \mathbb{R}^{n \times m}$, where each row corresponds to a vertex embedding. These matrices are initialized at random and constrained to have unit $\ell_2$ norm row-wise.

Vertex similarity is modeled as a scaled bilinear matrix
\[
Z = \tau \, ST^\top,
\]
where $\tau>0$ is a scaling (inverse-temperature). This parameter rescales all pairwise scores prior to normalization and thereby controls the concentration of the resulting distributions: smaller values of $\tau$ lead to flatter, more diffuse similarities, while larger values produce increasingly peaked scores.

A row-wise softmax is then applied to obtain a soft assignment matrix $\hat{P}$
\[
\hat{P}_{ij} = \frac{\exp(Z_{ij})}{\sum_{k=1}^n \exp(Z_{ik})},
\qquad i,j=1,\dots,n.
\]
Each row of $\hat{P}$ defines a probability distribution over possible images of vertex $i$. Note, however, that the row-wise softmax enforces only row stochasticity; in general, $\hat{P}$ is \emph{not} doubly stochastic, as column sums are unconstrained and may deviate substantially from one. The matrix $\hat{P}$ can therefore be interpreted as a relaxed permutation matrix that is subsequently regularized toward approximate double stochasticity through the objective function.

\myparagraph{Objective function}
\Ac{BASIN} minimizes a composite objective that balances approximate symmetry with several regularization constraints. Specifically, the optimization problem is defined as
\[
\mathcal{L} =
\mathcal{L}_{\mathrm{sym}}+
\lambda_{\mathrm{fp}}\mathcal{L}_{\mathrm{fp}}+
\lambda_{\mathrm{ds}}\mathcal{L}_{\mathrm{ds}}+
\lambda_{\mathrm{ortho}}\mathcal{L}_{\mathrm{ortho}},
\]
where $\mathcal{L}_{\mathrm{sym}}$ measures the discrepancy between the adjacency matrix and its softly permuted version $\hat{P}$, $\mathcal{L}_{\mathrm{fp}}$ penalizes fixed points, $\mathcal{L}_{\mathrm{ds}}$ penalizes deviation from double stochasticity of $\hat{P}$, and
$\mathcal{L}_{\mathrm{ortho}}$ promotes diversity of embedding dimensions. The coefficients $\lambda_{\cdot}$ control the relative contribution of each term. All components of the objective function are differentiable with respect to the learnable embedding matrices $S$ and $T$. In practice, some regularization components can be optionally disabled during hyperparameter optimization; when a constraint is inactive, the corresponding weight $\lambda_{\cdot}$ is set to zero. We now define each component in detail.

\myparagraph{Symmetry loss}
The main objective is to minimize the discrepancy between adjacency matrix $A$ and its permuted version. Since $\hat{P}$ is not a true permutation matrix, the transformed adjacency $\hat{P}A\hat{P}^\top$ is a continuous relaxation. We therefore use the fixed-point-normalized symmetry score of Eq.~\ref{eq:s_fp_value} with the soft fixed-point count $\hat{F}=\sum_i \hat{P}_{ii}^2$, $\mathcal{L}_{\mathrm{sym}}=\lVert A-\hat{P}A\hat{P}^\top\rVert_F^2/\max\bigl(n(n-1)-\hat{F}(\hat{F}-1),\,4n-6\bigr)$.

\myparagraph{Fixed-point penalty}
To suppress near-identity solutions in which many vertices map to themselves, we penalize such fixed points with a barrier that diverges at the identity:
\[
\mathcal{L}_{\mathrm{fp}} = \frac{f}{1-f}, \qquad f=\frac{1}{n}\sum_{i=1}^n \hat{P}_{ii}^2,
\]

\myparagraph{Double-stochasticity constraint}
While rows of $\hat{P}$ sum to one by construction, its columns need not sum to one. Thus we encourage approximate double stochasticity via
\[
\mathcal{L}_{\mathrm{ds}}=
\sum_{i=1}^n \Bigl(\sum_{j=1}^n\hat{P}_{ij} -1\Bigr)^2+
\sum_{j=1}^n \Bigl(\sum_{i=1}^n\hat{P}_{ij} -1\Bigr)^2.
\]

\myparagraph{Embedding orthogonality}
To prevent degenerate solutions, we encourage approximate orthogonality between embedding dimensions by minimizing
\[
\mathcal{L}_{\mathrm{ortho}}=
\|S^\top S -I\|_F + \|T^\top T - I\|_F,
\]
where $I$ is an identity matrix.

\myparagraph{Optimization and discretization}
The embedding matrices $S$ and $T$ are iteratively updated using the Adam optimizer, a popular first-order gradient-based optimizer that adaptively rescales step sizes based on running estimates of gradient moments. The optimization performed with learning rate $\eta$. After each update, rows of $S$ and $T$ are renormalized to unit norm.

Optimization is run for a maximum of $4000$ iterations, with early stopping applied if no improvement in the objective function $\mathcal{L}$ is observed for 50 consecutive steps. The globally best soft assignment matrix $\hat{P}^\star$ observed during the entire optimization run is retained.

To obtain a discrete permutation, $\hat{P}^\star$ is projected onto the space of permutation matrices in three ways: the Hungarian algorithm, a greedy row-wise maximization, and the Hungarian algorithm applied with the diagonal forbidden. In the greedy projection, rows are processed sequentially and each row is assigned to the column with the largest value among the columns that have not yet been assigned. Once a column is selected, it is excluded from all subsequent assignments. This ensures that each row and each column is selected exactly once and therefore yields a valid permutation matrix. The candidate with the lowest $\mathcal{S}'(A,P)$ is returned.

\myparagraph{Hyperparameters}
\Ac{BASIN} is controlled by the embedding dimension $m$, the temperature parameter $\tau$, and learning rate $\eta$. The embedding dimension $m$ can be specified either directly as the dimensionality of the vertex embeddings or as a graph-size--dependent ratio, in which case $m$ is set proportionally to the number of vertices. The ratio form is used here. Additional hyperparameters control the inclusion and weighting of regularization terms, including the fixed-point penalty $\lambda_{\mathrm{fp}}$, double-stochasticity constraint $\lambda_{\mathrm{ds}}$, and the orthogonality penalty $\lambda_{\mathrm{ortho}}$.

\section{Testing pipeline}\label{s:testing}

This section describes a testing pipeline for comparing strategies for calculating approximate symmetry. The basic idea is to utilize the various versions of random models described in Section~\ref{ss:data} to create datasets, on which are the optimization strategies then tested using the procedure outlined in Section~\ref{ss:pipeline}.

\subsection{Data}\label{ss:data}

To test the effectiveness of algorithms for detecting approximate symmetry, we employed several widely used models in complex networks that capture various aspects of these networks and exhibit different forms of symmetry. To maintain comparability with previous studies, we partially relied on a set of models used in a recently published study~\cite{pidnebesna2025computing}. First, let us note that all models have a parameter that can be used to control the density of the graph.

The first model is the \acf{ER} model~\cite{erdos1959random}, which represents the most random type of network and is not typically a good model for real networks. However, it often serves as an important reference model; moreover, it is almost surely asymmetric, with the probability of a non-trivial automorphism vanishing as $n$ grows~\cite{erdos1963asymmetric}.

The model that better represents real complex networks, primarily by showing a scale-free degree distribution, is the Barab\'{a}si-Albert model (BA)~\cite{barabasi1999emergence}. The BA model also tends to have fewer symmetries than the corresponding real network it models. For this reason, subsequent extensions have been developed to improve symmetry representation~\cite{xiao2008emergence}. It can therefore be assumed that the realizations of the BA model are relatively asymmetric but may contain non-trivial approximate symmetries.

The third model used is the {\em lateral random model} (LRM)~\cite{pidnebesna2025computing} inspired by the lateral symmetry of the brain~\cite{Jajcay2022}. In its basic configuration, this model contains two parts that are mutually perfectly symmetrical, which is also, with high probability, the only symmetry in the entire system. By adjusting the parameter, this symmetry can be modified to approximate symmetry in a controllable manner—it is possible to set how far this approximate symmetry will be from absolute symmetry. Since the principle of setting this distance is based on rewiring a given number of edges, this parameter is referred to as rewiring. The rewiring level $\mathrm{rew} \in \{0, 10, 20, 30\}$ is the percentage of the edges of the unrewired model that are rewired, $k = \max(1, \mathrm{round}(\mathrm{rew}\,m/100))$. At the same time, this model includes a parameter that allows for setting the network density. The purpose of this model is to represent a family of random networks inspired by the brain as a real system with a well-defined approximate symmetry (lateral symmetry of the brain hemispheres). Similar symmetries between modules/parts can be expected in other real-world complex systems~\cite{sanchez2020exploiting}.

The fourth model is the \acf{MMM}, which we introduce here. Like the \ac{LRM}, it consists of two mutually perfectly symmetrical halves, but each half is a stochastic block model~\cite{holland1983stochastic} of four equal-size modules, whose internal densities increase from one module to the next and which are only sparsely connected to one another. The modules therefore differ in eigenvector and betweenness centrality, while the degrees within a half are made as similar as possible, so that degree alone does not reveal which vertices are mirror images. Each vertex is joined to its mirror image with probability $0.85$. The symmetry is made approximate by the parameter $\mathrm{pert}$, the number of double-edge swaps applied to the graph, each of which replaces two edges $\{u,v\}$ and $\{x,y\}$ by $\{u,y\}$ and $\{x,v\}$ and so leaves the degree of every vertex unchanged, and the model again includes a parameter that sets the network density. The purpose of this model is a symmetry that maps whole modules onto each other, on which centrality carries information about the symmetry.

Every graph in both sets is connected, enforced by rejection sampling at generation, since isolated vertices are interchangeable at no structural cost.

\subsection{Real networks}\label{ss:data_real}

\subsubsection{Structural connectomes}\label{sss:data_sc}
We also evaluate on $88$ structural brain connectomes of healthy adults released with~\cite{skoch2022human}, parcellated into the $90$ cortical and subcortical regions of the \ac{AAL} atlas~\cite{tzouriomazoyer2002automated}. The diagonal is zeroed and the matrices, which are row-normalized, are symmetrized as $(A + A^{\mathrm{T}})/2$.

The matrices are binarized by two thresholding schemes. The first is proportional thresholding at density $d \in \{5, 10, 15, 40\}\,\%$: to keep every graph connected, we first take the edges of the maximum spanning tree and then add the strongest remaining edges until the density $d$ is reached, so all subjects have the same number of edges at a given density. The second is the \acf{OMST} construction~\cite{dimitriadis2017omst} at $k \in \{2, 4, 7, 15\}$ trees, which retains $k(n-1)$ edges; at $k=1$ the graph is a tree.

The homotopic permutation $\pi_{\mathrm{LR}}$ maps every region to its counterpart in the other hemisphere. It is the hemispheric symmetry of the brain, which in a connectome holds only approximately. Because the \ac{AAL} ordering interleaves the hemispheres, it has the closed form $\pi_{\mathrm{LR}} = [1,0,3,2,\dots]$. It serves as a reference, not a target: the searches are not informed about it.

\subsubsection{Benchmark networks with known automorphism groups}\label{sss:data_bench}
Three public networks whose automorphism groups can be computed exactly are taken from the Netzschleuder repository~\cite{peixoto2020netzschleuder}: \textsc{jazz}, a collaboration network of jazz bands~\cite{gleiser2003community}; \textsc{pdz}, a protein-interaction network of PDZ domains~\cite{beuming2005pdzbase}; and \textsc{netscience}, a co-authorship network of network scientists~\cite{newman2006finding}. Each is reduced to its largest connected component as a simple undirected graph, giving $n = 198$, $m = 2742$ for \textsc{jazz}; $n = 161$, $m = 209$ for \textsc{pdz}; and $n = 379$, $m = 914$ for \textsc{netscience}.

The automorphism group of each network is computed exactly with BLISS~\cite{junttila2007engineering}; it has $128 \approx 10^{2.11}$, $10^{34.85}$ and $10^{41.78}$ elements for \textsc{jazz}, \textsc{pdz} and \textsc{netscience}. Much of it consists of exchanges of twins, vertices with the same neighbors, which can be swapped without changing the graph. The permutations that only exchange twins form a subgroup of $10^{2.11}$, $10^{34.55}$ and $10^{40.57}$ elements, the whole group on \textsc{jazz}. The group is then used to evaluate the permutations found by the search algorithms (Section~\ref{sss:evaluation}).

On these networks the strategies that carry the cap on $\mathrm{tr}(P)$ (Section~\ref{ss:hyperparameter_optimization}) run with it relaxed to $\mathrm{tr}(P) \le n-2$, because their automorphisms are local, a single twin swap in \textsc{jazz} having $F/n = 0.990$. Thus a cap at $0.25n$ would forbid every automorphism they have. Every strategy is run for $20$ restarts on each network.

\subsection{Pipeline}\label{ss:pipeline}

We use a two-stage pipeline. The first stage performs hyperparameter optimization on a small, independent set of graphs, i.e., the training set. In the second stage, all strategies are evaluated on a much larger testing set of graphs. The two sets are strictly separated to ensure that hyperparameter tuning and final evaluation are performed on different graph instances.

\subsubsection{Hyperparameter optimization}\label{ss:hyperparameter_optimization}
To form the hyperparameter-search set (the training set), we generate graphs from all four generative models of Section~\ref{ss:data}. We vary their parameters as follows:
\begin{itemize}
    \item \textbf{\ac{ER} and \ac{BA}}
    \begin{itemize}
        \item Size: $n \in \{20, 50, 100\}$
        \item Edge density: $d \in \{0.15, 0.40\}$
    \end{itemize}
    \item \textbf{\ac{LRM}:}
    \begin{itemize}
        \item Size: $n \in \{50, 100\}$
        \item Edge density: $d \in \{0.15, 0.40\}$
        \item Rewiring: $\mathrm{rew} \in \{10, 30\}$
    \end{itemize}
    \item \textbf{\ac{MMM}:}
    \begin{itemize}
        \item Size: $n = 100$
        \item Edge density: $d \in \{0.15, 0.30\}$
        \item Perturbations: $\mathrm{pert} \in \{4, 16\}$
    \end{itemize}
\end{itemize}
Each combination of model $\times$ size $\times$ density $\times$ difficulty defines one graph type. In total, this yields twenty-four graph types. For each type, we sample three independent realizations, giving a training set of 72 graphs.

Hyperparameter search is then performed jointly on this set of 72 graphs. For each hyperparameter configuration suggested by the Tree-Parzen optimizer (a sequential model-based search method) using default implementation provided by Optuna~\cite{optuna_2019}, we run the given symmetry-optimization strategy on all training graphs, compute the resulting $\mathcal{S}'(A,P)$ value (Eq.~\ref{eq:s_fp_value}) for each graph (taking the minimum across four random initializations), and use the average $\mathcal{S}'(A,P)$ across the 72 graphs as the objective value to minimize.

The trial budget is 500 for \ac{BASIN} and for \ac{SA}, \ac{QSA}, \ac{TS} and their guided variants and 200 for the costlier \ac{GA}, \ac{PSO}, \ac{DE} and their guided variants (Table~\ref{tab:hyperparam-searches}). Each strategy is tuned once, and its configuration is held fixed for every experiment reported.

\myparagraph{Feasibility}
A permutation is flagged as degenerate when $F > 0.25n$ while $D > 0$. Where a strategy allows for a cap on $\mathrm{tr}(P)$ --- \ac{SA}, \ac{TS} and the \ac{GA} fitness --- the cap is set to $\lfloor 0.25n \rfloor$ in both tuning and at evaluation. For the remaining strategies the rule constrains the sampler: each training graph contributes a hinge violation $\max(0, (F - 0.25n)/n)$, the trial's violation is the maximum over graphs, and the objective remains $\mathcal{S}'(A,P)$. At evaluation, we do not discard a solution for its fixed-point count.

\subsubsection{Evaluation on the testing set}\label{sss:evaluation}
After hyperparameters are selected, we evaluate all optimization methods on a separate testing set. This testing set is generated from the same families of graph models as the training set but includes more granular parameter values:
\begin{itemize}
    \item \textbf{\ac{ER} and \ac{BA}:}
    \begin{itemize}
        \item Size: $n \in \{20, 50, 100\}$
        \item Edge density: $d \in \{0.15, 0.40\}$
    \end{itemize}
    \item \textbf{\ac{LRM}:}
    \begin{itemize}
        \item Size: $n \in \{50, 100\}$
        \item Edge density: $d \in \{0.15, 0.40\}$
        \item Rewiring: $\mathrm{rew} \in \{0, 10, 20, 30\}$
    \end{itemize}
    \item \textbf{\ac{MMM}:}
    \begin{itemize}
        \item Size: $n \in \{50, 100\}$
        \item Edge density: $d \in \{0.15, 0.30\}$
        \item Perturbations: $\mathrm{pert} \in \{0, 4, 8, 16\}$
    \end{itemize}
\end{itemize}
This gives forty-four graph types of $100$ graphs each, $4400$ in total. The test set generators use seeds distinct from the training set so there is no risk of identical graphs appearing in both sets.

For each graph-strategy pair, the symmetry search is repeated five times using independent random initialization. This results in five candidate solutions, each with their associated values of $\mathcal{S}(A,P)$ (Eq.~\ref{eq:s_value}) and $\mathcal{S}'(A,P)$ (Eq.~\ref{eq:s_fp_value}). Unless stated otherwise, the best (i.e. lowest) value across five runs is reported as the performance of the strategy for given graph. Throughout the evaluation, we use the fixed-point–normalized symmetry score $\mathcal{S}'$ as the primary metric. This choice reflects the setting considered in this work, where permutations are allowed to contain fixed points but near-identity mappings should be penalized, ensuring that detected symmetries correspond to non-trivial structural rearrangements.

\myparagraph{Statistical comparison}
To evaluate the differences between the tested optimization methods, we use Nemenyi rank-based tests and visualize the results using critical difference plots~\cite{nemenyi1963distribution, demvsar2006statistical}.

Where one condition per graph family is shown, it is the hardest condition of that graph family: $n=100$, the higher density ($d=40~\%$, $30~\%$ for the \ac{MMM}) and, where the graph family has a perturbation axis, its highest level ($\mathit{rew}=30$ for the \ac{LRM}, $\mathit{pert}=16$ for the \ac{MMM}).

\myparagraph{Distance to known symmetries}
On the \ac{LRM} and the \ac{MMM} the planted symmetry map $\sigma$ is known, and each returned permutation $\pi$ is also compared with it. At zero perturbation, where $\sigma$ is an exact automorphism, we report the share of restarts returning it. At higher perturbation we report the Cayley distance $d_{\mathrm{C}}(\sigma,\pi) = n - c(\sigma\pi^{-1})$, where $c$ counts the cycles of $\sigma\pi^{-1}$: the minimum number of transpositions taking one permutation to the other, $0$ only when $\pi = \sigma$. It is normalized by its maximum $n-1$.

On the benchmark networks (Section~\ref{sss:data_bench}) the automorphism group is known, and each permutation returned by a search strategy is checked against it. A permutation is exact when it is an automorphism. For one that is not, we measure how close it came as the smallest number of vertices whose image must be changed to turn it into an automorphism. Against the whole group this requires comparing the permutation with every element, which is feasible only on \textsc{jazz}, so we compute it against the twin subgroup, where it has a closed form: it counts the vertices mapped outside their own twin class, plus the vertices outside any twin class that are not fixed. It equals the distance to the full group on \textsc{jazz} and is an upper bound on it elsewhere.

\myparagraph{Hardware}
All experiments ran on a cluster with 16 nodes, each with one 32-core AMD EPYC 9374F processor and 376~GiB of memory, under Rocky Linux 9. \ac{BASIN} runs on the CPU although it could be run using GPU as well.

\section{Results}\label{s:results}
In this section, we evaluate the optimization strategies introduced in Section~\ref{s:methods} using the fixed-point--normalized symmetry score $\mathcal{S}'(A,P)$ (Eq.~\ref{eq:s_fp_value}) as the primary performance metric. Results obtained using the size-normalized score $\mathcal{S}(A,P)$ (Eq.~\ref{eq:s_value}) are reported in Appendix~\ref{app:size-normalized_S} (Figs.~\ref{fig:box_ER_S_random}--\ref{fig:box_MMM30_S}), and the guidance comparison under $\mathcal{S}(A,P)$ in Appendix~\ref{app:guidance_S} (Figs.~\ref{fig:gain_ER_S}--\ref{fig:gain_MMM30_S}); the qualitative performance patterns remain consistent across both evaluation metrics. Box figures show $\mathcal{S}'(A,P)$ for \ac{ER} at both edge densities, for \ac{LRM} at $40\,\%$ and for \ac{MMM} at $30\,\%$, one panel per condition; \ac{BA} and the $15\,\%$ density of \ac{LRM} and \ac{MMM} are in the Appendix (Figs.~\ref{fig:box_BA}--\ref{fig:box_MMM15}). Each algorithm is represented by its similarity-guided variant; both variants are compared in Appendix~\ref{app:comparison_without_one-variant_reduction} (Figs.~\ref{fig:box_ER_all}--\ref{fig:box_MMM30_all}).

The random-search baseline is omitted from the figures of this section, as it is consistently worse than every optimization-based strategy and would obscure the differences among them; the figures with random-search results are reported in Appendix~\ref{app:includint_random_search} (Figs.~\ref{fig:box_ER_random}--\ref{fig:box_MMM30_random}).

\subsection{Impact of node-similarity guidance}\label{ss:similarity_guidance}
\begin{figure}[ht]
    \centering
    \includegraphics[width=0.75\linewidth]{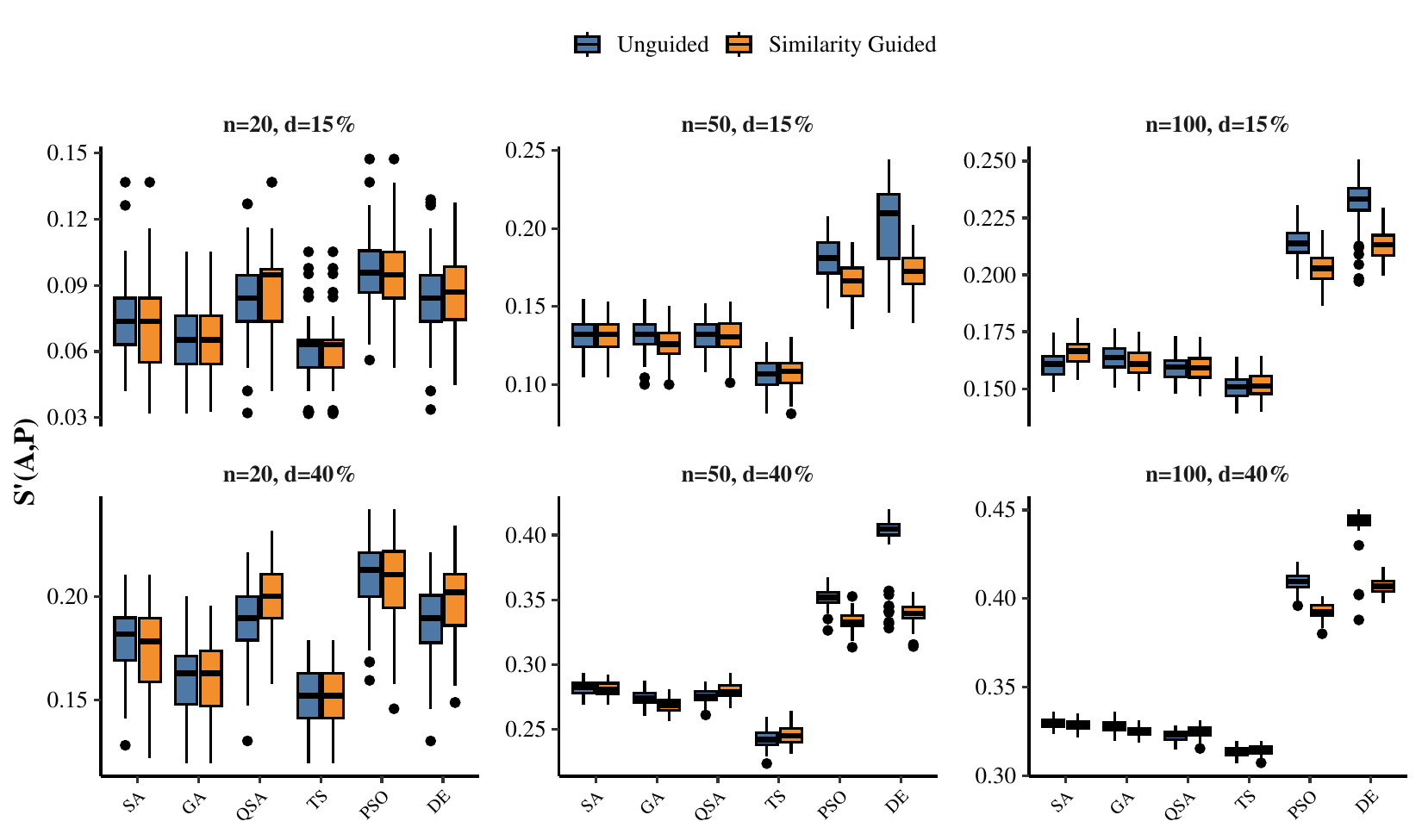}
    \caption{Node-similarity guidance on \ac{ER} graphs. Boxes show $\mathcal{S}'(A,P)$ over the 100 test graphs of each condition, one solution per graph, for the unguided and the similarity-guided variant of each algorithm; rows are edge densities $d$ of 15\% and 40\%, columns graph sizes $n$. \ac{BASIN} has no guided variant and is not shown.}
    \label{fig:gain_ER}
\end{figure}

\begin{figure}[ht]
    \centering
    \includegraphics[width=0.75\linewidth]{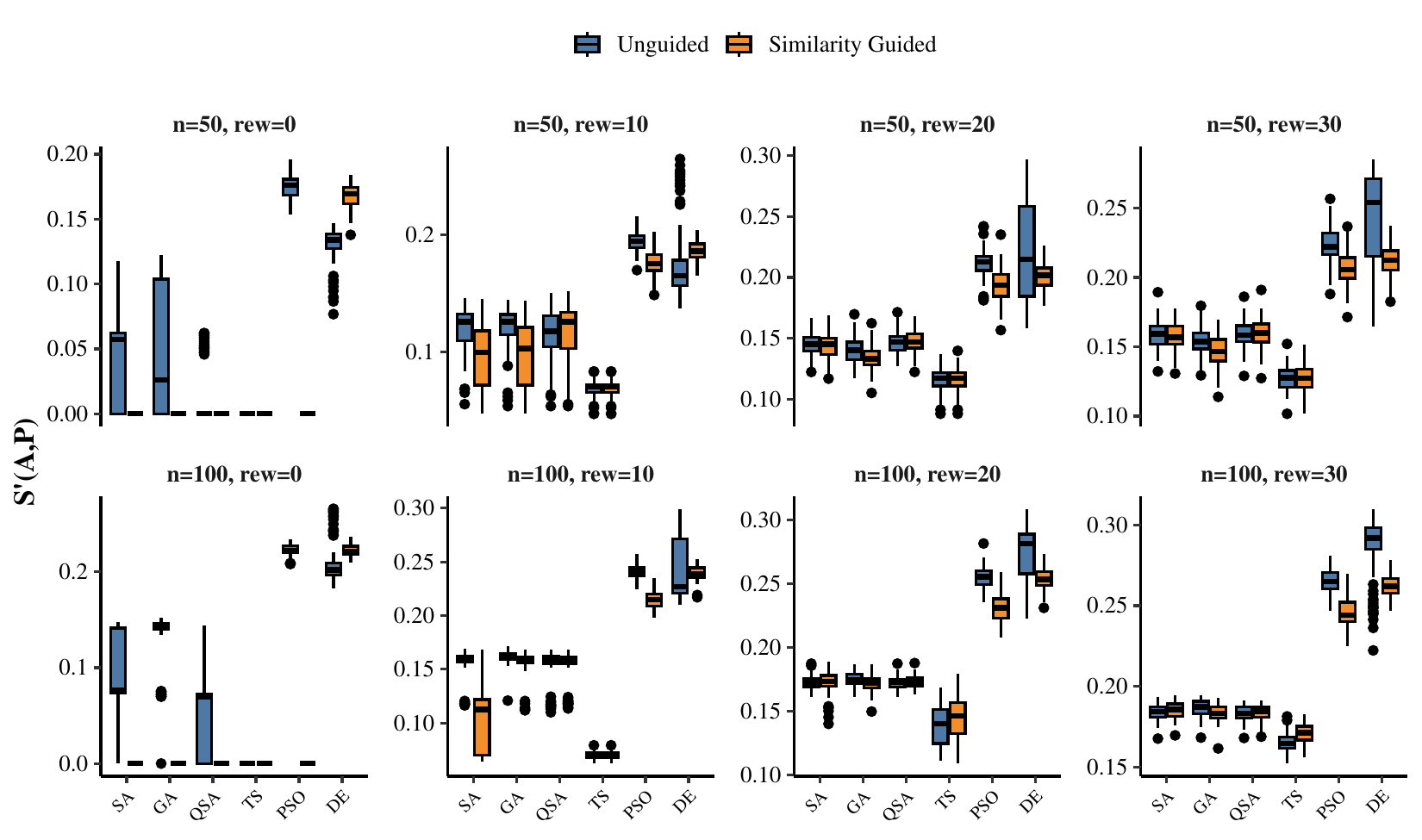}
    \caption{Node-similarity guidance on \ac{LRM} graphs at edge density $40~\%$. Boxes show $\mathcal{S}'(A,P)$ over the 100 test graphs of each condition, one solution per graph, for the unguided and the similarity-guided variant of each algorithm; one panel per graph size $n$ and share $\mathit{rew}$ of rewired edges. \ac{BASIN} has no guided variant and is not shown.}
    \label{fig:gain_LRM40}
\end{figure}

\begin{figure}[ht]
    \centering
    \includegraphics[width=0.75\linewidth]{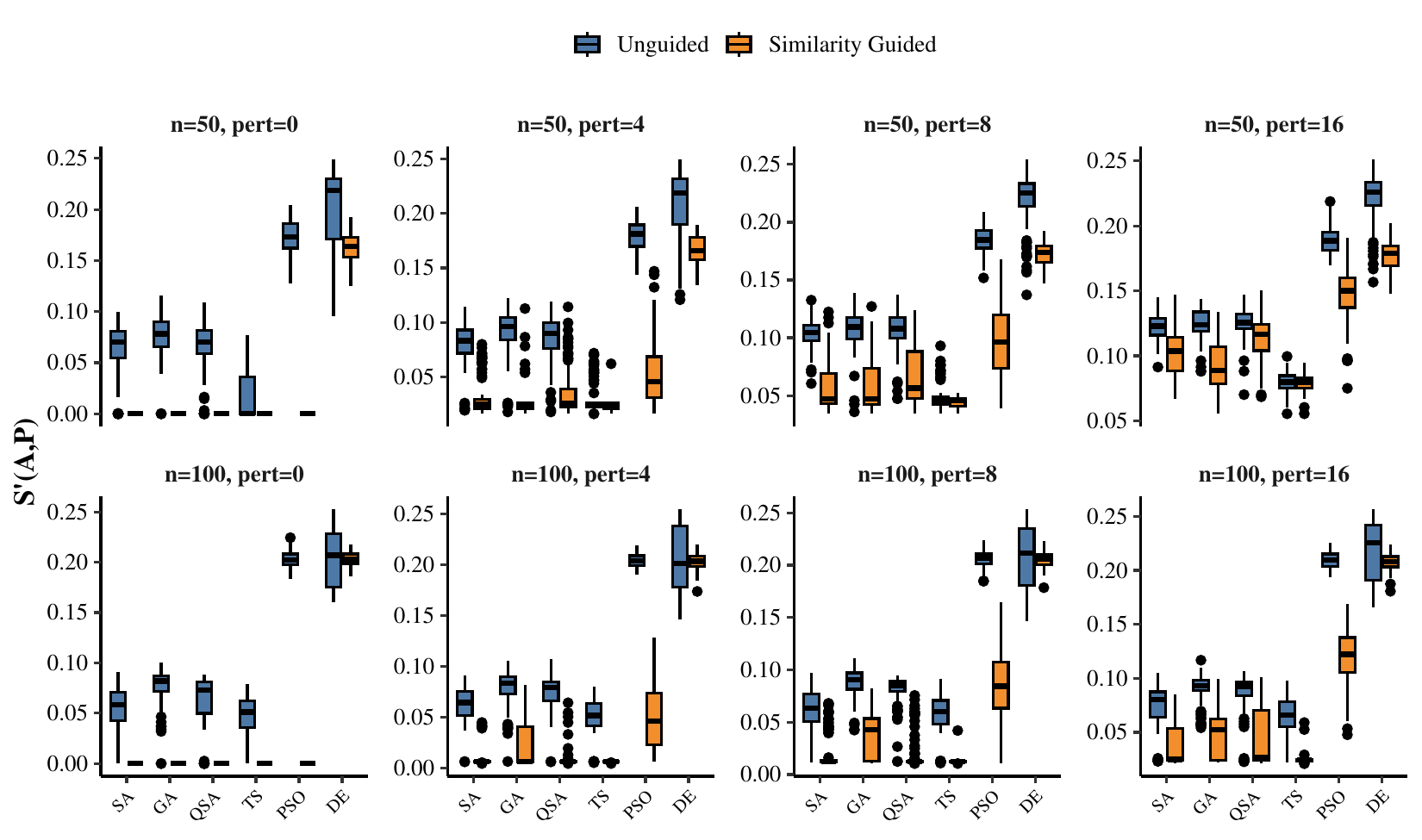}
    \caption{Node-similarity guidance on \ac{MMM} graphs at edge density $30~\%$. Boxes show $\mathcal{S}'(A,P)$ over the 100 test graphs of each condition, one solution per graph, for the unguided and the similarity-guided variant of each algorithm; one panel per graph size $n$ and number $\mathit{pert}$ of perturbing edge swaps. \ac{BASIN} has no guided variant and is not shown.}
    \label{fig:gain_MMM30}
\end{figure}

\begin{figure}[ht]
    \centering
    \includegraphics[width=\linewidth]{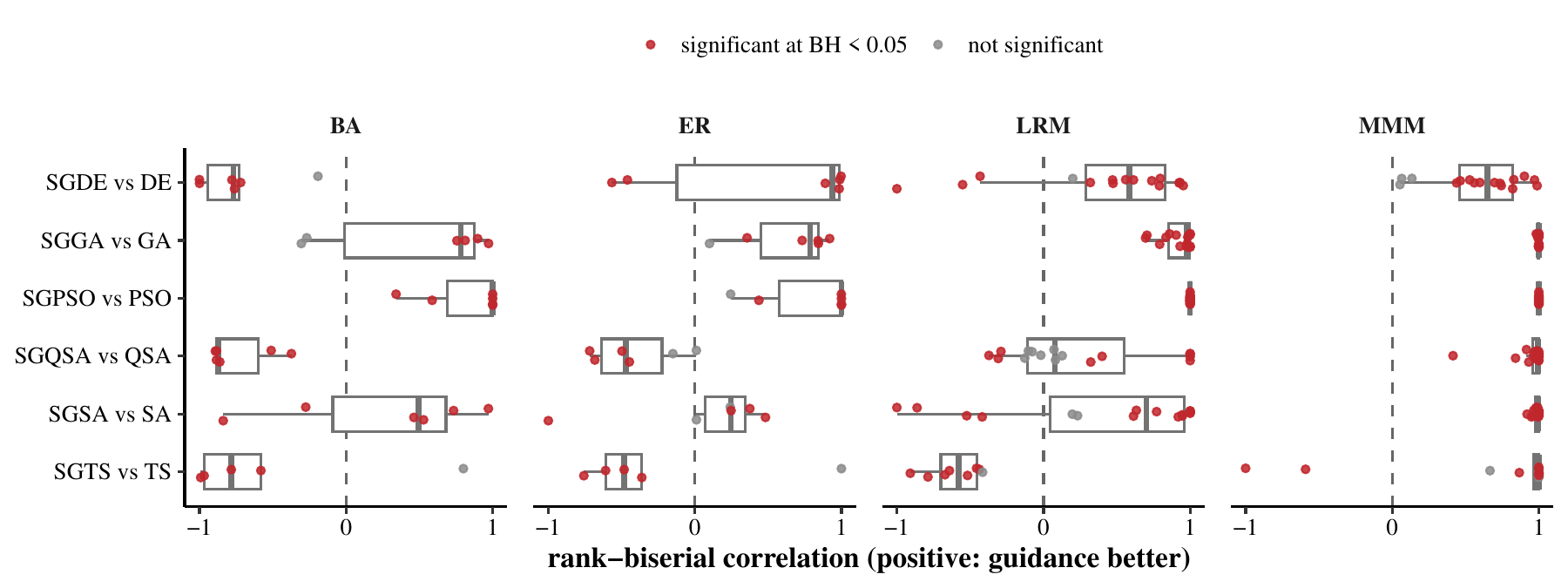}
    \caption{\label{fig:ablation-summary}Guidance ablation. Each point is one condition: the matched-pairs rank-biserial correlation of the paired Wilcoxon signed-rank test of guided against unguided $\mathcal{S}'(A,P)$ over its 100 graphs, positive where guidance is better. Boxes summarize the conditions of each graph family; colored points are significant after Benjamini--Hochberg correction at $0.05$.}
\end{figure}

Guided and unguided variants are compared in each condition, that is, each combination of size, density and perturbation level, by a Wilcoxon signed-rank test over its 100 graphs, Benjamini--Hochberg corrected within each contrast (Fig.~\ref{fig:ablation-summary}; distributions in Figs.~\ref{fig:gain_ER}, \ref{fig:gain_LRM40} and \ref{fig:gain_MMM30}, and for \ac{BA} and the $15\,\%$ densities in Appendix Figs.~\ref{fig:gain_BA}--\ref{fig:gain_MMM15}).

Guidance is decisive on the \ac{MMM}. \ac{SGGA}, \ac{SGQSA}, \ac{SGSA} and \ac{SGPSO} are significantly better in 16 conditions of 16, at a median rank-biserial effect of $+0.99$ or above in each case, and \ac{SGDE} in 13 at $+0.65$; \ac{SGTS} is better in 7 and worse in 2.

At zero perturbation, where the planted map is an exact automorphism, the guided variants of \ac{SA}, \ac{GA}, \ac{QSA}, \ac{TS} and \ac{PSO} return it on at least $0.854$ of restarts on the \ac{MMM}, and the plain ones on at most $0.408$ (Appendix Table~\ref{tab:planted-recovery} and Fig.~\ref{fig:planted-exact}).

With perturbation the planted map is no longer an automorphism. On the \ac{MMM} the median Cayley distance to it of each guided variant is at or below that of its unguided counterpart in every condition (Appendix Fig.~\ref{fig:planted-distance-mmm-all}).

Outside the \ac{MMM}, guidance helps some algorithms and harms others. \ac{SGTS} is significantly better in no condition of \ac{BA}, \ac{ER} or \ac{LRM} and significantly worse in 4 of 6, 4 of 6 and 7 of 16 respectively, with 8 of the \ac{LRM} conditions decided entirely by ties. \ac{SGDE} is significantly worse in 5 of 6 \ac{BA} conditions and better in none, but better in 4 of 6 \ac{ER} and 12 of 16 \ac{LRM} conditions against 2 and 3 worse. \ac{SGGA} and \ac{SGPSO} improve on every graph family and lose in no condition of any, \ac{SGGA} at 4 of 6 on \ac{BA}, 5 of 6 on \ac{ER} and 16 of 16 on both \ac{LRM} and the \ac{MMM}, \ac{SGPSO} at 6 of 6, 5 of 6, 16 of 16 and 16 of 16. \ac{SGQSA} is significantly worse in all 6 \ac{BA} and 4 of 6 \ac{ER} conditions, better in none, and on the \ac{LRM} better in 6 and worse in 3. \ac{SGSA} is better more often than worse on every graph family, in 4, 3 and 10 conditions of \ac{BA}, \ac{ER} and \ac{LRM} against 2, 1 and 4.

\subsection{Performance comparison across optimization strategies}\label{ss:optimization_strategies}

\begin{figure}[ht]
    \centering
    \includegraphics[width=0.75\linewidth]{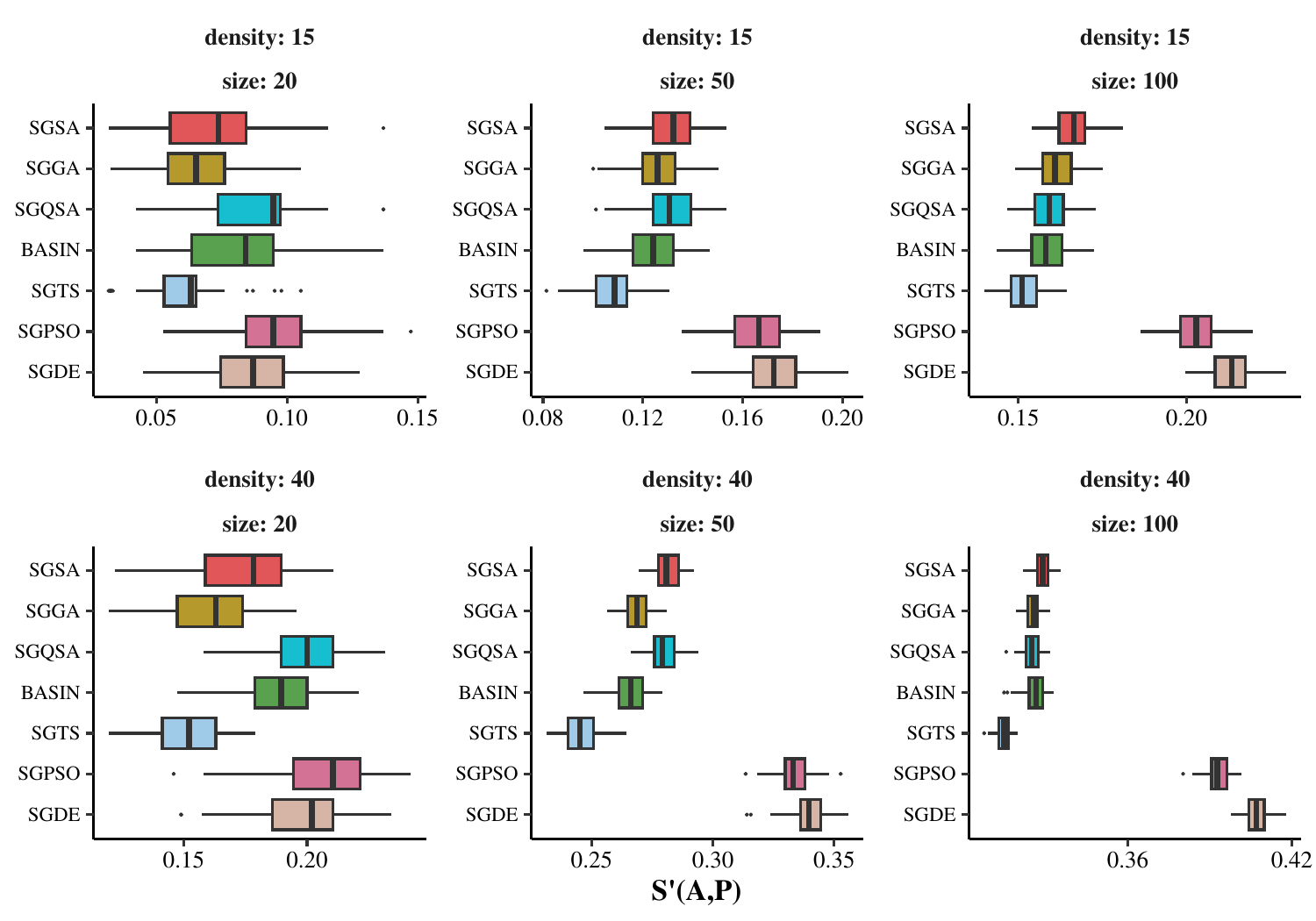}
    \caption{Performance on \ac{ER} graphs, one variant per algorithm. Boxes show $\mathcal{S}'(A,P)$ over the 100 test graphs of each condition, one solution per graph; one panel per graph size $n$ and edge density $d$, each with its own horizontal axis.}
    \label{fig:box_ER}
\end{figure}

\begin{figure}[ht]
    \centering
    \includegraphics[width=0.75\linewidth]{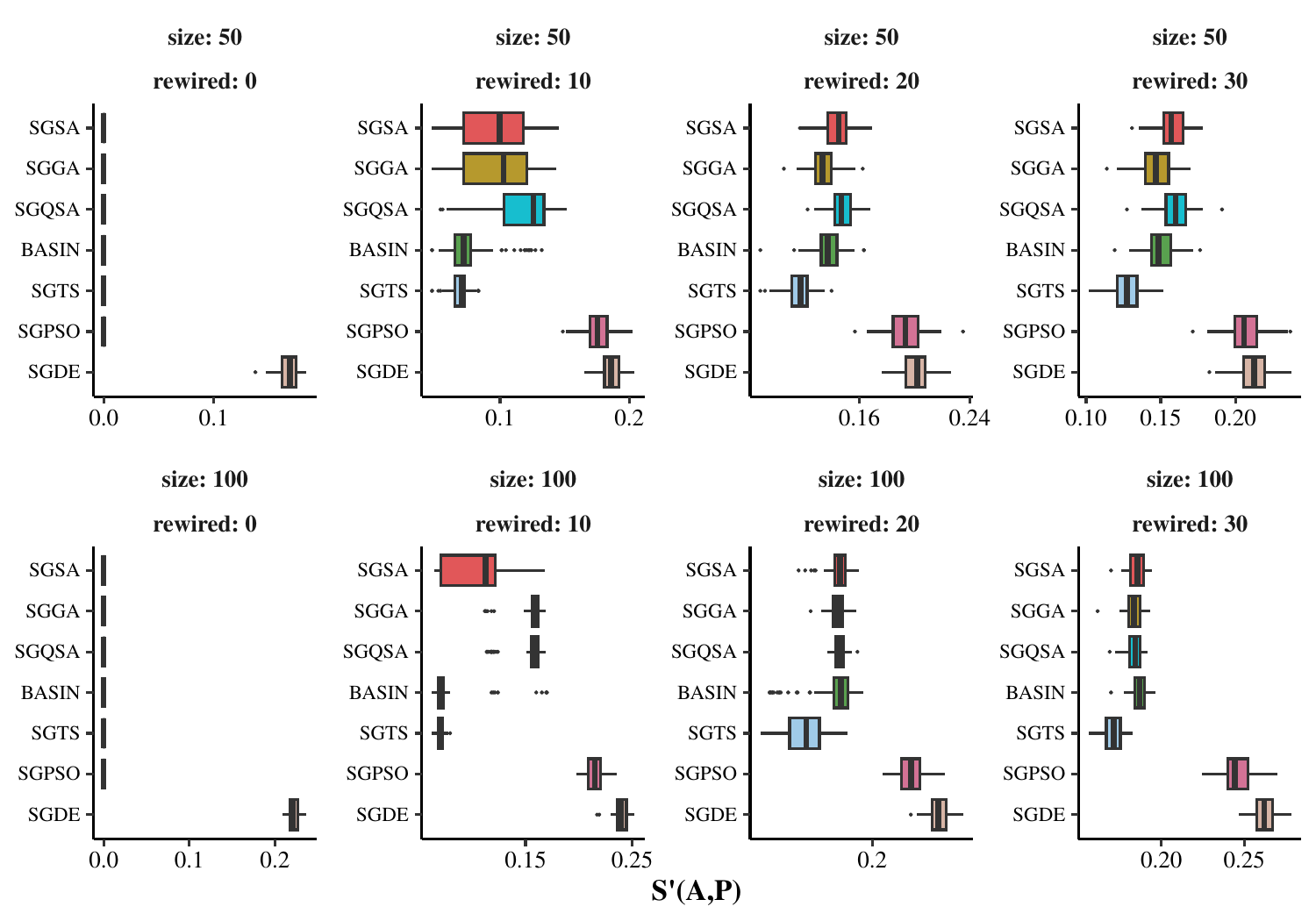}
    \caption{Performance on \ac{LRM} graphs at edge density $40~\%$, one variant per algorithm. Boxes show $\mathcal{S}'(A,P)$ over the 100 test graphs of each condition, one solution per graph; one panel per graph size $n$ and share $\mathit{rew}$ of rewired edges, each with its own horizontal axis.}
    \label{fig:box_LRM40}
\end{figure}

\begin{figure}[ht]
    \centering
    \includegraphics[width=0.75\linewidth]{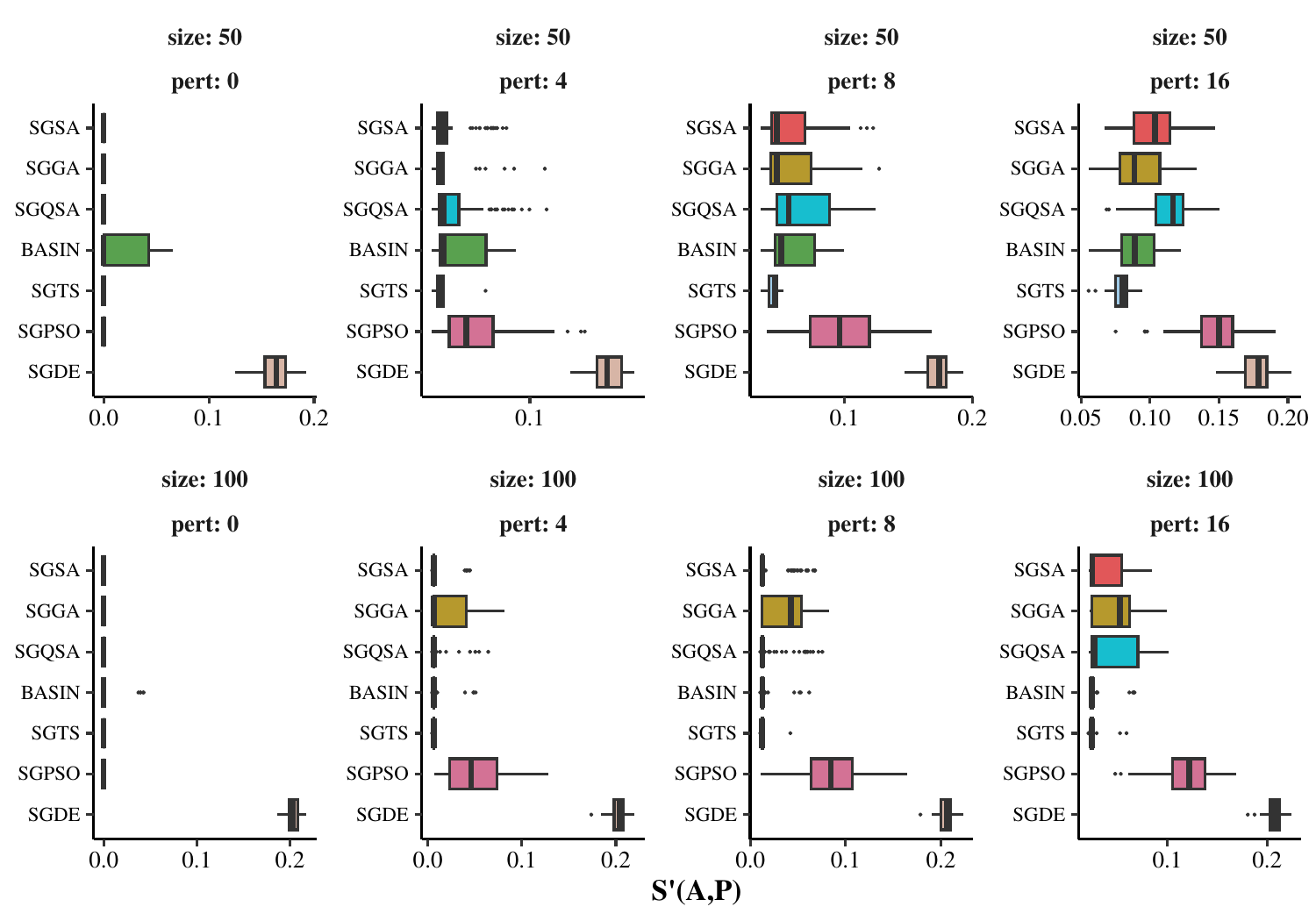}
    \caption{Performance on \ac{MMM} graphs at edge density $30~\%$, one variant per algorithm. Boxes show $\mathcal{S}'(A,P)$ over the 100 test graphs of each condition, one solution per graph; one panel per graph size $n$ and number $\mathit{pert}$ of perturbing edge swaps, each with its own horizontal axis.}
    \label{fig:box_MMM30}
\end{figure}

\begin{table}

\caption{\label{tab:family-ranks}Mean rank per graph family, with the number of conditions led in parentheses. Algorithms are ranked within a condition on its median $\mathcal{S}'(A,P)$, ties averaged, rank 1 lowest.}
\centering
\begin{tabular}[t]{lllll}
\toprule
Method & BA & ER & LRM & MMM\\
\midrule
Random & 8.00 (0) & 8.00 (0) & 8.00 (0) & 8.00 (0)\\
SGSA & 3.58 (0) & 4.33 (0) & 3.62 (0) & 3.19 (0)\\
SGGA & 2.50 (0) & 2.83 (0) & 3.38 (0) & 2.97 (0)\\
SGQSA & 4.75 (0) & 4.08 (0) & 4.06 (0) & 4.16 (0)\\
BASIN & 4.08 (0) & 3.00 (0) & 2.91 (0) & 3.25 (0)\\
SGTS & 1.00 (6) & 1.00 (6) & 1.66 (11) & 2.06 (6)\\
SGPSO & 5.92 (0) & 6.25 (0) & 5.38 (0) & 5.38 (0)\\
SGDE & 6.17 (0) & 6.50 (0) & 7.00 (0) & 7.00 (0)\\
\bottomrule
\end{tabular}
\end{table}

\begin{figure}[ht]
    \centering
    \includegraphics[width=\linewidth]{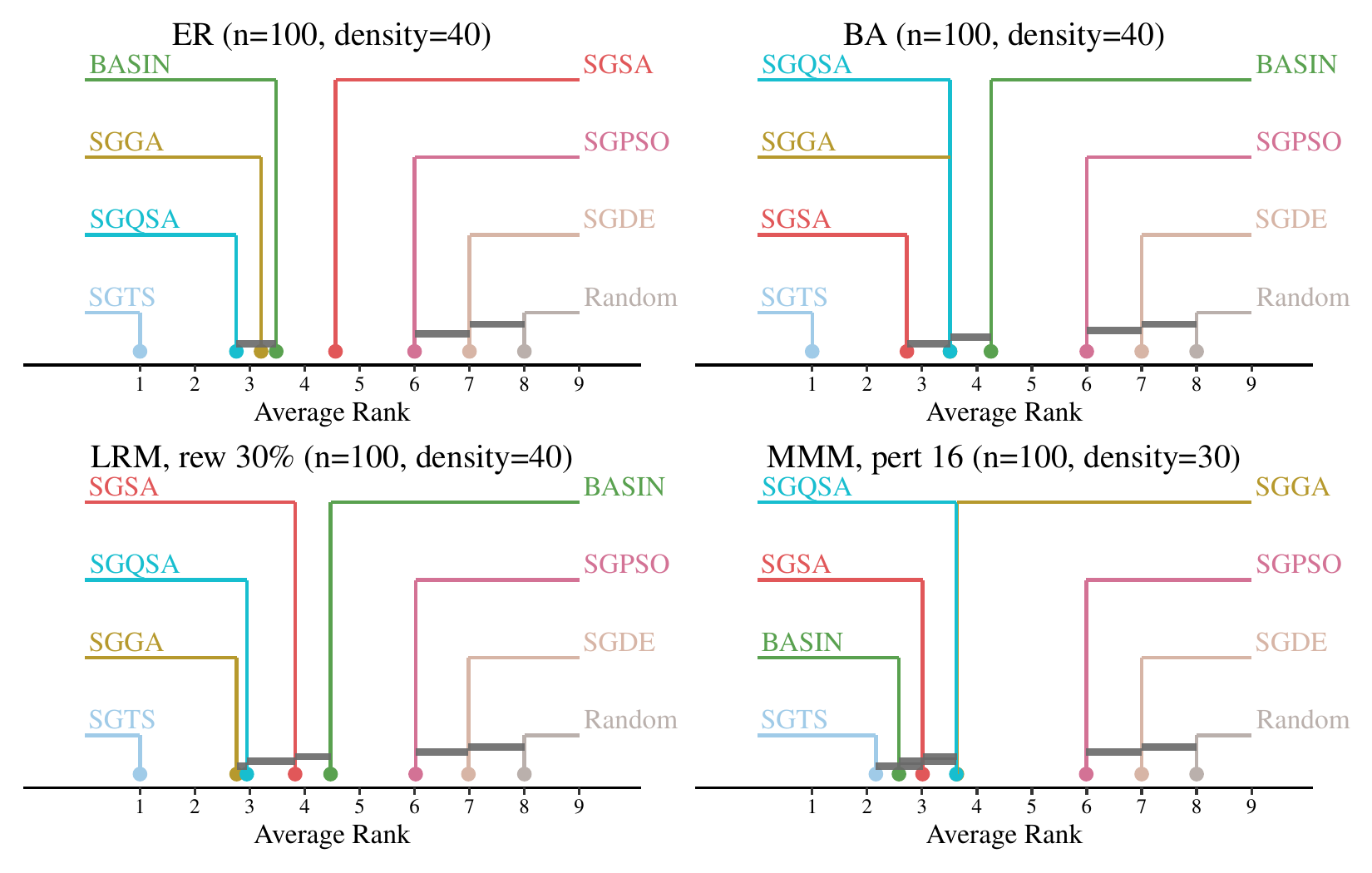}
    \caption{Critical-difference diagram (Friedman with post-hoc Nemenyi) at each graph family's hardest condition ($n=100$, $d=40~\%$, or $30~\%$ for the \ac{MMM}, \ac{LRM} $\mathit{rew}=30$, \ac{MMM} $\mathit{pert}=16$), one panel per graph family. Each algorithm is placed at its mean rank on $\mathcal{S}'(A,P)$ over the 100 test graphs of the condition, rank 1 lowest; algorithms joined by a bar do not differ at the $5~\%$ level.}
    \label{fig:CD_headline}
\end{figure}

\begin{figure}[ht]
    \centering
    \includegraphics[width=\linewidth]{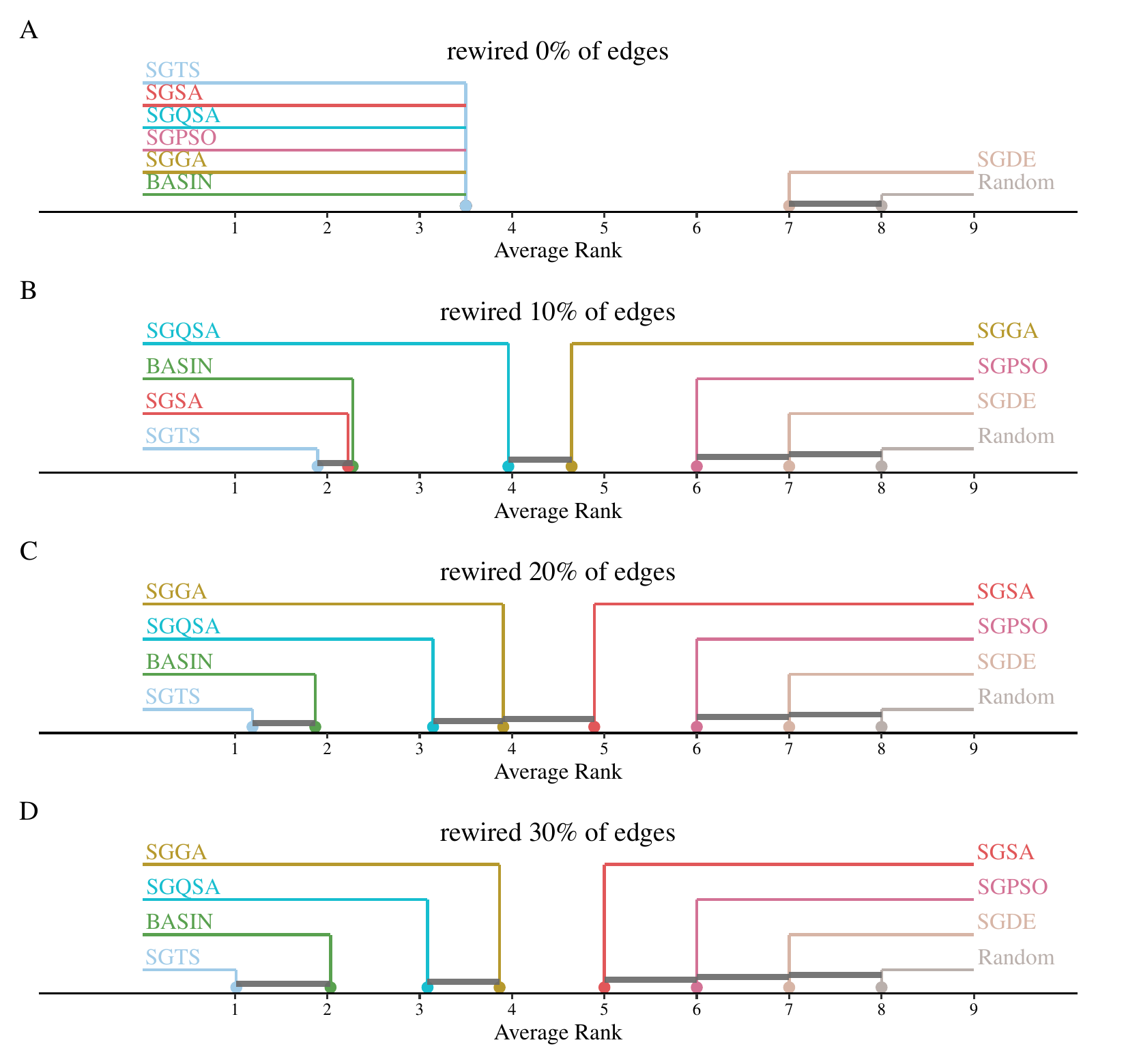}
    \caption{Critical-difference diagrams (Friedman with post-hoc Nemenyi) on \ac{LRM} graphs at $n=100$ and edge density $15~\%$, \textbf{(A)}--\textbf{(D)} at $\mathit{rew} = 0$, $10$, $20$ and $30~\%$ of edges rewired, one variant per algorithm and the random-search baseline. Each algorithm is placed at its mean rank on $\mathcal{S}'(A,P)$ over the 100 test graphs of the condition, rank 1 lowest; algorithms joined by a bar do not differ at the $5~\%$ level.}
    \label{fig:CD_LRM}
\end{figure}

\ac{SGTS} has the best mean rank in all four graph families (Table~\ref{tab:family-ranks}; Figs.~\ref{fig:box_ER}--\ref{fig:box_MMM30}). It attains the lowest median in all 44 conditions and is the sole leader in 29 (Appendix Table~\ref{tab:comparison}); the other 15 are ties it shares, 8 of them the unperturbed conditions in which six algorithms reach $\mathcal{S}'(A,P)=0$ together. The random baseline is last in every condition.

Behind \ac{SGTS} the field separates into two groups. \ac{SGGA}, \ac{SGQSA}, \ac{BASIN} and \ac{SGSA} have medians a few thousandths of $\mathcal{S}'(A,P)$ apart: at $n=100$ and $d=40~\%$ the hardest \ac{ER} condition separates \ac{SGTS} at $0.3147$ from \ac{SGSA} at $0.3289$, with the other three between them, against $0.4513$ for random search. \ac{SGPSO} and \ac{SGDE} follow.

At the hardest condition of each graph family (Fig.~\ref{fig:CD_headline}; the \ac{LRM} rewiring axis in Fig.~\ref{fig:CD_LRM}; every condition in Appendix~\ref{app:critical_difference}), on \ac{BA}, \ac{ER} and \ac{LRM}, \ac{SGTS} is significantly better than every other algorithm. On the \ac{MMM} it shares the group with \ac{SGSA} and \ac{BASIN}, whose medians lie between $0.0242$ and $0.0246$.

\begin{figure}[ht]
    \centering
    \includegraphics[width=0.95\linewidth]{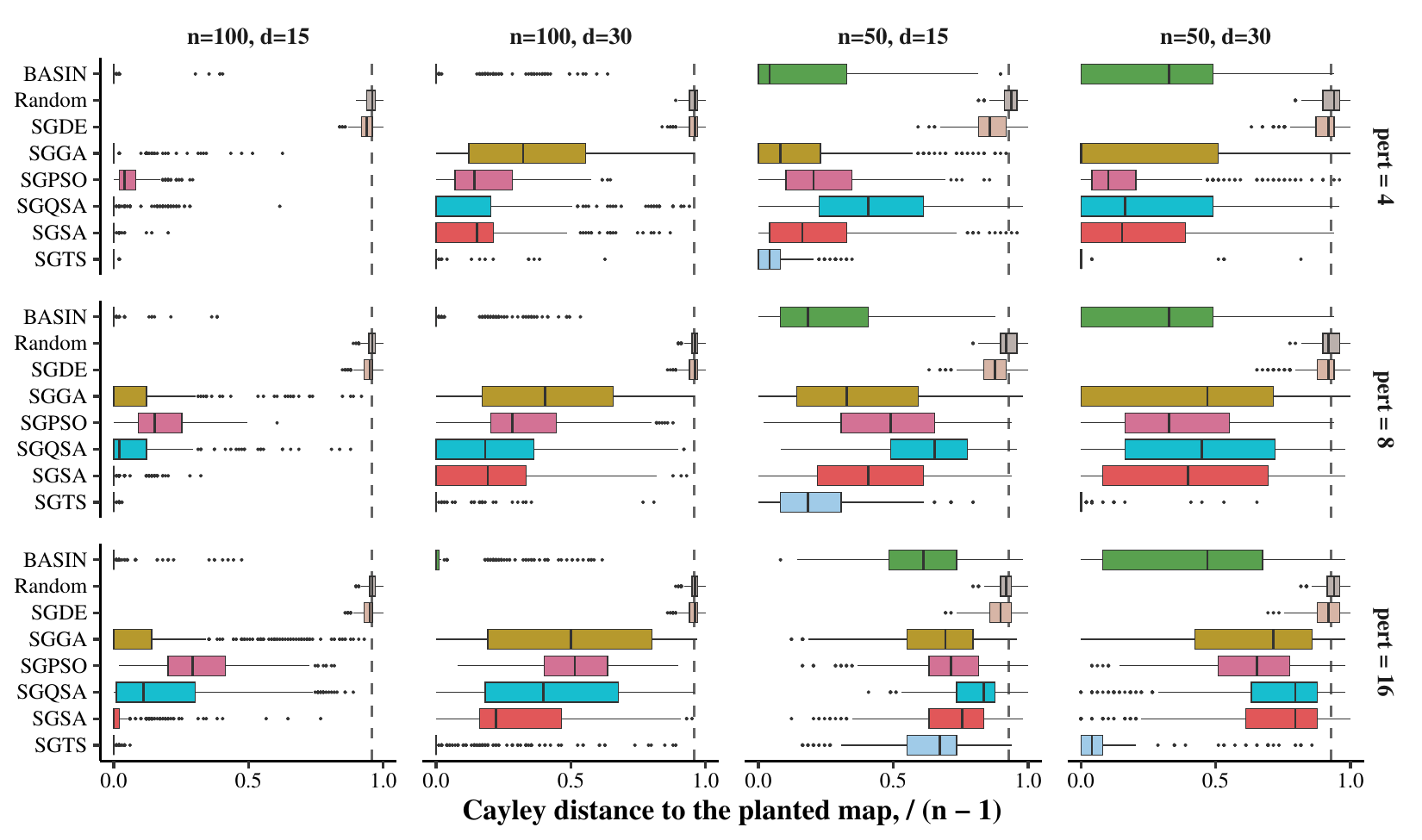}
    \caption{Cayley distance to the planted map over its maximum $n-1$ on \ac{MMM} graphs, one variant per algorithm and the random-search baseline. Boxes span every restart on the 100 test graphs of each condition, $0$ being the planted map itself; one panel per graph size $n$, edge density $d$ and number $\mathit{pert}\in\{4,8,16\}$ of perturbing edge swaps. The dashed line is the median distance of a random permutation without fixed points.}
    \label{fig:planted-distance-mmm}
\end{figure}

\myparagraph{Cayley distance to the planted map on the \ac{MMM}}
The planted map on the \ac{MMM} (Fig.~\ref{fig:planted-distance-mmm}; \ac{LRM} in Appendix Fig.~\ref{fig:planted-distance}) is recovered in every condition at $n=100$ by \ac{SGTS} and \ac{BASIN}, whose median Cayley distance to it is $0$. \ac{SGSA} and \ac{SGGA} recover it at $d=15~\%$ only, \ac{SGQSA} at $\mathit{pert}=4$ only, and the three lie up to $0.500$ from it at $d=30~\%$. \ac{SGPSO} recovers it at $\mathit{pert}=0$ only and lies up to $0.515$ from it at $d=30~\%$, and \ac{SGDE} misses it in every condition by about $0.95$, against $0.958$ for a random permutation without fixed points.

\subsection{Comparison of fixed-point ratios across optimization strategies}\label{ss:FP_across_optimization}

\begin{figure}[htbp]
    \centering
    \includegraphics[width=1\linewidth]{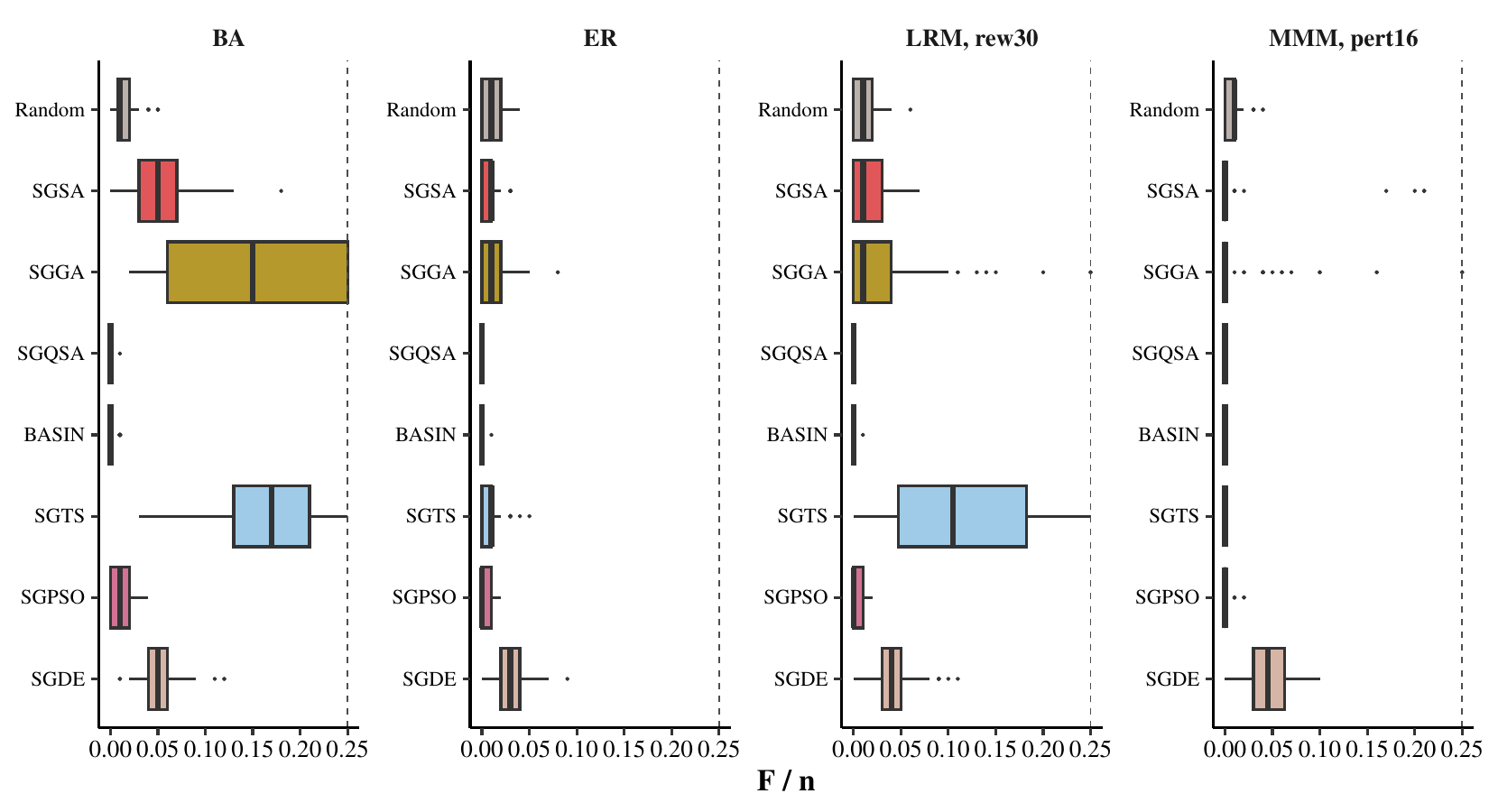}
    \caption{Fixed-point ratio $F/n$ of each graph's solution, over the 100 test graphs of the hardest condition of each graph family ($n=100$, $d=40~\%$, and $30~\%$ for the \ac{LRM} and \ac{MMM} respectively, \ac{LRM} $\mathit{rew}=30$, \ac{MMM} $\mathit{pert}=16$). The dashed line is the likely degeneracy threshold $0.25n$.}
    \label{fig:fp_hard}
\end{figure}

At the hardest condition of each graph family (Fig.~\ref{fig:fp_hard}), almost every algorithm returns few fixed points, with a median $F/n$ of at most $0.05$. The exceptions are \ac{SGTS} and \ac{SGGA} on \ac{BA}, at medians of $0.17$ and $0.15$, and \ac{SGTS} on the \ac{LRM} at $\mathit{rew}=30$, at $0.105$. \ac{SGGA} sits exactly on the threshold $F/n = 0.25$ on half of the \ac{BA} graphs, \ac{SGTS} on at most $7~\%$ of the graphs of either graph family. Fig.~\ref{fig:fp_all} pools all conditions, and Fig.~\ref{fig:fp_unnormalised} plots $F/n$ against $\mathcal{S}(A,P)$ on the \ac{LRM} at $d=40~\%$.

Over all testing graphs, only algorithms without the cap in their search exceed the threshold. \ac{SGSA}, \ac{SGGA} and \ac{SGTS}, which carry it, never do. Of the algorithms constrained during tuning only (Section~\ref{ss:hyperparameter_optimization}), \ac{SGPSO} exceeds it on a single graph, reaching $F/n = 0.40$ against the random baseline's $0.35$, while \ac{SGDE}, \ac{SGQSA} and \ac{BASIN} stay at or below $0.25$, $0.01$ and $0.10$ throughout.

\subsection{Ablation of \ac{BASIN} components}\label{ss:ablation_BASIN}

Every ablation worsens $\mathcal{S}'(A,P)$ on more graphs than it improves (Appendix Fig.~\ref{fig:basin-ablation-delta}). Removing the double-stochasticity penalty $\mathcal{L}_{\mathrm{ds}}$ worsens it on $66~\%$ of the graphs and improves it on $2~\%$, significantly in 40 of the 44 conditions and in none the other way. Reducing the embedding dimension to $n/4$ worsens it on $62~\%$ against $6~\%$, significantly in 36 conditions and again in none the other way. Removing the fixed-point penalty $\mathcal{L}_{\mathrm{fp}}$ worsens it on $60~\%$ against $9~\%$, significantly in 31 conditions. It improves it in 4 \ac{MMM} conditions, three of them at $\mathit{pert}=0$, where it recovers the planted map on graphs the adopted configuration misses, with a lower mismatch $D$ in every improved pair. Without $\mathcal{L}_{\mathrm{fp}}$, however, $5.5~\%$ of all solutions exceed $F>0.25n$, in 20 of the 44 conditions, against none of the adopted configuration's, whose $F/n$ never exceeds $0.10$. Rounding by the Hungarian assignment alone worsens $\mathcal{S}'(A,P)$ on $6~\%$ of the graphs and improves none, significantly in 12 conditions. At dimension $n$ the loss is smaller, $30~\%$ against $18~\%$, significant in 15 conditions and reversed in 5, and no larger fixed dimension is significantly better in any condition.

\subsection{Computational efficiency, stability, and hyperparameters}\label{ss:computational_parameters}

In addition to solution quality, we assess computational efficiency, stability across repeated runs, and the complexity of hyperparameter selection. Cost is measured on a separate timing corpus of one graph per condition, so that every algorithm is timed on the same instances. The tuned hyperparameter configurations are reported in Appendix~\ref{app:hyper}. Stability analysis (Appendix Fig.~\ref{fig:stability}) shows that all optimization-based strategies exhibit comparable variability across independent runs, with no method displaying systematically unstable behavior relative to the others.

\begin{figure}[htbp]
    \centering
    \includegraphics[width=\linewidth]{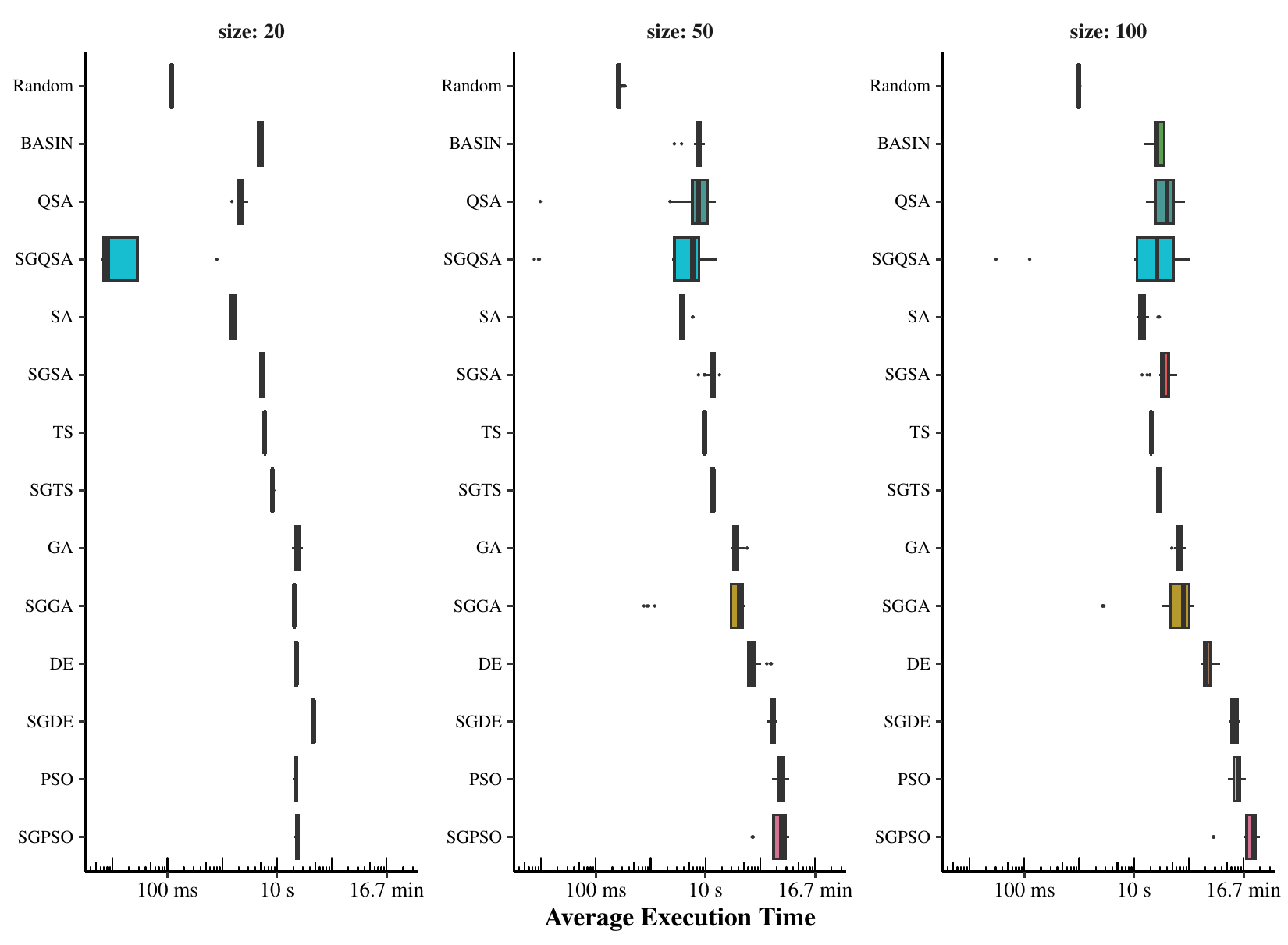}
    \caption{Execution time of a single restart by graph size, every variant of every algorithm. Each box spans the mean time per restart of the conditions at that size, not individual graphs. The horizontal axis is logarithmic.}
    \label{fig:execution}
\end{figure}

\myparagraph{Execution time at $n=100$}
At $n=100$ (Fig.~\ref{fig:execution}) the algorithms fall into three cost bands. \ac{SGTS}, \ac{BASIN}, \ac{SGQSA} and \ac{SGSA} take about half a minute per restart, \ac{SGGA} two to three times as long, and \ac{SGDE} and \ac{SGPSO} more than an order of magnitude longer, against $1.0$~s for the random baseline. \Ac{SGQSA} varies most within a size, from $0.03$~s to $100$~s per restart at $n=100$. Cost grows least with size for \ac{SGTS} and \ac{SGGA}, by a factor of about $4$ from $n=20$ to $n=100$, and most for \ac{SGPSO} and \ac{SGDE}, by $64$ and $14$, apart from \ac{SGQSA}, whose median rises from $0.01$~s to $26$~s.

\myparagraph{Scaling to $n=2000$}
Figures~\ref{fig:scal_cost} and~\ref{fig:scal_mem} extend the measurement to $n = 200$, $500$, $1000$ and $2000$, on a probe of $20$ graphs per size. Throughput is measured relative to each algorithm's own value at $n=200$, as budget units differ between algorithms. \Ac{SGTS}, \ac{SGSA} and \ac{BASIN} lose factors of $252$, $331$ and $622$ over that range, between the $O(n^{2})$ and $O(n^{3})$ reference slopes, \ac{SGPSO} a factor of $993$, on the $O(n^{3})$ slope, and \ac{SGDE} a factor of $3598$, steeper than it. \Ac{SGGA} did not complete the probe at $n \ge 1000$, and \ac{SGQSA} converges before exhausting its budget below $n=2000$, so it is absent from both figures.

Resident memory separates the algorithms more sharply. \Ac{SGSA} and \ac{SGTS} stay at or below $732$~MB at every size, whereas \ac{SGDE} and \ac{BASIN} reach $2.9$~GB and $3.4$~GB at $n=2000$, and \ac{SGPSO} reaches $16.9$~GB.

\subsection{Evaluation on structural brain connectomes}\label{ss:brain_connectomes}
\begin{figure}[ht]
    \centering
    \includegraphics[width=0.9\linewidth]{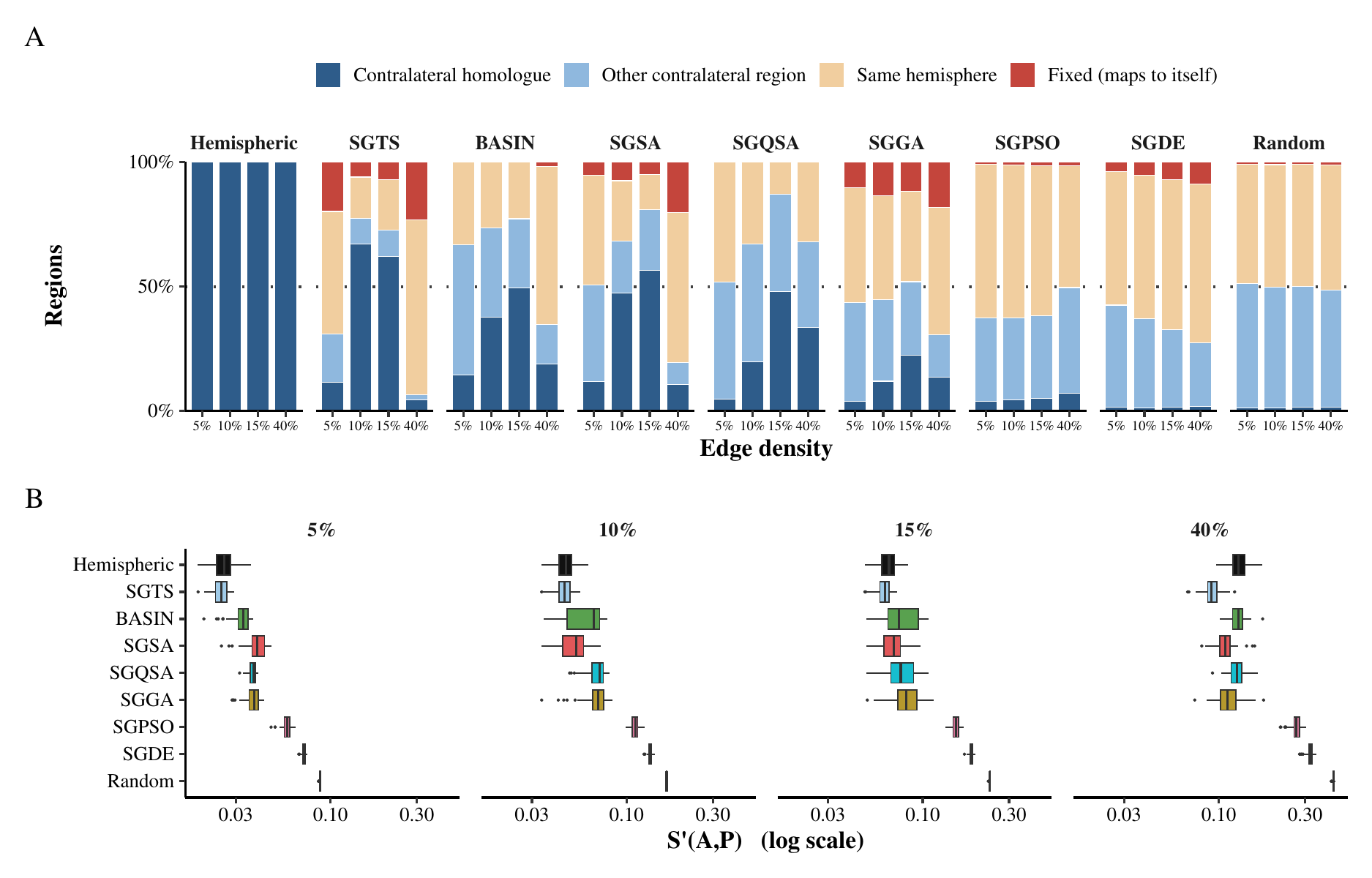}
    \caption{The 88 structural connectomes at four proportional edge densities. \textbf{(A)} Share of the 90 regions, averaged over subjects, that the solution sends to the contralateral homologue, to another contralateral region, within the same hemisphere, or to themselves; \emph{Hemispheric} is $\pi_{LR}$, and dotted is the share sent across hemispheres by the random baseline. \textbf{(B)} $\mathcal{S}'(A,P)$ of each subject's solution, logarithmic axis.}
    \label{fig:real-headline}
\end{figure}

Each method returns one solution per subject, the lowest of 20 restarts rather than five (Fig.~\ref{fig:real-headline}; Appendix Tables~\ref{tab:real-sc-tex} and~\ref{tab:real-sc-map}). Whether the hemispheric mapping $\pi_{LR}$ is recovered depends on density (Fig.~\ref{fig:real-headline}A). At $5\%$ homologues make up a small share of every algorithm's permutation, and only \ac{SGTS} scores comparable $\mathcal{S}'(A,P)$ with $\pi_{LR}$ (Fig.~\ref{fig:real-headline}B), but by permuting most regions within a hemisphere or fixing them instead. At $10\%$ and $15\%$ homologues are the largest share of the \ac{SGTS} permutations, and \ac{SGTS} scores on par or below $\pi_{LR}$. \Ac{SGSA}, and \ac{BASIN} and \ac{SGQSA} at $15\%$, recover a smaller share and score above $\pi_{LR}$. At $40\%$ the homologue share falls for every algorithm that recovers it at $15\%$, and same-hemisphere pairs and, for the algorithms that carry the $0.25n$ bound, fixed regions take its place. \Ac{SGTS}, \ac{SGSA} and \ac{SGGA} score below $\pi_{LR}$, \ac{SGTS} furthest, while \ac{SGQSA} and \ac{BASIN} score level with it.

\begin{figure}[ht]
    \centering
    \includegraphics[width=0.9\linewidth]{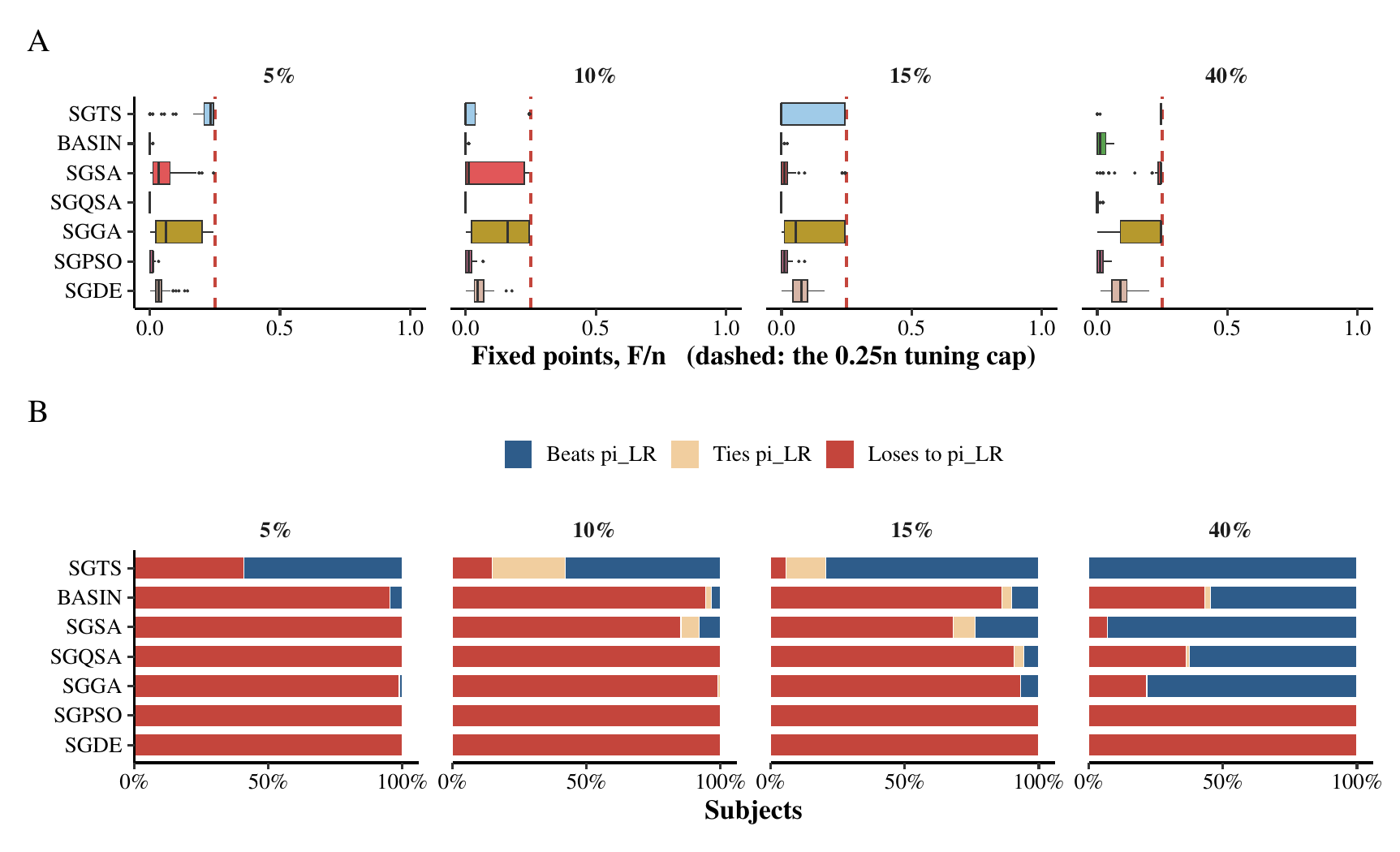}
    \caption{The 88 structural connectomes at four proportional edge densities, one solution per subject. \textbf{(A)} Fixed-point ratio $F/n$ over subjects; dashed is the $0.25n$ bound. \textbf{(B)} Share of subjects whose solution scores below (beats), equal to (ties) or above (loses to) $\pi_{LR}$ on $\mathcal{S}'(A,P)$.}
    \label{fig:real-fp-cost}
\end{figure}

Counted subject by subject (Fig.~\ref{fig:real-fp-cost}, B), \ac{SGTS} beats $\pi_{LR}$ on most subjects at every density and on all 88 at $40\%$, and at $10\%$ ties it exactly on a further quarter. \ac{BASIN} beats it on few subjects below $40\%$ and on about half at $40\%$, and the random baseline never does.

\begin{figure}[ht]
    \centering
    \includegraphics[width=0.9\linewidth]{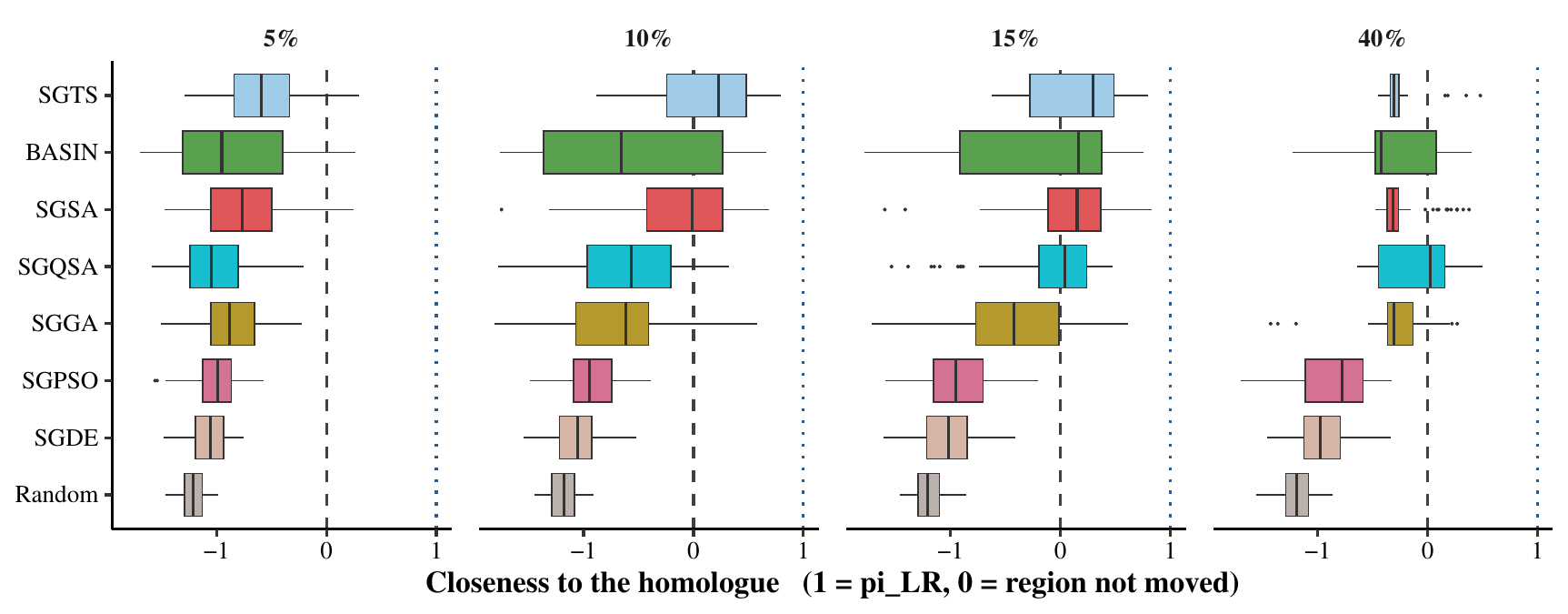}
    \caption{Where the regions not mapped to their contralateral homologue are sent. Each boxplot is across the 88 subjects, and a subject contributes the average over those regions in its solution. The x-axis compares the physical distance in the brain between a region's image and its homologue with the distance between the region itself and its homologue: $1$ (dotted) is the homologue itself, $0$ (dashed) is as far from it as leaving the region in place, and negative values lie further from it than the region already does.}
    \label{fig:real-nearmiss}
\end{figure}

Where the returned map is not the hemispheric mirror, it is often near it. Nearness is measured in physical space: every node of the \ac{AAL} parcellation is a region of the brain with a fixed centre point in a standard anatomical coordinate system, and $d$ denotes the Euclidean distance between these points. For a node $i$ whose image $\sigma(i)$ is not its homologue $\pi_{LR}(i)$, the proximity $\mathrm{prox}(i) = 1 - d(\sigma(i), \pi_{LR}(i)) / d(i, \pi_{LR}(i))$ compares how far the image lies from the homologue with how far $i$ itself lies from it. It is $1$ when the image is the homologue, $0$ when the image is as far from the homologue as $i$ is, and negative when it is further. At $10\%$ and $15\%$ the non-homologous images of \ac{SGTS} land closer to the homologue than their source regions, and those of \ac{SGSA} and \ac{BASIN}, and barely those of \ac{SGQSA}, do at $15\%$ (Fig.~\ref{fig:real-nearmiss}). At $5\%$ and $40\%$ every algorithm's images lie further from it, those of \ac{SGQSA} at $40\%$ about as far, though all lie closer than the random baseline's, at a median of $-1.2$.

\myparagraph{Fixed-point ratio on the connectomes}
Comparable scores are reached through different fixed-point behavior (Fig.~\ref{fig:real-fp-cost}A). At $40\%$ every algorithm that carries the $0.25n$ bound in its search sits on it, and \ac{SGTS} does so at $5\%$ too, while at $10\%$ and $15\%$, where it recovers $\pi_{LR}$, it fixes no region. \ac{BASIN} and \ac{SGQSA} fix almost none at any density.

\myparagraph{Guided and unguided variants on the connectomes}
On the connectomes guidance helps the weaker searches and leaves \ac{TS} unchanged (Appendix Table~\ref{tab:real-sc-tex}). \ac{SGSA} and \ac{SGGA} score lower than their unguided variants at $10\%$, $15\%$ and $40\%$; at $5\%$ only \ac{SGGA} does, and \ac{SGSA} scores higher than \ac{SA}. For \ac{SA}, guidance also recovers much of the hemispheric map: at $10\%$ and $15\%$ the homologue share of its permutations rises from under a tenth to over half, whereas for \ac{GA} it stays near zero with or without guidance. The two variants of \ac{TS} are level in score and in homologue share at every density. For \ac{QSA} guidance lowers the score at $5\%$ only, is level at $10\%$ and raises it at $15\%$, but raises the homologue share at $15\%$ and $40\%$, from $0.18$ to $0.52$ and from $0$ to $0.42$; at $40\%$ \ac{QSA} fixes $86$ of the $90$ regions and \ac{SGQSA} none. What guidance changes consistently is agreement between restarts (Appendix Table~\ref{tab:real-sc-agreement}): the guided variant returns more similar permutations across restarts for every algorithm at every density, \ac{SGTS} restarts agreeing on $0.42$ to $0.70$ of the regions against $0.08$ to $0.23$ for \ac{TS}. The one exception is \ac{QSA} at $40\%$, whose restarts agree trivially because it fixes $86$ of the $90$ regions.

\subsection{Benchmark networks with known automorphism groups}\label{ss:benchmarks}

On JAZZ an exact automorphism is reachable, but only by the discrete searches (Table~\ref{tab:bench}). \ac{SGTS} reaches one most often, on $0.80$ of restarts, level with \ac{TS}. Guidance helps \ac{GA}, raising its rate from $0.35$ to $0.65$, but worsens \ac{SA}, from $0.70$ to none. Neither the continuous relaxations nor \ac{PSO} and \ac{DE} ever reach one, \ac{QSA} coming closest.

\myparagraph{Distance to the twin subgroup on PDZ and NETSCIENCE}
On PDZ and NETSCIENCE no configuration reaches an automorphism, so the distance to the twin subgroup (Section~\ref{sss:evaluation}) is reported instead. Under it the guided variants land significantly closer on both networks for \ac{GA}, \ac{TS} and \ac{PSO}, whereas \ac{SGDE} lands further than \ac{DE} on both; the difference for \ac{SA} is marginal (Table~\ref{tab:bench-guidance}). \ac{SGQSA}, \ac{SGGA} and \ac{SGTS} attain the lowest $\mathcal{S}'(A,P)$ on both networks (Table~\ref{tab:bench}). The distance to the twin subgroup does not follow $\mathcal{S}'(A,P)$: the solutions closest to it are those of \ac{SGPSO} on both networks, whose $\mathcal{S}'(A,P)$ is about twice that of \ac{SGQSA}. \ac{SGSA} and \ac{BASIN} end about as far from the twin subgroup as the random baseline.

\begin{table}[!h]
\centering
\caption{\label{tab:bench-guidance}Unguided against similarity-guided variants on PDZ and NETSCIENCE, where no configuration reaches an automorphism. Distance to the twin subgroup in vertices, median over the restarts with $F \leq 0.25n$ of 20 per variant (\emph{Restarts}, unguided/guided). Two-sided Mann--Whitney test, $p$ uncorrected for the 12 comparisons; the rank-biserial $r$ is negative where the guided variant lands closer.}
\centering
\fontsize{8}{10}\selectfont
\begin{tabular}[t]{llllll}
\toprule
Variants & Restarts & Median unguided & Median guided & $r$ & $p$\\
\midrule
\addlinespace[0.3em]
\multicolumn{6}{l}{\textbf{PDZ}}\\
\hspace{1em}SA / SGSA & 20/20 & 157.0 & 156.0 & $-0.02$ & $0.913$\\
\hspace{1em}GA / SGGA & 20/20 & 157.5 & 130.0 & $-0.95$ & $<0.001$\\
\hspace{1em}QSA / SGQSA & 20/20 & 161.0 & 161.0 & $+0.13$ & $0.451$\\
\hspace{1em}TS / SGTS & 20/20 & 156.5 & 129.0 & $-0.95$ & $<0.001$\\
\hspace{1em}PSO / SGPSO & 19/20 & 137.0 & 107.5 & $-0.94$ & $<0.001$\\
\hspace{1em}DE / SGDE & 20/20 & 133.0 & 149.0 & $+0.93$ & $<0.001$\\
\addlinespace[0.3em]
\multicolumn{6}{l}{\textbf{NETSCIENCE}}\\
\hspace{1em}SA / SGSA & 20/20 & 377.5 & 376.5 & $-0.36$ & $0.049$\\
\hspace{1em}GA / SGGA & 20/20 & 375.5 & 337.5 & $-1.00$ & $<0.001$\\
\hspace{1em}QSA / SGQSA & 20/20 & 373.5 & 371.5 & $-0.09$ & $0.645$\\
\hspace{1em}TS / SGTS & 20/20 & 372.5 & 357.0 & $-0.96$ & $<0.001$\\
\hspace{1em}PSO / SGPSO & 20/20 & 347.0 & 304.0 & $-1.00$ & $<0.001$\\
\hspace{1em}DE / SGDE & 19/20 & 339.0 & 366.5 & $+0.97$ & $<0.001$\\
\bottomrule
\end{tabular}
\end{table}

Because the twin subgroup contains the identity, a solution that maps most nodes to themselves scores as close to it, so only runs with $F \leq 0.25n$ are compared. This removes 1 of 20 \ac{DE} restarts on NETSCIENCE and 1 of 20 plain \ac{PSO} restarts on PDZ. Every other configuration, \ac{BASIN}, \ac{QSA} and \ac{SGQSA} included, keeps all 20 restarts on both networks (distributions over restarts in Figs.~\ref{fig:bench-sprime} and~\ref{fig:bench-distance}).

\section{Discussion}\label{s:discussion}
The present study revisits approximate symmetry detection from an optimization perspective, focusing on how node-level structural information can be used to bias the search process. By evaluating local search (\ac{SA}), population-based optimization (\ac{GA}), and continuous relaxations (\ac{BASIN} and \ac{QSA}), together with \ac{TS}, \ac{PSO} and \ac{DE}, within a unified validation framework, the results highlight both the strengths and limitations of node-similarity guidance and what practical trade-offs emerge between solution quality, robustness, and computational cost.

\subsection{Impact of node-similarity guidance}\label{ss:disc_node-similarity_guidance}
Node-similarity guidance does not always help. On the \ac{MMM}, whose planted orbit the unguided searches miss, guidance improves every algorithm but \ac{TS} in nearly every condition and, for five of the six, the recovery of the planted map. On \ac{BA} and \ac{ER}, which have no planted orbit, guidance still improves \ac{GA} and \ac{PSO} in most conditions and \ac{SA} in about half, but worsens \ac{TS} and \ac{QSA} in most conditions and \ac{DE} on \ac{BA}. A planted orbit therefore explains where guidance helps every algorithm, not where it helps at all. On the \ac{LRM} the effect of guidance depends on the algorithm. This behavior reflects the dual role of such guidance: it biases optimization toward structurally plausible correspondences, yet may introduce unwanted bias when the structural signal is weak or noisy. Node similarity should therefore be interpreted as a guiding mechanism rather than a strict constraint.

The benchmark networks show the same dependence. On PDZ and NETSCIENCE, where no configuration reaches an automorphism, the guided variants of \ac{GA}, \ac{TS} and \ac{PSO} end closer to one on both networks and that of \ac{DE} further; on JAZZ, where one is reachable, guidance leaves \ac{TS} at its ceiling, helps \ac{GA} and worsens \ac{SA}. On PDZ and NETSCIENCE the maps of lowest $\mathcal{S}'(A,P)$ are not those closest to the twin subgroup. The twin subgroup is only part of the automorphism group, so a map of low $\mathcal{S}'(A,P)$ that lies far from it may still lie close to an automorphism outside it, i.e.\ approximate a different symmetry equally well.

These findings align with known limitations of centrality-based heuristics. Structural descriptors assume sufficient heterogeneity in node roles, which fails in regular or centrality-uniform graphs~\cite{gago2013betweenness,ghanbari2023structure,hartman2024connectivity}. Although automorphic vertices necessarily share centrality values, asymmetric graphs may still exhibit uniform centrality profiles. In complex networks, however, centrality distributions typically remain sufficiently diverse to provide informative guidance.

On the connectomes, which carry no planted map, guidance lowers the score of \ac{SGSA} and \ac{SGGA} at most densities but leaves that of \ac{SGTS} unchanged. Only with guidance does \ac{SA} send most regions to their contralateral homologue, at $10\%$ and $15\%$. For every algorithm it changes which map is returned, returning nearly the same one at every restart. That helps in cases when a single answer is wanted but hurts when the spread of approximate symmetries is itself the object, as it is when no correct answer is known. On connectomes this spread may carry information of its own. The restarts of an unguided search, and the solutions of different algorithms with and without guidance, each sample approximate symmetries of the same brain, and their distribution, which regions are paired consistently and which vary, could describe its hemispheric organization more fully than any single map.

\subsection{Performance across optimization paradigms}\label{ss:disc_optimization_paradigms}
Methods maintaining population diversity (\ac{SGGA}) or employing differentiable relaxations (\ac{BASIN}) achieve stronger performance than every strategy but \ac{SGTS}, \ac{SGGA} on \ac{BA} and \ac{ER} and \ac{BASIN} on rewired \ac{LRM} graphs, where deviations from perfect symmetry create complex optimization landscapes. Transposition search that samples one move at a time, such as \ac{SGSA}, becomes less effective in such regimes, whereas global exploration or continuous relaxation escape unfavorable search-space regions and local minima more reliably. Similar trends have been reported in earlier work comparing annealing-based and relaxation-based approaches~\cite{Liu2020approx,pidnebesna2025computing}.

\ac{SGTS} shares this transposition neighborhood with \ac{SA} yet remains first at the highest rewiring level; what differs is that the maintained delta table evaluates a swap in $O(1)$, so the full $O(n^{2})$ neighborhood is scanned at every iteration and the move is chosen on $\mathcal{S}'(A,P)$ itself, at a cost per restart in the band of \ac{SA} and \ac{BASIN}.

\Ac{BASIN} provides a particularly favorable balance between efficiency and solution quality among the continuous relaxations. Unlike explicit centrality-guided strategies, \ac{BASIN} implements data-driven node-similarity guidance through learnable embeddings shaped directly by the symmetry objective, avoiding reliance on predefined descriptors while maintaining computational stability.

The population-based searches sit less favorably on the same balance of cost and quality. \ac{SGGA} often achieves strong objective values but at substantially higher computational cost than the other discrete searches (Fig.~\ref{fig:execution}): it is second on \ac{BA} and \ac{ER}, and does not complete the scalability probe at $n \ge 1000$. \ac{SGPSO} and \ac{SGDE} rank sixth and seventh of eight in every graph family and are among the slowest per restart; a population of $n \times n$ real matrices offers no advantage when the objective can be evaluated exactly and incrementally on permutations.

\subsection{Interpreting approximate symmetries}\label{ss:disc_interpretation}
Similar objective values may correspond to qualitatively different permutations, especially when fixed points are present. Figure~\ref{fig:fp_hard} illustrates this variability: for a given symmetry score, the proportion of fixed vertices differs markedly across methods. The fixed-point--normalized score $\mathcal{S}'(A,P)$ therefore plays an important role in distinguishing informative approximate symmetries from near-identity solutions. Because the normalization does not penalize fixed points strongly enough, only the strategies carrying the bound inside their search stay under it, and on the connectomes at $40\%$ \ac{QSA} attains the lowest $\mathcal{S}'(A,P)$ of any algorithm from a map that fixes $86$ of the $90$ regions. The normalized score is therefore necessary but not sufficient, and an explicit feasibility bound is required with it. Explicit penalties or alternative normalization strategies, as explored in~\cite{hartman2025local}, may further improve interpretability in settings where partial symmetries are expected.

The fixed-point bound is a modeling choice, not a property of symmetry. On JAZZ it would forbid every correct answer, which is why \ac{SA}, \ac{TS} and \ac{GA}, the strategies that carry the cap, run without it on the benchmark networks. Whether near-identity maps are an artifact or the object of interest must thus be decided for each network.

\myparagraph{Stability across restarts}
Stability analyses (Fig.~\ref{fig:stability}) show broadly comparable variability across independent runs for all optimization-based methods, with no strategy displaying systematically unstable behavior. The observed variability therefore primarily reflects optimization dynamics rather than excessive stochastic sensitivity.

\myparagraph{Recovery of the hemispheric mirror}
On the connectomes the hemispheric mirror is recovered in a narrow density band. At $5\%$ the best maps contain no more homologues than a random permutation; at $10\%$ and $15\%$ \ac{SGTS} sends most regions to their homologue; at $40\%$ \ac{SGTS} beats the mirror on every subject while sending almost no region to its homologue, so the anatomical mirror is not the $\mathcal{S}'(A,P)$-optimal symmetry of a dense connectome. Where \ac{SGTS} misses the homologue at $10\%$ and $15\%$ it lands near it, whereas at $5\%$ it lands further from it than the region itself; an exact-recovery rate alone would score both as zero.

\subsection{Limitations and future directions}\label{ss:disc_limitations_future}
Several limitations suggest concrete directions for follow-up work. Explicit node-similarity guidance may be uninformative in graphs with degenerate structural descriptors (e.g. regular graphs), motivating exploration of richer automorphism-invariant descriptors such as spectral embeddings, motif-based representations, or learned graph neural network descriptors.

On the optimization side, \ac{BASIN} naturally invites some principled extensions including tighter enforcement of doubly-stochastic structure (e.g. Sinkhorn normalization) and coupling discretization with the continuous objective to reduce projection-induced artifacts. More broadly, gradient-based techniques from deep learning, including reinforcement learning for discrete search and graph neural networks for continuous search and learning structural descriptors, offer promising directions for scaling approximate symmetry detection to larger networks while preserving efficiency and robustness.

The connectome analysis extends the evaluation to an empirical dataset and to the recovered node correspondences, not only objective values. The connectomes are one cohort of 88 subjects in one parcellation, binarized at a fixed edge count, and the outcome depends on the density. Whether the band at $10\%$ to $15\%$ is a property of structural connectivity or of the parcellation cannot be settled from one atlas. Extending the analysis to further network classes and parcellations would strengthen the interpretation of approximate symmetries as meaningful structural patterns. Investigation of how approximate symmetry relates to dynamical properties of real systems, building on connections established for exact automorphisms~\cite{Hart2019topological,Mondal2024symmetry}, remains an important open question.

Hyperparameters tuned at $n \leq 100$ are not re-tuned for the benchmark networks, so comparisons there are within-network at equal budget rather than of each algorithm at its best. The distance to an exact automorphism is computed against the twin subgroup, an upper bound on the distance to the full group that contains the identity, so it must be read together with the fixed-point count. The graph family on which guidance is decisive was constructed to carry a planted orbit; the benchmark networks reduce but do not remove that circularity.

Both $\mathcal{S}'(A,P)$ and every algorithm here are defined for unweighted graphs. The objective extends to weighted matrices unchanged, but the feasibility bound, the fixed-point normalization and exact-hit recovery would each have to be restated.

\section{Conclusion}\label{s:conclusions}
This work extends the optimization-based framework for approximate network symmetry by making the role of node-level structural information explicit. Building on Liu's formulation~\cite{Liu2020approx}, we introduce a gradient-based soft-assignment model (\ac{BASIN}) that learns a continuous surrogate followed by discretization, adapt a genetic algorithm, tabu search, particle swarm optimization and differential evolution to the problem, and add node-similarity guidance to each of them, including \ac{SGTS}, which selects its guidance per instance. Together with similarity-guided simulated annealing (\ac{SGSA}) and the quadratic symmetry approximator (\ac{QSA})~\cite{pidnebesna2025computing}, these methods enable unified comparison across local, global, and continuous optimization paradigms.

Across extensive simulations on \ac{ER}, \ac{BA}, \ac{LRM} and \ac{MMM} graphs, structural connectomes and networks with known automorphism groups, node-similarity guidance improves solutions where the unguided search misses a planted orbit, and elsewhere its effect depends on the algorithm. \Ac{TS}, guided or not, reaches the lowest scores, and \ac{BASIN} is the stronger continuous relaxation on rewired lateral graphs. A normalized score alone does not exclude near-identity solutions, so a fixed-point bound is needed with it.

In summary, this article demonstrates that the effectiveness of node-similarity guidance for approximate symmetry detection is regime-dependent across multiple algorithmic paradigms, and introduces \ac{BASIN} and similarity-guided variants of six optimization strategies, of which tabu search, guided or not, attains the lowest scores among the strategies compared. The improved accuracy and robustness of these methods may prove particularly valuable where approximate symmetries govern system behavior, including the analysis of synchronization patterns in dynamical networks and the characterization of asymmetries in brain connectivity, where even subtle deviations from exact symmetry can have functional consequences.

\section{Declarations}\label{s:declarations}

\noindent
{\bf Conflicts of interest}
The authors have no conflicts of interest to declare that are relevant to the content of this article.\newline

\noindent
{\bf Data availability statement}
The synthetic graphs are generated by the models described in Section~\ref{ss:data}. The generator, the per-graph derived seeds and a manifest of SHA-256 hashes for every file are part of the released code, so both splits reconstruct byte-identically. The 88 structural connectivity matrices are taken from the publicly released dataset of~\cite{skoch2022human}. The three empirical networks with known automorphism groups are taken from the Netzschleuder repository~\cite{peixoto2020netzschleuder}, with their primary sources cited in Section~\ref{sss:data_bench}. No data were newly acquired from human participants for this study.\newline

\noindent
{\bf Code availability statement}
All algorithms, hyperparameter searches, analysis scripts and the adopted configuration of every reported method are available on GitHub at \url{https://github.com/dominikklepl/Centrality-guided-Search-of-Graph-Symmetries-}. The software environment is pinned by a lock file resolving every dependency to an exact version under Python 3.11--3.13.\newline

\noindent
{\bf Ethics approval and consent to participate}
This study is a secondary analysis of previously released, anonymized datasets and involved no new data collection. Ethical approval and participant consent for the structural connectivity dataset are documented in~\cite{skoch2022human}.\newline

\noindent
{\bf Acknowledgements}
The authors were supported by the Czech Science Foundation Grant No. 23-07074S and by ERDF-Project No. CZ.02.01.01/00/22\_008/0004643.

\bibliographystyle{unsrtnat}
\bibliography{references}

\clearpage
\appendix
\begin{center}{\LARGE Supplementary material}\end{center}
\addcontentsline{toc}{section}{Supplementary material}
\setcounter{figure}{0}
\setcounter{table}{0}
\setcounter{equation}{0}
\setcounter{algorithm}{0}
\renewcommand{\thefigure}{S\arabic{figure}}
\renewcommand{\thetable}{S\arabic{table}}
\renewcommand{\theequation}{S\arabic{equation}}
\renewcommand{\thealgorithm}{S\arabic{algorithm}}

\section{Impact of node-similarity guidance: Additional results}
\begin{figure}[ht]
    \centering
    \includegraphics[width=0.75\linewidth]{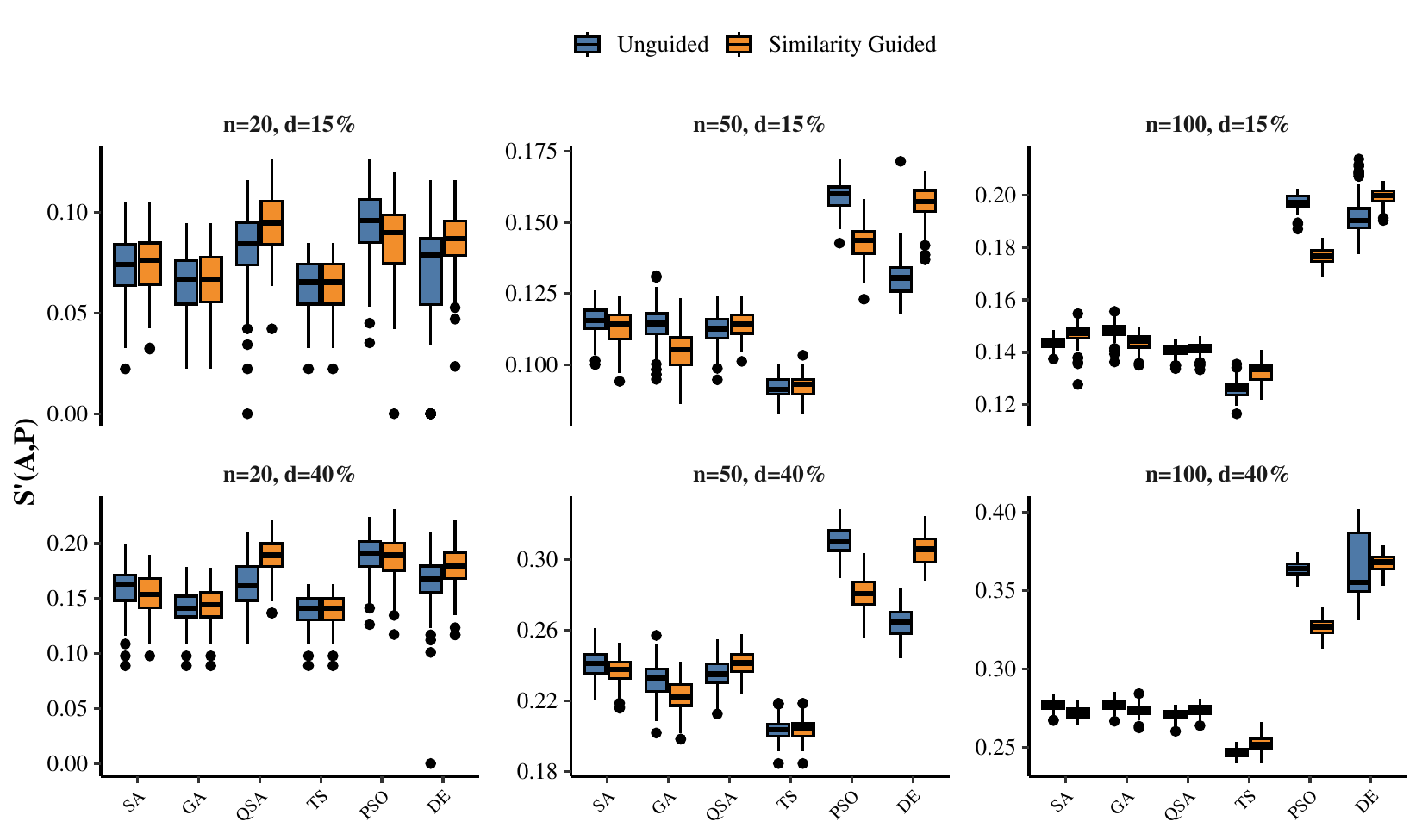}
 \caption{Node-similarity guidance on \ac{BA} graphs. Boxes show $\mathcal{S}'(A,P)$ over the 100 test graphs of each condition, one solution per graph, for the unguided and the similarity-guided variant of each algorithm; rows are edge densities $d$ of 15\% and 40\%, columns graph sizes $n$. \ac{BASIN} has no guided variant and is not shown.}
    \label{fig:gain_BA}
\end{figure}

\begin{figure}[ht]
    \centering
    \includegraphics[width=0.75\linewidth]{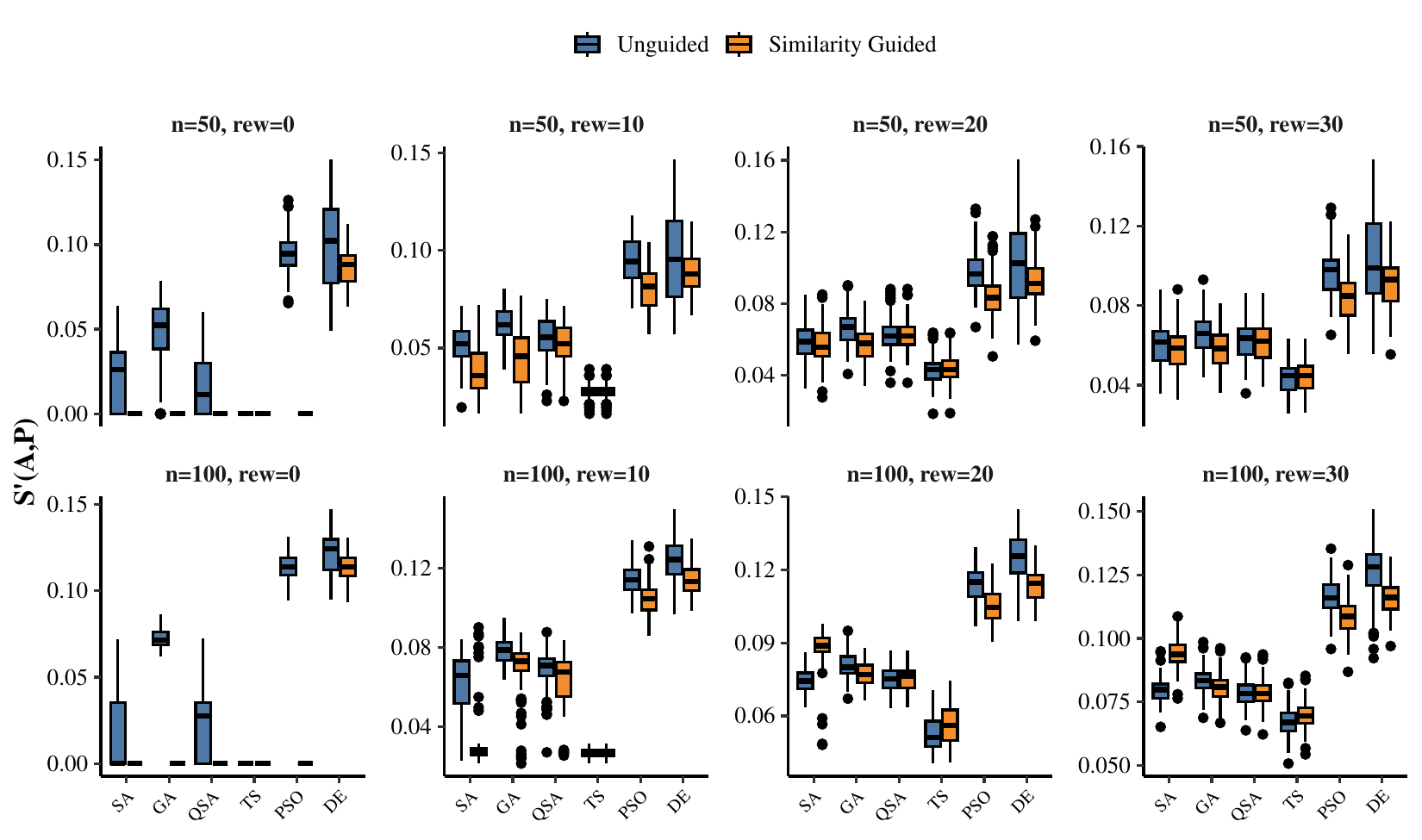}
 \caption{Node-similarity guidance on \ac{LRM} graphs at edge density $15~\%$. Boxes show $\mathcal{S}'(A,P)$ over the 100 test graphs of each condition, one solution per graph, for the unguided and the similarity-guided variant of each algorithm; one panel per graph size $n$ and share $\mathit{rew}$ of rewired edges. \ac{BASIN} has no guided variant and is not shown.}
    \label{fig:gain_LRM15}
\end{figure}

\begin{figure}[ht]
    \centering
    \includegraphics[width=0.75\linewidth]{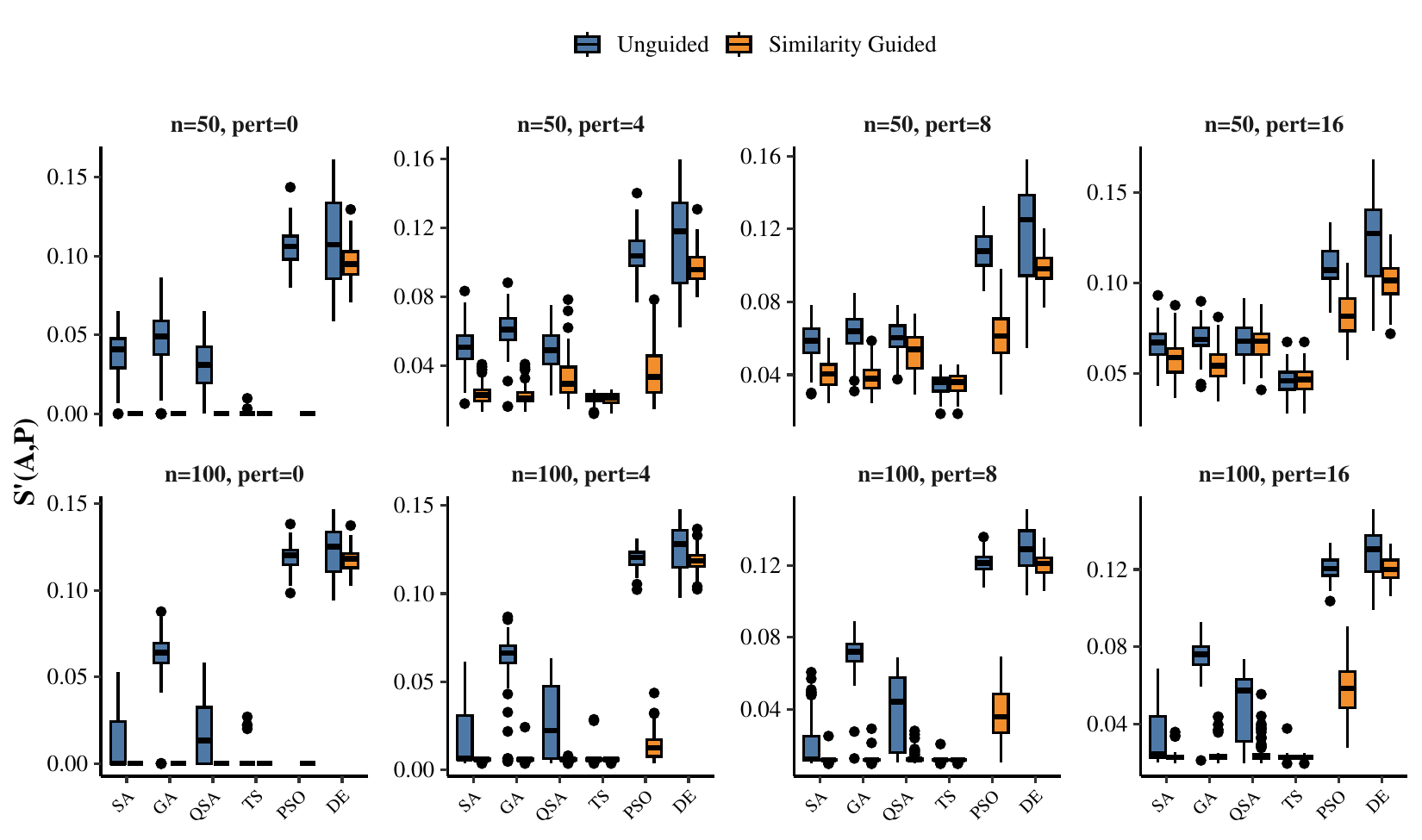}
 \caption{Node-similarity guidance on \ac{MMM} graphs at edge density $15~\%$. Boxes show $\mathcal{S}'(A,P)$ over the 100 test graphs of each condition, one solution per graph, for the unguided and the similarity-guided variant of each algorithm; one panel per graph size $n$ and number $\mathit{pert}$ of perturbing edge swaps. \ac{BASIN} has no guided variant and is not shown.}
    \label{fig:gain_MMM15}
\end{figure}

\subsection{The guidance comparison under the size-normalized score}\label{app:guidance_S}

\begin{figure}[ht]
    \centering
    \includegraphics[width=0.75\linewidth]{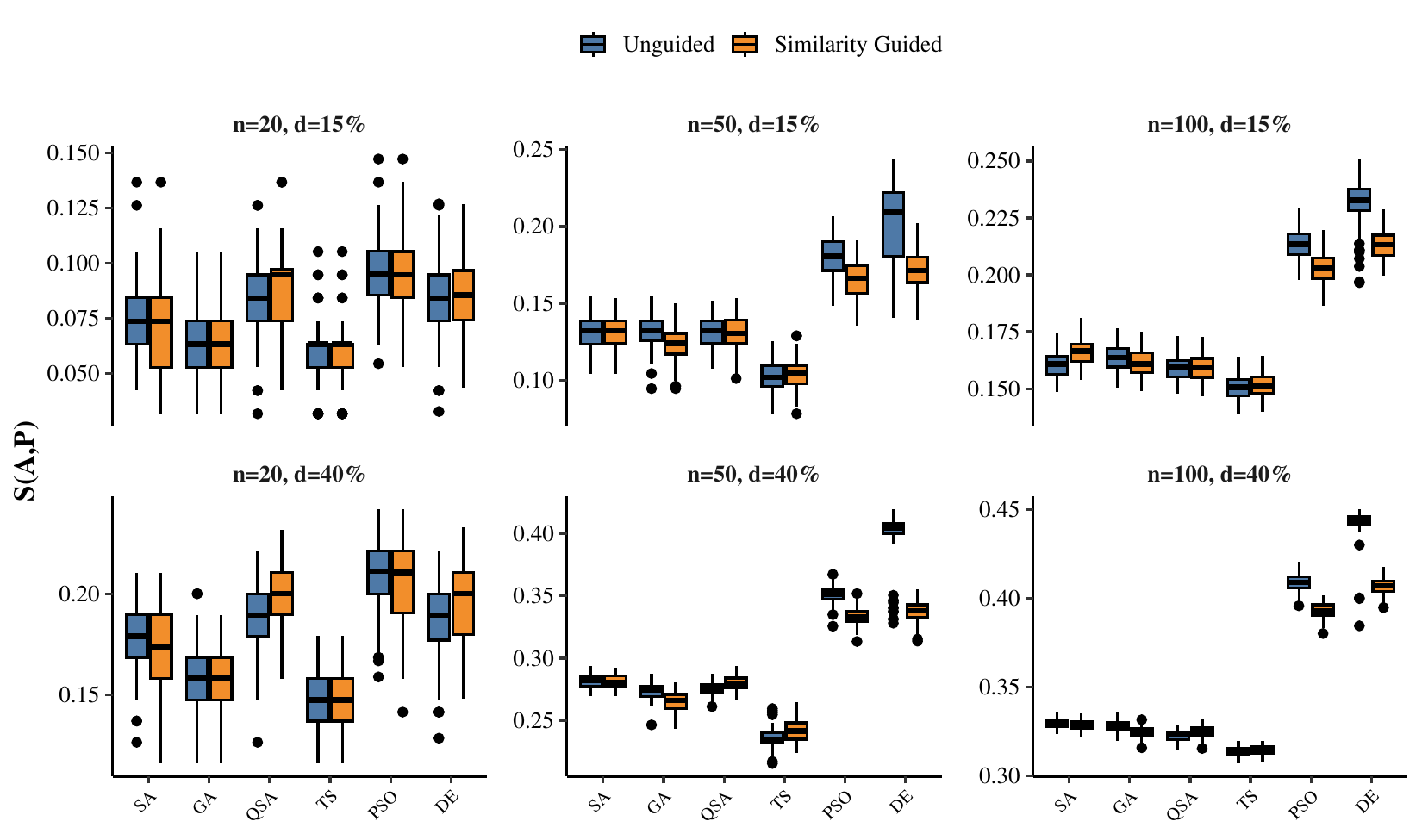}
 \caption{Node-similarity guidance on \ac{ER} graphs, under $\mathcal{S}(A,P)$. Boxes show the size-normalized score $\mathcal{S}(A,P)$, which does not correct for fixed points over the 100 test graphs of each condition, one solution per graph, for the unguided and the similarity-guided variant of each algorithm; rows are densities $d$ of 15\% and 40\%, columns graph sizes $n$. \ac{BASIN} has no guided variant and is not shown.}
    \label{fig:gain_ER_S}
\end{figure}

\begin{figure}[ht]
    \centering
    \includegraphics[width=0.75\linewidth]{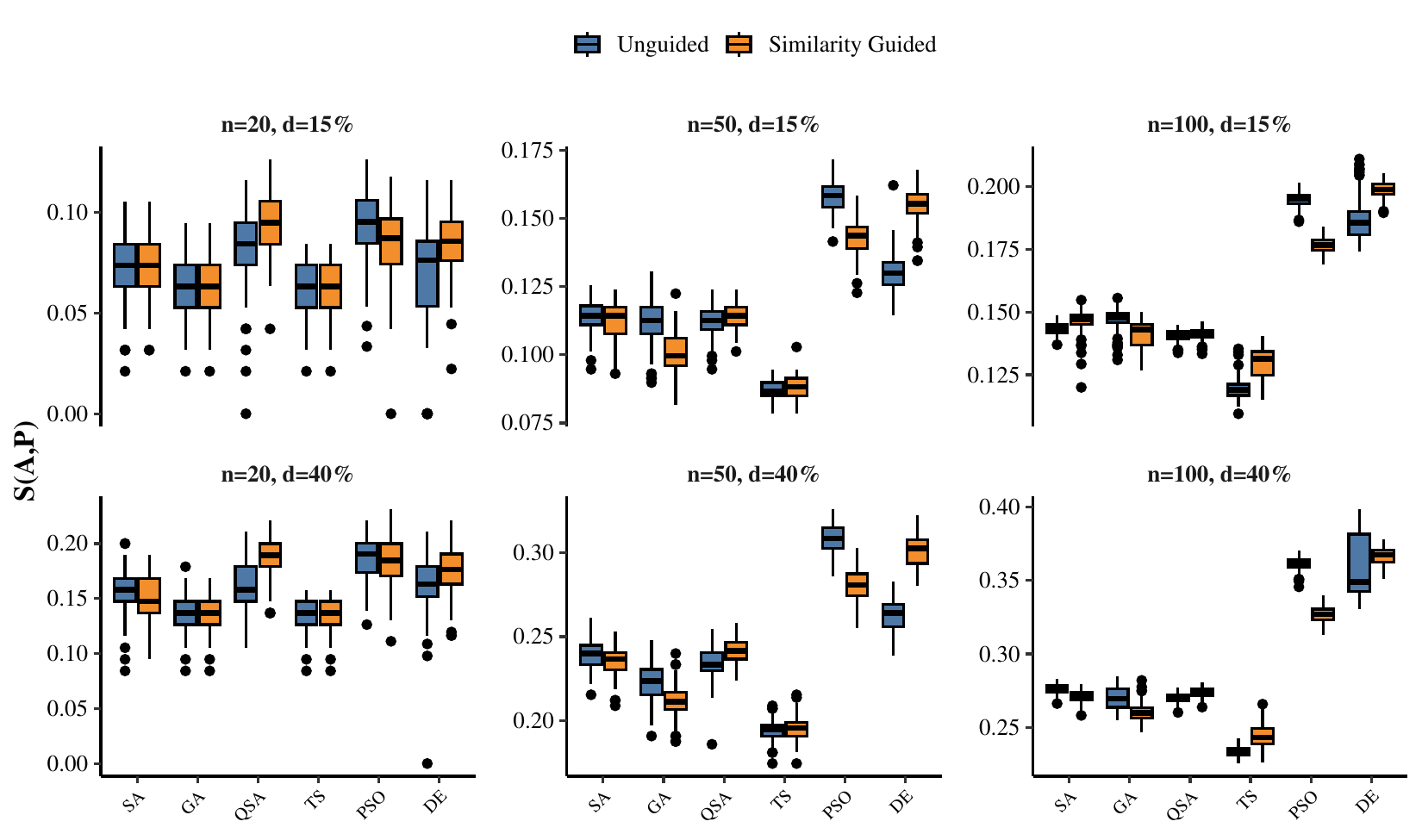}
 \caption{Node-similarity guidance on \ac{BA} graphs, under $\mathcal{S}(A,P)$. Boxes show the size-normalized score $\mathcal{S}(A,P)$, which does not correct for fixed points over the 100 test graphs of each condition, one solution per graph, for the unguided and the similarity-guided variant of each algorithm; rows are densities $d$ of 15\% and 40\%, columns graph sizes $n$. \ac{BASIN} has no guided variant and is not shown.}
    \label{fig:gain_BA_S}
\end{figure}

\begin{figure}[ht]
    \centering
    \includegraphics[width=0.75\linewidth]{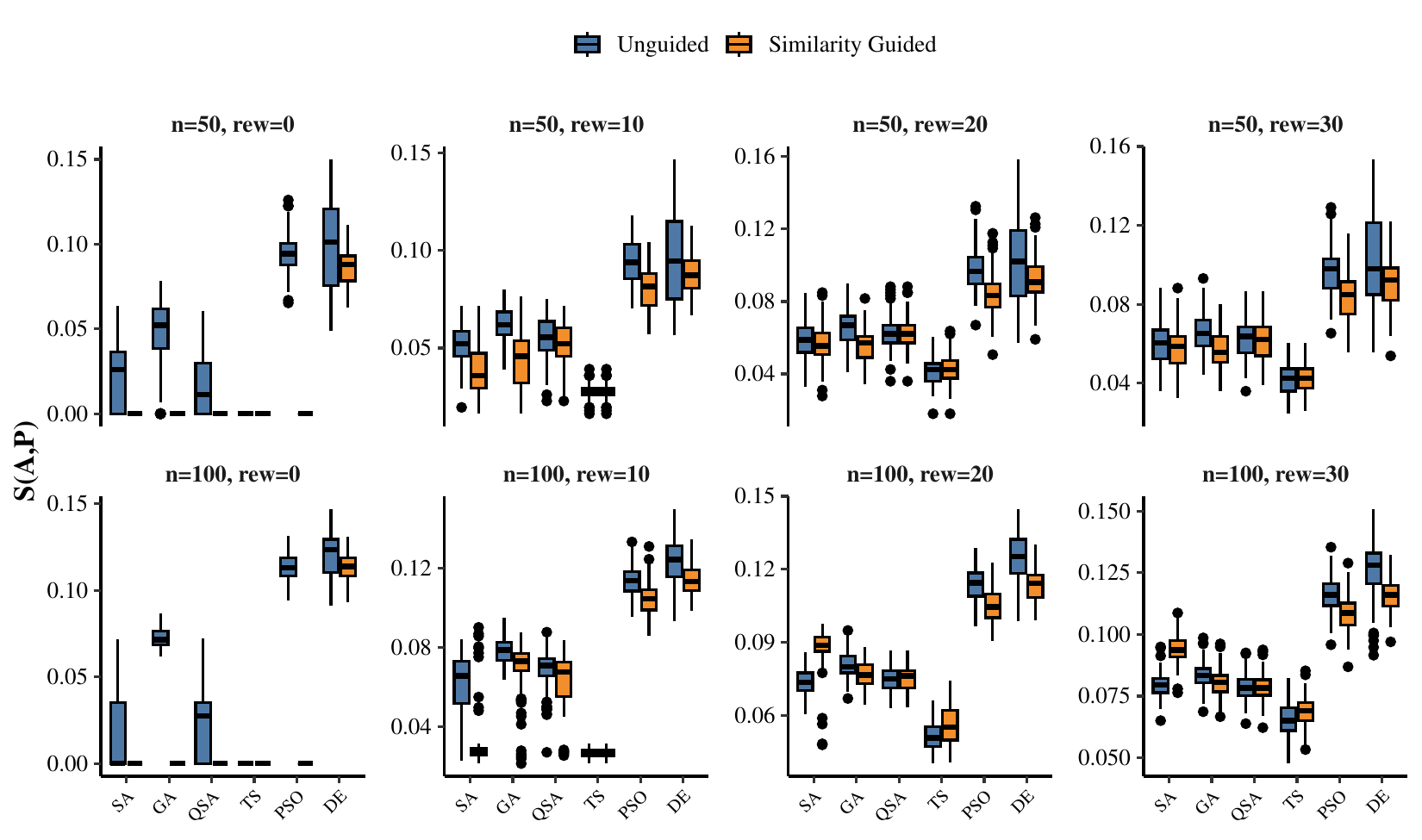}
 \caption{Node-similarity guidance on \ac{LRM} graphs at edge density $15~\%$, under $\mathcal{S}(A,P)$. Boxes show the size-normalized score $\mathcal{S}(A,P)$, which does not correct for fixed points over the 100 test graphs of each condition, one solution per graph, for the unguided and the similarity-guided variant of each algorithm; one panel per graph size $n$ and share $\mathit{rew}$ of rewired edges. \ac{BASIN} has no guided variant and is not shown.}
    \label{fig:gain_LRM15_S}
\end{figure}

\begin{figure}[ht]
    \centering
    \includegraphics[width=0.75\linewidth]{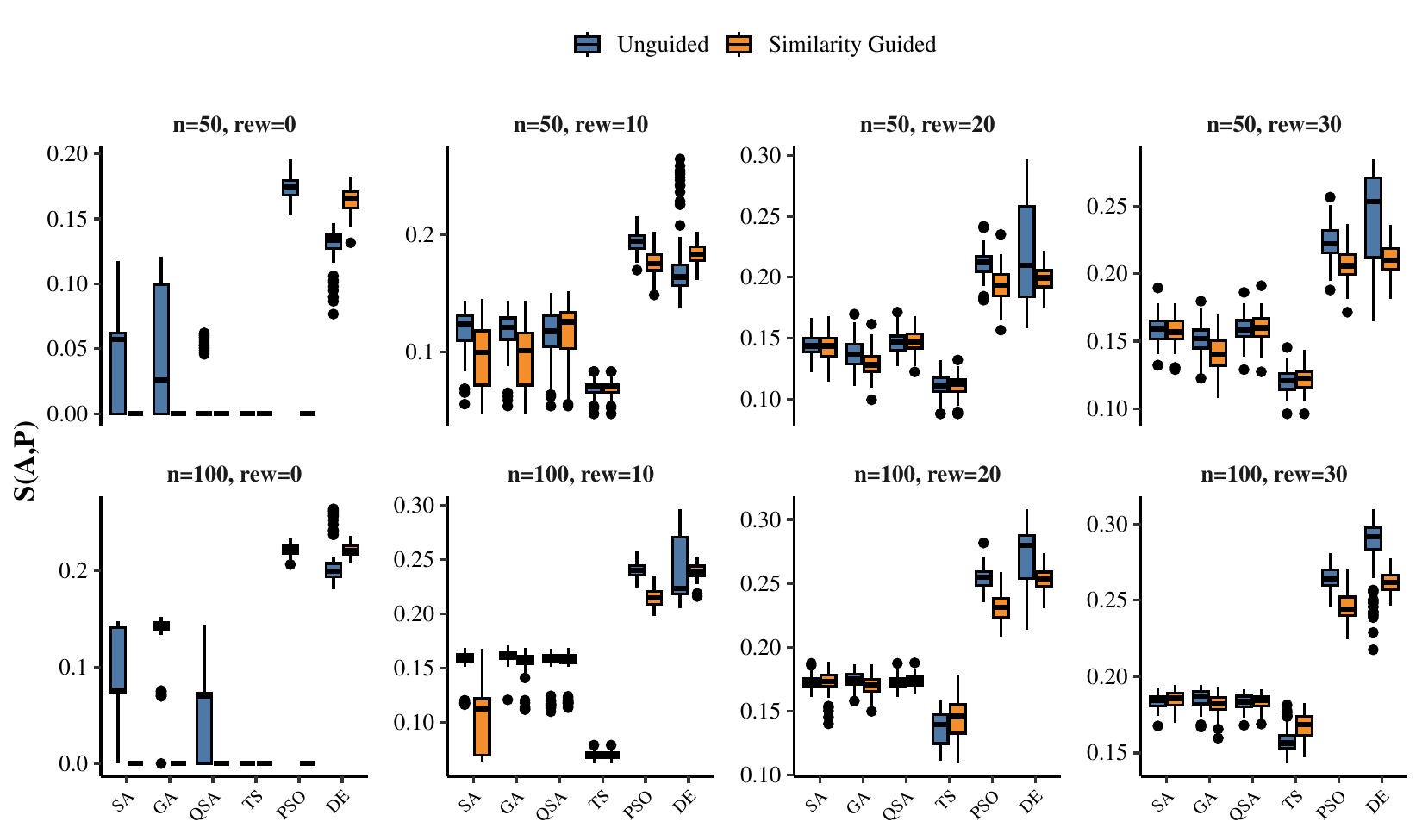}
 \caption{Node-similarity guidance on \ac{LRM} graphs at edge density $40~\%$, under $\mathcal{S}(A,P)$. Boxes show the size-normalized score $\mathcal{S}(A,P)$, which does not correct for fixed points over the 100 test graphs of each condition, one solution per graph, for the unguided and the similarity-guided variant of each algorithm; one panel per graph size $n$ and share $\mathit{rew}$ of rewired edges. \ac{BASIN} has no guided variant and is not shown.}
    \label{fig:gain_LRM40_S}
\end{figure}

\begin{figure}[ht]
    \centering
    \includegraphics[width=0.75\linewidth]{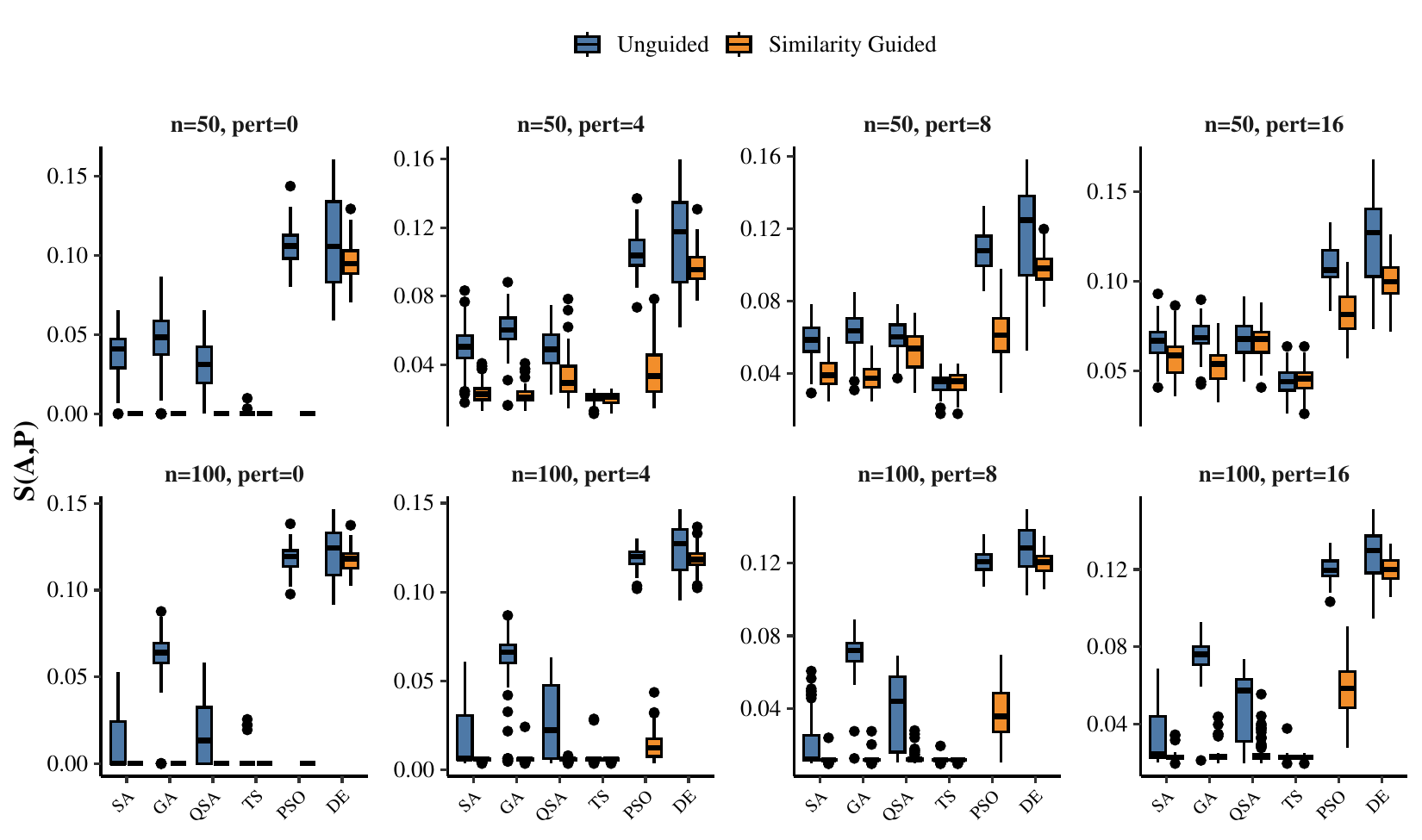}
 \caption{Node-similarity guidance on \ac{MMM} graphs at edge density $15~\%$, under $\mathcal{S}(A,P)$. Boxes show the size-normalized score $\mathcal{S}(A,P)$, which does not correct for fixed points over the 100 test graphs of each condition, one solution per graph, for the unguided and the similarity-guided variant of each algorithm; one panel per graph size $n$ and number $\mathit{pert}$ of perturbing edge swaps. \ac{BASIN} has no guided variant and is not shown.}
    \label{fig:gain_MMM15_S}
\end{figure}

\begin{figure}[ht]
    \centering
    \includegraphics[width=0.75\linewidth]{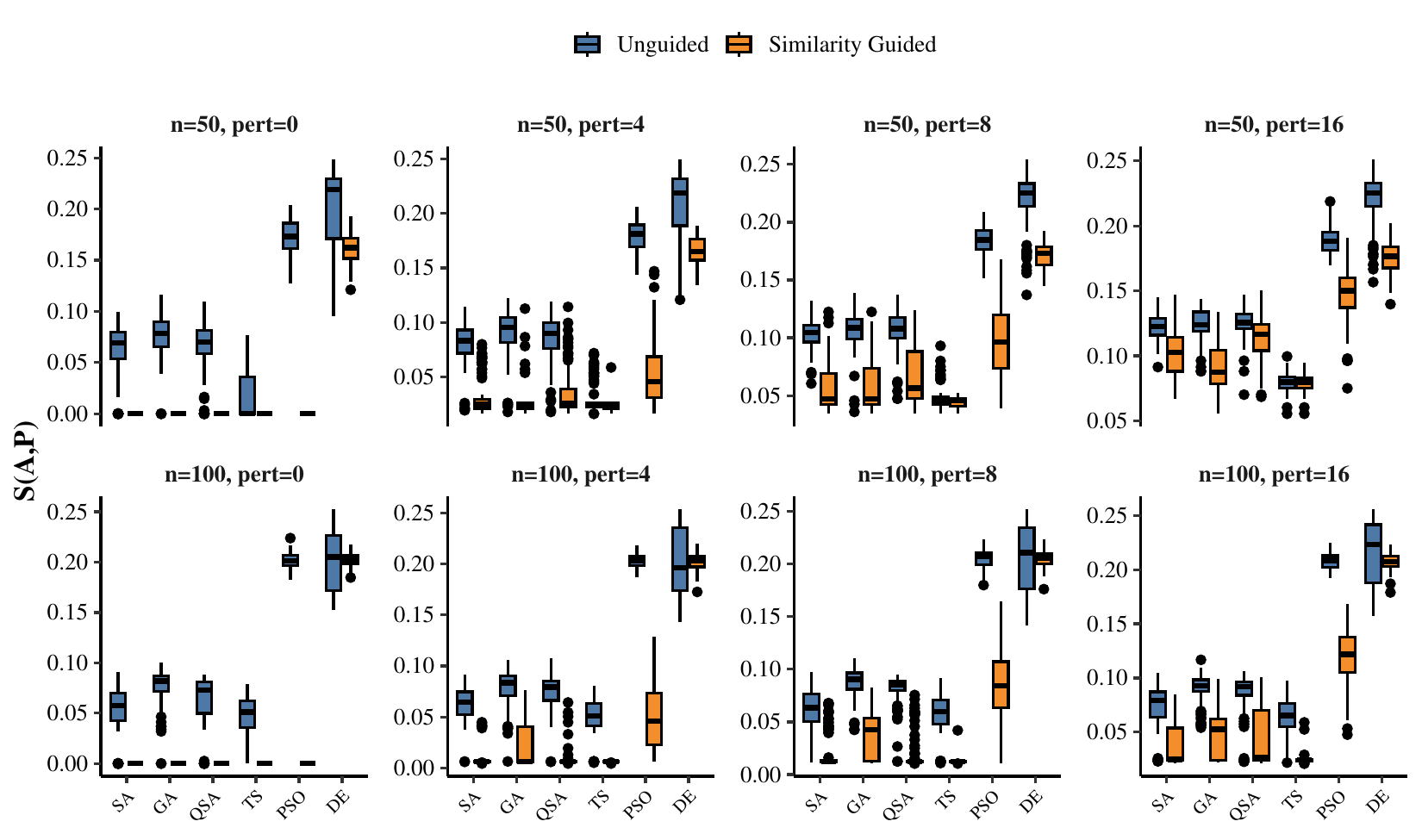}
 \caption{Node-similarity guidance on \ac{MMM} graphs at edge density $30~\%$, under $\mathcal{S}(A,P)$. Boxes show the size-normalized score $\mathcal{S}(A,P)$, which does not correct for fixed points over the 100 test graphs of each condition, one solution per graph, for the unguided and the similarity-guided variant of each algorithm; one panel per graph size $n$ and number $\mathit{pert}$ of perturbing edge swaps. \ac{BASIN} has no guided variant and is not shown.}
    \label{fig:gain_MMM30_S}
\end{figure}
\clearpage

\section{Performance comparison: Additional results}
\begin{figure}[ht]
    \centering
    \includegraphics[width=0.75\linewidth]{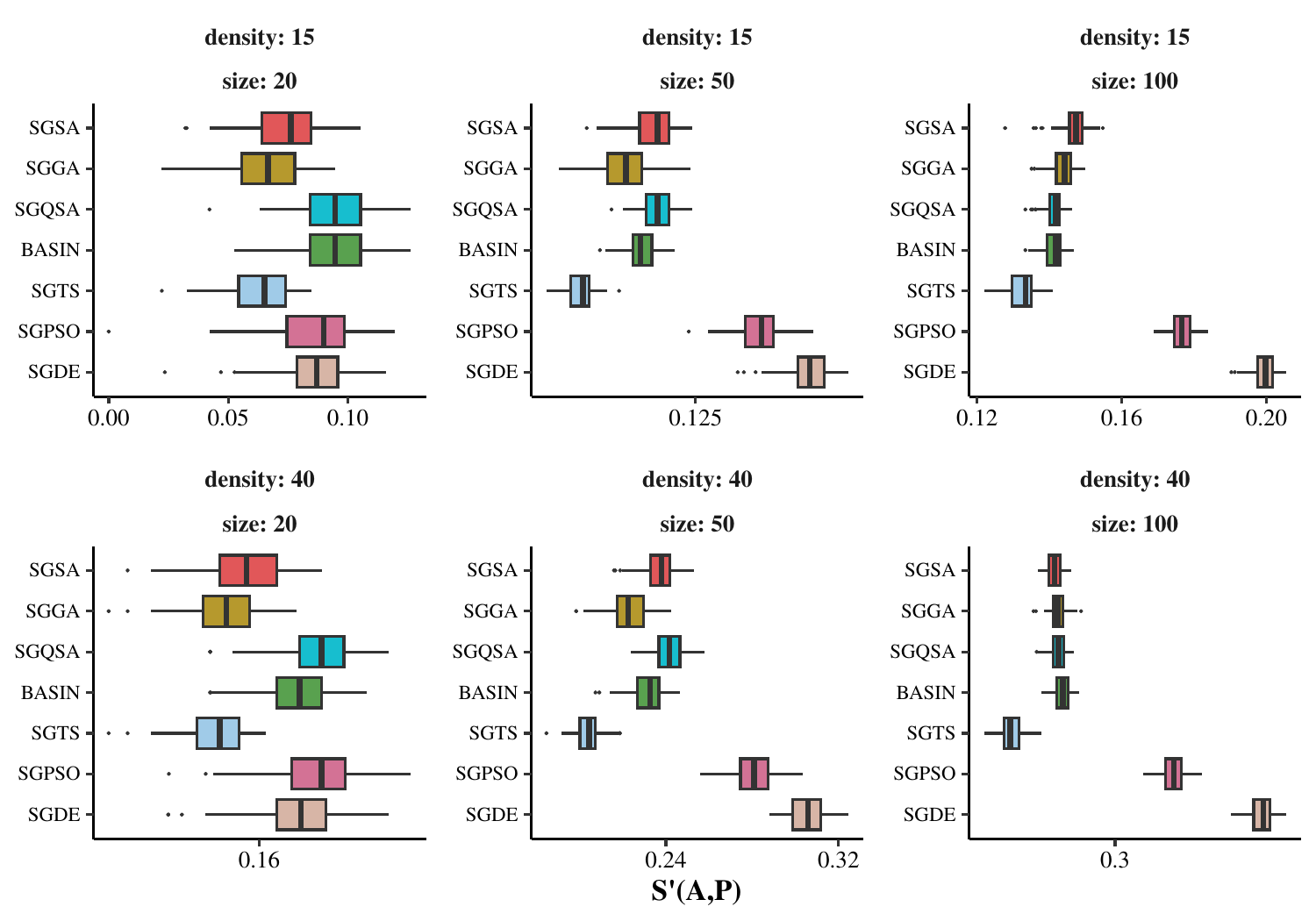}
 \caption{Performance on \ac{BA} graphs, one variant per algorithm. Boxes show $\mathcal{S}'(A,P)$ over the 100 test graphs of each condition, one solution per graph; one panel per graph size $n$ and edge density $d$, each with its own horizontal axis.}
    \label{fig:box_BA}
\end{figure}

\begin{figure}[ht]
    \centering
    \includegraphics[width=0.75\linewidth]{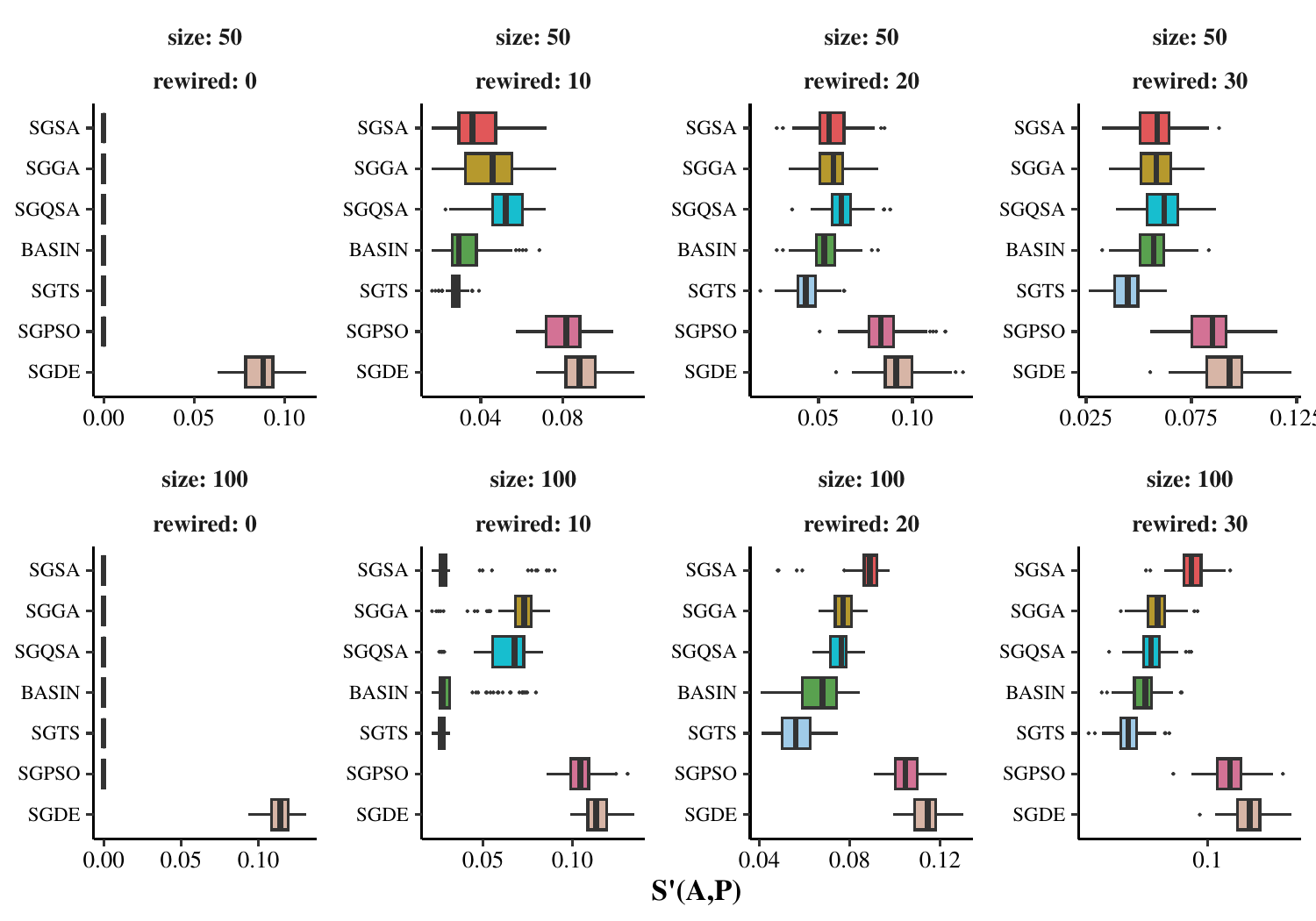}
 \caption{Performance on \ac{LRM} graphs at edge density $15~\%$, one variant per algorithm. Boxes show $\mathcal{S}'(A,P)$ over the 100 test graphs of each condition, one solution per graph; one panel per graph size $n$ and share $\mathit{rew}$ of rewired edges, each with its own horizontal axis.}
    \label{fig:box_LRM15}
\end{figure}

\begin{figure}[ht]
    \centering
    \includegraphics[width=0.75\linewidth]{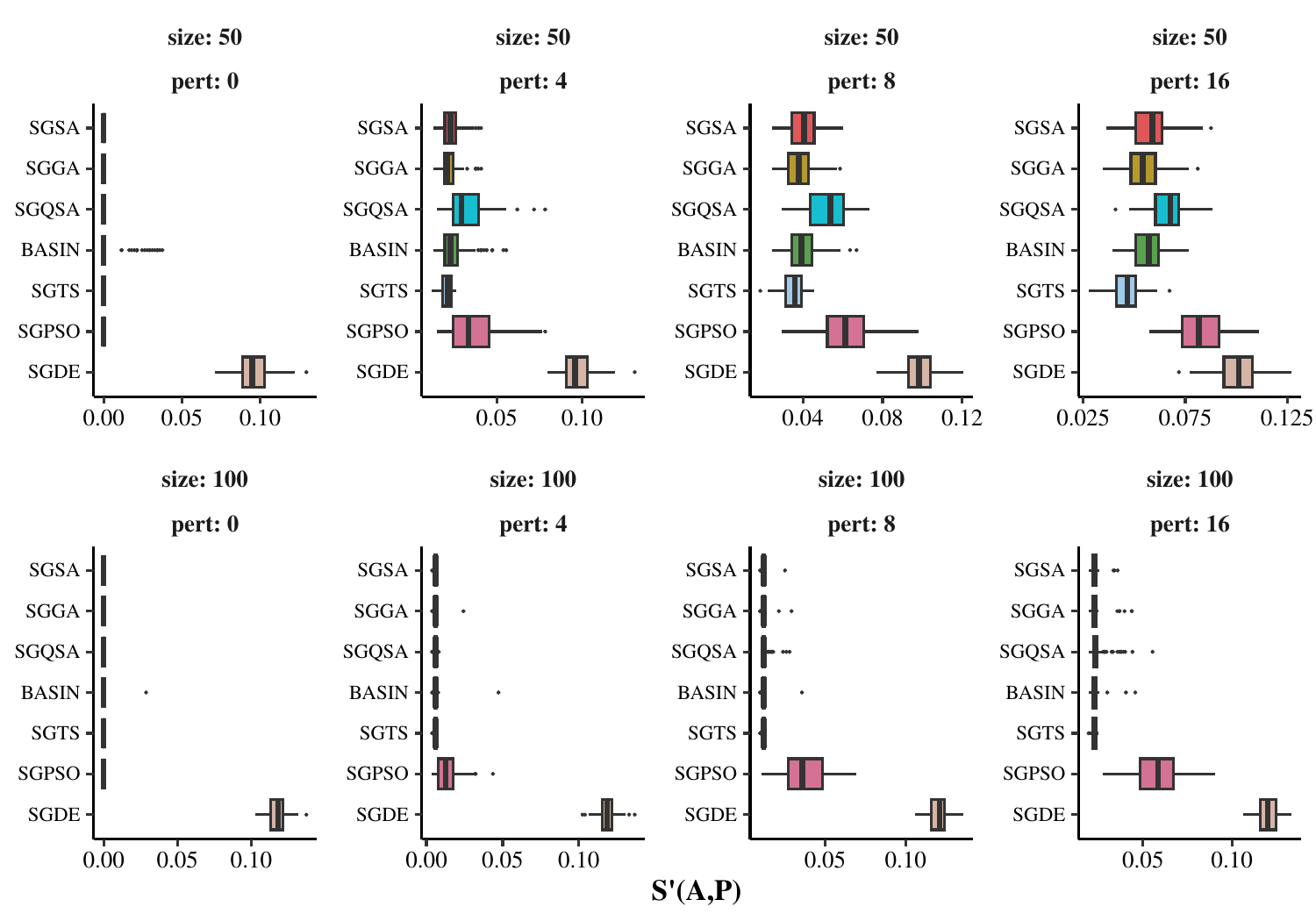}
 \caption{Performance on \ac{MMM} graphs at edge density $15~\%$, one variant per algorithm. Boxes show $\mathcal{S}'(A,P)$ over the 100 test graphs of each condition, one solution per graph; one panel per graph size $n$ and number $\mathit{pert}$ of perturbing edge swaps, each with its own horizontal axis.}
    \label{fig:box_MMM15}
\end{figure}

\subsection{Recovery of the planted map}\label{app:planted}

\begingroup\fontsize{8}{10}\selectfont

\begin{longtable}[t]{llll}
\caption{\label{tab:planted-recovery}Recovery of the planted map at zero perturbation (LRM rew0, MMM pert0), where it is an exact automorphism; four conditions of 100 graphs per graph family, against $0.015$ for a random permutation without fixed points. \emph{Restarts exact} and \emph{Graphs hit} are the shares of restarts and of graphs returning it.}\\
\toprule
Method & Restarts exact & Graphs hit & Nodes correct\\
\midrule
\endfirsthead
\caption[]{Recovery of the planted map at zero perturbation (LRM rew0, MMM pert0), where it is an exact automorphism. \textit{(continued)}}\\
\toprule
Method & Restarts exact & Graphs hit & Nodes correct\\
\midrule
\endhead

\endfoot
\bottomrule
\endlastfoot
\addlinespace[0.3em]
\multicolumn{4}{l}{\textbf{LRM}}\\
\hspace{1em}SGQSA & 0.999 & 0.998 & 1.000\\
\hspace{1em}BASIN & 0.983 & 0.988 & 1.000\\
\hspace{1em}SGTS & 0.981 & 0.980 & 1.000\\
\hspace{1em}SGGA & 0.980 & 0.980 & 1.000\\
\hspace{1em}SGPSO & 0.978 & 0.978 & 1.000\\
\hspace{1em}SGSA & 0.975 & 0.978 & 1.000\\
\hspace{1em}TS & 0.946 & 0.980 & 1.000\\
\hspace{1em}QSA & 0.184 & 0.545 & 0.326\\
\hspace{1em}SA & 0.086 & 0.307 & 0.110\\
\hspace{1em}GA & 0.041 & 0.163 & 0.013\\
\hspace{1em}DE & 0.000 & 0.000 & 0.013\\
\hspace{1em}PSO & 0.000 & 0.000 & 0.007\\
\hspace{1em}Random & 0.000 & 0.000 & 0.015\\
\hspace{1em}SGDE & 0.000 & 0.000 & 0.007\\
\addlinespace[0.3em]
\multicolumn{4}{l}{\textbf{MMM}}\\
\hspace{1em}SGQSA & 0.948 & 0.945 & 1.000\\
\hspace{1em}SGGA & 0.930 & 0.930 & 1.000\\
\hspace{1em}SGPSO & 0.929 & 0.925 & 1.000\\
\hspace{1em}SGTS & 0.923 & 0.910 & 1.000\\
\hspace{1em}SGSA & 0.854 & 0.935 & 1.000\\
\hspace{1em}BASIN & 0.709 & 0.802 & 0.860\\
\hspace{1em}TS & 0.408 & 0.625 & 0.642\\
\hspace{1em}SA & 0.057 & 0.200 & 0.160\\
\hspace{1em}QSA & 0.035 & 0.125 & 0.105\\
\hspace{1em}GA & 0.009 & 0.043 & 0.015\\
\hspace{1em}DE & 0.000 & 0.000 & 0.015\\
\hspace{1em}PSO & 0.000 & 0.000 & 0.015\\
\hspace{1em}Random & 0.000 & 0.000 & 0.015\\
\hspace{1em}SGDE & 0.000 & 0.000 & 0.013\\*
\end{longtable}
\endgroup{}{}

\begin{figure}[ht]
    \centering
    \includegraphics[width=0.95\linewidth]{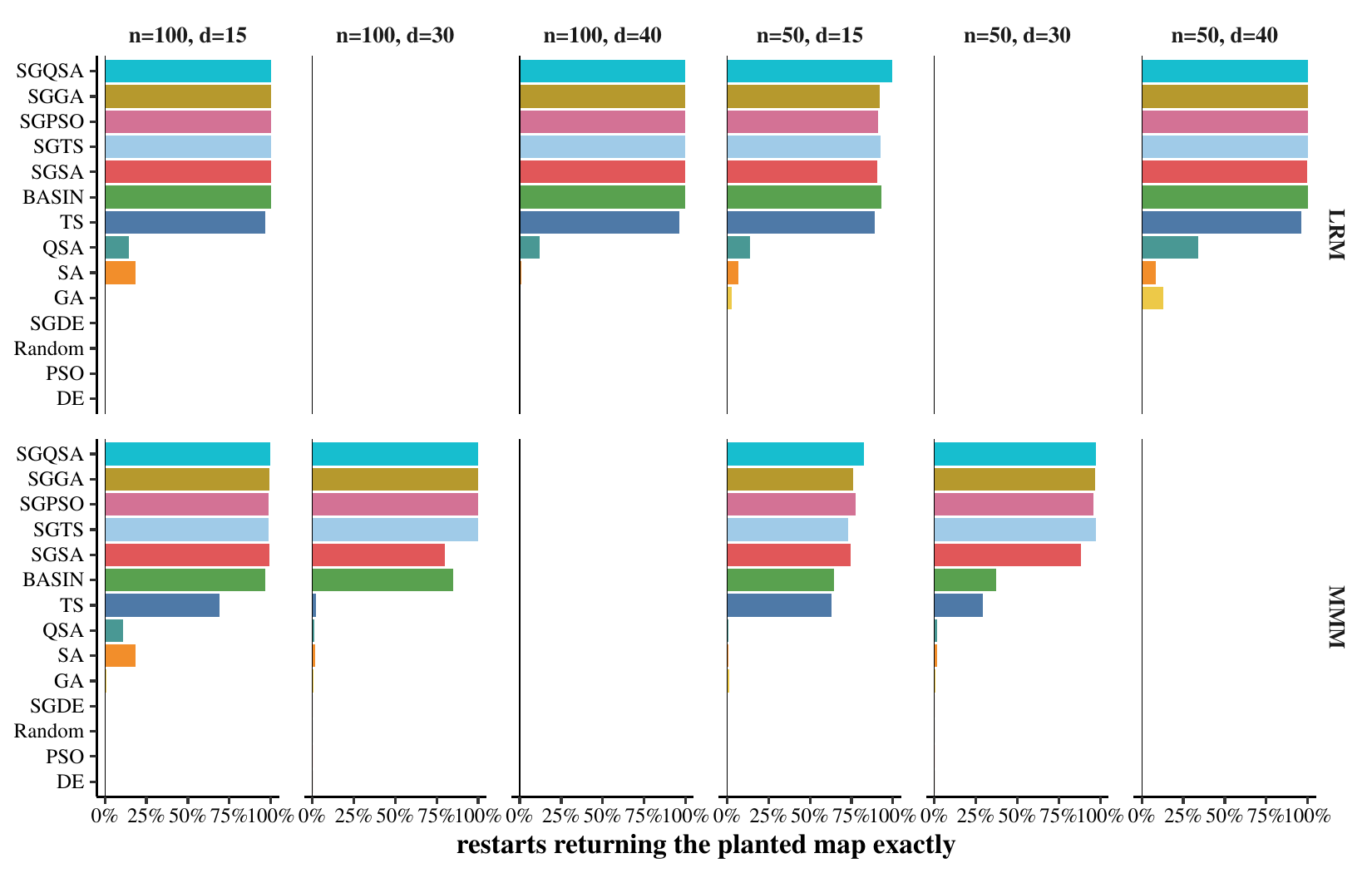}
 \caption{Share of restarts returning the planted map exactly where it is an exact automorphism (\ac{LRM} $\mathit{rew}=0$, \ac{MMM} $\mathit{pert}=0$), over every restart on the 100 test graphs of each condition; every variant of every algorithm and the random-search baseline. One panel per graph size $n$, edge density $d$ and graph family.}
    \label{fig:planted-exact}
\end{figure}

\begin{figure}[ht]
    \centering
    \includegraphics[width=0.95\linewidth]{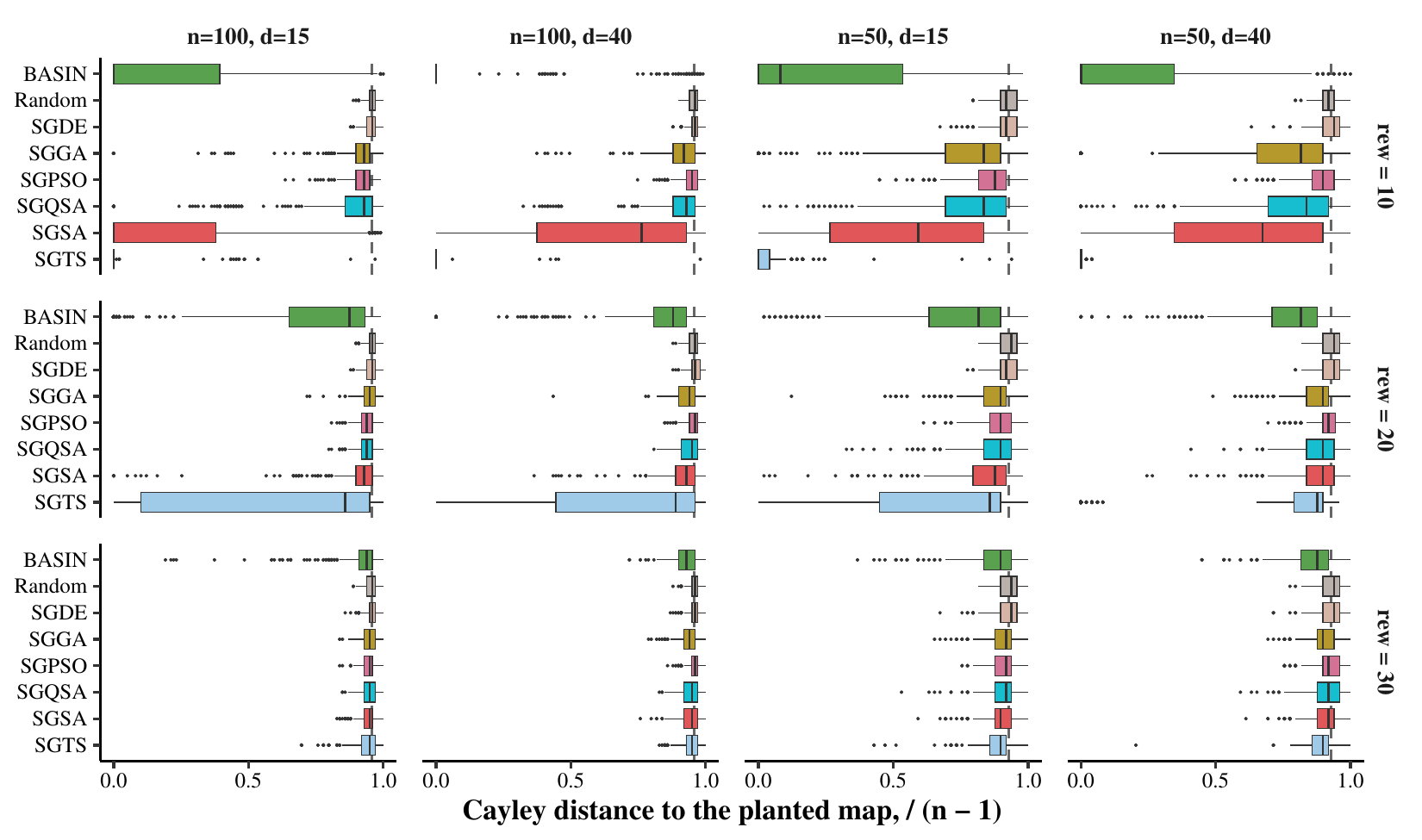}
 \caption{Cayley distance to the planted map over its maximum $n-1$ on \ac{LRM} graphs, one variant per algorithm and the random-search baseline. Boxes span every restart on the 100 test graphs of each condition, $0$ being the planted map itself; one panel per graph size $n$, edge density $d$ and share $\mathit{rew}\in\{10,20,30\}~\%$ of rewired edges. The dashed line is the median distance of a random permutation without fixed points.}
    \label{fig:planted-distance}
\end{figure}

\begin{figure}[ht]
    \centering
    \includegraphics[width=0.95\linewidth]{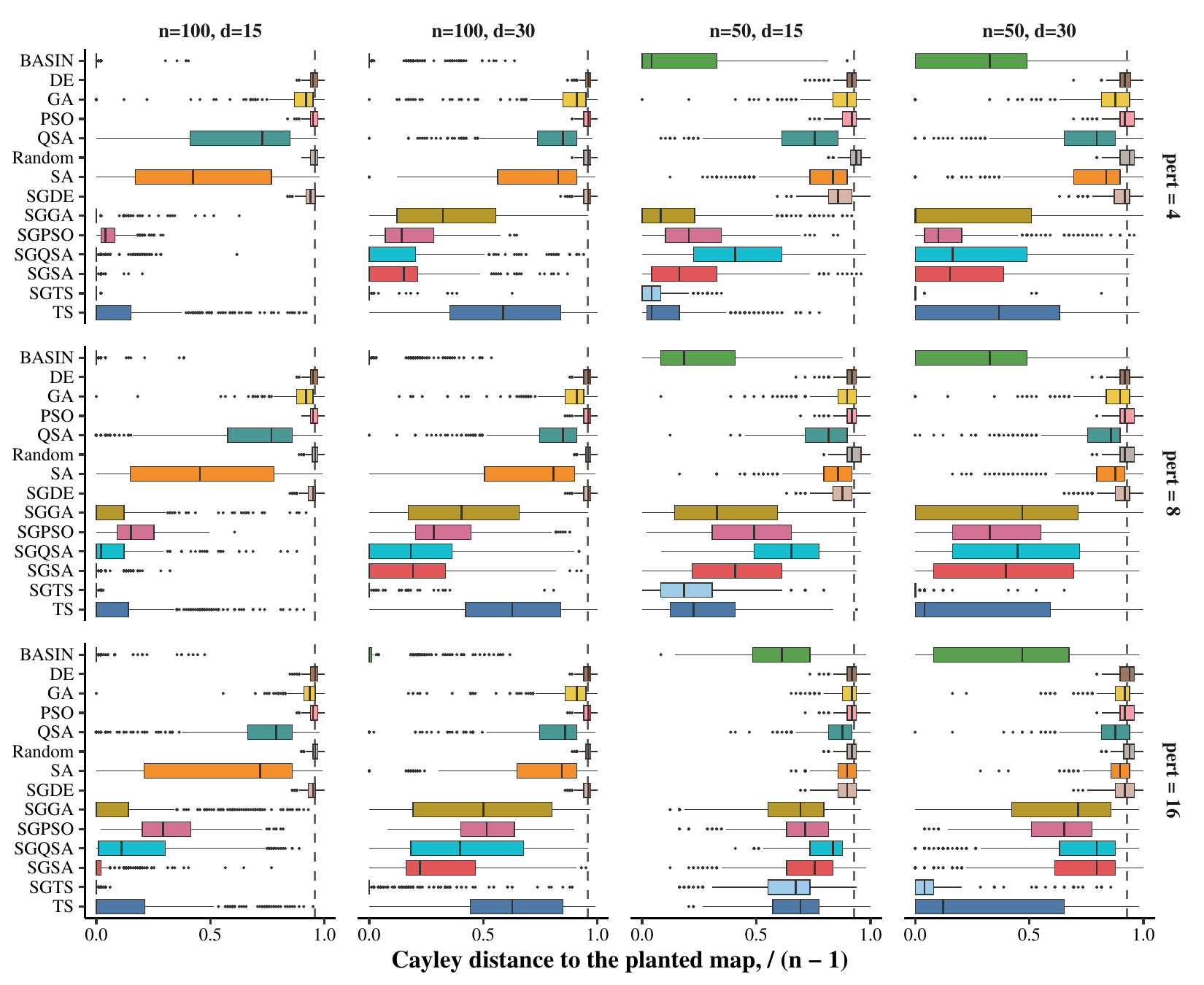}
    \caption{Cayley distance to the planted map over its maximum $n-1$ on \ac{MMM} graphs, every variant of every algorithm and the random-search baseline. Boxes span every restart on the 100 test graphs of each condition, $0$ being the planted map itself; one panel per graph size $n$, edge density $d$ and number $\mathit{pert}\in\{4,8,16\}$ of perturbing edge swaps. The dashed line is the median distance of a random permutation without fixed points.}
    \label{fig:planted-distance-mmm-all}
\end{figure}

\subsection{BASIN component ablation}\label{app:basin-ablation}

\begin{figure}[ht]
    \centering
    \includegraphics[width=0.95\linewidth]{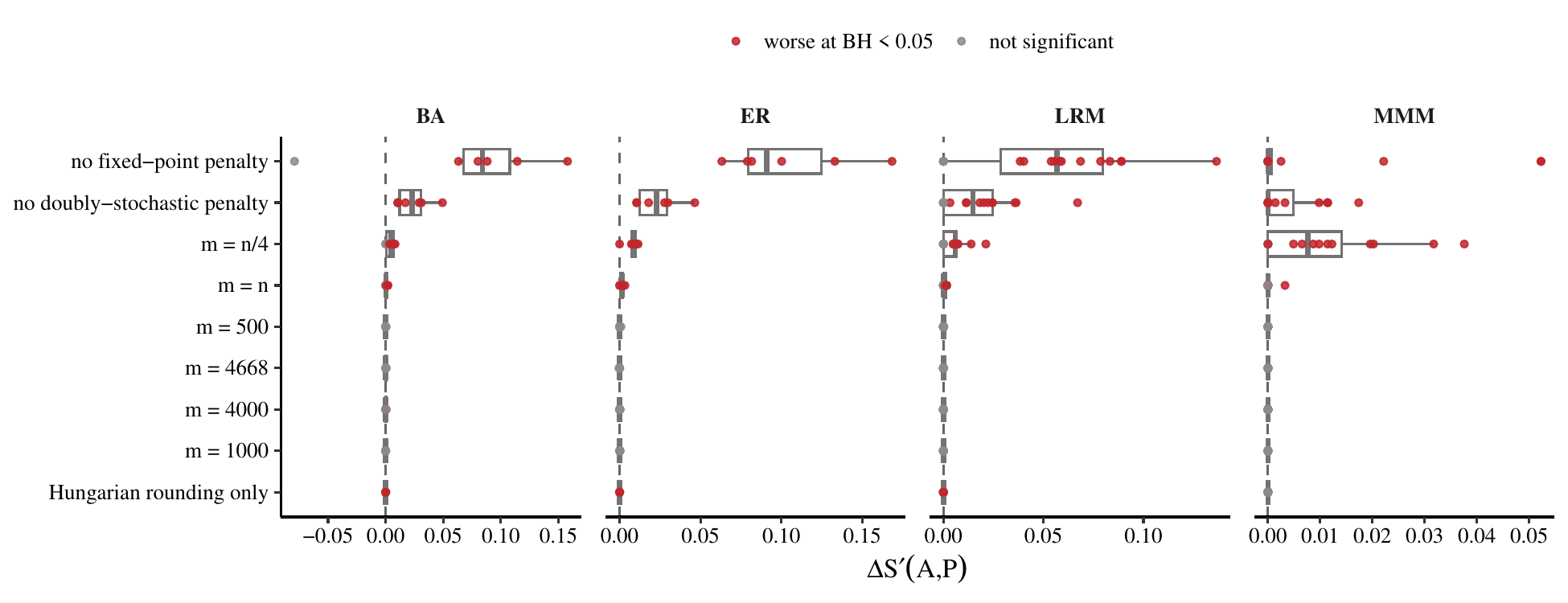}
 \caption{\Ac{BASIN} component ablation, one row per ablated or varied component and one panel per graph family. Each point is one condition: the median $\mathcal{S}'(A,P)$ of the ablated configuration minus that of the adopted one over the 100 test graphs, so that right of the dashed zero line the change hurt; boxes summarize the conditions. Red points are worse by a paired Wilcoxon signed-rank test after Benjamini--Hochberg correction at $0.05$.}
    \label{fig:basin-ablation-delta}
\end{figure}

\subsection{The orthogonal minimum spanning tree thresholding}\label{app:threshold}

\begin{figure}[ht]
    \centering
    \includegraphics[width=0.9\linewidth]{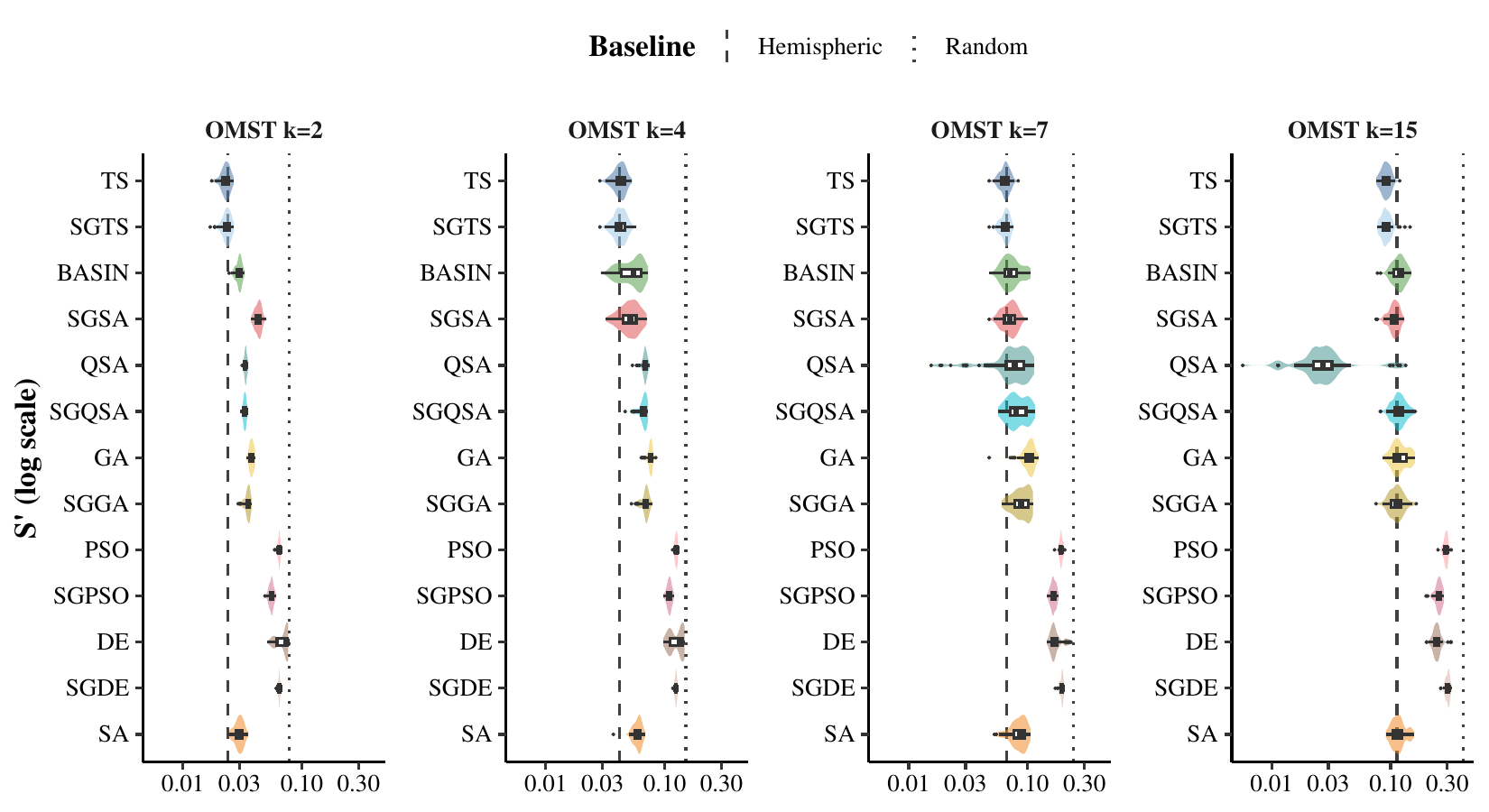}
 \caption{$\mathcal{S}'(A,P)$ on the 88 structural connectomes thresholded by orthogonal minimum spanning trees, one panel per $k \in \{2,4,7,15\}$, every variant of every algorithm. Violins and boxes span the subjects, one solution per subject; dashed is the median of $\pi_{LR}$ and dotted that of the random baseline. Logarithmic axis.}
    \label{fig:real-omst}
\end{figure}

Beside the proportional densities, each connectome is also thresholded by orthogonal minimum spanning trees at $k \in \{2,4,7,15\}$, which retain $k(n-1)$ edges. The progression of the proportional densities is reproduced at these edge counts (Fig.~\ref{fig:real-omst}). \ac{OMST} $k=4$ and $k=7$ behave as the $10\%$ and $15\%$ densities do, at median homologue rates of $0.856$ and $0.800$ for \ac{TS}, while \ac{OMST} $k=2$ behaves as $5\%$ does, at $0.000$.

\subsection{Structural connectomes: additional results}\label{app:structural_connectomes}

\begingroup\fontsize{8}{10}\selectfont

\begin{longtable}[t]{llll}
\caption{\label{tab:real-sc-tex}Structural connectomes at the configuration tuned under $F \leq 0.25n$, one solution per subject. Median $\mathcal{S}'(A,P)$ with interquartile range in brackets. Win / tie / loss are shares of subjects against $\pi_{LR}$, exact ties their own category.}\\
\toprule
Method & Median $S'$ [IQR] & $F/n$ & Win / tie / loss vs $\pi_{LR}$\\
\midrule
\endfirsthead
\caption[]{Structural connectomes at the tuned configuration ($F \leq 0.25n$), 88 subjects, one solution each. \textit{(continued)}}\\
\toprule
Method & Median $S'$ [IQR] & $F/n$ & Win / tie / loss vs $\pi_{LR}$\\
\midrule
\endhead

\endfoot
\bottomrule
\endlastfoot
\addlinespace[0.3em]
\multicolumn{4}{l}{\textbf{5\%}}\\
\hspace{1em}Hemispheric & 0.0260 [0.0235, 0.0281] & 0.000 & --\\
\hspace{1em}SA & 0.0311 [0.0289, 0.0328] & 0.206 & 0.12 / 0.00 / 0.88\\
\hspace{1em}SGSA & 0.0395 [0.0368, 0.0431] & 0.033 & 0.00 / 0.00 / 1.00\\
\hspace{1em}GA & 0.0432 [0.0423, 0.0444] & 0.000 & 0.00 / 0.00 / 1.00\\
\hspace{1em}SGGA & 0.0380 [0.0355, 0.0399] & 0.061 & 0.01 / 0.00 / 0.99\\
\hspace{1em}PSO & 0.0712 [0.0700, 0.0726] & 0.056 & 0.00 / 0.00 / 1.00\\
\hspace{1em}SGPSO & 0.0574 [0.0554, 0.0594] & 0.011 & 0.00 / 0.00 / 1.00\\
\hspace{1em}DE & 0.0744 [0.0649, 0.0801] & 0.111 & 0.00 / 0.00 / 1.00\\
\hspace{1em}SGDE & 0.0719 [0.0705, 0.0725] & 0.033 & 0.00 / 0.00 / 1.00\\
\hspace{1em}TS & 0.0244 [0.0223, 0.0257] & 0.244 & 0.69 / 0.05 / 0.26\\
\hspace{1em}SGTS & 0.0249 [0.0232, 0.0267] & 0.233 & 0.59 / 0.00 / 0.41\\
\hspace{1em}QSA & 0.0385 [0.0373, 0.0390] & 0.000 & 0.00 / 0.00 / 1.00\\
\hspace{1em}SGQSA & 0.0375 [0.0360, 0.0385] & 0.000 & 0.00 / 0.00 / 1.00\\
\hspace{1em}BASIN & 0.0330 [0.0310, 0.0350] & 0.000 & 0.05 / 0.00 / 0.95\\
\hspace{1em}Random & 0.0879 [0.0874, 0.0879] & 0.011 & 0.00 / 0.00 / 1.00\\
\addlinespace[0.3em]
\multicolumn{4}{l}{\textbf{10\%}}\\
\hspace{1em}Hemispheric & 0.0464 [0.0424, 0.0496] & 0.000 & --\\
\hspace{1em}SA & 0.0605 [0.0563, 0.0648] & 0.183 & 0.00 / 0.02 / 0.98\\
\hspace{1em}SGSA & 0.0527 [0.0444, 0.0578] & 0.011 & 0.08 / 0.07 / 0.85\\
\hspace{1em}GA & 0.0814 [0.0769, 0.0844] & 0.011 & 0.00 / 0.00 / 1.00\\
\hspace{1em}SGGA & 0.0697 [0.0651, 0.0747] & 0.161 & 0.00 / 0.01 / 0.99\\
\hspace{1em}PSO & 0.1348 [0.1319, 0.1375] & 0.078 & 0.00 / 0.00 / 1.00\\
\hspace{1em}SGPSO & 0.1119 [0.1074, 0.1153] & 0.011 & 0.00 / 0.00 / 1.00\\
\hspace{1em}DE & 0.1216 [0.1166, 0.1298] & 0.167 & 0.00 / 0.00 / 1.00\\
\hspace{1em}SGDE & 0.1359 [0.1332, 0.1374] & 0.044 & 0.00 / 0.00 / 1.00\\
\hspace{1em}TS & 0.0455 [0.0419, 0.0489] & 0.000 & 0.51 / 0.36 / 0.12\\
\hspace{1em}SGTS & 0.0455 [0.0423, 0.0488] & 0.000 & 0.58 / 0.27 / 0.15\\
\hspace{1em}QSA & 0.0709 [0.0670, 0.0745] & 0.000 & 0.00 / 0.00 / 1.00\\
\hspace{1em}SGQSA & 0.0712 [0.0644, 0.0744] & 0.000 & 0.00 / 0.00 / 1.00\\
\hspace{1em}BASIN & 0.0659 [0.0469, 0.0709] & 0.000 & 0.03 / 0.02 / 0.94\\
\hspace{1em}Random & 0.1663 [0.1653, 0.1668] & 0.011 & 0.00 / 0.00 / 1.00\\
\addlinespace[0.3em]
\multicolumn{4}{l}{\textbf{15\%}}\\
\hspace{1em}Hemispheric & 0.0652 [0.0594, 0.0699] & 0.000 & --\\
\hspace{1em}SA & 0.0782 [0.0694, 0.0847] & 0.083 & 0.15 / 0.00 / 0.85\\
\hspace{1em}SGSA & 0.0694 [0.0609, 0.0754] & 0.011 & 0.24 / 0.08 / 0.68\\
\hspace{1em}GA & 0.0975 [0.0885, 0.1045] & 0.056 & 0.00 / 0.01 / 0.99\\
\hspace{1em}SGGA & 0.0811 [0.0731, 0.0930] & 0.056 & 0.07 / 0.00 / 0.93\\
\hspace{1em}PSO & 0.1834 [0.1792, 0.1868] & 0.111 & 0.00 / 0.00 / 1.00\\
\hspace{1em}SGPSO & 0.1523 [0.1473, 0.1578] & 0.011 & 0.00 / 0.00 / 1.00\\
\hspace{1em}DE & 0.1585 [0.1539, 0.1665] & 0.178 & 0.00 / 0.00 / 1.00\\
\hspace{1em}SGDE & 0.1855 [0.1824, 0.1895] & 0.078 & 0.00 / 0.00 / 1.00\\
\hspace{1em}TS & 0.0615 [0.0584, 0.0655] & 0.000 & 0.84 / 0.11 / 0.05\\
\hspace{1em}SGTS & 0.0617 [0.0579, 0.0652] & 0.000 & 0.80 / 0.15 / 0.06\\
\hspace{1em}QSA & 0.0704 [0.0544, 0.0817] & 0.022 & 0.28 / 0.00 / 0.72\\
\hspace{1em}SGQSA & 0.0757 [0.0667, 0.0889] & 0.000 & 0.06 / 0.03 / 0.91\\
\hspace{1em}BASIN & 0.0742 [0.0643, 0.0948] & 0.000 & 0.10 / 0.03 / 0.86\\
\hspace{1em}Random & 0.2352 [0.2342, 0.2365] & 0.011 & 0.00 / 0.00 / 1.00\\
\addlinespace[0.3em]
\multicolumn{4}{l}{\textbf{40\%}}\\
\hspace{1em}Hemispheric & 0.1281 [0.1199, 0.1395] & 0.000 & --\\
\hspace{1em}SA & 0.1166 [0.1060, 0.1251] & 0.244 & 0.76 / 0.00 / 0.24\\
\hspace{1em}SGSA & 0.1082 [0.1011, 0.1157] & 0.244 & 0.93 / 0.00 / 0.07\\
\hspace{1em}GA & 0.1155 [0.1073, 0.1255] & 0.244 & 0.73 / 0.00 / 0.27\\
\hspace{1em}SGGA & 0.1117 [0.1021, 0.1246] & 0.244 & 0.78 / 0.00 / 0.22\\
\hspace{1em}PSO & 0.3067 [0.3002, 0.3137] & 0.178 & 0.00 / 0.00 / 1.00\\
\hspace{1em}SGPSO & 0.2679 [0.2607, 0.2793] & 0.011 & 0.00 / 0.00 / 1.00\\
\hspace{1em}DE & 0.2587 [0.2475, 0.2728] & 0.200 & 0.00 / 0.00 / 1.00\\
\hspace{1em}SGDE & 0.3205 [0.3137, 0.3277] & 0.089 & 0.00 / 0.00 / 1.00\\
\hspace{1em}TS & 0.0901 [0.0862, 0.0961] & 0.244 & 1.00 / 0.00 / 0.00\\
\hspace{1em}SGTS & 0.0909 [0.0868, 0.0975] & 0.244 & 1.00 / 0.00 / 0.00\\
\hspace{1em}QSA & 0.0239 [0.0226, 0.0308] & 0.956 & 1.00 / 0.00 / 0.00\\
\hspace{1em}SGQSA & 0.1266 [0.1169, 0.1342] & 0.000 & 0.62 / 0.01 / 0.36\\
\hspace{1em}BASIN & 0.1280 [0.1192, 0.1352] & 0.011 & 0.55 / 0.02 / 0.43\\
\hspace{1em}Random & 0.4320 [0.4290, 0.4341] & 0.011 & 0.00 / 0.00 / 1.00\\*
\end{longtable}
\endgroup

\begingroup\fontsize{8}{10}\selectfont

\begin{longtable}[t]{llll}
\caption{\label{tab:real-sc-map}The map returned on the 88 structural connectomes at the configuration tuned under $F \leq 0.25n$, one solution per subject. \emph{Homologue rate} is the share of the $90$ regions sent to their homologue, \emph{Share $>0.9$ / $<0.1$} the subjects beyond those rates, \emph{Cayley (norm.)} the Cayley distance to $\pi_{LR}$ (Sec.~\ref{sss:evaluation}) over its maximum $n-1$.}\\
\toprule
Method & Homologue rate & Share $>0.9$ / $<0.1$ & Cayley (norm.)\\
\midrule
\endfirsthead
\caption[]{The map returned on the 88 structural connectomes, one solution per subject. \textit{(continued)}}\\
\toprule
Method & Homologue rate & Share $>0.9$ / $<0.1$ & Cayley (norm.)\\
\midrule
\endhead

\endfoot
\bottomrule
\endlastfoot
\addlinespace[0.3em]
\multicolumn{4}{l}{\textbf{5\%}}\\
\hspace{1em}Hemispheric & 1.000 & 1.00 / 0.00 & \vphantom{3} 0.000\\
\hspace{1em}SA & 0.011 & 0.00 / 0.91 & 0.921\\
\hspace{1em}SGSA & 0.061 & 0.00 / 0.57 & 0.865\\
\hspace{1em}GA & 0.011 & 0.00 / 0.99 & 0.944\\
\hspace{1em}SGGA & 0.022 & 0.00 / 0.90 & 0.899\\
\hspace{1em}PSO & 0.011 & 0.00 / 1.00 & \vphantom{1} 0.944\\
\hspace{1em}SGPSO & 0.028 & 0.00 / 0.92 & 0.933\\
\hspace{1em}DE & 0.011 & 0.00 / 1.00 & 0.949\\
\hspace{1em}SGDE & 0.011 & 0.00 / 0.99 & 0.955\\
\hspace{1em}TS & 0.017 & 0.02 / 0.65 & 0.854\\
\hspace{1em}SGTS & 0.011 & 0.00 / 0.73 & 0.854\\
\hspace{1em}QSA & 0.028 & 0.00 / 0.88 & 0.916\\
\hspace{1em}SGQSA & 0.028 & 0.00 / 0.82 & 0.910\\
\hspace{1em}BASIN & 0.033 & 0.00 / 0.61 & 0.893\\
\hspace{1em}Random & 0.011 & 0.00 / 1.00 & \vphantom{2} 0.955\\
\addlinespace[0.3em]
\multicolumn{4}{l}{\textbf{10\%}}\\
\hspace{1em}Hemispheric & 1.000 & 1.00 / 0.00 & \vphantom{2} 0.000\\
\hspace{1em}SA & 0.033 & 0.00 / 0.60 & 0.843\\
\hspace{1em}SGSA & 0.583 & 0.09 / 0.31 & 0.343\\
\hspace{1em}GA & 0.011 & 0.00 / 0.95 & 0.933\\
\hspace{1em}SGGA & 0.028 & 0.01 / 0.70 & 0.876\\
\hspace{1em}PSO & 0.006 & 0.00 / 1.00 & 0.944\\
\hspace{1em}SGPSO & 0.033 & 0.00 / 0.86 & 0.910\\
\hspace{1em}DE & 0.011 & 0.00 / 1.00 & 0.933\\
\hspace{1em}SGDE & 0.011 & 0.00 / 1.00 & 0.955\\
\hspace{1em}TS & 0.844 & 0.34 / 0.19 & 0.090\\
\hspace{1em}SGTS & 0.833 & 0.33 / 0.22 & 0.112\\
\hspace{1em}QSA & 0.128 & 0.00 / 0.44 & 0.781\\
\hspace{1em}SGQSA & 0.178 & 0.00 / 0.38 & 0.708\\
\hspace{1em}BASIN & 0.106 & 0.09 / 0.47 & 0.781\\
\hspace{1em}Random & 0.011 & 0.00 / 1.00 & 0.949\\
\addlinespace[0.3em]
\multicolumn{4}{l}{\textbf{15\%}}\\
\hspace{1em}Hemispheric & 1.000 & 1.00 / 0.00 & \vphantom{1} 0.000\\
\hspace{1em}SA & 0.056 & 0.03 / 0.52 & 0.798\\
\hspace{1em}SGSA & 0.661 & 0.07 / 0.19 & 0.270\\
\hspace{1em}GA & 0.022 & 0.01 / 0.76 & 0.876\\
\hspace{1em}SGGA & 0.022 & 0.01 / 0.61 & 0.831\\
\hspace{1em}PSO & 0.011 & 0.00 / 1.00 & 0.944\\
\hspace{1em}SGPSO & 0.022 & 0.00 / 0.80 & 0.921\\
\hspace{1em}DE & 0.011 & 0.00 / 1.00 & \vphantom{1} 0.921\\
\hspace{1em}SGDE & 0.000 & 0.00 / 1.00 & 0.944\\
\hspace{1em}TS & 0.794 & 0.27 / 0.32 & 0.140\\
\hspace{1em}SGTS & 0.811 & 0.22 / 0.27 & 0.129\\
\hspace{1em}QSA & 0.178 & 0.01 / 0.49 & 0.562\\
\hspace{1em}SGQSA & 0.522 & 0.00 / 0.12 & 0.365\\
\hspace{1em}BASIN & 0.689 & 0.05 / 0.35 & 0.225\\
\hspace{1em}Random & 0.011 & 0.00 / 1.00 & \vphantom{1} 0.955\\
\addlinespace[0.3em]
\multicolumn{4}{l}{\textbf{40\%}}\\
\hspace{1em}Hemispheric & 1.000 & 1.00 / 0.00 & 0.000\\
\hspace{1em}SA & 0.022 & 0.00 / 0.77 & 0.820\\
\hspace{1em}SGSA & 0.022 & 0.00 / 0.83 & 0.787\\
\hspace{1em}GA & 0.022 & 0.00 / 0.81 & 0.809\\
\hspace{1em}SGGA & 0.039 & 0.00 / 0.72 & 0.787\\
\hspace{1em}PSO & 0.011 & 0.00 / 1.00 & 0.933\\
\hspace{1em}SGPSO & 0.056 & 0.00 / 0.67 & 0.876\\
\hspace{1em}DE & 0.011 & 0.00 / 1.00 & 0.921\\
\hspace{1em}SGDE & 0.011 & 0.00 / 1.00 & 0.933\\
\hspace{1em}TS & 0.000 & 0.00 / 0.98 & 0.775\\
\hspace{1em}SGTS & 0.000 & 0.00 / 0.94 & 0.775\\
\hspace{1em}QSA & 0.000 & 0.00 / 0.98 & 0.528\\
\hspace{1em}SGQSA & 0.417 & 0.00 / 0.32 & 0.489\\
\hspace{1em}BASIN & 0.022 & 0.00 / 0.65 & 0.770\\
\hspace{1em}Random & 0.011 & 0.00 / 1.00 & 0.955\\*
\end{longtable}
\endgroup{}{}{}

\begin{table}[!h]
\centering
\caption{\label{tab:real-sc-agreement}Agreement between restarts on the structural connectomes: the share of the $90$ nodes that two restarts of a subject send to the same image, averaged over restart pairs and then over the $88$ subjects. \ac{QSA} at $40\%$ fixes $86$ of the $90$ regions, so its restarts agree trivially.}
\centering
\fontsize{7}{9}\selectfont
\begin{tabular}[t]{@{}lllllllll@{}}
\toprule
\multicolumn{1}{c}{ } & \multicolumn{2}{c}{5\%} & \multicolumn{2}{c}{10\%} & \multicolumn{2}{c}{15\%} & \multicolumn{2}{c}{40\%} \\
\cmidrule(l{3pt}r{3pt}){2-3} \cmidrule(l{3pt}r{3pt}){4-5} \cmidrule(l{3pt}r{3pt}){6-7} \cmidrule(l{3pt}r{3pt}){8-9}
Algorithms & unguided & guided & unguided & guided & unguided & guided & unguided & guided\\
\midrule
SA / SGSA & 0.03 & 0.17 & 0.04 & 0.28 & 0.05 & 0.25 & 0.11 & 0.22\\
GA / SGGA & 0.01 & 0.05 & 0.02 & 0.05 & 0.02 & 0.08 & 0.08 & 0.17\\
QSA / SGQSA & 0.02 & 0.03 & 0.04 & 0.07 & 0.09 & 0.16 & 0.49 & 0.26\\
TS / SGTS & 0.08 & 0.42 & 0.21 & 0.63 & 0.21 & 0.70 & 0.23 & 0.69\\
\bottomrule
\end{tabular}

\end{table}

\begin{figure}[ht]
    \centering
    \includegraphics[width=1.0\linewidth]{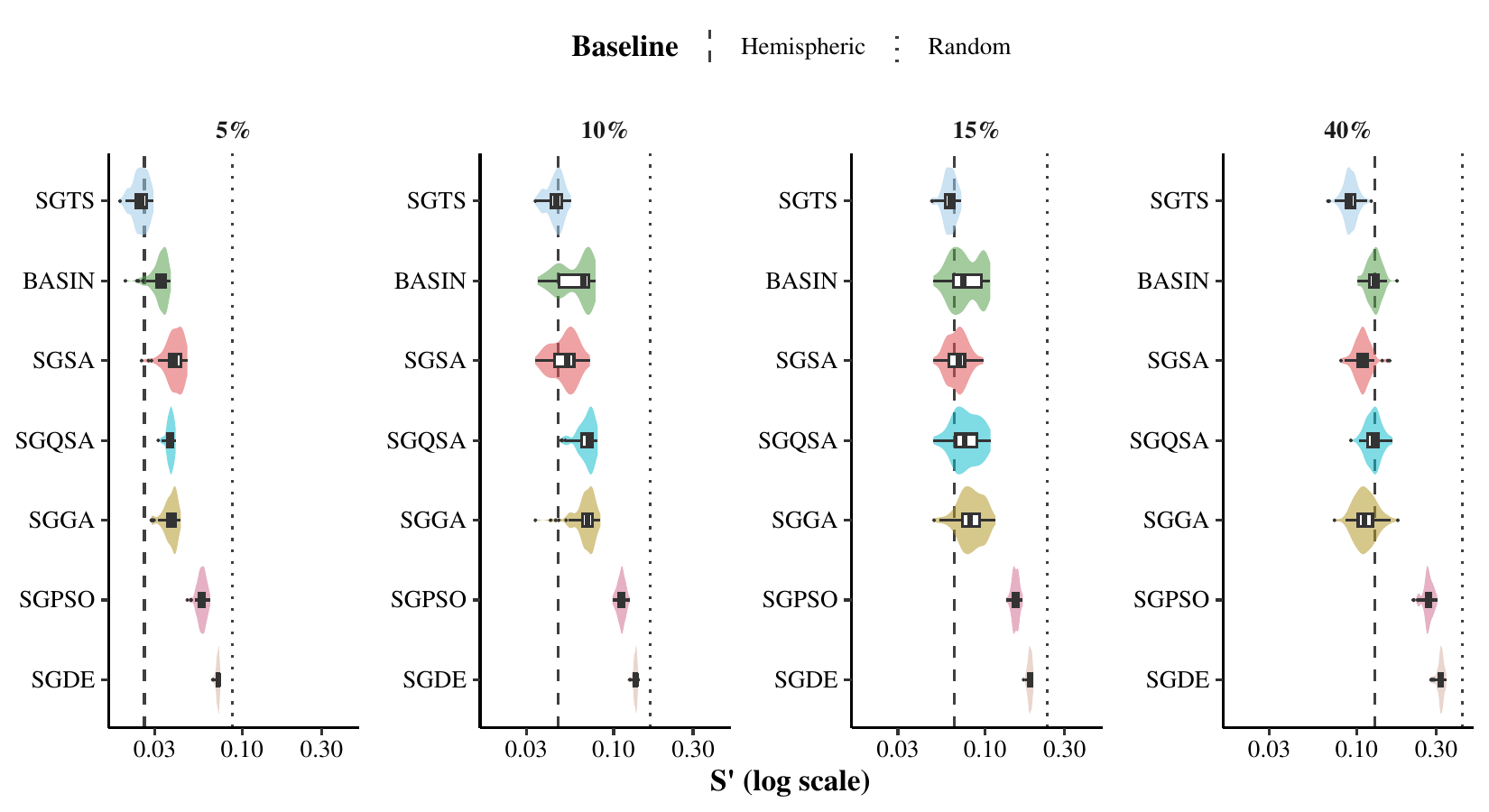}
 \caption{$\mathcal{S}'(A,P)$ on the 88 structural connectomes, one panel per proportional edge density, one variant per algorithm. Violins and boxes span the subjects, one solution per subject; dashed is the median of $\pi_{LR}$ and dotted that of the random baseline. Logarithmic axis.}
    \label{fig:real-sprime}
\end{figure}

\begin{figure}[ht]
    \centering
    \includegraphics[width=0.75\linewidth]{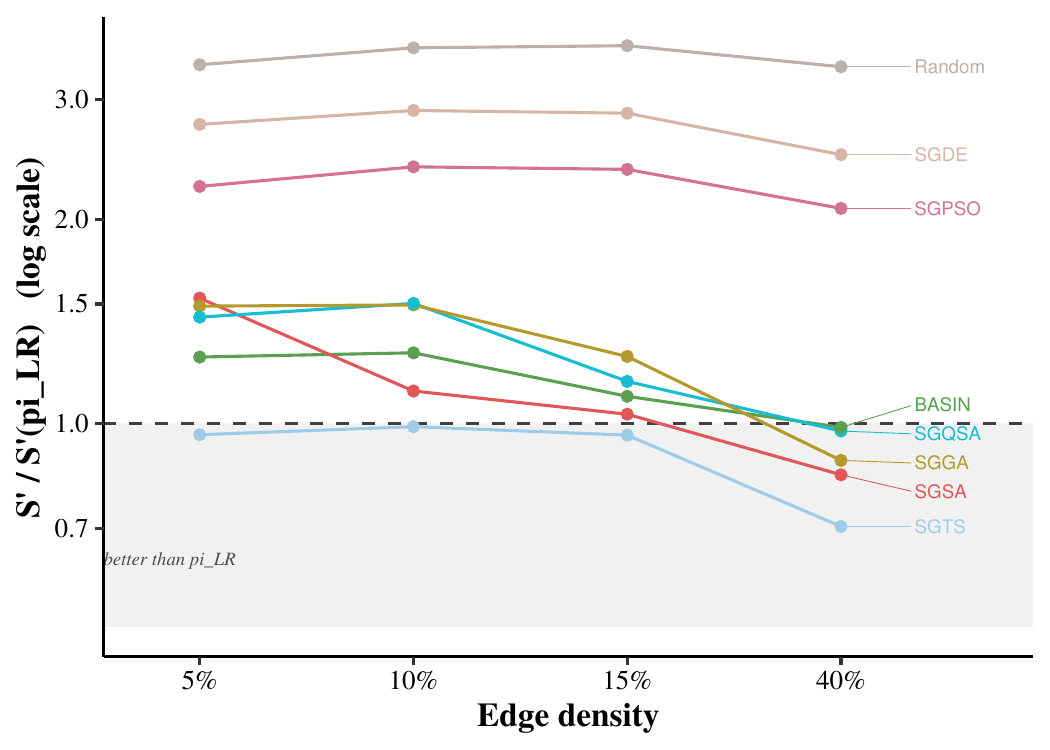}
 \caption{Median over the 88 structural connectomes of $\mathcal{S}'(A,P)$ divided by that of $\pi_{LR}$ on the same subject, against proportional edge density, one variant per algorithm and the random baseline, one solution per subject. Below the dashed line at one (shaded) the solution scores better than $\pi_{LR}$. Logarithmic vertical axis.}
    \label{fig:real-ratio-ladder}
\end{figure}

\begin{figure}[ht]
    \centering
    \includegraphics[width=0.95\linewidth]{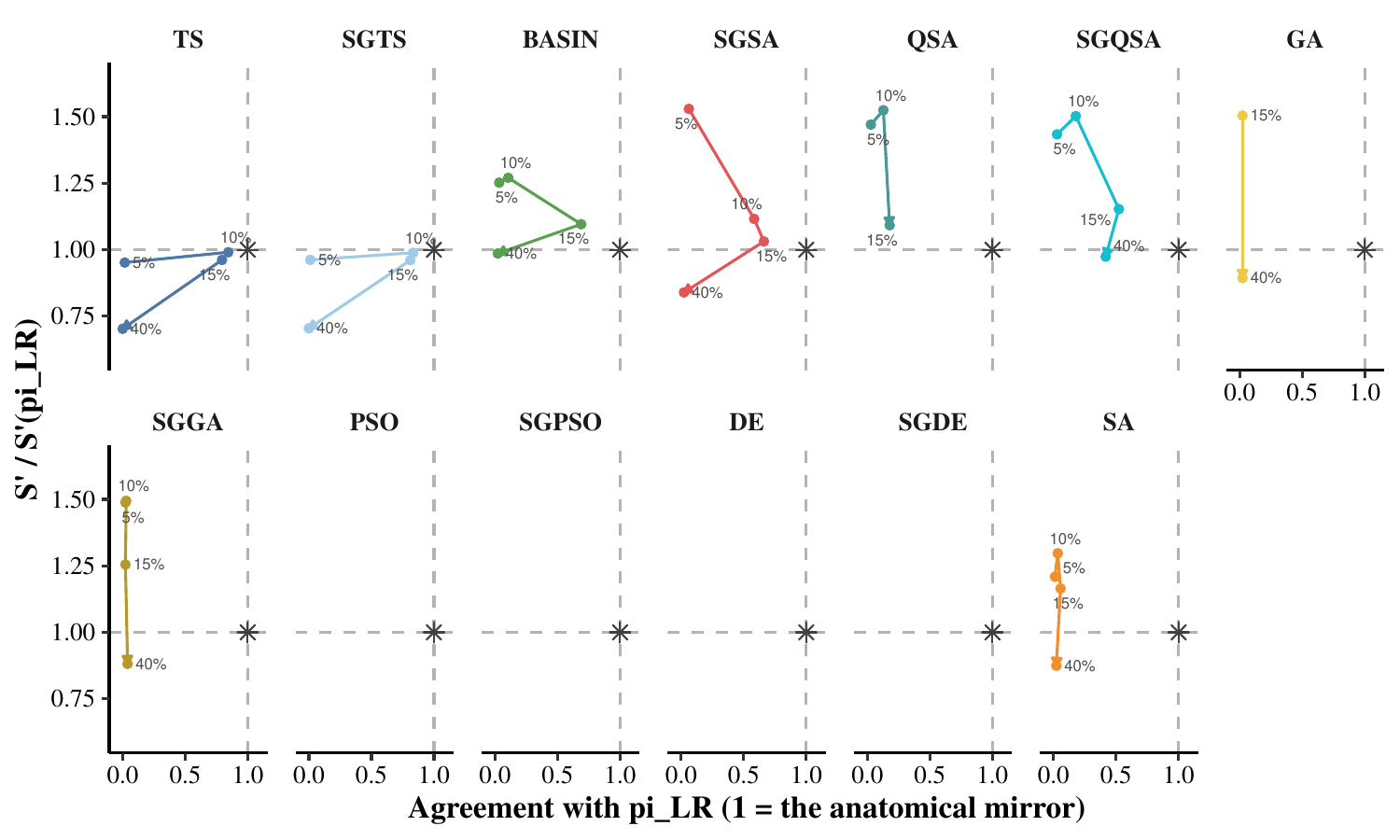}
 \caption{Score against agreement with $\pi_{LR}$ on the 88 structural connectomes, one panel per algorithm and variant, one point per proportional edge density (labeled). The vertical axis is the median $\mathcal{S}'(A,P)/\mathcal{S}'(\pi_{LR})$ and the horizontal axis the median share of the 90 regions sent to their homologue, over subjects, one solution each; the star is $\pi_{LR}$ itself.}
    \label{fig:real-quality-homology}
\end{figure}

\begin{figure}[ht]
    \centering
    \includegraphics[width=1.0\linewidth]{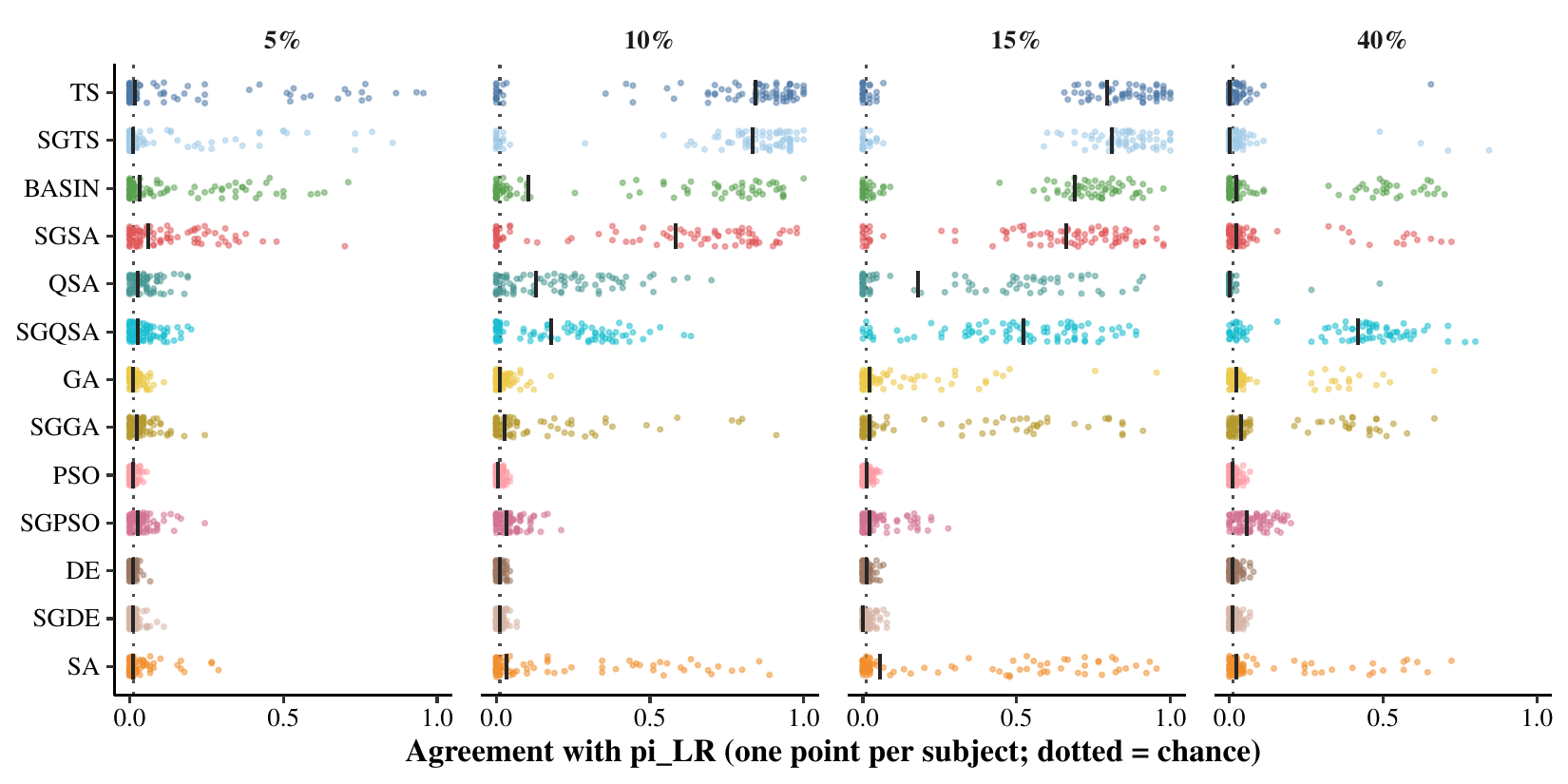}
 \caption{Share of the 90 regions that the solution sends to their contralateral homologue, one point per subject of the 88 structural connectomes, one panel per proportional edge density, every variant of every algorithm. The black bar is the median; dotted is the chance level $1/89$ of sending a region to its homologue.}
    \label{fig:real-homologue-spread}
\end{figure}

\begin{figure}[ht]
    \centering
    \includegraphics[width=1.0\linewidth]{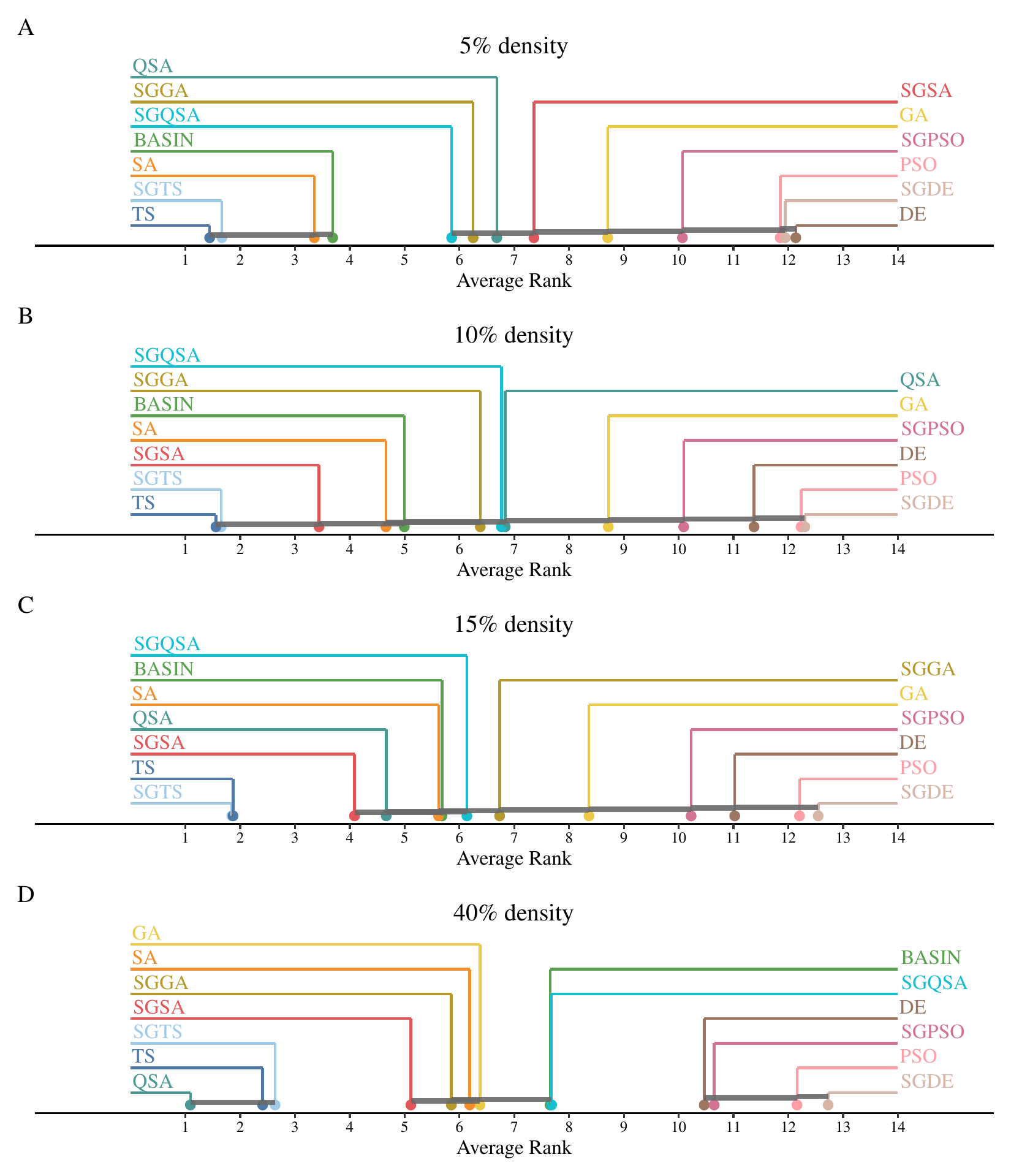}
 \caption{Critical-difference diagrams (Friedman with post-hoc Nemenyi) on the 88 structural connectomes, \textbf{(A)}--\textbf{(D)} at proportional edge densities $5$, $10$, $15$ and $40~\%$, every variant of every algorithm. Each algorithm is placed at its mean rank on $\mathcal{S}'(A,P)$ over the 88 subjects, one solution per subject, rank 1 lowest; algorithms joined by a bar do not differ at the $5~\%$ level.}
    \label{fig:real-cd-all}
\end{figure}

Score distributions: Fig.~\ref{fig:real-sprime}; score relative to $\pi_{LR}$ along the densities: Fig.~\ref{fig:real-ratio-ladder}; score against homologue rate: Fig.~\ref{fig:real-quality-homology}; agreement with $\pi_{LR}$ by subject: Fig.~\ref{fig:real-homologue-spread}; critical differences per density: Fig.~\ref{fig:real-cd-all}.

\clearpage

\subsection{Benchmark networks: additional results}\label{app:benchmarks}

\begingroup\fontsize{8}{10}\selectfont

\begin{longtable}[t]{lllllr}
\caption{\label{tab:bench}Benchmark networks, 20 restarts per configuration, fixed points capped at $n-2$. \emph{Exact rate}: share of restarts returning an automorphism. \emph{Orbit recovery}: mean share of vertices in non-trivial orbits mapped within their own orbit. \emph{Twin distance}: minimum over restarts of the distance to the twin subgroup in vertices, exact for JAZZ. Other columns are medians over restarts.}\\
\toprule
Method & Exact rate & Median $S'$ & Median $F/n$ & Orbit recovery & Twin distance\\
\midrule
\endfirsthead
\caption[]{Benchmark networks, 20 restarts per configuration, fixed points capped at $n-2$. \textit{(continued)}}\\
\toprule
Method & Exact rate & Median $S'$ & Median $F/n$ & Orbit recovery & Twin distance\\
\midrule
\endhead

\endfoot
\bottomrule
\endlastfoot
\addlinespace[0.3em]
\multicolumn{6}{l}{\textbf{JAZZ}}\\
\hspace{1em}TS & 0.80 & 0.00000 & 0.929 & 0.832 & 0\\
\hspace{1em}SGTS & 0.80 & 0.00000 & 0.929 & 0.907 & 0\\
\hspace{1em}SA & 0.70 & 0.00000 & 0.939 & 0.679 & 0\\
\hspace{1em}SGGA & 0.65 & 0.00000 & 0.929 & 0.896 & 0\\
\hspace{1em}GA & 0.35 & 0.09670 & 0.265 & 0.425 & 0\\
\hspace{1em}QSA & 0.00 & 0.00680 & 0.798 & 0.750 & 16\\
\hspace{1em}SGSA & 0.00 & 0.03910 & 0.467 & 0.714 & 60\\
\hspace{1em}SGQSA & 0.00 & 0.06959 & 0.051 & 0.793 & 168\\
\hspace{1em}BASIN & 0.00 & 0.08656 & 0.015 & 0.382 & 187\\
\hspace{1em}SGPSO & 0.00 & 0.14179 & 0.008 & 0.864 & 180\\
\hspace{1em}PSO & 0.00 & 0.18401 & 0.139 & 0.025 & 141\\
\hspace{1em}DE & 0.00 & 0.20595 & 0.101 & 0.011 & 130\\
\hspace{1em}SGDE & 0.00 & 0.20846 & 0.033 & 0.029 & 185\\
\hspace{1em}Random & 0.00 & 0.23072 & 0.010 & 0.007 & 193\\
\addlinespace[0.3em]
\multicolumn{6}{l}{\textbf{PDZ}}\\
\hspace{1em}SGGA & 0.00 & 0.00968 & 0.075 & 0.235 & 79\\
\hspace{1em}SGTS & 0.00 & 0.00997 & 0.056 & 0.233 & 105\\
\hspace{1em}SGQSA & 0.00 & 0.01079 & 0.000 & 0.029 & 157\\
\hspace{1em}QSA & 0.00 & 0.01079 & 0.000 & 0.027 & 148\\
\hspace{1em}BASIN & 0.00 & 0.01335 & 0.000 & 0.001 & 160\\
\hspace{1em}TS & 0.00 & 0.01336 & 0.006 & 0.048 & 142\\
\hspace{1em}GA & 0.00 & 0.01638 & 0.000 & 0.051 & 146\\
\hspace{1em}SGPSO & 0.00 & 0.01865 & 0.025 & 0.576 & 97\\
\hspace{1em}PSO & 0.00 & 0.02426 & 0.118 & 0.067 & 108\\
\hspace{1em}DE & 0.00 & 0.02488 & 0.137 & 0.071 & 118\\
\hspace{1em}SGDE & 0.00 & 0.02583 & 0.043 & 0.052 & 137\\
\hspace{1em}SGSA & 0.00 & 0.02756 & 0.006 & 0.039 & 152\\
\hspace{1em}SA & 0.00 & 0.02873 & 0.009 & 0.035 & 148\\
\hspace{1em}Random & 0.00 & 0.03043 & 0.006 & 0.031 & 153\\
\addlinespace[0.3em]
\multicolumn{6}{l}{\textbf{NETSCIENCE}}\\
\hspace{1em}SGTS & 0.00 & 0.00899 & 0.011 & 0.105 & 346\\
\hspace{1em}SGQSA & 0.00 & 0.00949 & 0.000 & 0.055 & 349\\
\hspace{1em}SGGA & 0.00 & 0.00950 & 0.032 & 0.172 & 316\\
\hspace{1em}QSA & 0.00 & 0.00981 & 0.003 & 0.041 & 353\\
\hspace{1em}TS & 0.00 & 0.01046 & 0.003 & 0.029 & 363\\
\hspace{1em}GA & 0.00 & 0.01263 & 0.001 & 0.016 & 365\\
\hspace{1em}SGPSO & 0.00 & 0.01953 & 0.005 & 0.418 & 275\\
\hspace{1em}PSO & 0.00 & 0.02242 & 0.080 & 0.012 & 321\\
\hspace{1em}DE & 0.00 & 0.02292 & 0.098 & 0.010 & 277\\
\hspace{1em}BASIN & 0.00 & 0.02329 & 0.000 & 0.040 & 369\\
\hspace{1em}SGDE & 0.00 & 0.02378 & 0.024 & 0.014 & 357\\
\hspace{1em}SGSA & 0.00 & 0.02443 & 0.000 & 0.012 & 373\\
\hspace{1em}SA & 0.00 & 0.02463 & 0.003 & 0.005 & 371\\
\hspace{1em}Random & 0.00 & 0.02478 & 0.003 & 0.005 & 375\\*
\end{longtable}
\endgroup{}{}

\begin{figure}[ht]
    \centering
    \includegraphics[width=\linewidth]{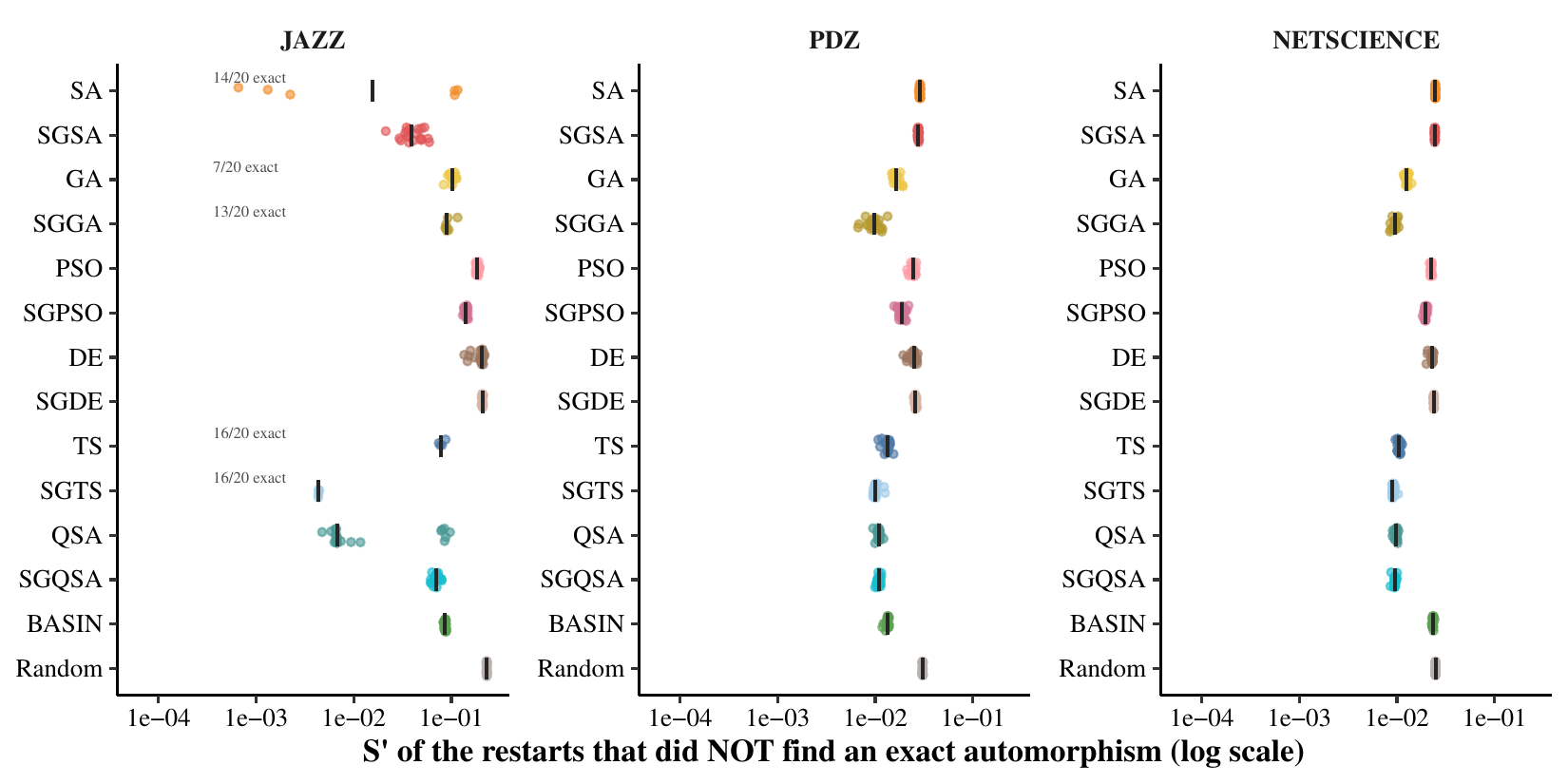}
 \caption{$\mathcal{S}'(A,P)$ on the three benchmark networks, one panel per network, every variant of every algorithm and the random baseline, fixed points uncapped. Each point is one of the 20 restarts that did not return an exact automorphism, the black bar their median; where some did, their count is printed beside the row. Logarithmic axis.}
    \label{fig:bench-sprime}
\end{figure}

\begin{figure}[ht]
    \centering
    \includegraphics[width=\linewidth]{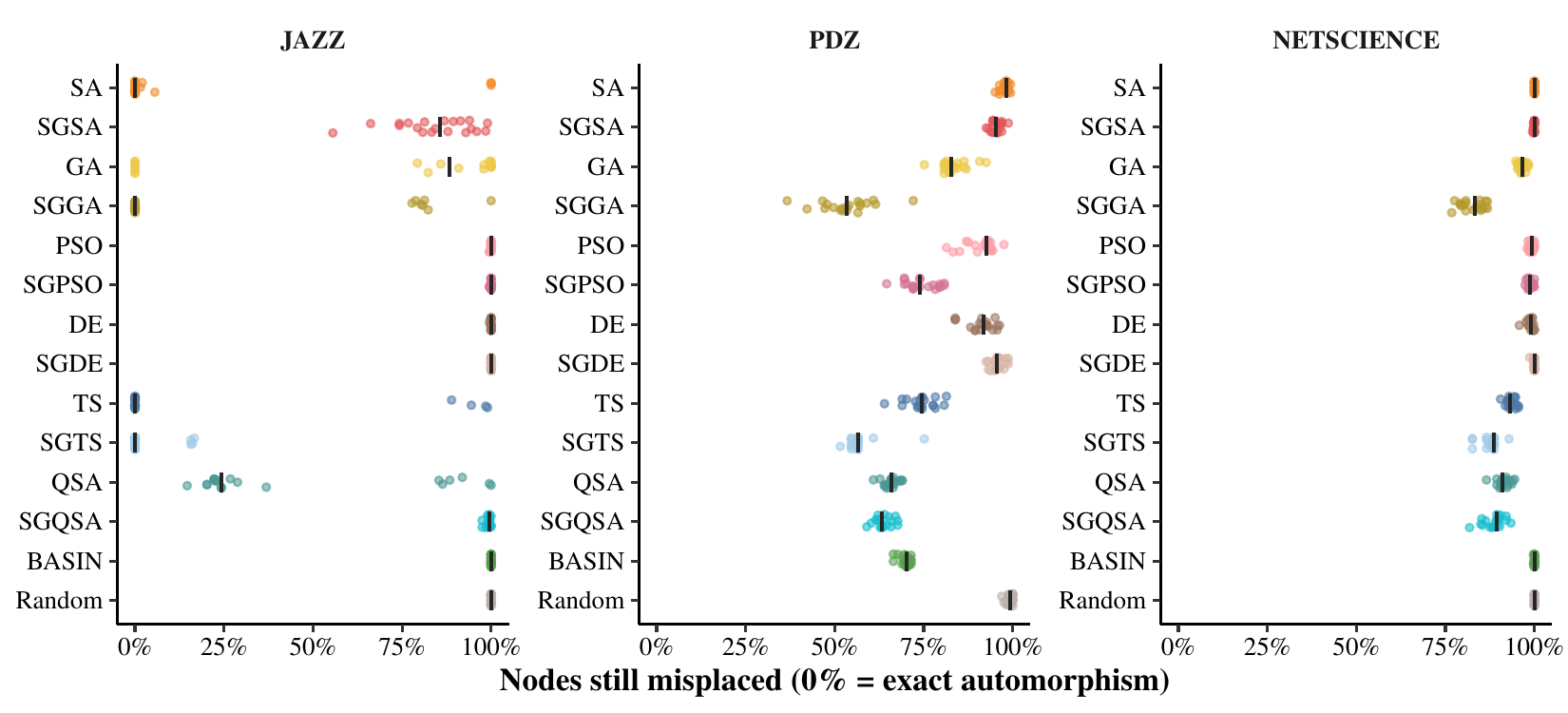}
 \caption{Distance from an exact automorphism on the three benchmark networks, one panel per network, every variant of every algorithm and the random baseline, fixed points uncapped: the share of nodes that must be reassigned to reach the twin subgroup, $0~\%$ being an exact automorphism. One point per restart of 20, the black bar their median.}
    \label{fig:bench-distance}
\end{figure}

\subsection{Fixed points at every condition}\label{app:FP}

\begin{figure}[ht]
    \centering
    \includegraphics[width=\linewidth]{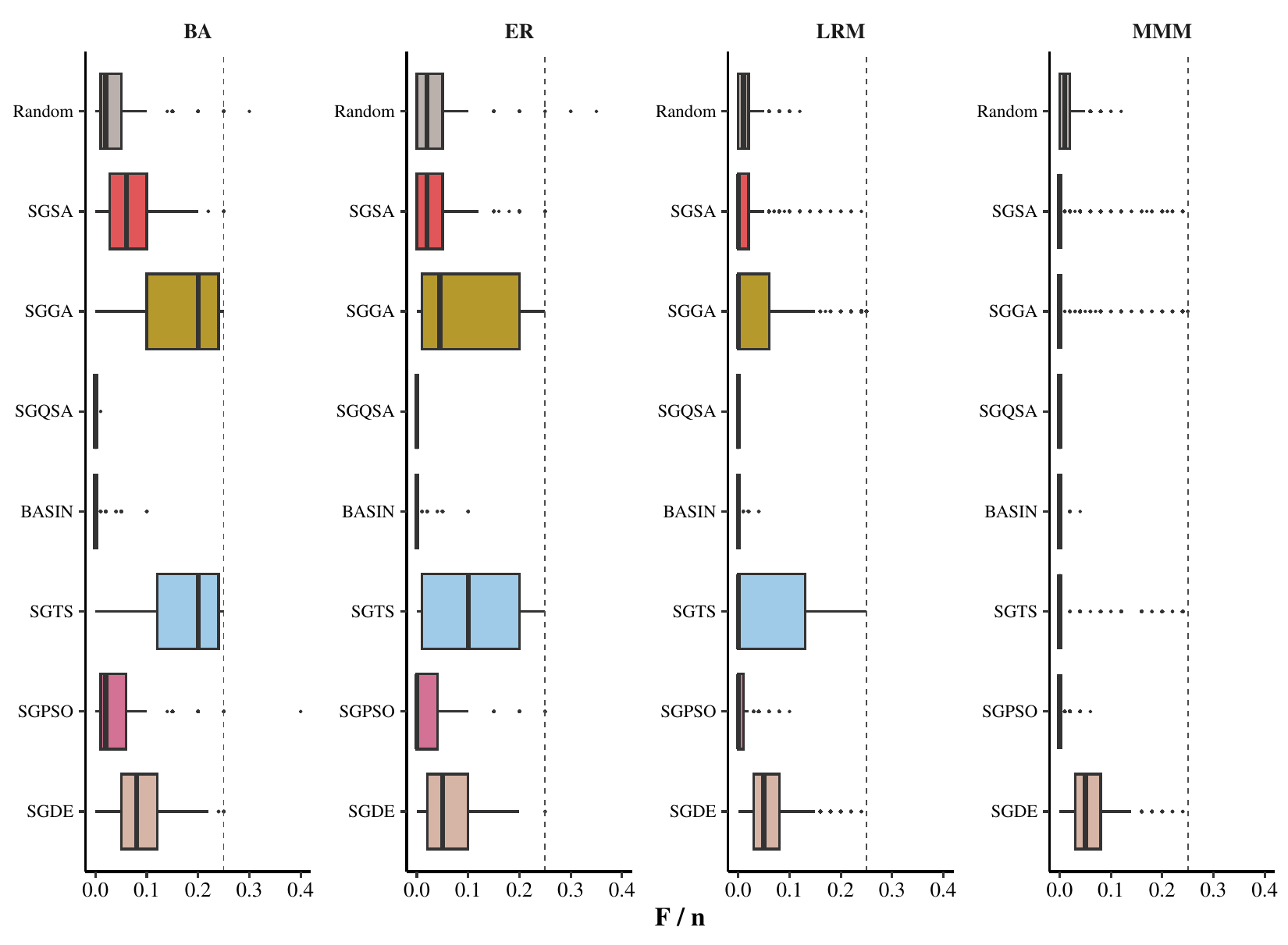}
 \caption{Fixed-point ratio $F/n$ of each graph's solution, pooled over every condition of the testing set, one panel per graph family, one variant per algorithm and the random-search baseline. The dashed line is the likely degeneracy threshold $0.25n$; the hardest condition of each graph family is in Figure~\ref{fig:fp_hard}.}
    \label{fig:fp_all}
\end{figure}

\begin{figure}[ht]
    \centering
    \includegraphics[width=1.0\linewidth]{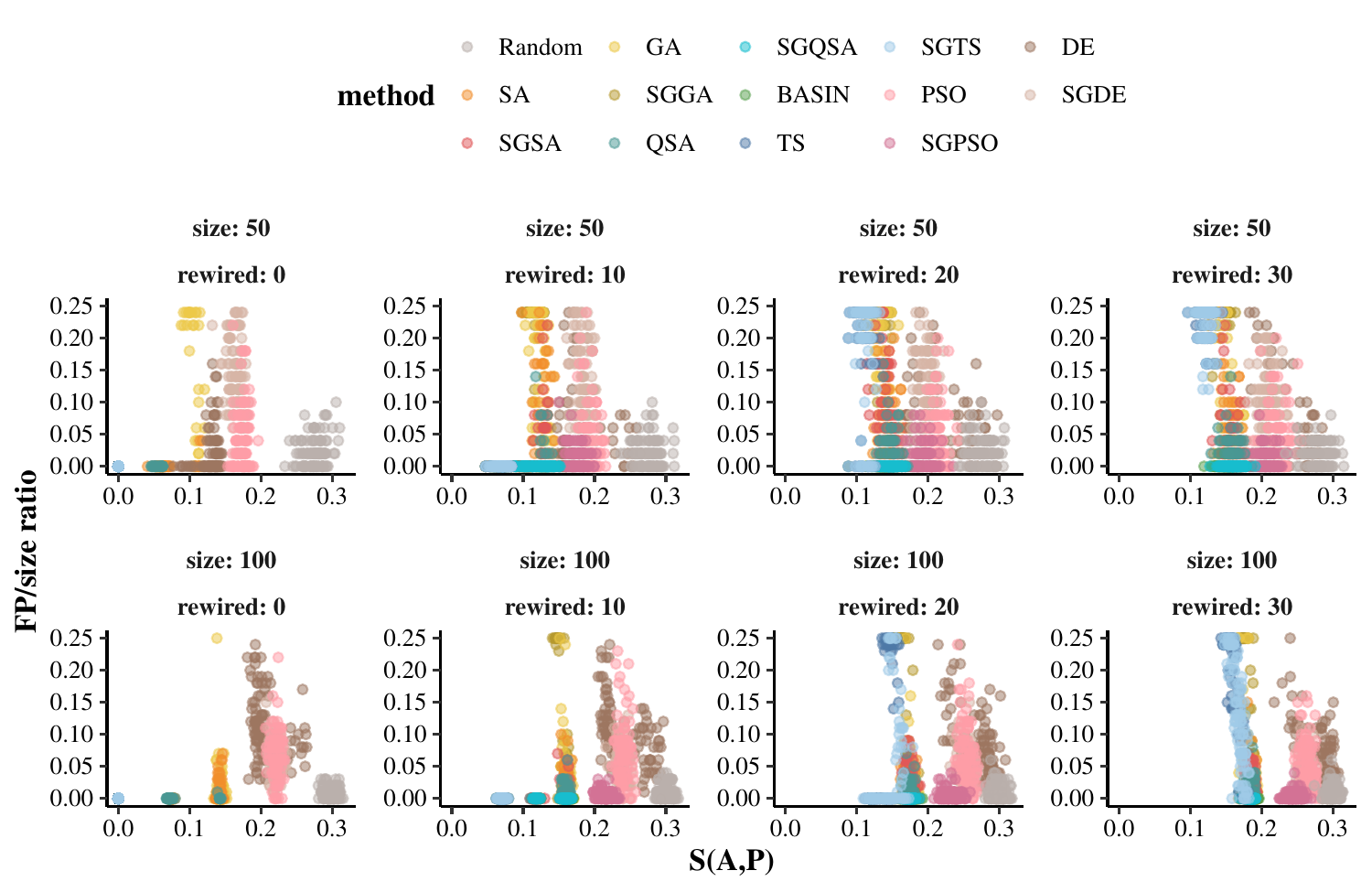}
 \caption{Fixed-point ratio $F/n$ against the unnormalized score $\mathcal{S}(A,P)$, which fixed points lower, on \ac{LRM} graphs at edge density $40~\%$; one point per graph and algorithm, its solution, every variant of every algorithm and the random-search baseline. One panel per graph size and share of rewired edges.}
    \label{fig:fp_unnormalised}
\end{figure}

\subsection{The comparison without the one-variant reduction}\label{app:comparison_without_one-variant_reduction}

\begin{figure}[ht]
    \centering
    \includegraphics[width=0.75\linewidth]{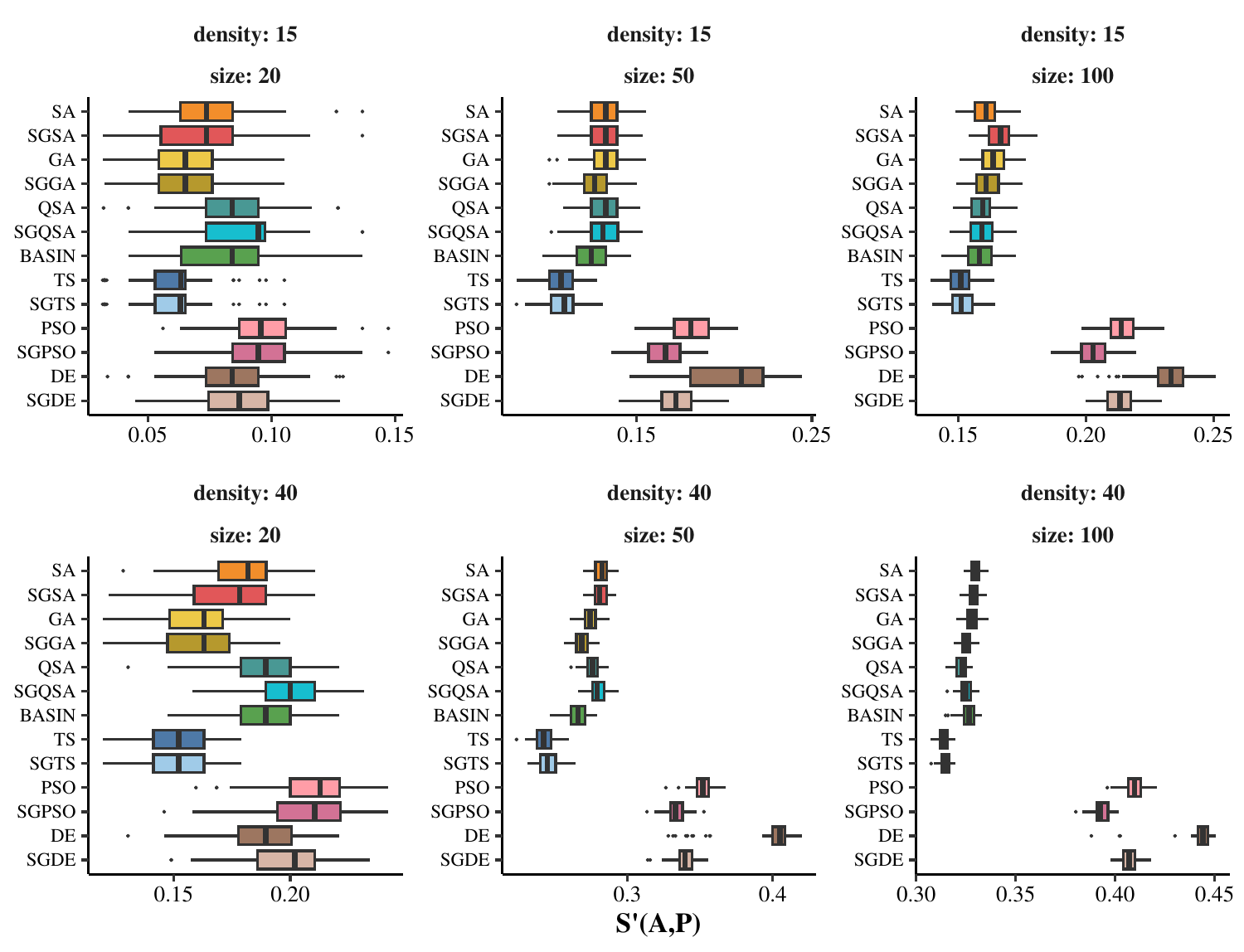}
 \caption{Performance on \ac{ER} graphs, every variant of every algorithm, unguided and similarity-guided. Boxes show $\mathcal{S}'(A,P)$ over the 100 test graphs of each condition, one solution per graph; one panel per graph size $n$ and edge density $d$, each with its own horizontal axis.}
    \label{fig:box_ER_all}
\end{figure}

\begin{figure}[ht]
    \centering
    \includegraphics[width=0.75\linewidth]{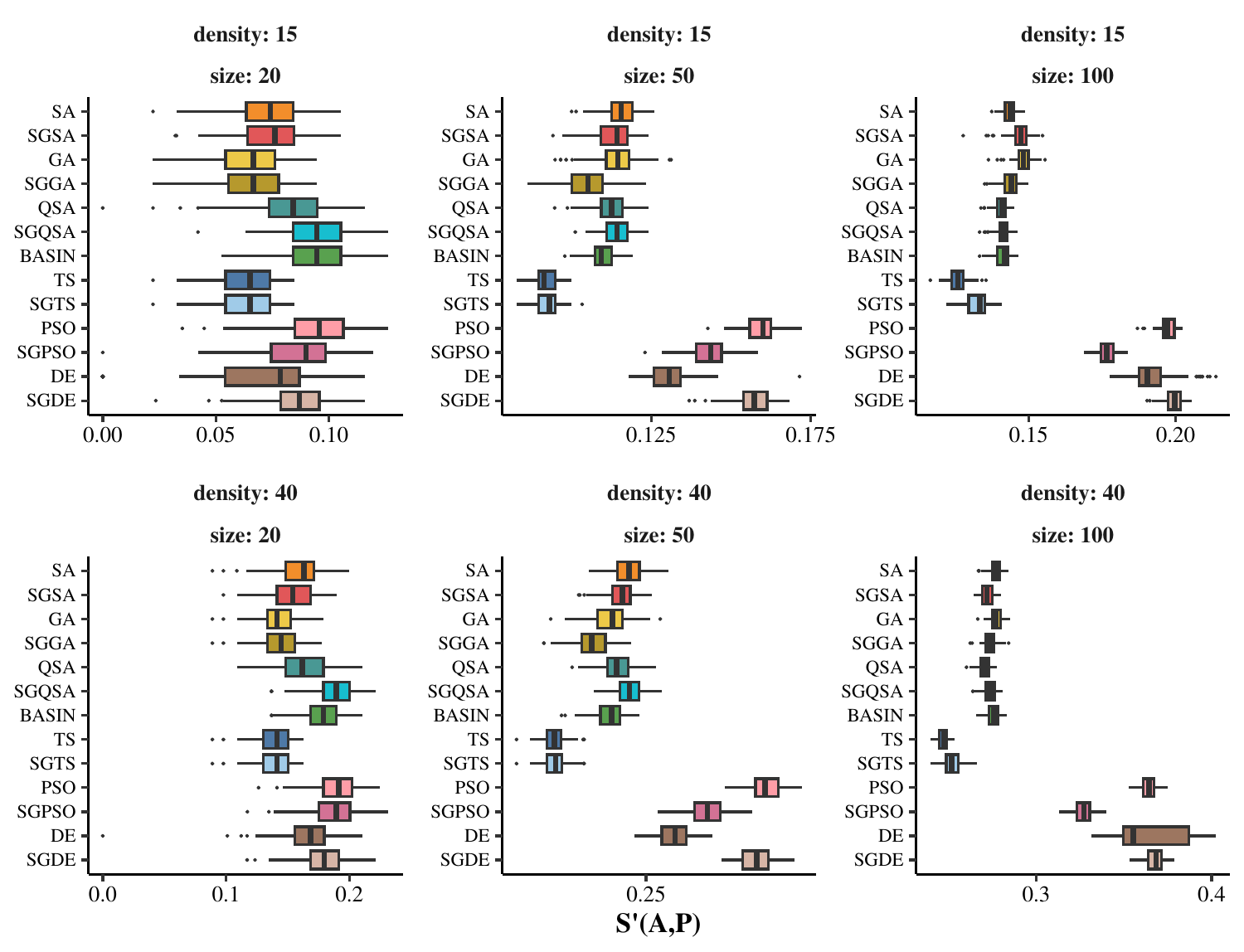}
 \caption{Performance on \ac{BA} graphs, every variant of every algorithm, unguided and similarity-guided. Boxes show $\mathcal{S}'(A,P)$ over the 100 test graphs of each condition, one solution per graph; one panel per graph size $n$ and edge density $d$, each with its own horizontal axis.}
    \label{fig:box_BA_all}
\end{figure}

\begin{figure}[ht]
    \centering
    \includegraphics[width=0.75\linewidth]{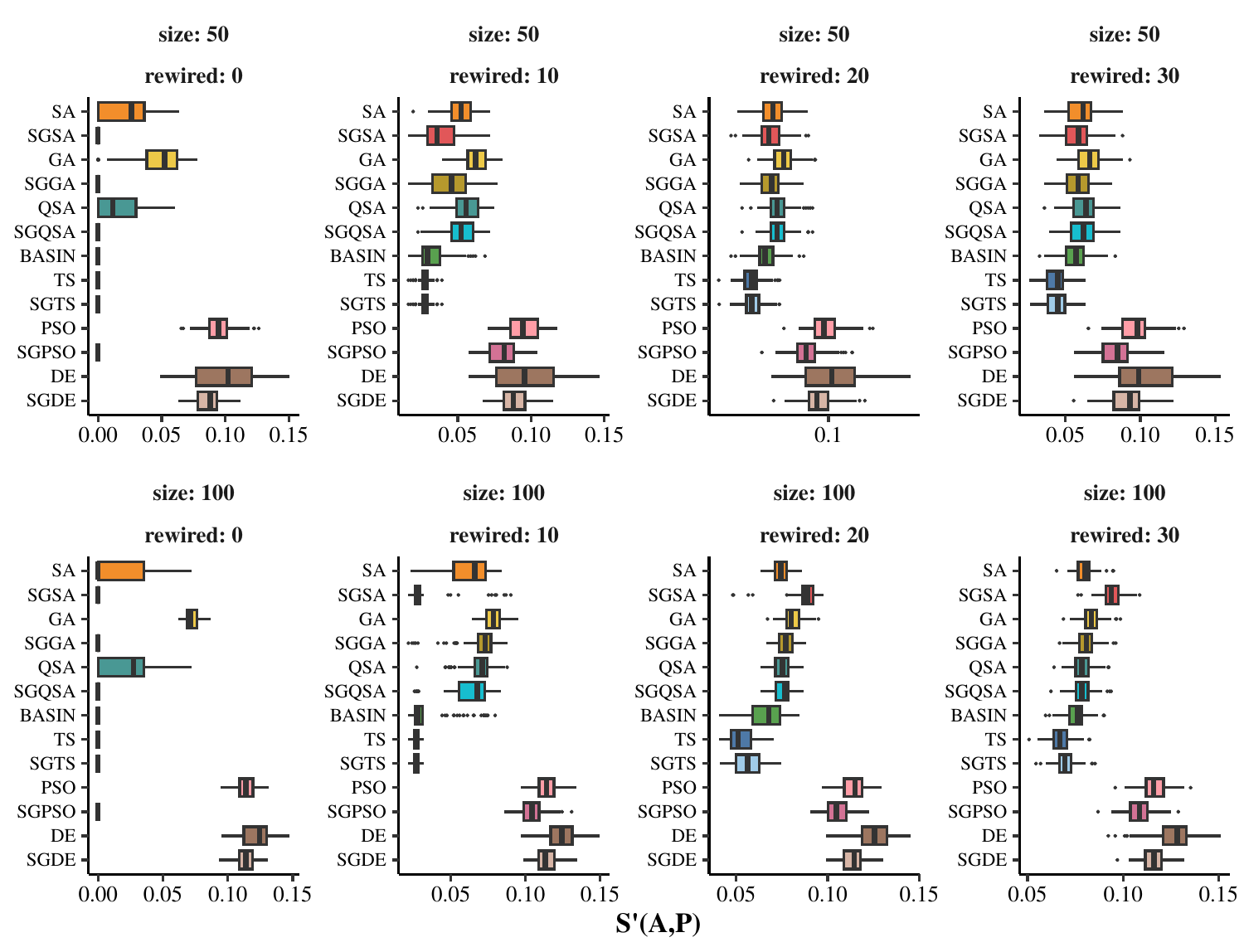}
 \caption{Performance on \ac{LRM} graphs at edge density $15~\%$, every variant of every algorithm, unguided and similarity-guided. Boxes show $\mathcal{S}'(A,P)$ over the 100 test graphs of each condition, one solution per graph; one panel per graph size $n$ and share $\mathit{rew}$ of rewired edges, each with its own horizontal axis.}
    \label{fig:box_LRM15_all}
\end{figure}

\begin{figure}[ht]
    \centering
    \includegraphics[width=0.75\linewidth]{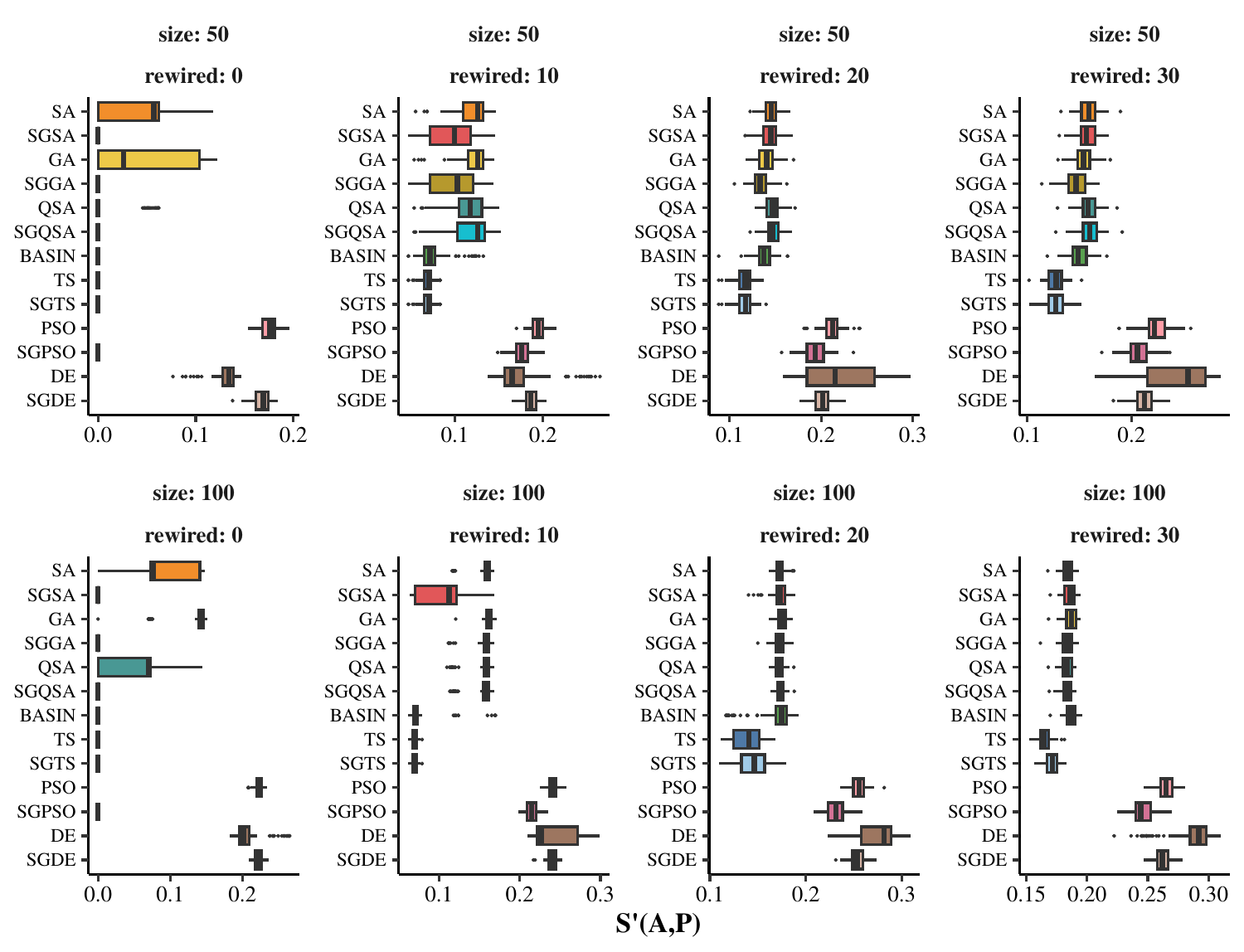}
 \caption{Performance on \ac{LRM} graphs at edge density $40~\%$, every variant of every algorithm, unguided and similarity-guided. Boxes show $\mathcal{S}'(A,P)$ over the 100 test graphs of each condition, one solution per graph; one panel per graph size $n$ and share $\mathit{rew}$ of rewired edges, each with its own horizontal axis.}
    \label{fig:box_LRM40_all}
\end{figure}

\begin{figure}[ht]
    \centering
    \includegraphics[width=0.75\linewidth]{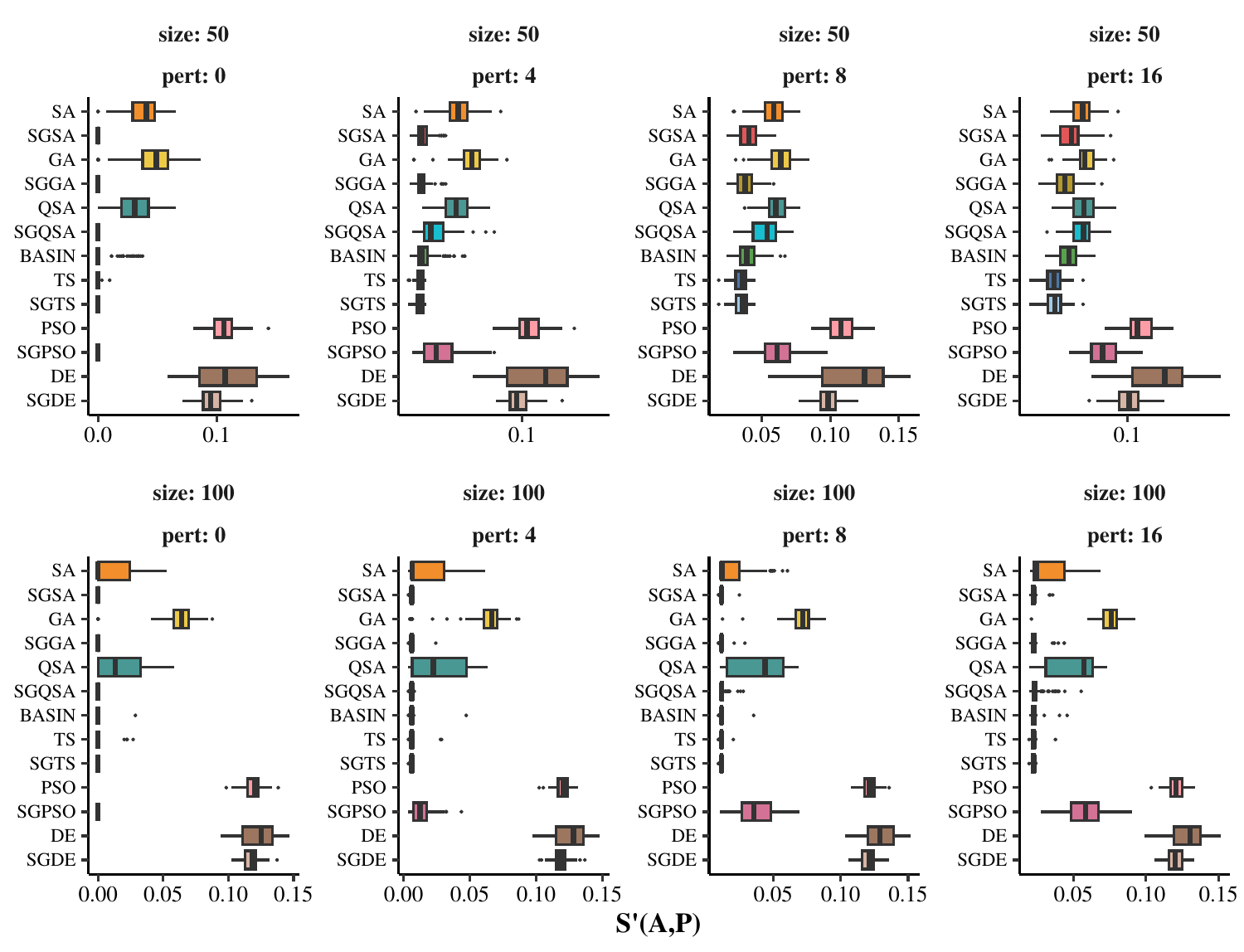}
 \caption{Performance on \ac{MMM} graphs at edge density $15~\%$, every variant of every algorithm, unguided and similarity-guided. Boxes show $\mathcal{S}'(A,P)$ over the 100 test graphs of each condition, one solution per graph; one panel per graph size $n$ and number $\mathit{pert}$ of perturbing edge swaps, each with its own horizontal axis.}
    \label{fig:box_MMM15_all}
\end{figure}

\begin{figure}[ht]
    \centering
    \includegraphics[width=0.75\linewidth]{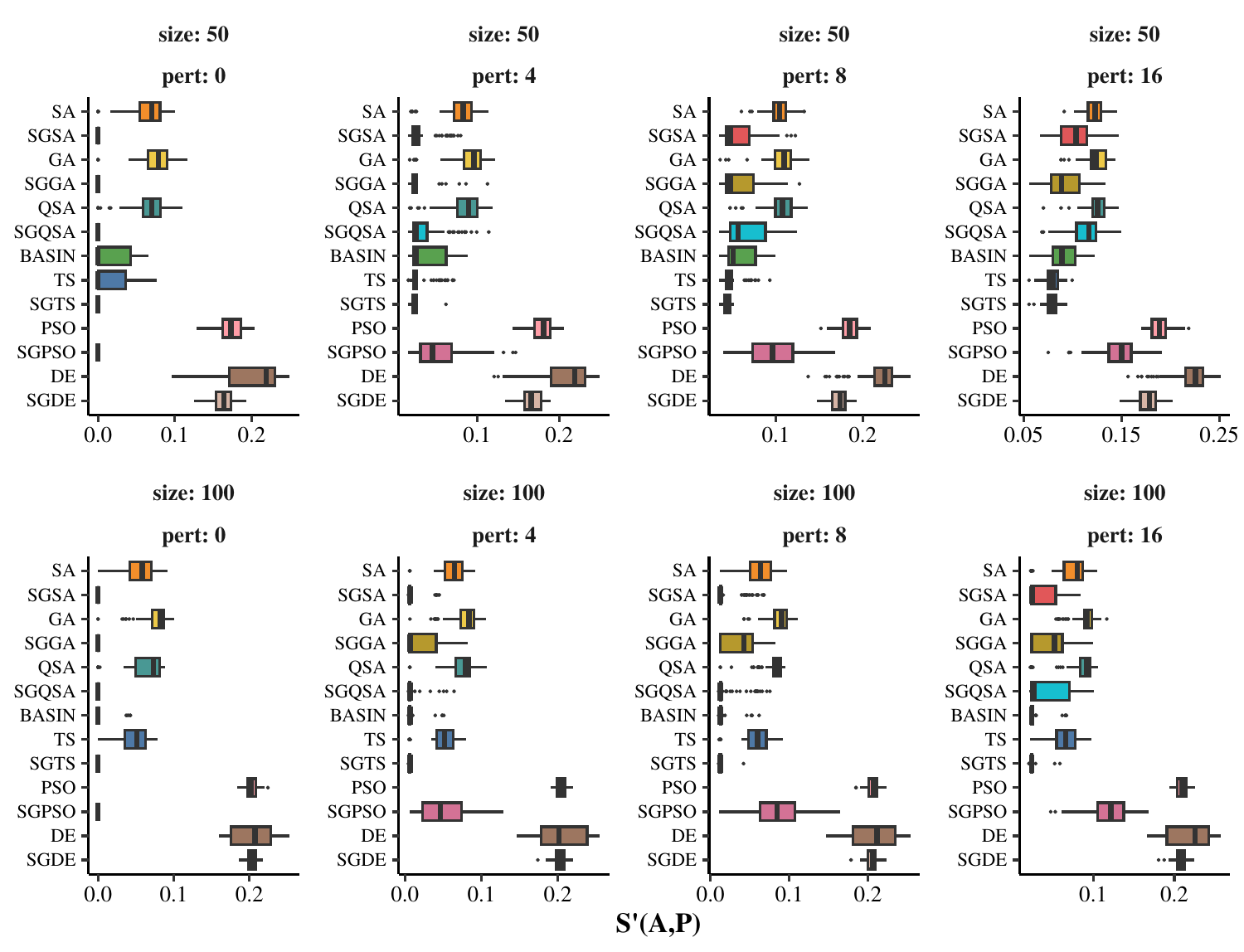}
 \caption{Performance on \ac{MMM} graphs at edge density $30~\%$, every variant of every algorithm, unguided and similarity-guided. Boxes show $\mathcal{S}'(A,P)$ over the 100 test graphs of each condition, one solution per graph; one panel per graph size $n$ and number $\mathit{pert}$ of perturbing edge swaps, each with its own horizontal axis.}
    \label{fig:box_MMM30_all}
\end{figure}
\clearpage

\subsection{Critical difference diagrams at every condition}\label{app:critical_difference}

\begin{figure}[ht]
    \centering
    \includegraphics[width=0.75\linewidth]{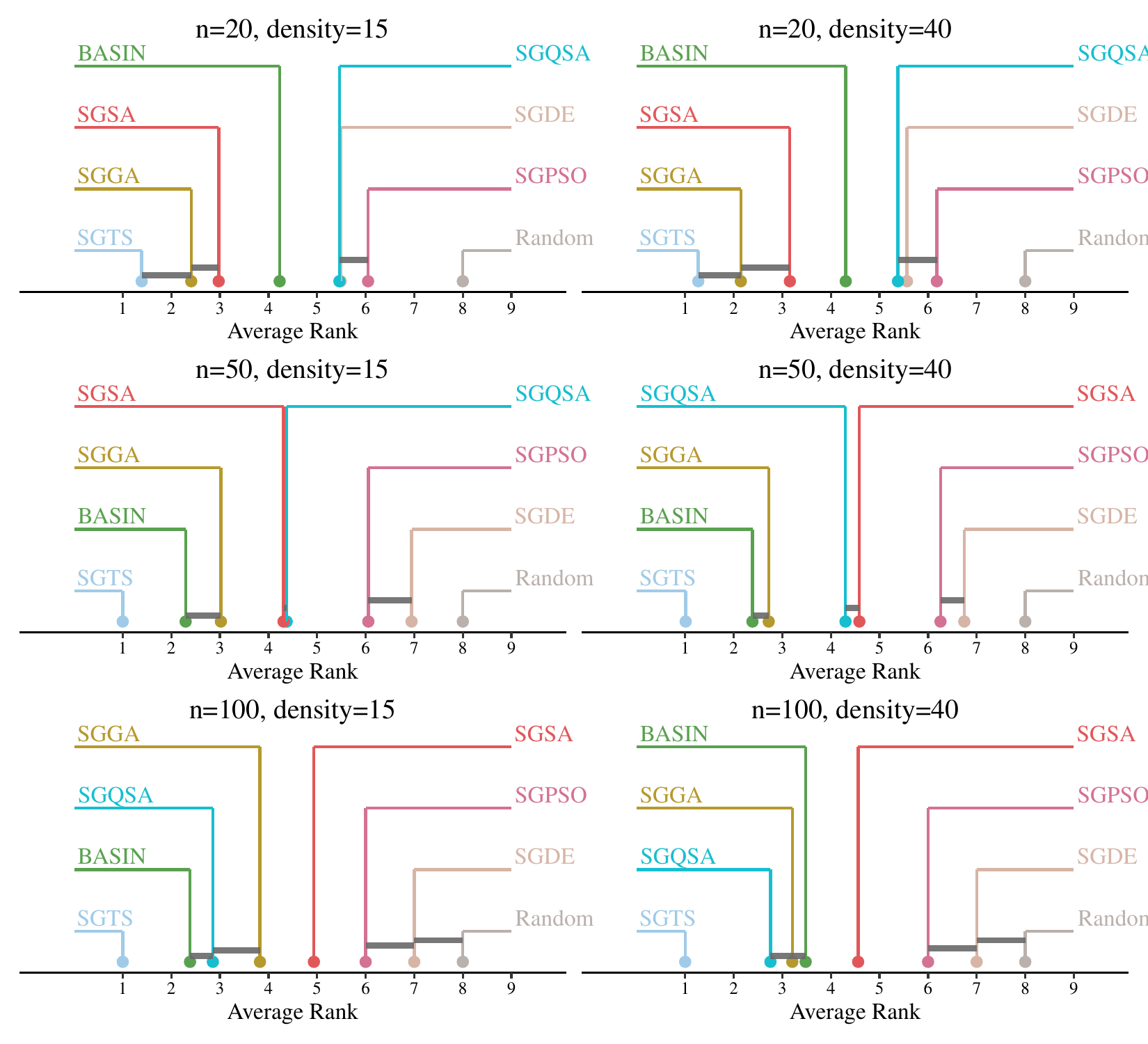}
 \caption{Critical-difference diagrams (Friedman with post-hoc Nemenyi) on \ac{ER} graphs, one panel per graph size $n$ and edge density $d$, one variant per algorithm and the random-search baseline. Each algorithm is placed at its mean rank on $\mathcal{S}'(A,P)$ over the 100 test graphs of the condition, rank 1 lowest; algorithms joined by a bar do not differ at the $5~\%$ level.}
    \label{fig:CD_ER}
\end{figure}

\begin{figure}[ht]
    \centering
    \includegraphics[width=0.75\linewidth]{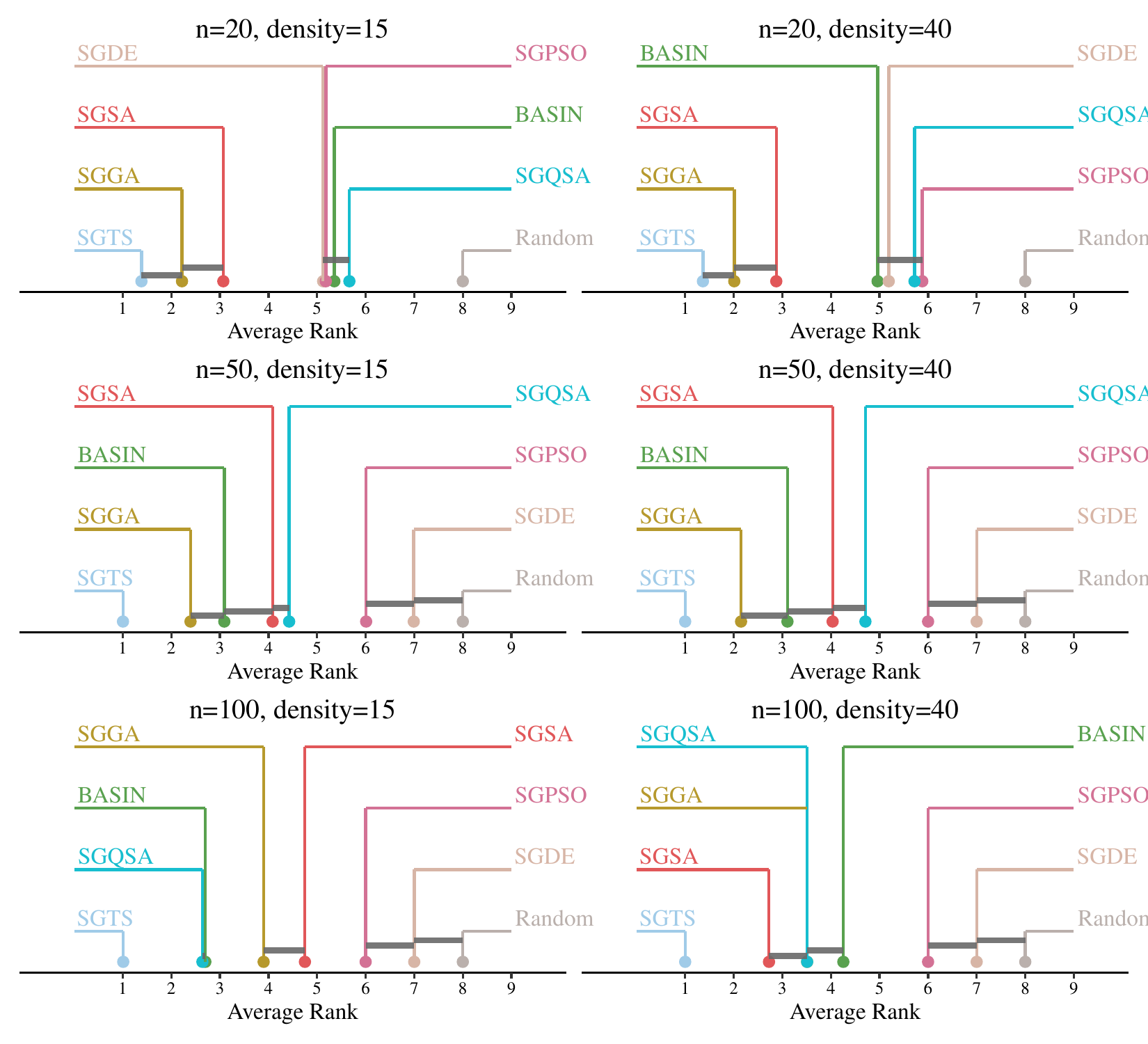}
 \caption{Critical-difference diagrams (Friedman with post-hoc Nemenyi) on \ac{BA} graphs, one panel per graph size $n$ and edge density $d$, one variant per algorithm and the random-search baseline. Each algorithm is placed at its mean rank on $\mathcal{S}'(A,P)$ over the 100 test graphs of the condition, rank 1 lowest; algorithms joined by a bar do not differ at the $5~\%$ level.}
    \label{fig:CD_BA}
\end{figure}

\begin{figure}[ht]
    \centering
    \includegraphics[width=0.75\linewidth]{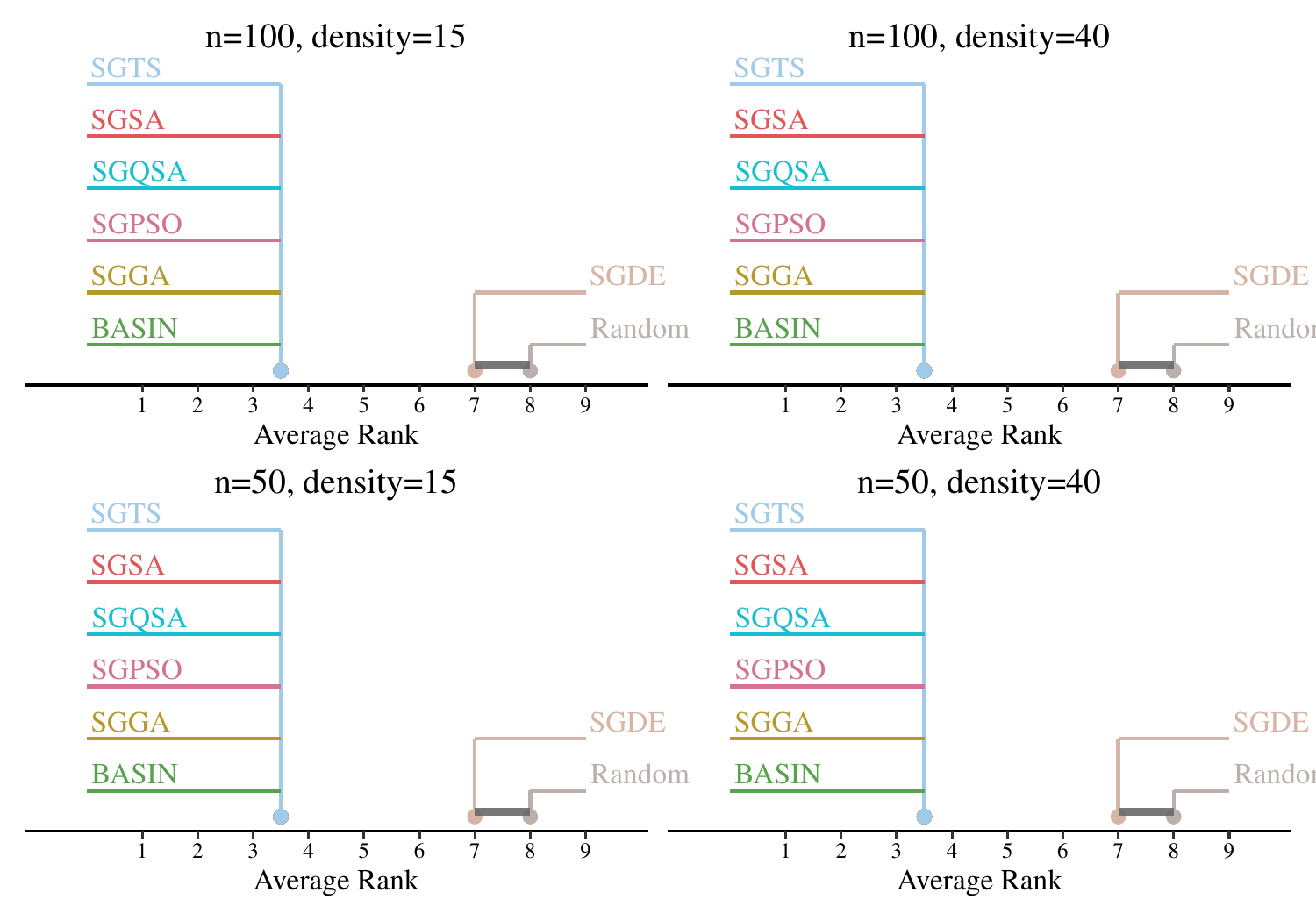}
 \caption{Critical-difference diagrams (Friedman with post-hoc Nemenyi) on \ac{LRM} graphs with $\mathit{rew}=0~\%$ of edges rewired, one panel per graph size $n$ and edge density $d$, one variant per algorithm and the random-search baseline. Each algorithm is placed at its mean rank on $\mathcal{S}'(A,P)$ over the 100 test graphs of the condition, rank 1 lowest; algorithms joined by a bar do not differ at the $5~\%$ level.}
    \label{fig:CD_LRM_rew0}
\end{figure}

\begin{figure}[ht]
    \centering
    \includegraphics[width=0.75\linewidth]{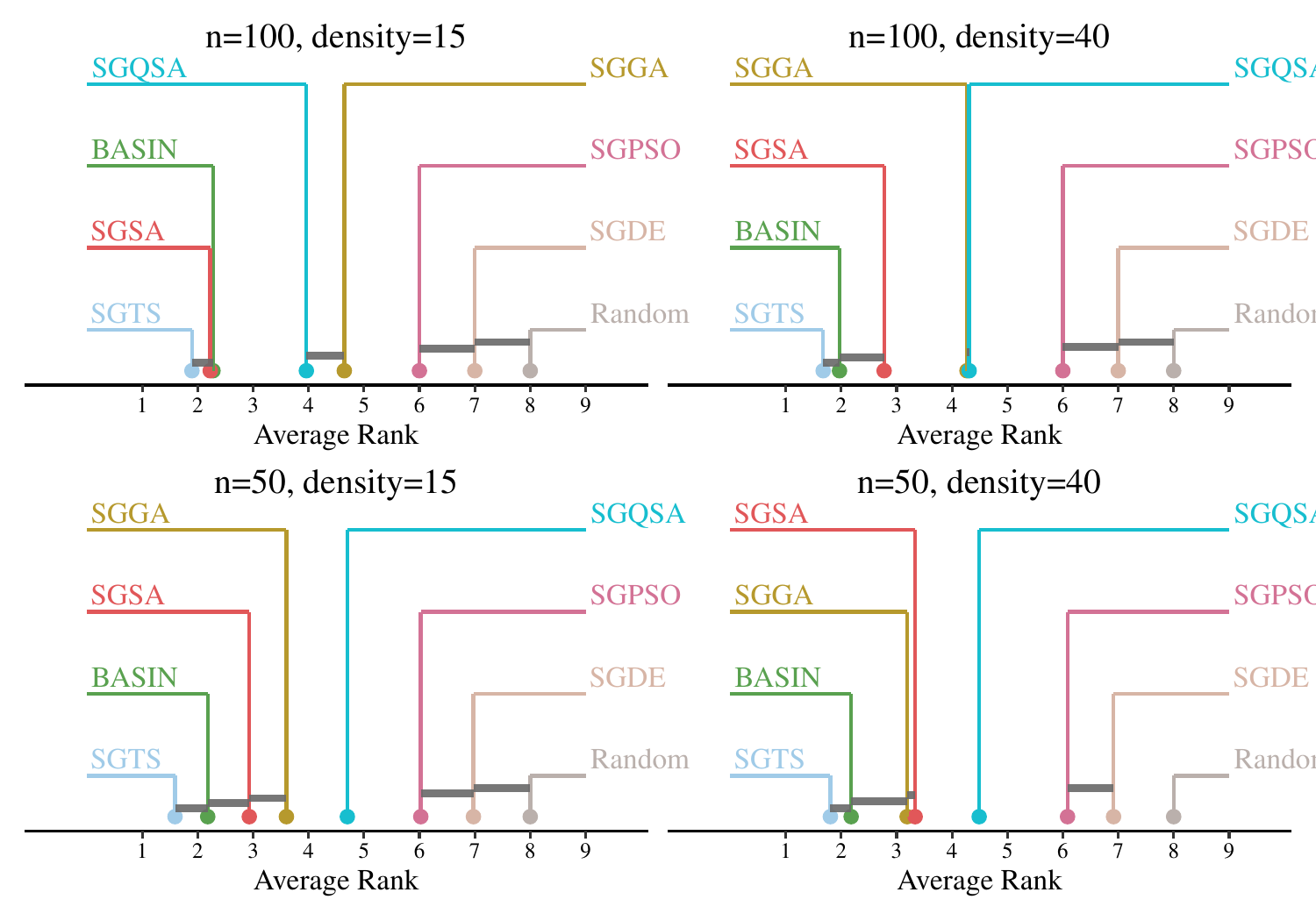}
 \caption{Critical-difference diagrams (Friedman with post-hoc Nemenyi) on \ac{LRM} graphs with $\mathit{rew}=10~\%$ of edges rewired, one panel per graph size $n$ and edge density $d$, one variant per algorithm and the random-search baseline. Each algorithm is placed at its mean rank on $\mathcal{S}'(A,P)$ over the 100 test graphs of the condition, rank 1 lowest; algorithms joined by a bar do not differ at the $5~\%$ level.}
    \label{fig:CD_LRM_rew10}
\end{figure}

\begin{figure}[ht]
    \centering
    \includegraphics[width=0.75\linewidth]{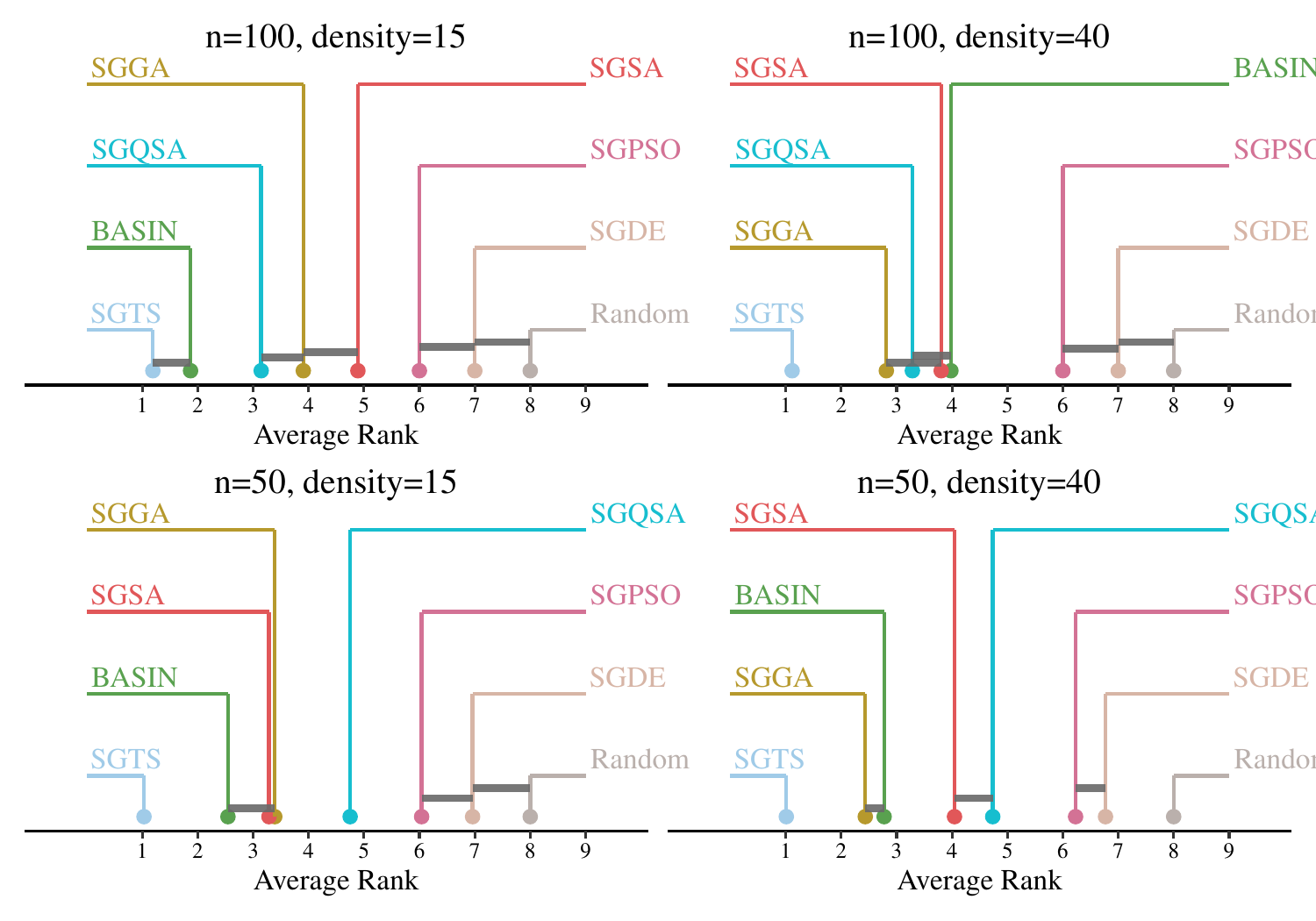}
 \caption{Critical-difference diagrams (Friedman with post-hoc Nemenyi) on \ac{LRM} graphs with $\mathit{rew}=20~\%$ of edges rewired, one panel per graph size $n$ and edge density $d$, one variant per algorithm and the random-search baseline. Each algorithm is placed at its mean rank on $\mathcal{S}'(A,P)$ over the 100 test graphs of the condition, rank 1 lowest; algorithms joined by a bar do not differ at the $5~\%$ level.}
    \label{fig:CD_LRM_rew20}
\end{figure}

\begin{figure}[ht]
    \centering
    \includegraphics[width=0.75\linewidth]{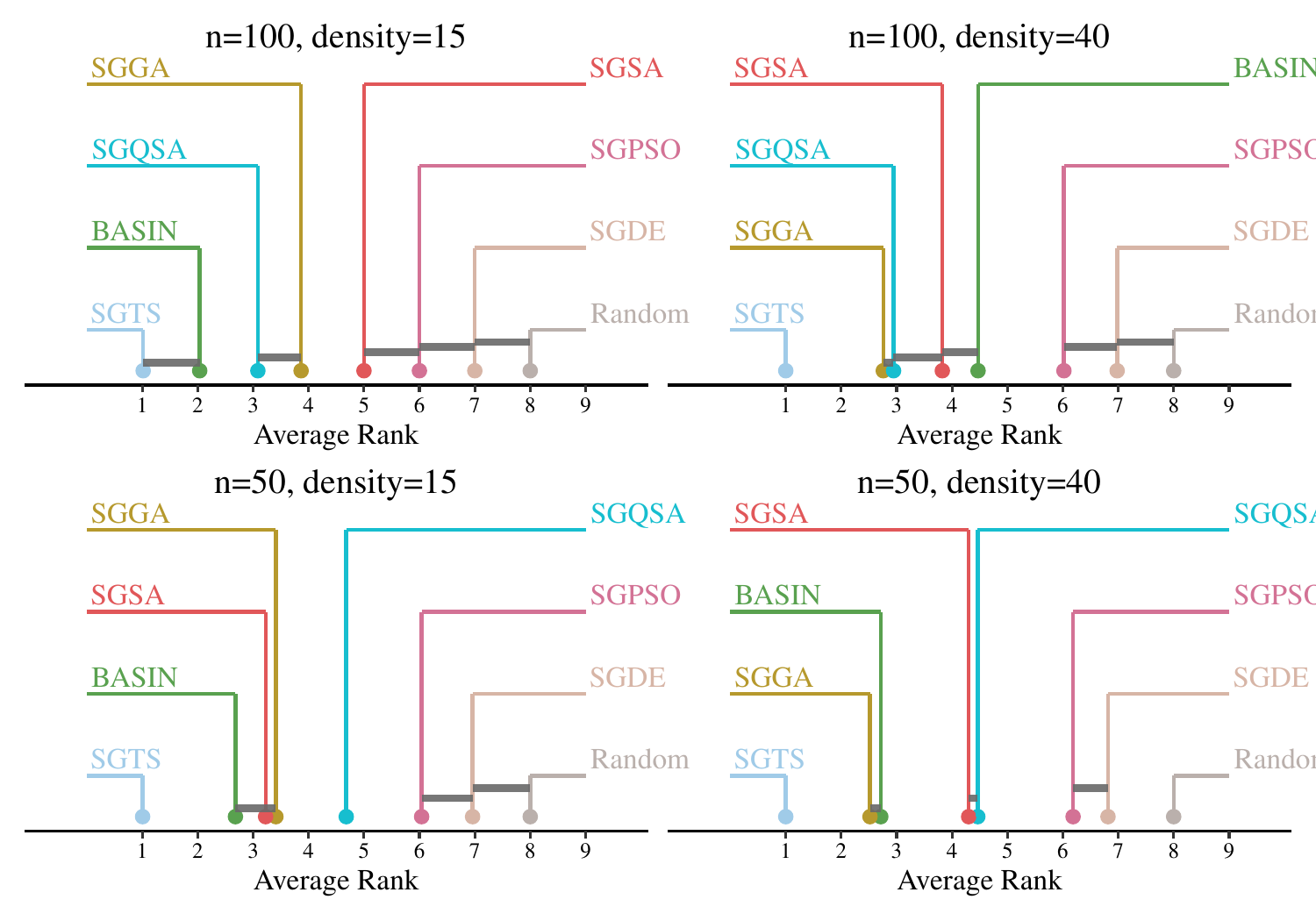}
 \caption{Critical-difference diagrams (Friedman with post-hoc Nemenyi) on \ac{LRM} graphs with $\mathit{rew}=30~\%$ of edges rewired, one panel per graph size $n$ and edge density $d$, one variant per algorithm and the random-search baseline. Each algorithm is placed at its mean rank on $\mathcal{S}'(A,P)$ over the 100 test graphs of the condition, rank 1 lowest; algorithms joined by a bar do not differ at the $5~\%$ level.}
    \label{fig:CD_LRM_rew30}
\end{figure}

\begin{figure}[ht]
    \centering
    \includegraphics[width=0.75\linewidth]{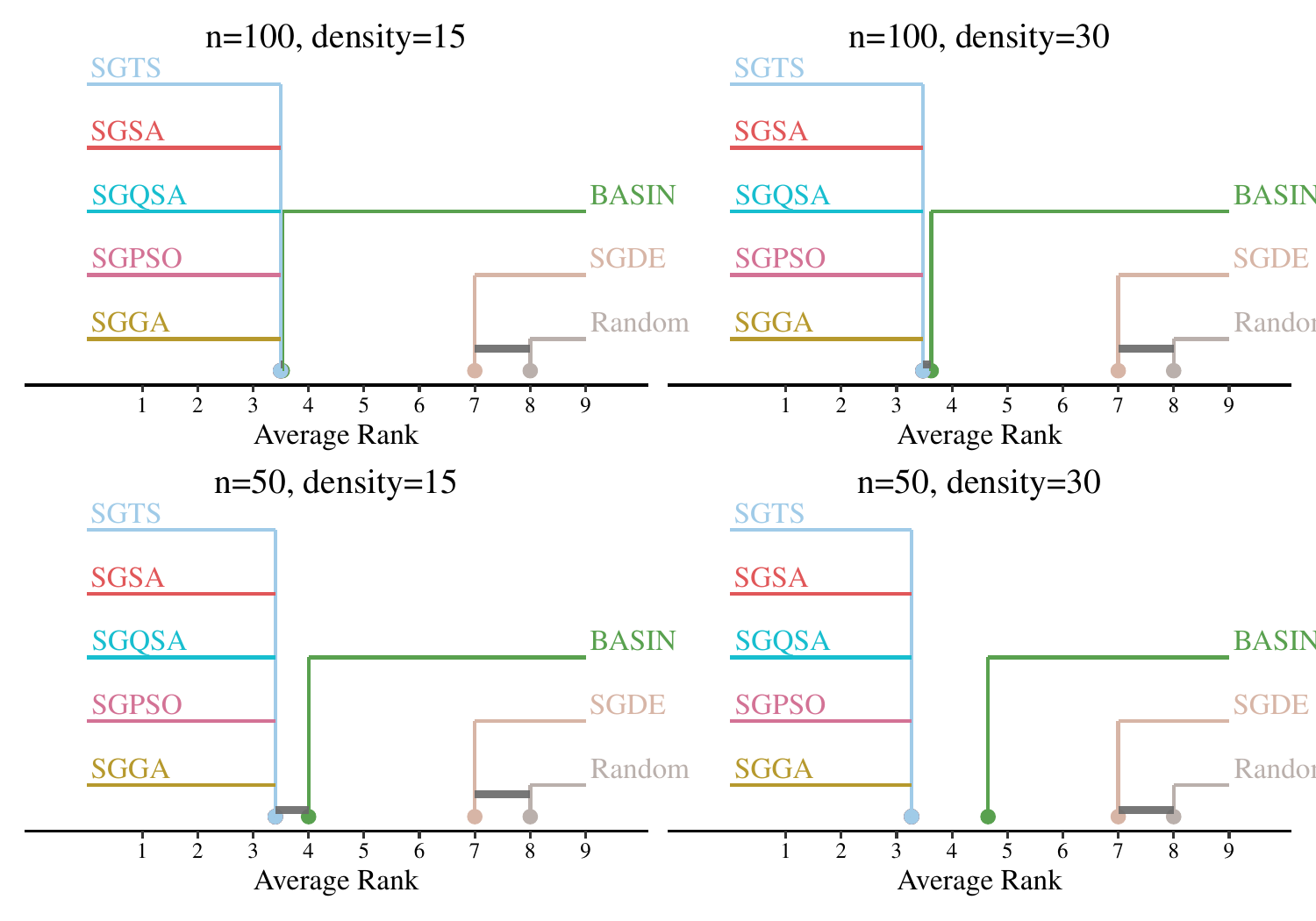}
 \caption{Critical-difference diagrams (Friedman with post-hoc Nemenyi) on \ac{MMM} graphs with $\mathit{pert}=0$ perturbing edge swaps, one panel per graph size $n$ and edge density $d$, one variant per algorithm and the random-search baseline. Each algorithm is placed at its mean rank on $\mathcal{S}'(A,P)$ over the 100 test graphs of the condition, rank 1 lowest; algorithms joined by a bar do not differ at the $5~\%$ level.}
    \label{fig:CD_MMM_pert0}
\end{figure}

\begin{figure}[ht]
    \centering
    \includegraphics[width=0.75\linewidth]{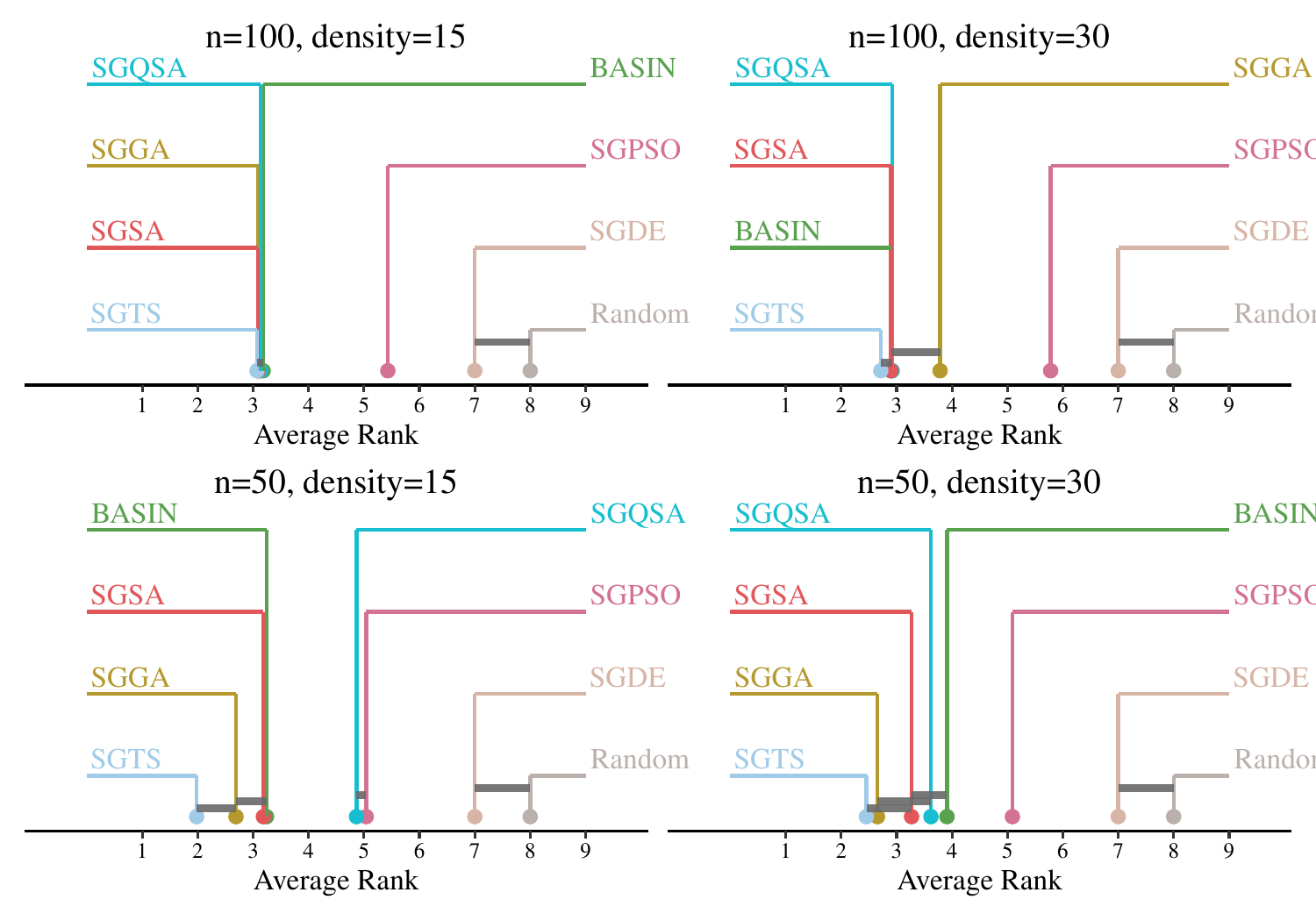}
 \caption{Critical-difference diagrams (Friedman with post-hoc Nemenyi) on \ac{MMM} graphs with $\mathit{pert}=4$ perturbing edge swaps, one panel per graph size $n$ and edge density $d$, one variant per algorithm and the random-search baseline. Each algorithm is placed at its mean rank on $\mathcal{S}'(A,P)$ over the 100 test graphs of the condition, rank 1 lowest; algorithms joined by a bar do not differ at the $5~\%$ level.}
    \label{fig:CD_MMM_pert4}
\end{figure}

\begin{figure}[ht]
    \centering
    \includegraphics[width=0.75\linewidth]{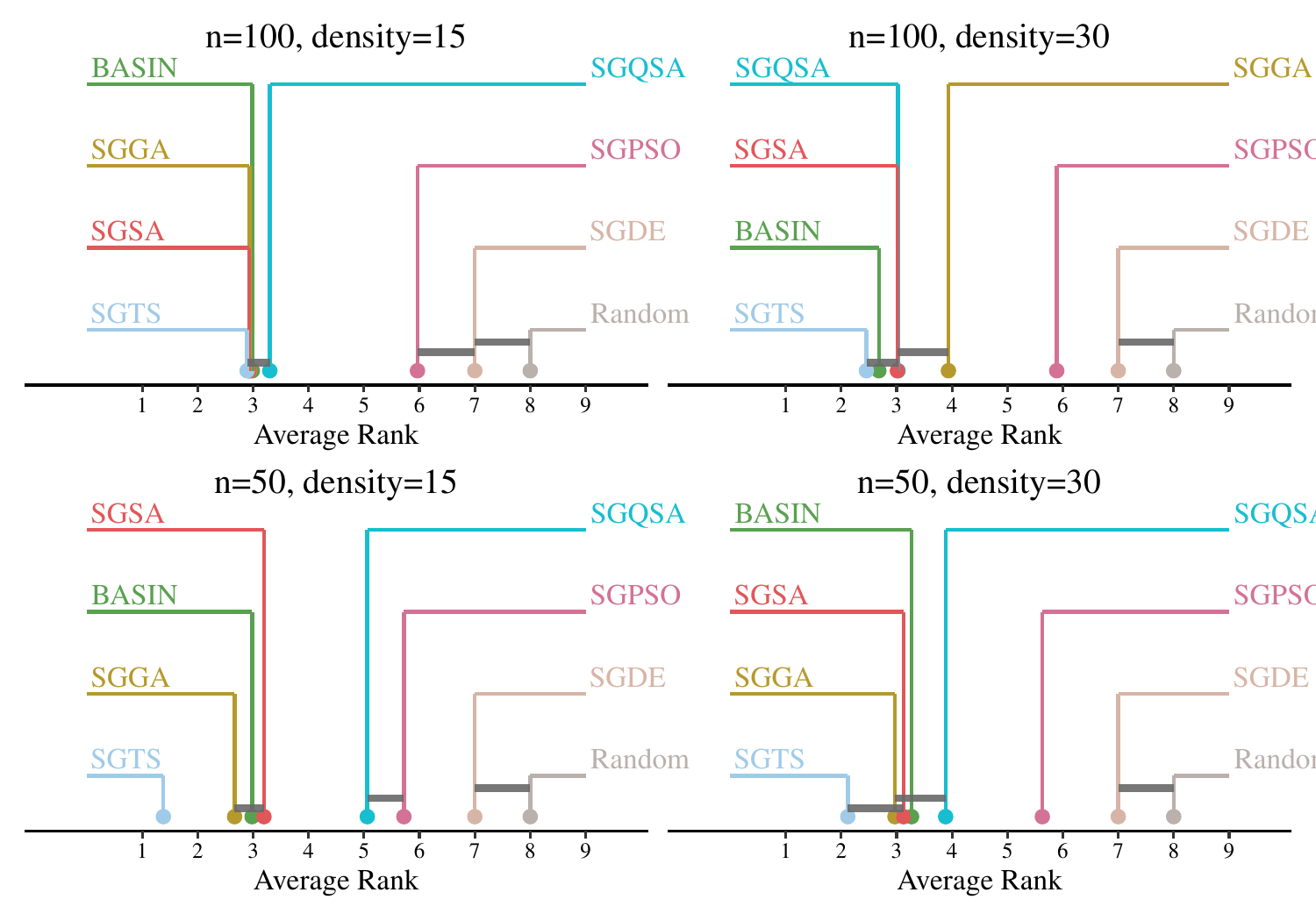}
 \caption{Critical-difference diagrams (Friedman with post-hoc Nemenyi) on \ac{MMM} graphs with $\mathit{pert}=8$ perturbing edge swaps, one panel per graph size $n$ and edge density $d$, one variant per algorithm and the random-search baseline. Each algorithm is placed at its mean rank on $\mathcal{S}'(A,P)$ over the 100 test graphs of the condition, rank 1 lowest; algorithms joined by a bar do not differ at the $5~\%$ level.}
    \label{fig:CD_MMM_pert8}
\end{figure}

\begin{figure}[ht]
    \centering
    \includegraphics[width=0.75\linewidth]{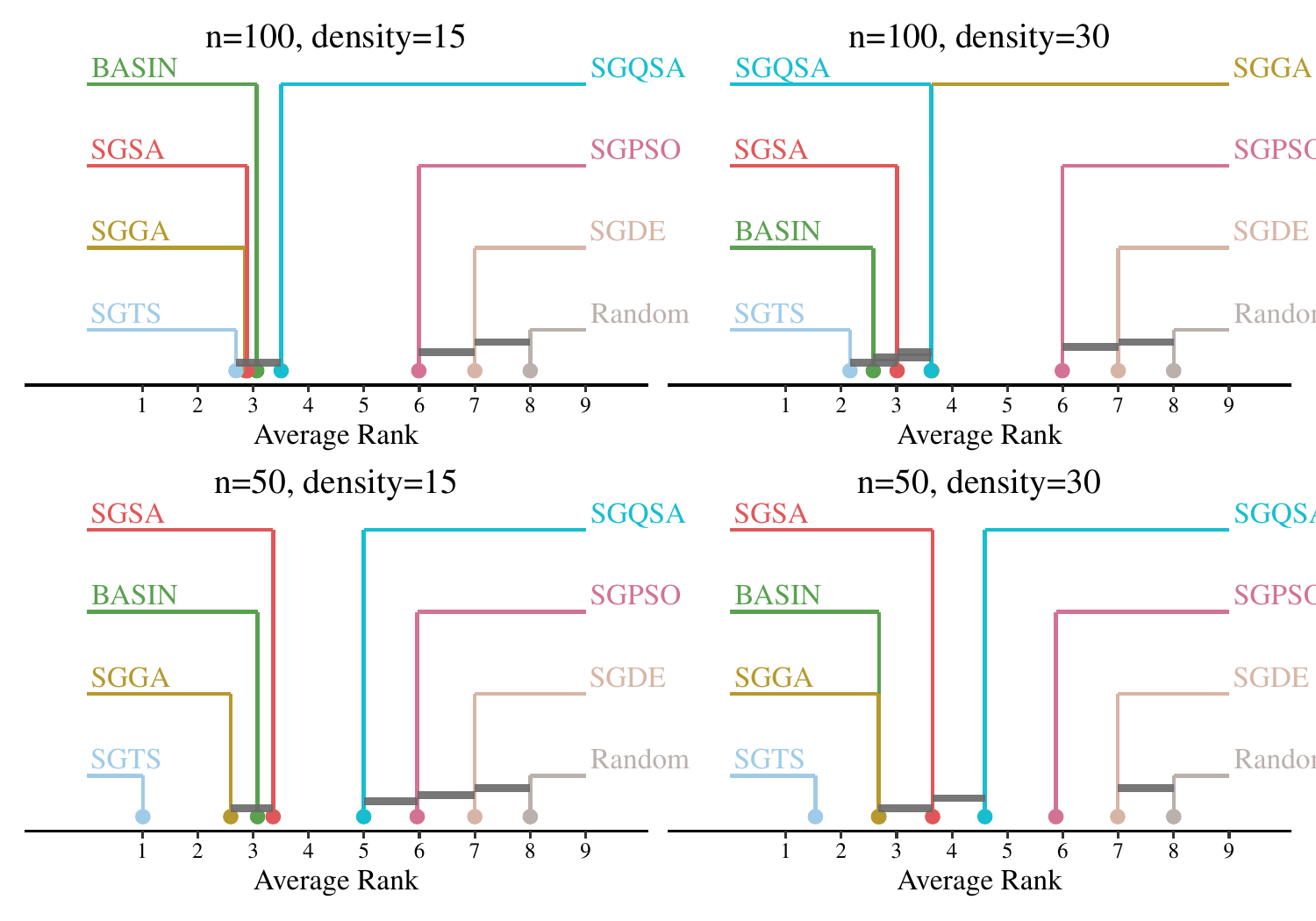}
 \caption{Critical-difference diagrams (Friedman with post-hoc Nemenyi) on \ac{MMM} graphs with $\mathit{pert}=16$ perturbing edge swaps, one panel per graph size $n$ and edge density $d$, one variant per algorithm and the random-search baseline. Each algorithm is placed at its mean rank on $\mathcal{S}'(A,P)$ over the 100 test graphs of the condition, rank 1 lowest; algorithms joined by a bar do not differ at the $5~\%$ level.}
    \label{fig:CD_MMM_pert16}
\end{figure}

\begin{figure}[ht]
    \centering
    \includegraphics[width=\linewidth]{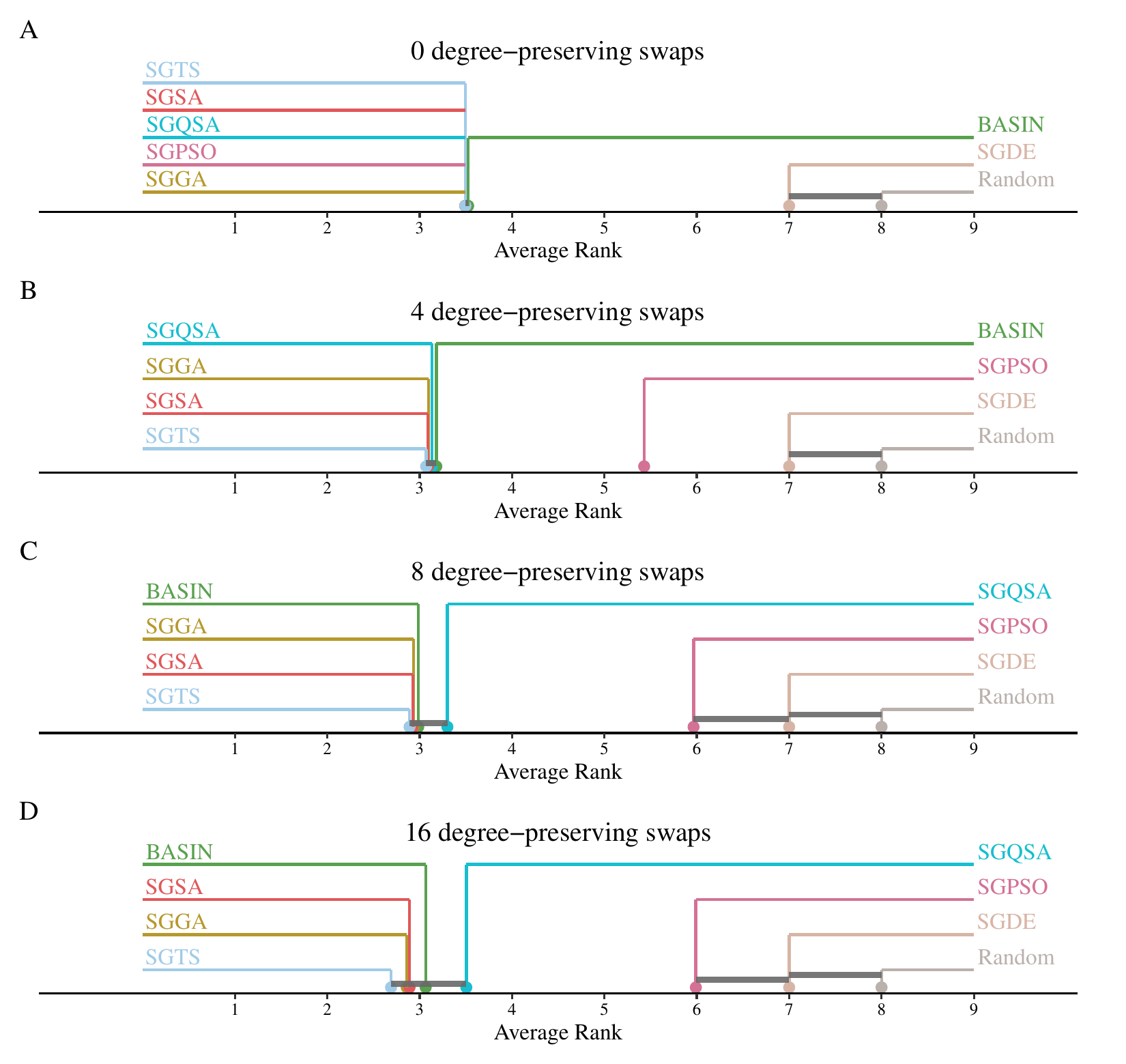}
 \caption{Critical-difference diagrams (Friedman with post-hoc Nemenyi) on \ac{MMM} graphs at $n=100$ and edge density $15~\%$, \textbf{(A)}--\textbf{(D)} at $\mathit{pert} = 0$, $4$, $8$ and $16$ perturbing edge swaps, one variant per algorithm and the random-search baseline. Each algorithm is placed at its mean rank on $\mathcal{S}'(A,P)$ over the 100 test graphs of the condition, rank 1 lowest; algorithms joined by a bar do not differ at the $5~\%$ level.}
    \label{fig:CD_ladder_MMM}
\end{figure}

In Figs.~\ref{fig:CD_ER}--\ref{fig:CD_ladder_MMM} each panel is one (size, density) condition; algorithms joined by a horizontal bar are not significantly different at the $5~\%$ level.

\clearpage

\subsection{Results including the random-search baseline}\label{app:includint_random_search}

Random search scores above every optimization strategy in every condition (Table~\ref{tab:comparison}), which is why the other figures omit it.

\begin{figure}[ht]
    \centering
    \includegraphics[width=0.75\linewidth]{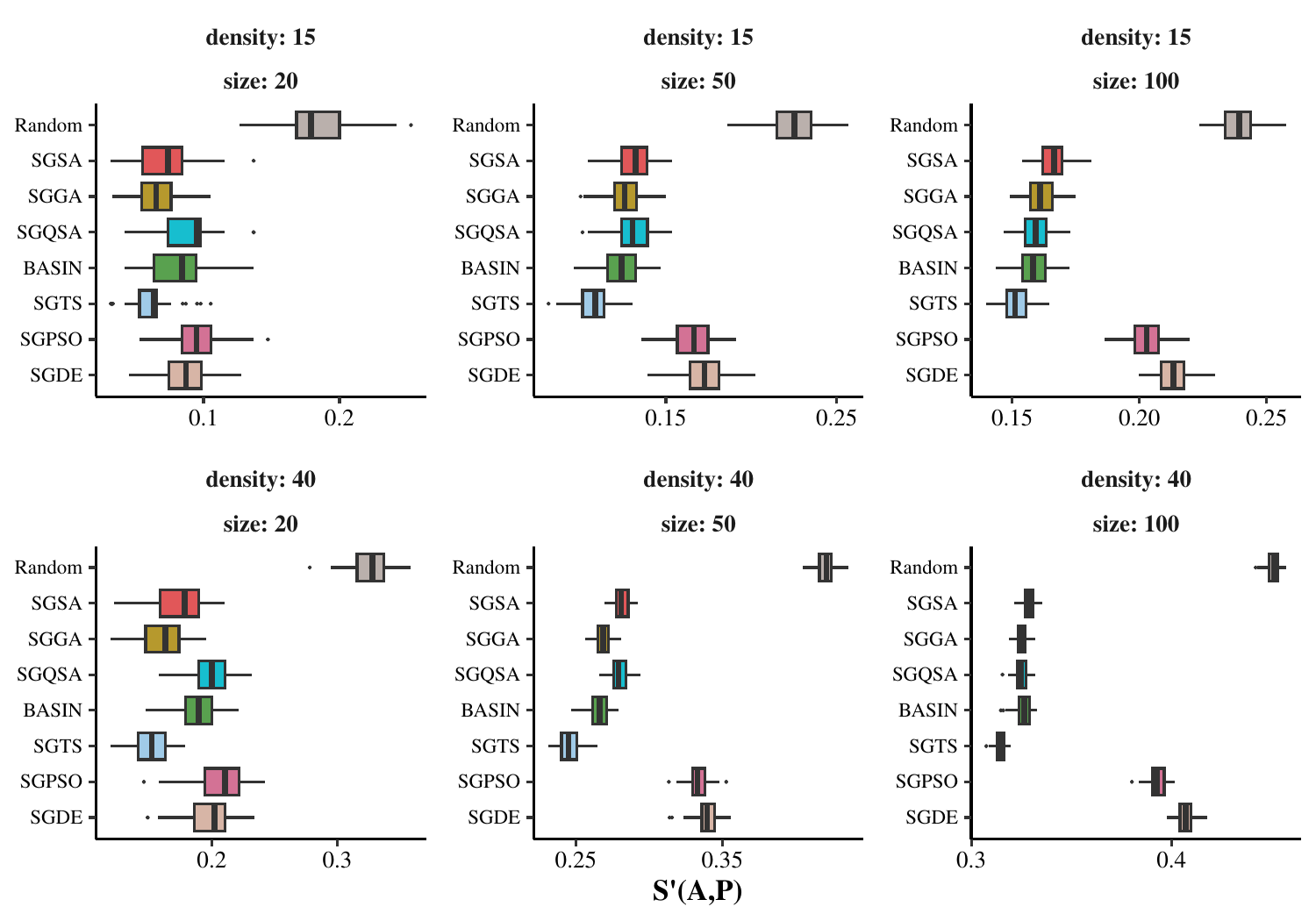}
 \caption{Performance on \ac{ER} graphs, one variant per algorithm, random-search baseline included. Boxes show $\mathcal{S}'(A,P)$ over the 100 test graphs of each condition, one solution per graph; one panel per graph size $n$ and edge density $d$, each with its own horizontal axis.}
    \label{fig:box_ER_random}
\end{figure}

\begin{figure}[ht]
    \centering
    \includegraphics[width=0.75\linewidth]{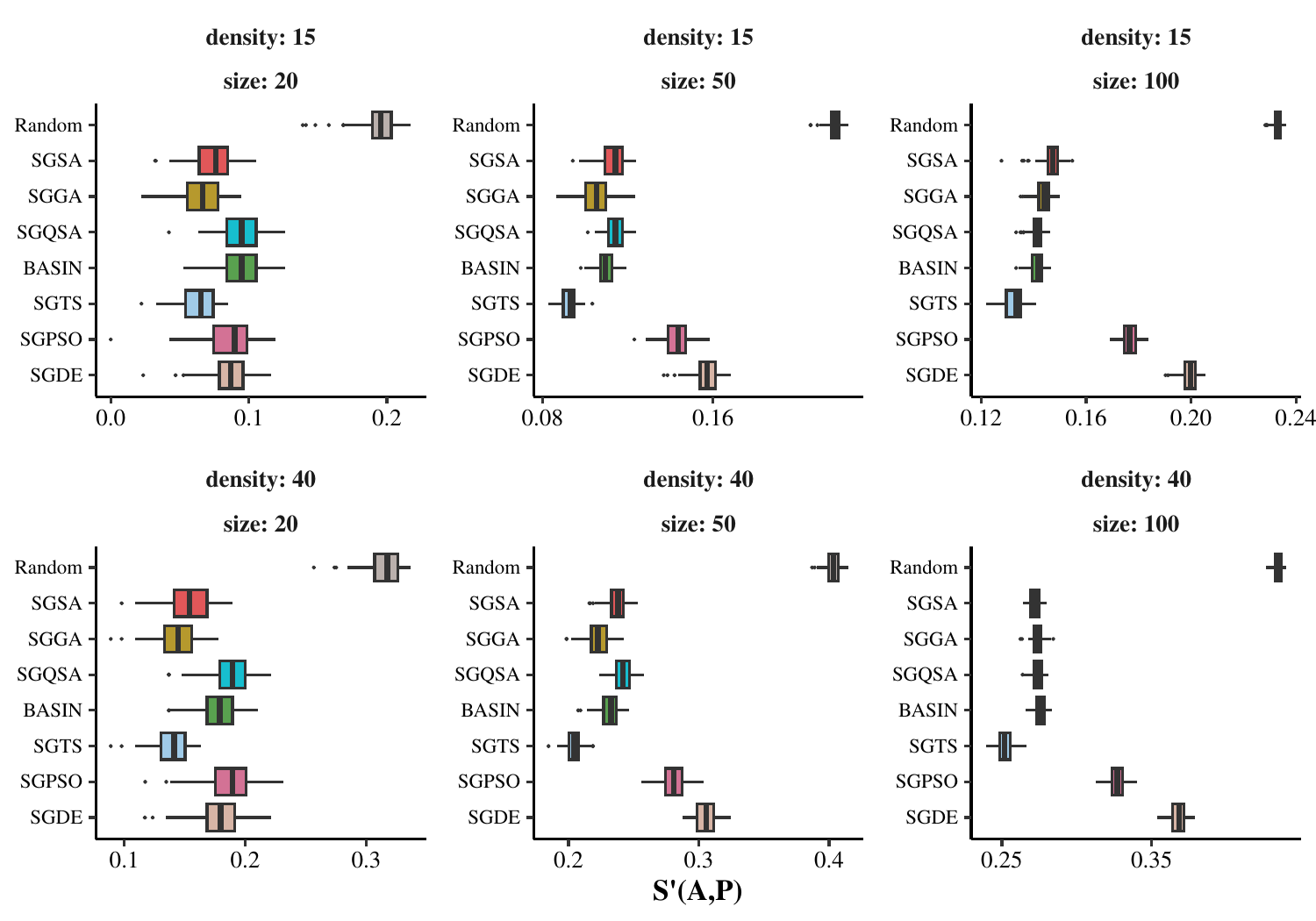}
 \caption{Performance on \ac{BA} graphs, one variant per algorithm, random-search baseline included. Boxes show $\mathcal{S}'(A,P)$ over the 100 test graphs of each condition, one solution per graph; one panel per graph size $n$ and edge density $d$, each with its own horizontal axis.}
    \label{fig:box_BA_random}
\end{figure}

\begin{figure}[ht]
    \centering
    \includegraphics[width=0.75\linewidth]{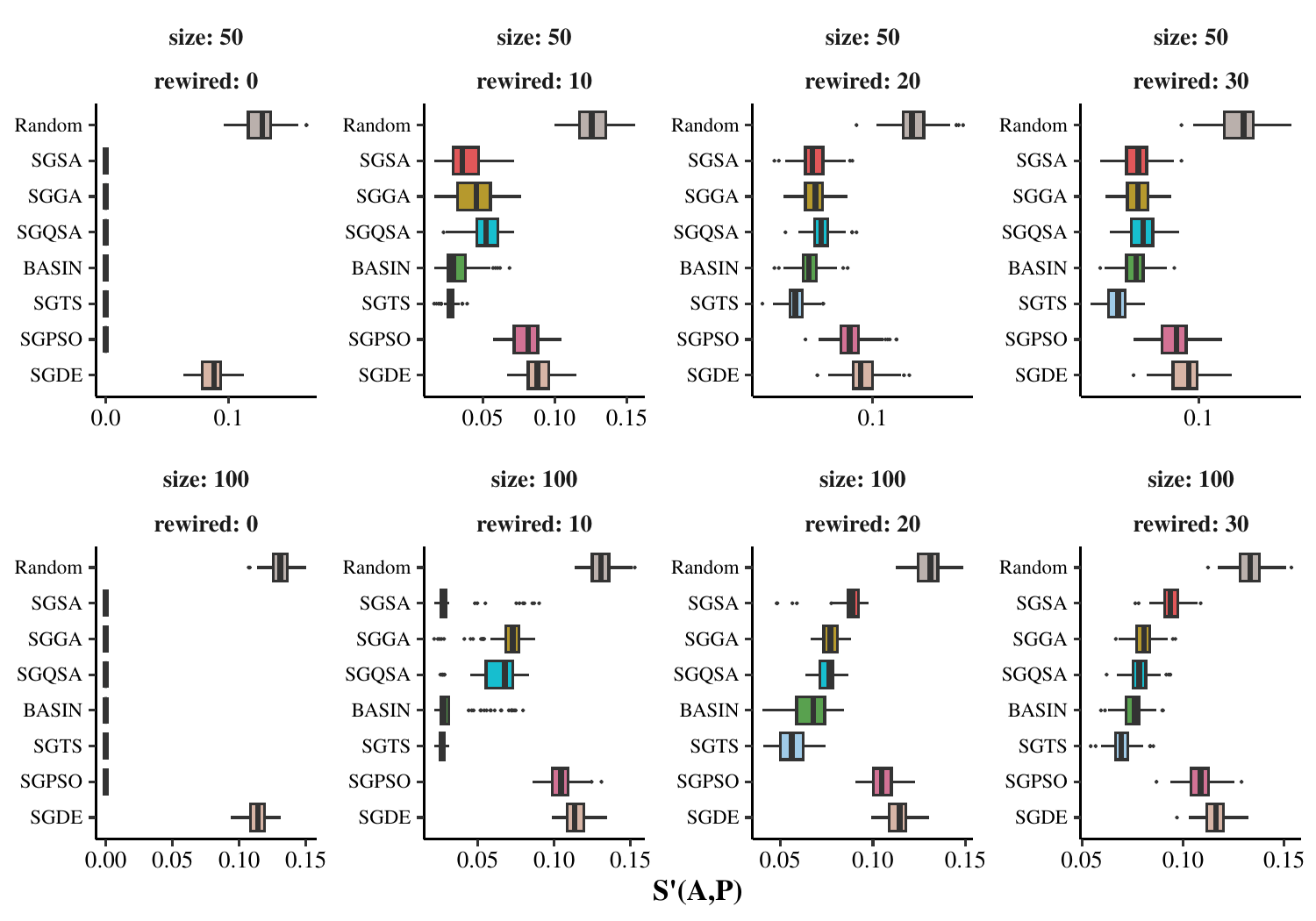}
 \caption{Performance on \ac{LRM} graphs at edge density $15~\%$, one variant per algorithm, random-search baseline included. Boxes show $\mathcal{S}'(A,P)$ over the 100 test graphs of each condition, one solution per graph; one panel per graph size $n$ and share $\mathit{rew}$ of rewired edges, each with its own horizontal axis.}
    \label{fig:box_LRM15_random}
\end{figure}

\begin{figure}[ht]
    \centering
    \includegraphics[width=0.75\linewidth]{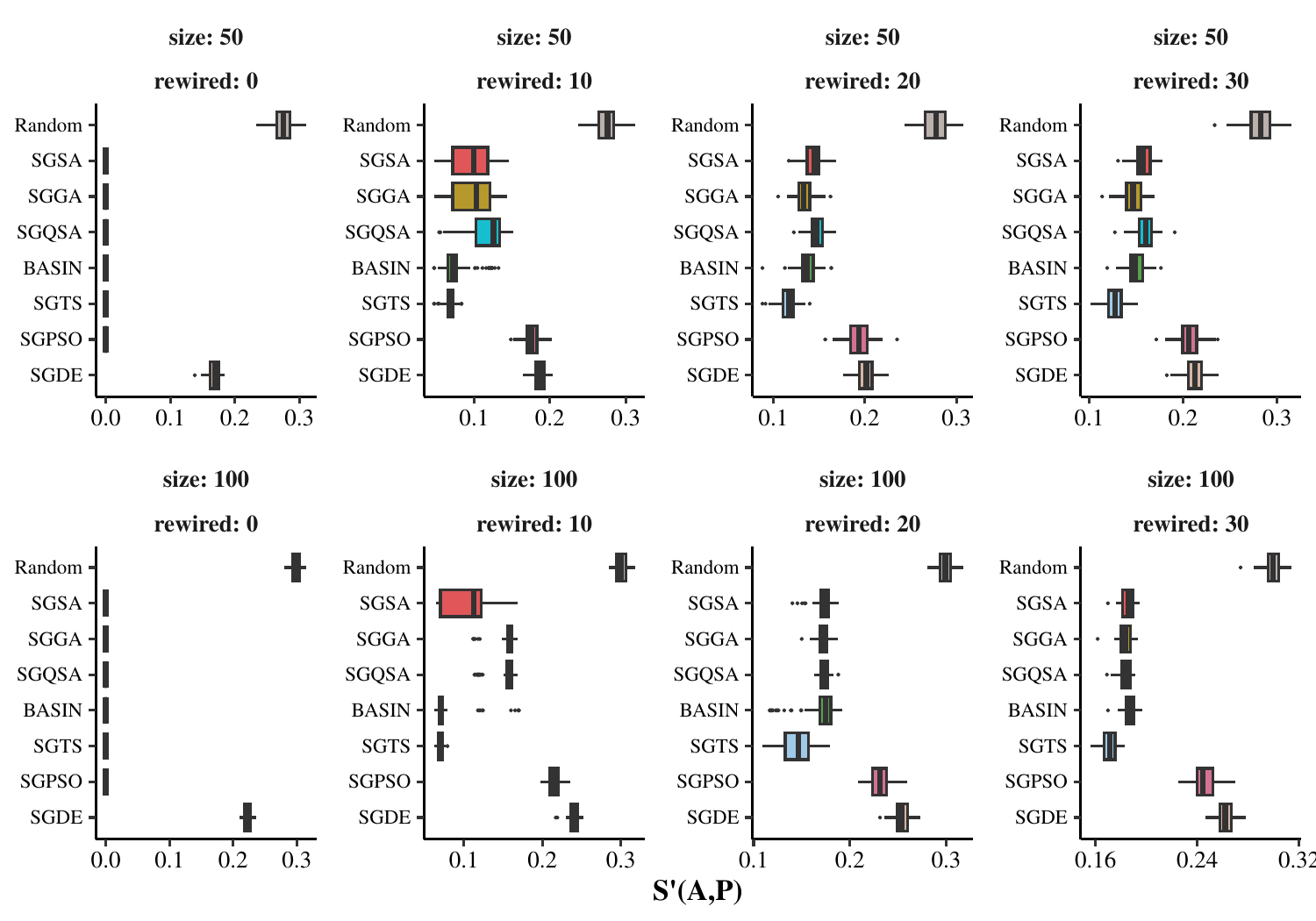}
 \caption{Performance on \ac{LRM} graphs at edge density $40~\%$, one variant per algorithm, random-search baseline included. Boxes show $\mathcal{S}'(A,P)$ over the 100 test graphs of each condition, one solution per graph; one panel per graph size $n$ and share $\mathit{rew}$ of rewired edges, each with its own horizontal axis.}
    \label{fig:box_LRM40_random}
\end{figure}

\begin{figure}[ht]
    \centering
    \includegraphics[width=0.75\linewidth]{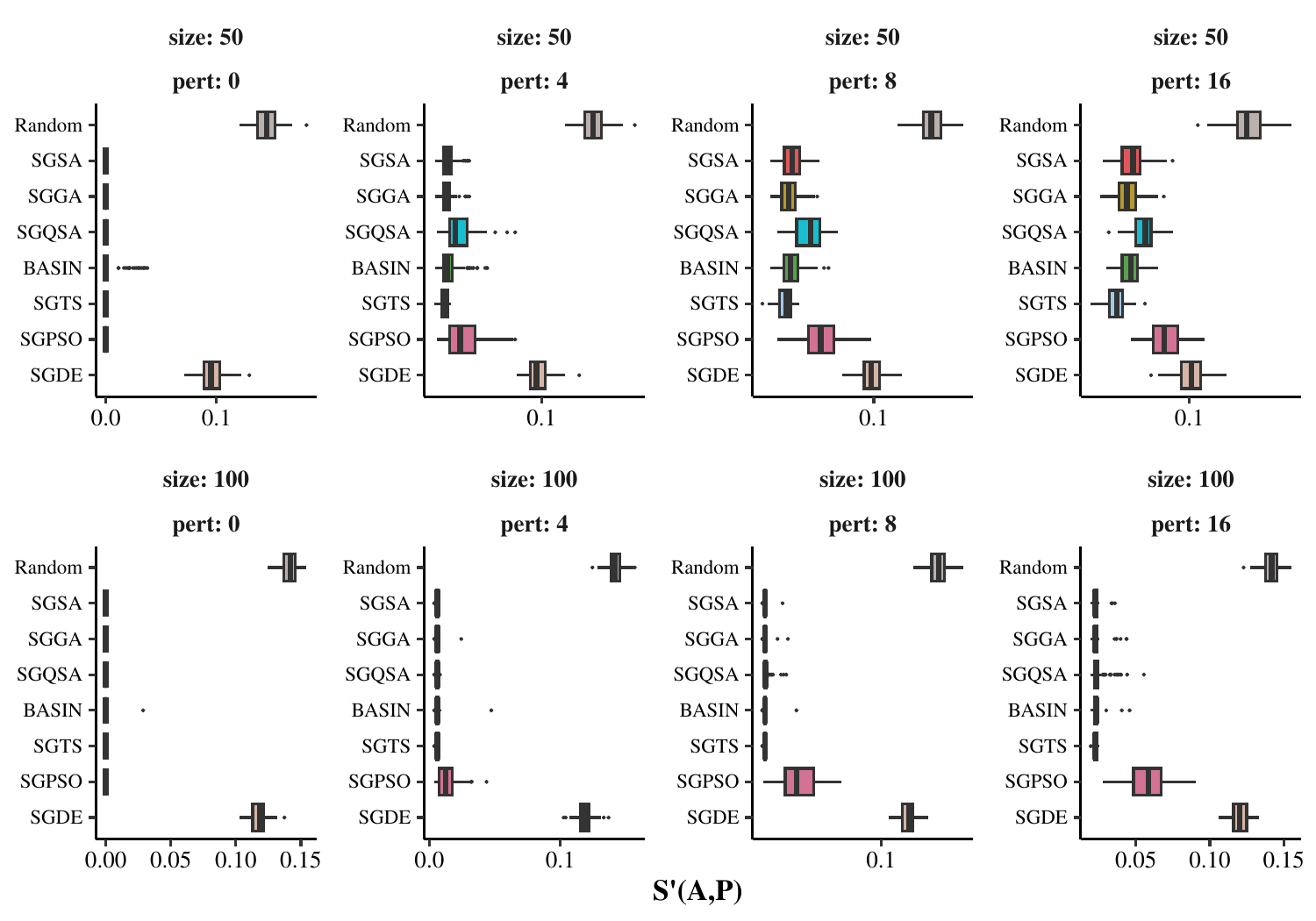}
 \caption{Performance on \ac{MMM} graphs at edge density $15~\%$, one variant per algorithm, random-search baseline included. Boxes show $\mathcal{S}'(A,P)$ over the 100 test graphs of each condition, one solution per graph; one panel per graph size $n$ and number $\mathit{pert}$ of perturbing edge swaps, each with its own horizontal axis.}
    \label{fig:box_MMM15_random}
\end{figure}

\begin{figure}[ht]
    \centering
    \includegraphics[width=0.75\linewidth]{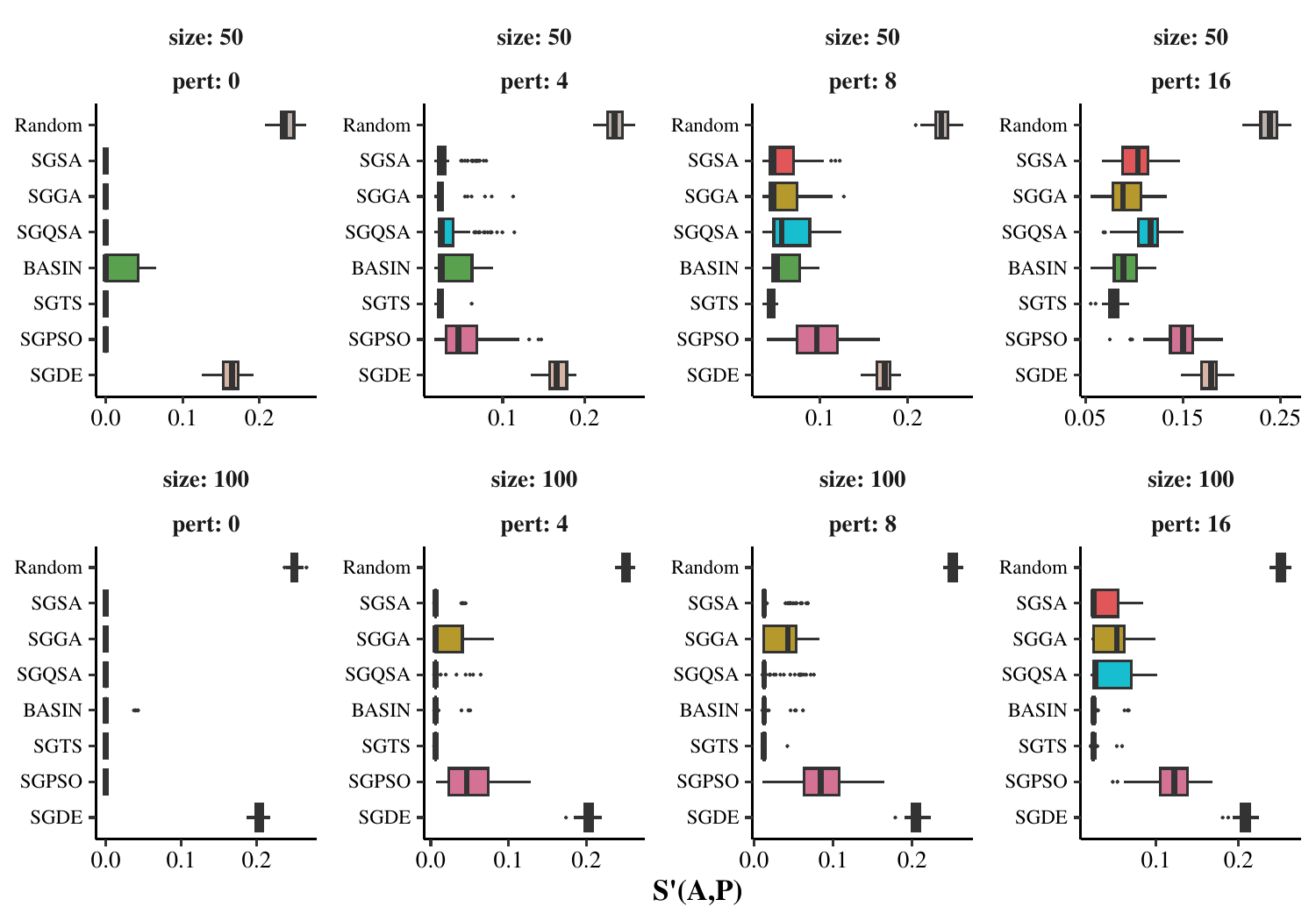}
 \caption{Performance on \ac{MMM} graphs at edge density $30~\%$, one variant per algorithm, random-search baseline included. Boxes show $\mathcal{S}'(A,P)$ over the 100 test graphs of each condition, one solution per graph; one panel per graph size $n$ and number $\mathit{pert}$ of perturbing edge swaps, each with its own horizontal axis.}
    \label{fig:box_MMM30_random}
\end{figure}
\clearpage

\subsection{Results using the size-normalized symmetry score $\mathcal{S}(A,P)$}\label{app:size-normalized_S}

\begin{figure}[ht]
    \centering
    \includegraphics[width=0.75\linewidth]{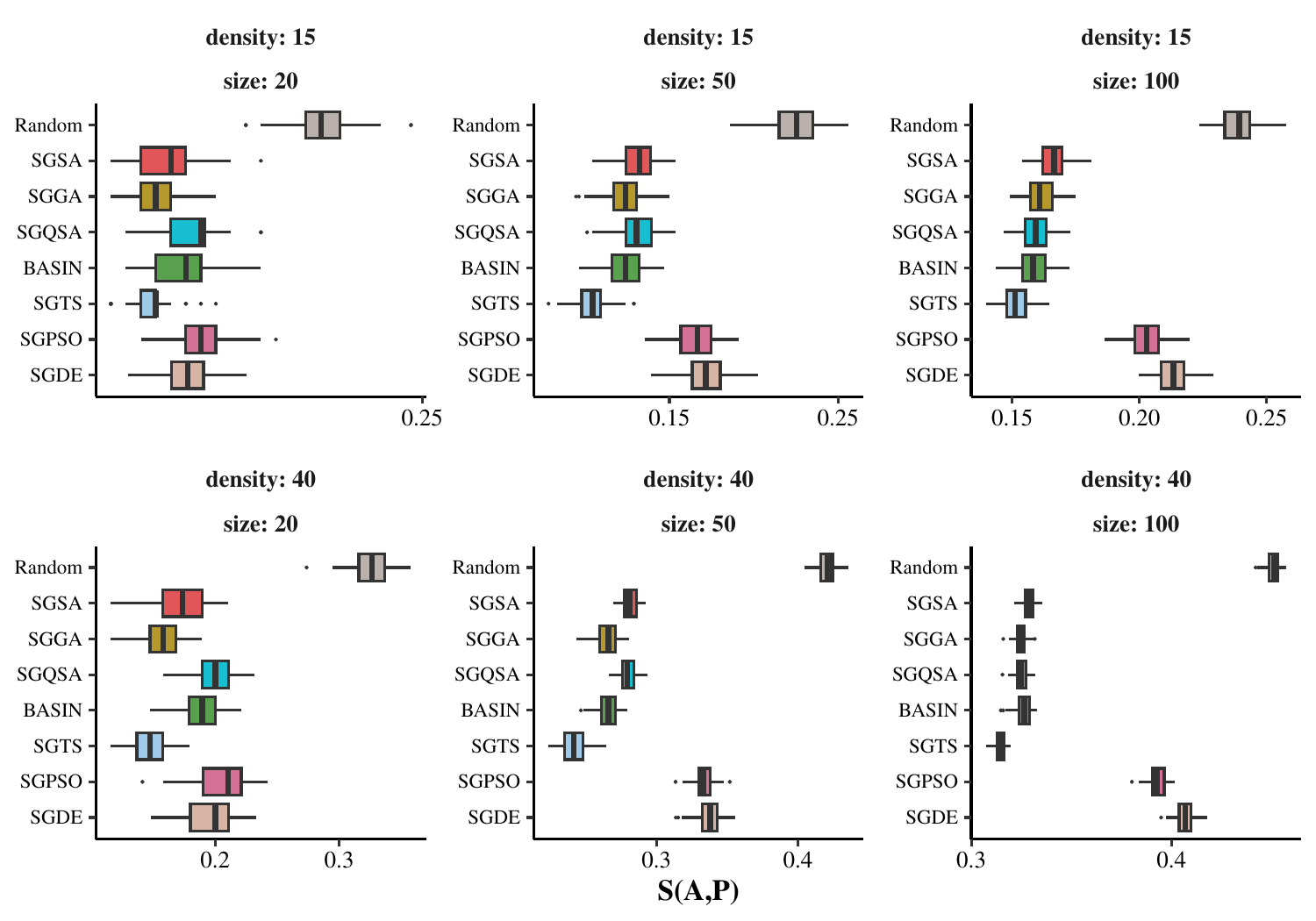}
 \caption{Performance on \ac{ER} graphs, under $\mathcal{S}(A,P)$, one variant per algorithm, random-search baseline included. Boxes show the size-normalized score $\mathcal{S}(A,P)$, which does not correct for fixed points over the 100 test graphs of each condition, one solution per graph; one panel per graph size $n$ and edge density $d$, each with its own horizontal axis.}
    \label{fig:box_ER_S_random}
\end{figure}

\begin{figure}[ht]
    \centering
    \includegraphics[width=0.75\linewidth]{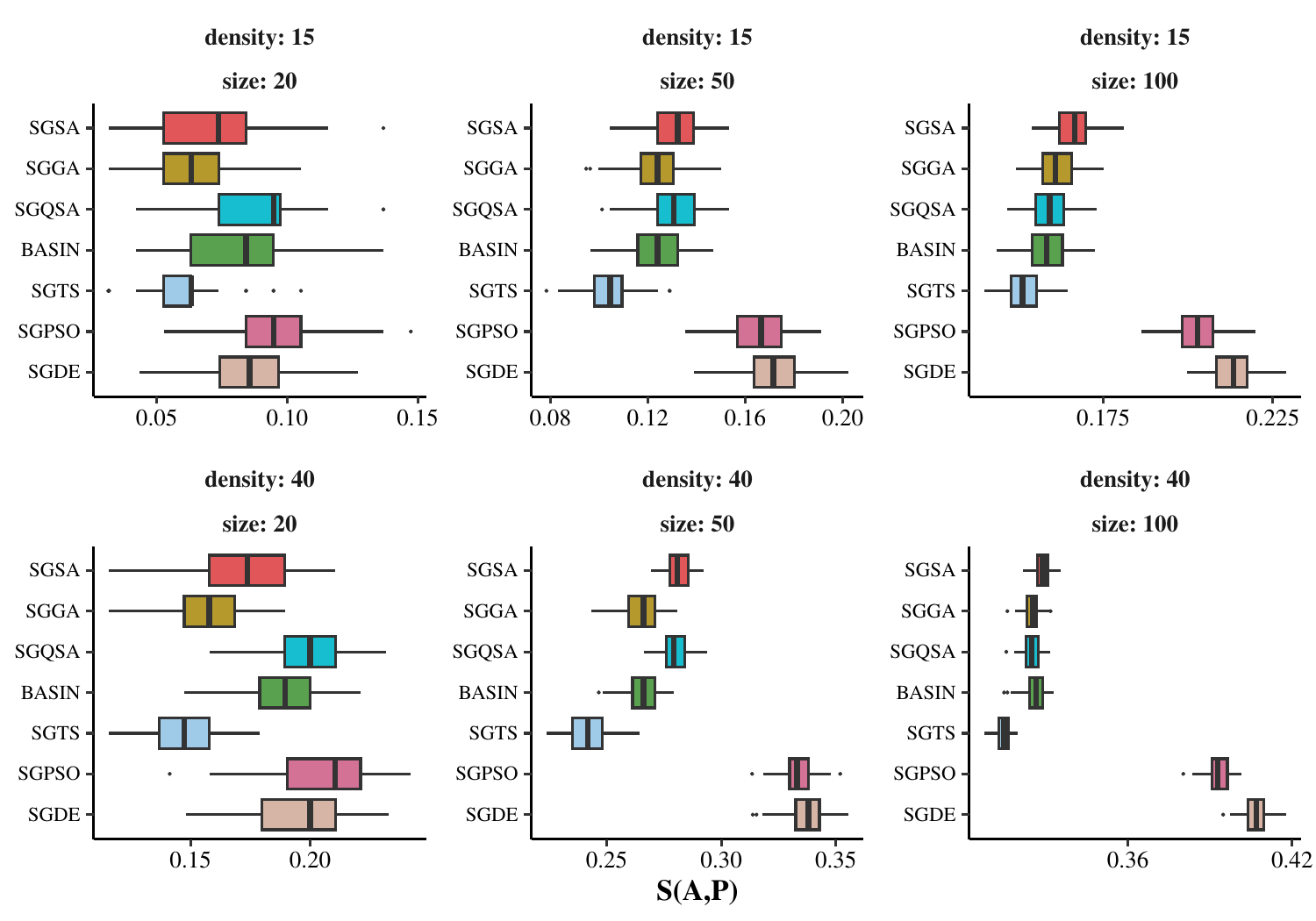}
 \caption{Performance on \ac{ER} graphs, under $\mathcal{S}(A,P)$, one variant per algorithm. Boxes show the size-normalized score $\mathcal{S}(A,P)$, which does not correct for fixed points over the 100 test graphs of each condition, one solution per graph; one panel per graph size $n$ and edge density $d$, each with its own horizontal axis.}
    \label{fig:box_ER_S}
\end{figure}

\begin{figure}[ht]
    \centering
    \includegraphics[width=0.75\linewidth]{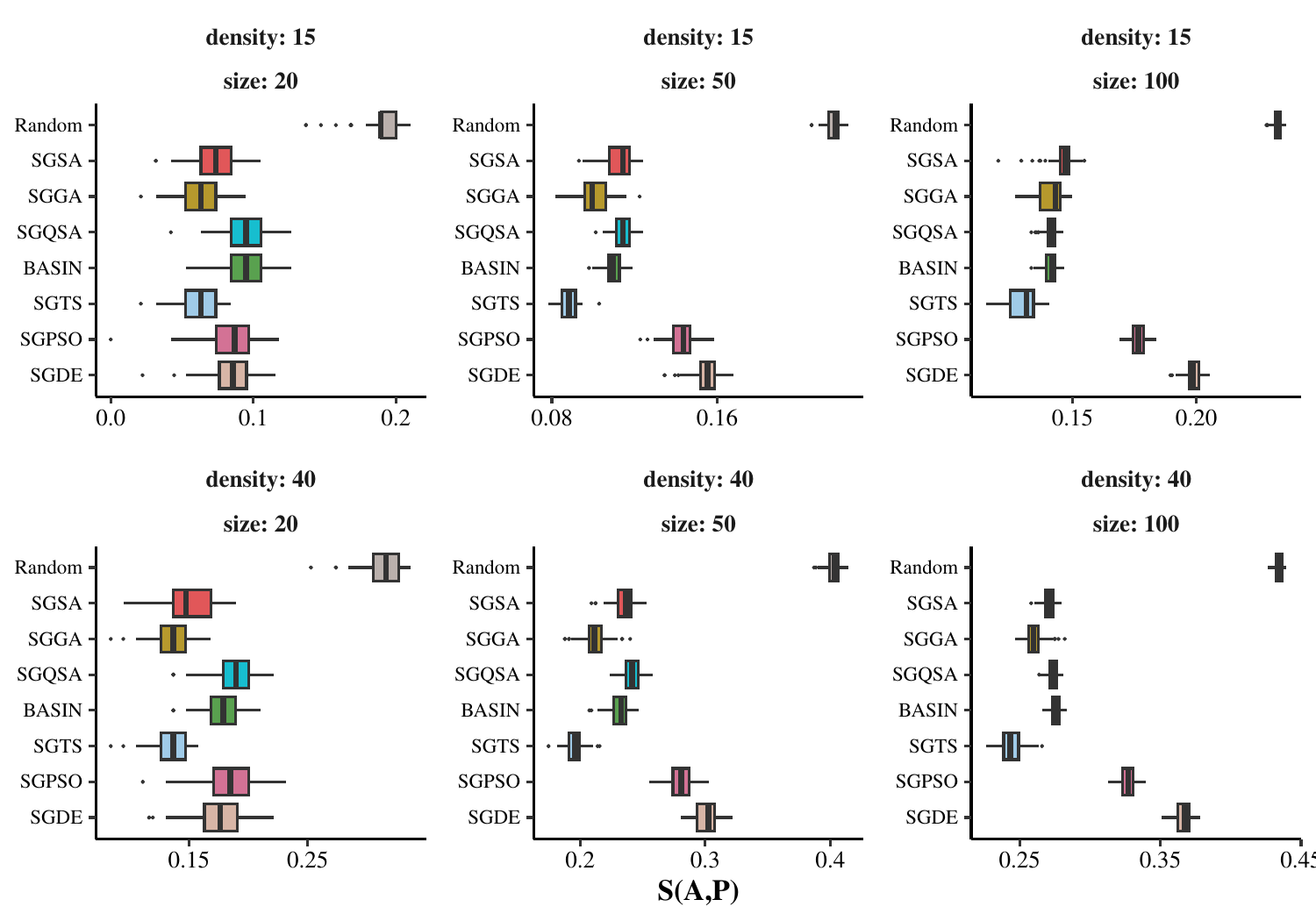}
 \caption{Performance on \ac{BA} graphs, under $\mathcal{S}(A,P)$, one variant per algorithm, random-search baseline included. Boxes show the size-normalized score $\mathcal{S}(A,P)$, which does not correct for fixed points over the 100 test graphs of each condition, one solution per graph; one panel per graph size $n$ and edge density $d$, each with its own horizontal axis.}
    \label{fig:box_BA_S_random}
\end{figure}

\begin{figure}[ht]
    \centering
    \includegraphics[width=0.75\linewidth]{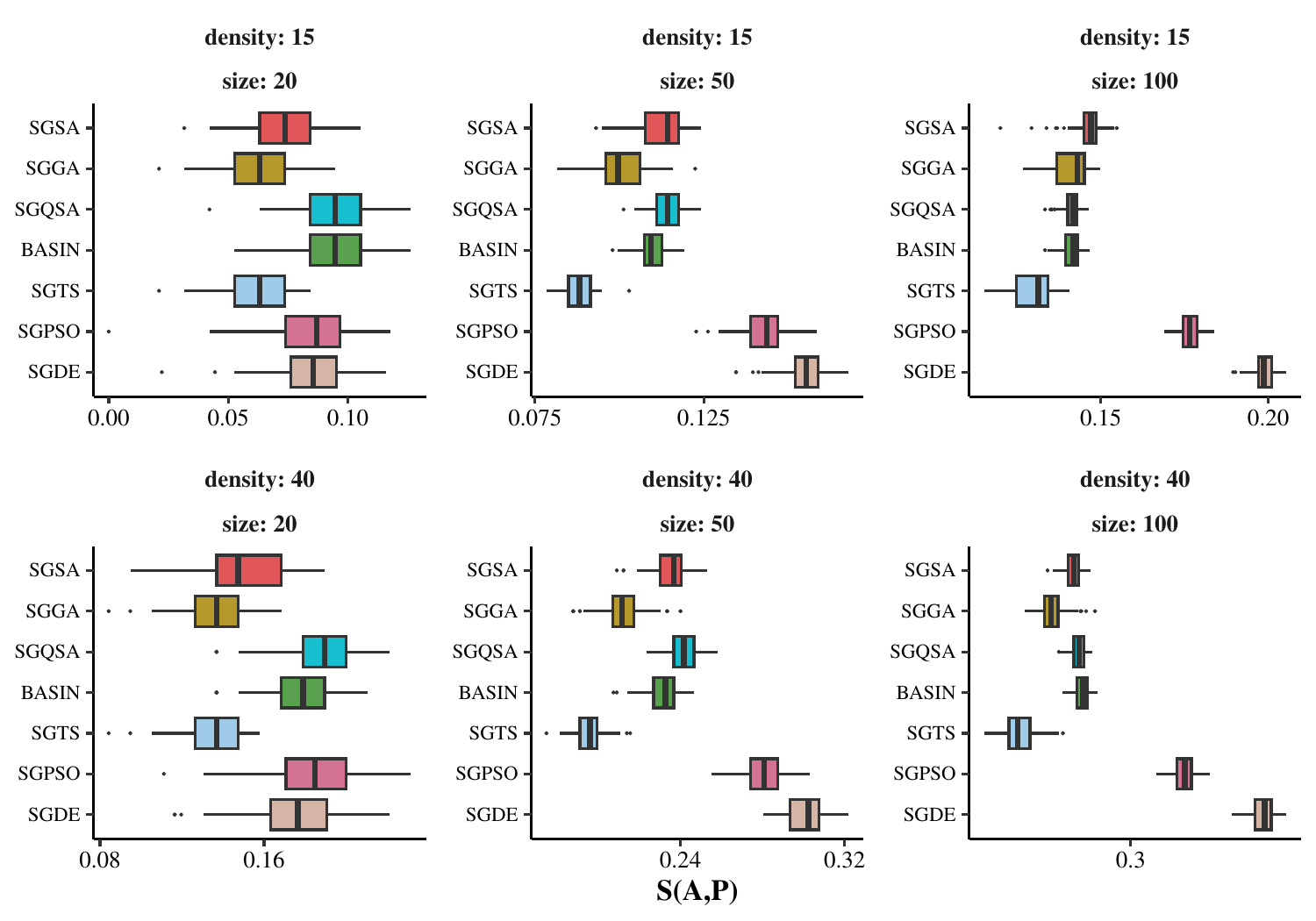}
 \caption{Performance on \ac{BA} graphs, under $\mathcal{S}(A,P)$, one variant per algorithm. Boxes show the size-normalized score $\mathcal{S}(A,P)$, which does not correct for fixed points over the 100 test graphs of each condition, one solution per graph; one panel per graph size $n$ and edge density $d$, each with its own horizontal axis.}
    \label{fig:box_BA_S}
\end{figure}

\begin{figure}[ht]
    \centering
    \includegraphics[width=0.75\linewidth]{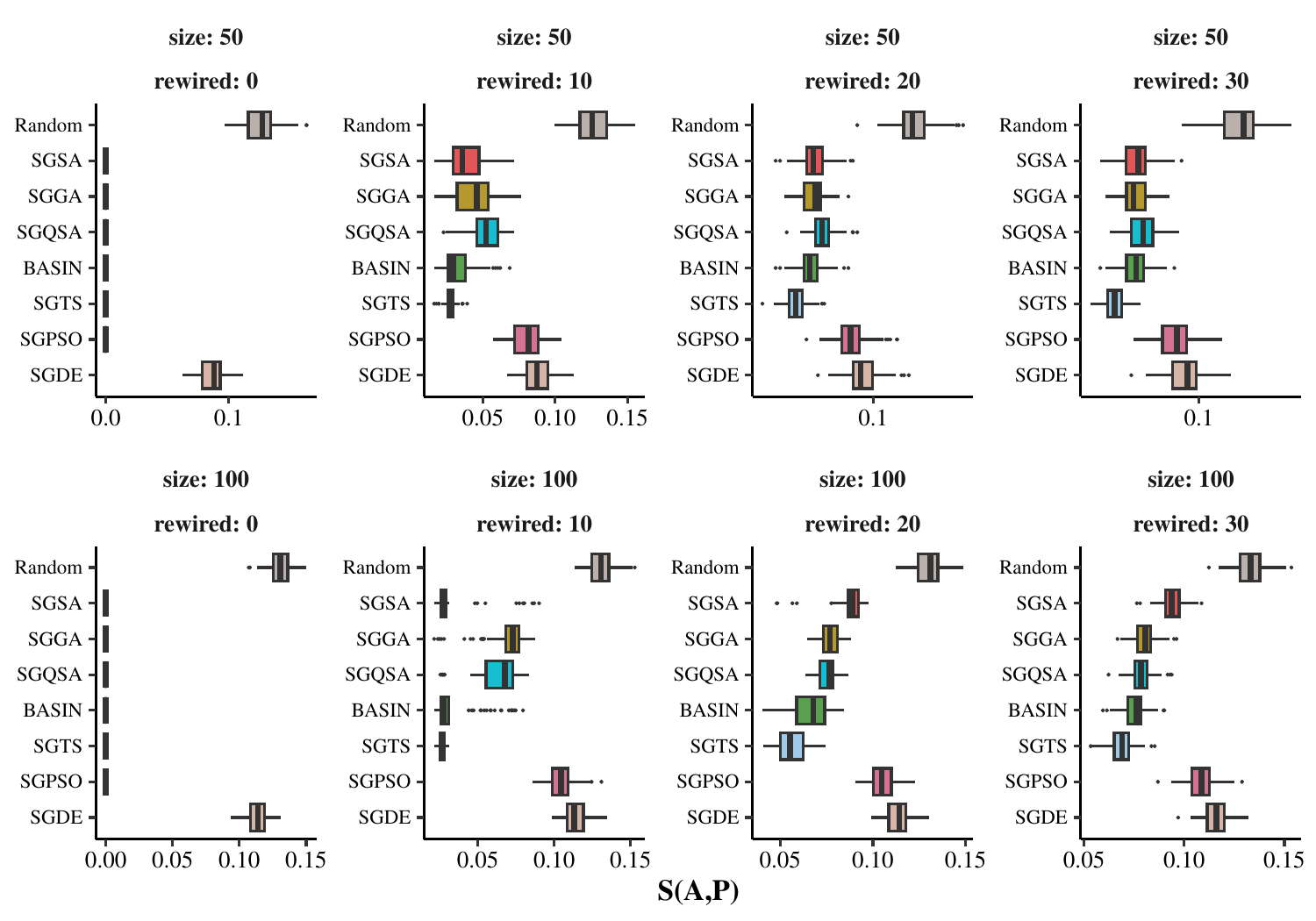}
 \caption{Performance on \ac{LRM} graphs at edge density $15~\%$, under $\mathcal{S}(A,P)$, one variant per algorithm, random-search baseline included. Boxes show the size-normalized score $\mathcal{S}(A,P)$, which does not correct for fixed points over the 100 test graphs of each condition, one solution per graph; one panel per graph size $n$ and share $\mathit{rew}$ of rewired edges, each with its own horizontal axis.}
    \label{fig:box_LRM15_S_random}
\end{figure}

\begin{figure}[ht]
    \centering
    \includegraphics[width=0.75\linewidth]{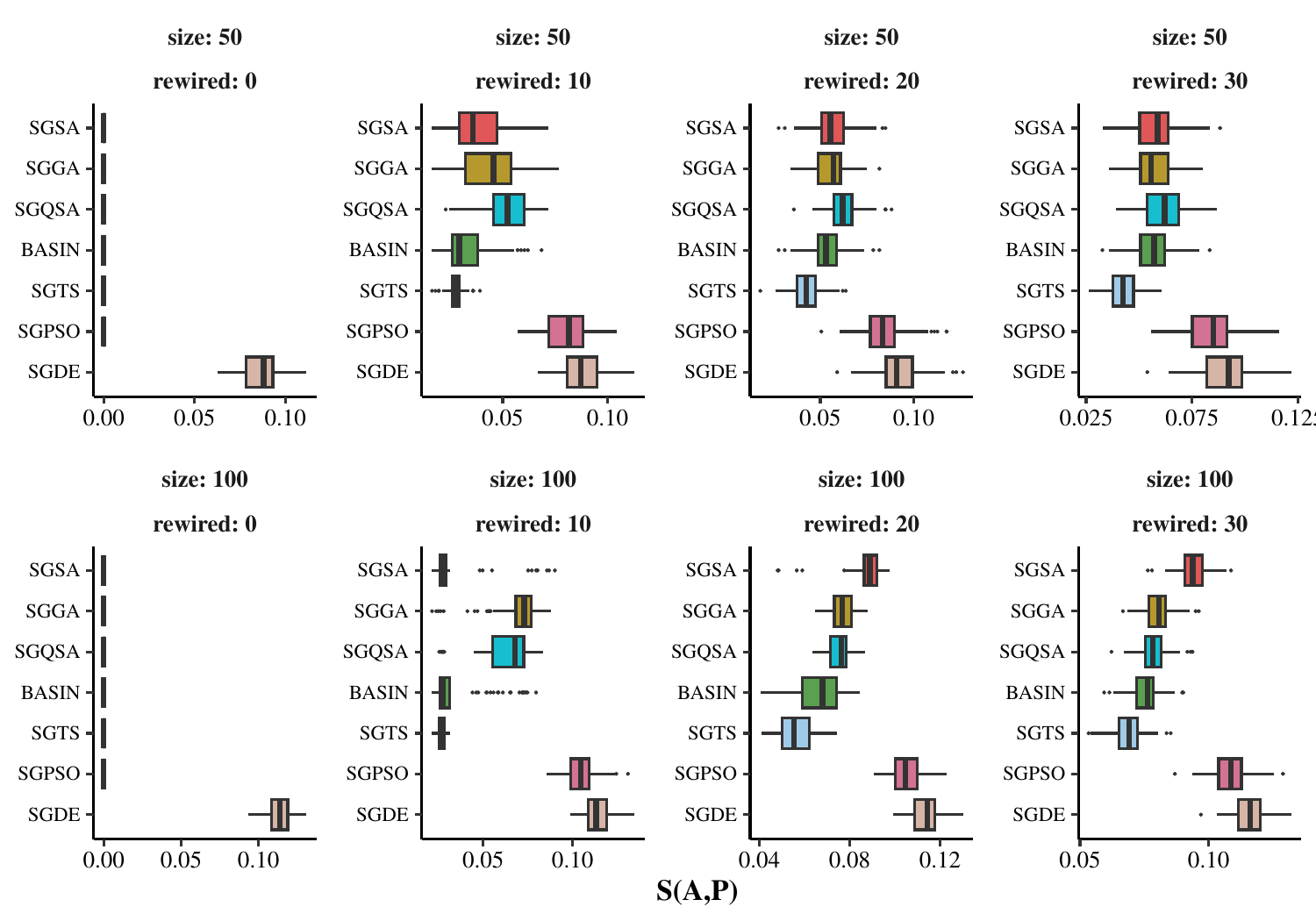}
 \caption{Performance on \ac{LRM} graphs at edge density $15~\%$, under $\mathcal{S}(A,P)$, one variant per algorithm. Boxes show the size-normalized score $\mathcal{S}(A,P)$, which does not correct for fixed points over the 100 test graphs of each condition, one solution per graph; one panel per graph size $n$ and share $\mathit{rew}$ of rewired edges, each with its own horizontal axis.}
    \label{fig:box_LRM15_S}
\end{figure}
\clearpage

\begin{figure}[ht]
    \centering
    \includegraphics[width=0.75\linewidth]{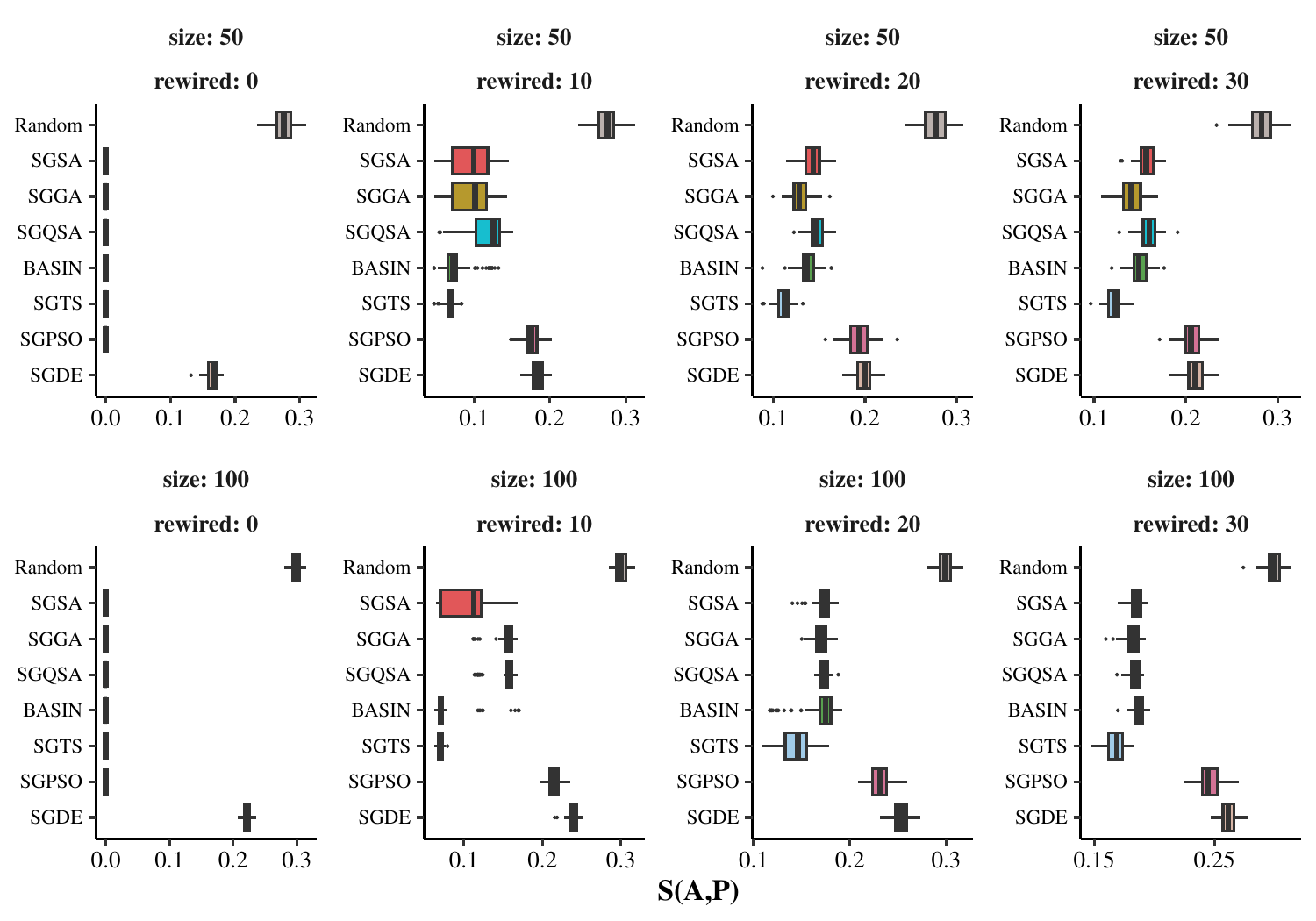}
 \caption{Performance on \ac{LRM} graphs at edge density $40~\%$, under $\mathcal{S}(A,P)$, one variant per algorithm, random-search baseline included. Boxes show the size-normalized score $\mathcal{S}(A,P)$, which does not correct for fixed points over the 100 test graphs of each condition, one solution per graph; one panel per graph size $n$ and share $\mathit{rew}$ of rewired edges, each with its own horizontal axis.}
    \label{fig:box_LRM40_S_random}
\end{figure}

\begin{figure}[ht]
    \centering
    \includegraphics[width=0.75\linewidth]{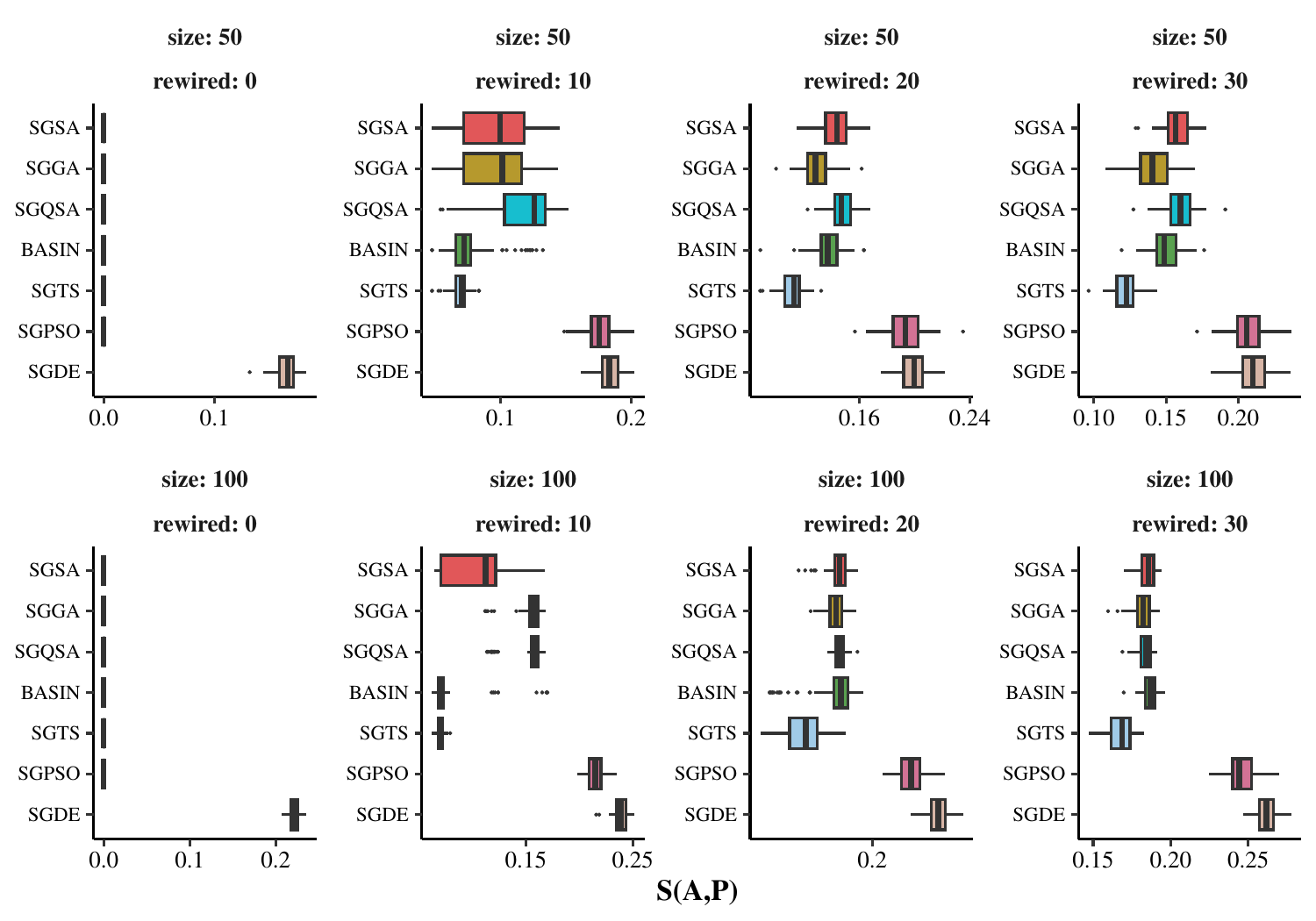}
 \caption{Performance on \ac{LRM} graphs at edge density $40~\%$, under $\mathcal{S}(A,P)$, one variant per algorithm. Boxes show the size-normalized score $\mathcal{S}(A,P)$, which does not correct for fixed points over the 100 test graphs of each condition, one solution per graph; one panel per graph size $n$ and share $\mathit{rew}$ of rewired edges, each with its own horizontal axis.}
    \label{fig:box_LRM40_S}
\end{figure}

\begin{figure}[ht]
    \centering
    \includegraphics[width=0.75\linewidth]{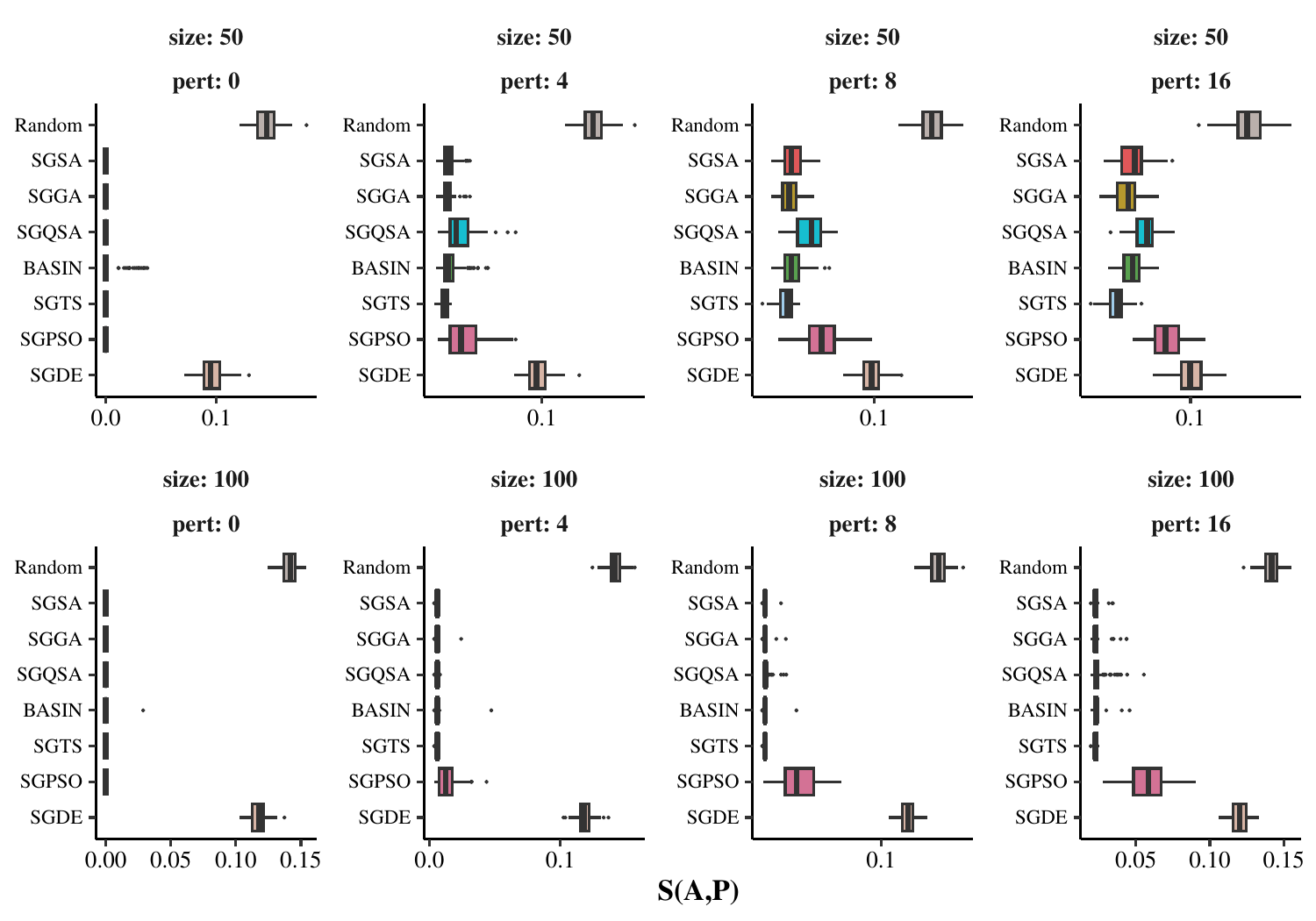}
 \caption{Performance on \ac{MMM} graphs at edge density $15~\%$, under $\mathcal{S}(A,P)$, one variant per algorithm, random-search baseline included. Boxes show the size-normalized score $\mathcal{S}(A,P)$, which does not correct for fixed points over the 100 test graphs of each condition, one solution per graph; one panel per graph size $n$ and number $\mathit{pert}$ of perturbing edge swaps, each with its own horizontal axis.}
    \label{fig:box_MMM15_S_random}
\end{figure}

\begin{figure}[ht]
    \centering
    \includegraphics[width=0.75\linewidth]{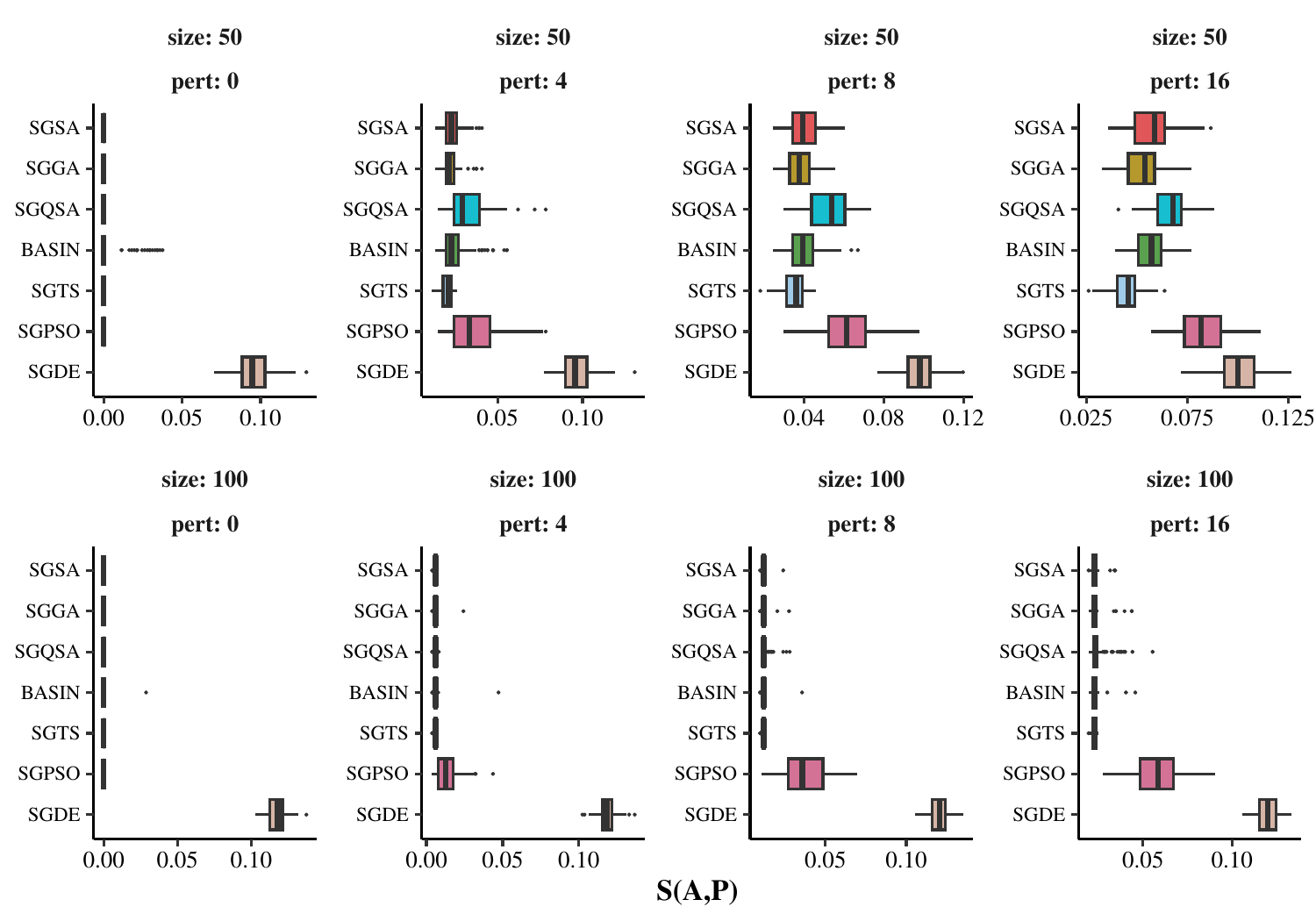}
 \caption{Performance on \ac{MMM} graphs at edge density $15~\%$, under $\mathcal{S}(A,P)$, one variant per algorithm. Boxes show the size-normalized score $\mathcal{S}(A,P)$, which does not correct for fixed points over the 100 test graphs of each condition, one solution per graph; one panel per graph size $n$ and number $\mathit{pert}$ of perturbing edge swaps, each with its own horizontal axis.}
    \label{fig:box_MMM15_S}
\end{figure}

\begin{figure}[ht]
    \centering
    \includegraphics[width=0.75\linewidth]{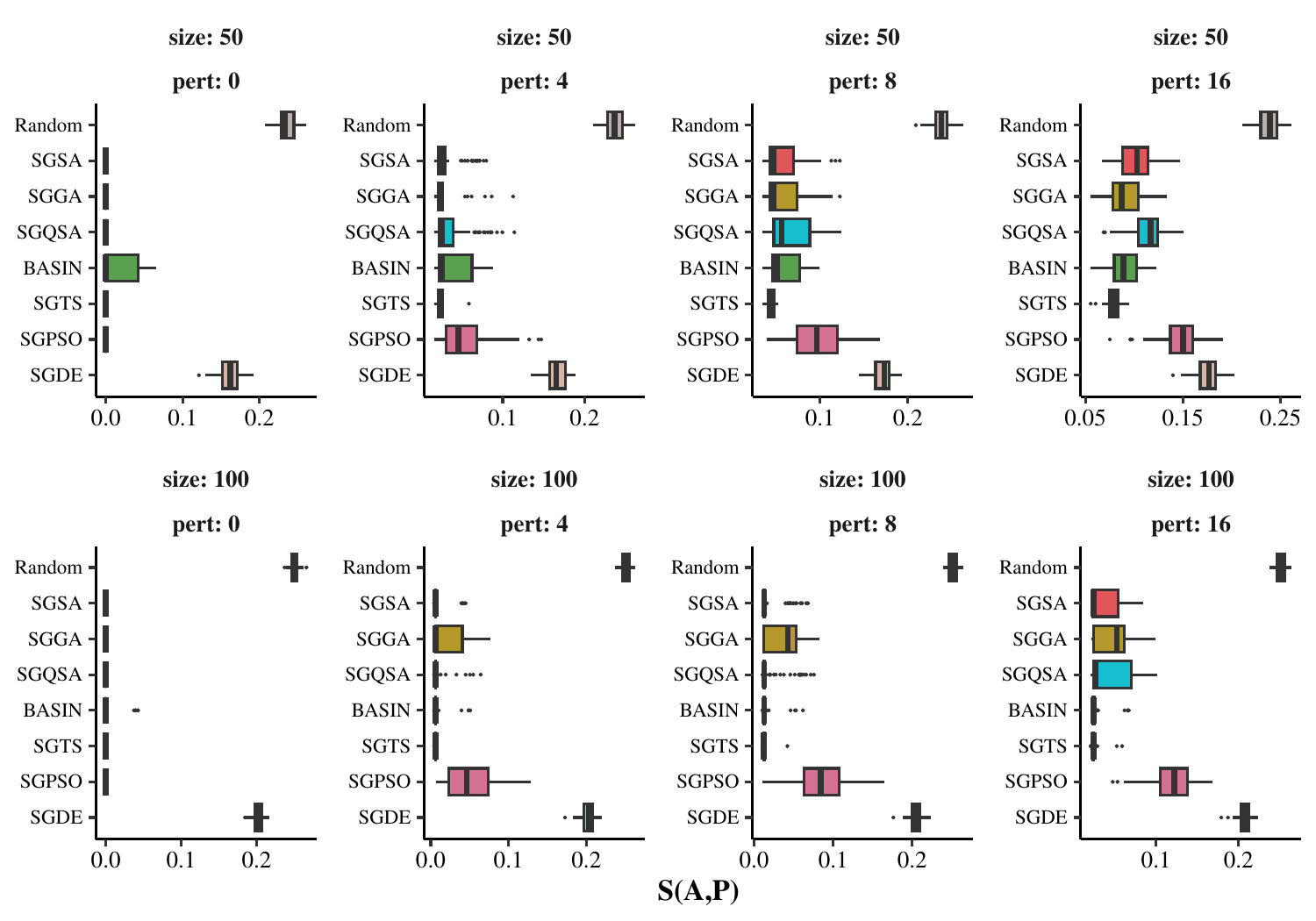}
 \caption{Performance on \ac{MMM} graphs at edge density $30~\%$, under $\mathcal{S}(A,P)$, one variant per algorithm, random-search baseline included. Boxes show the size-normalized score $\mathcal{S}(A,P)$, which does not correct for fixed points over the 100 test graphs of each condition, one solution per graph; one panel per graph size $n$ and number $\mathit{pert}$ of perturbing edge swaps, each with its own horizontal axis.}
    \label{fig:box_MMM30_S_random}
\end{figure}

\begin{figure}[ht]
    \centering
    \includegraphics[width=0.75\linewidth]{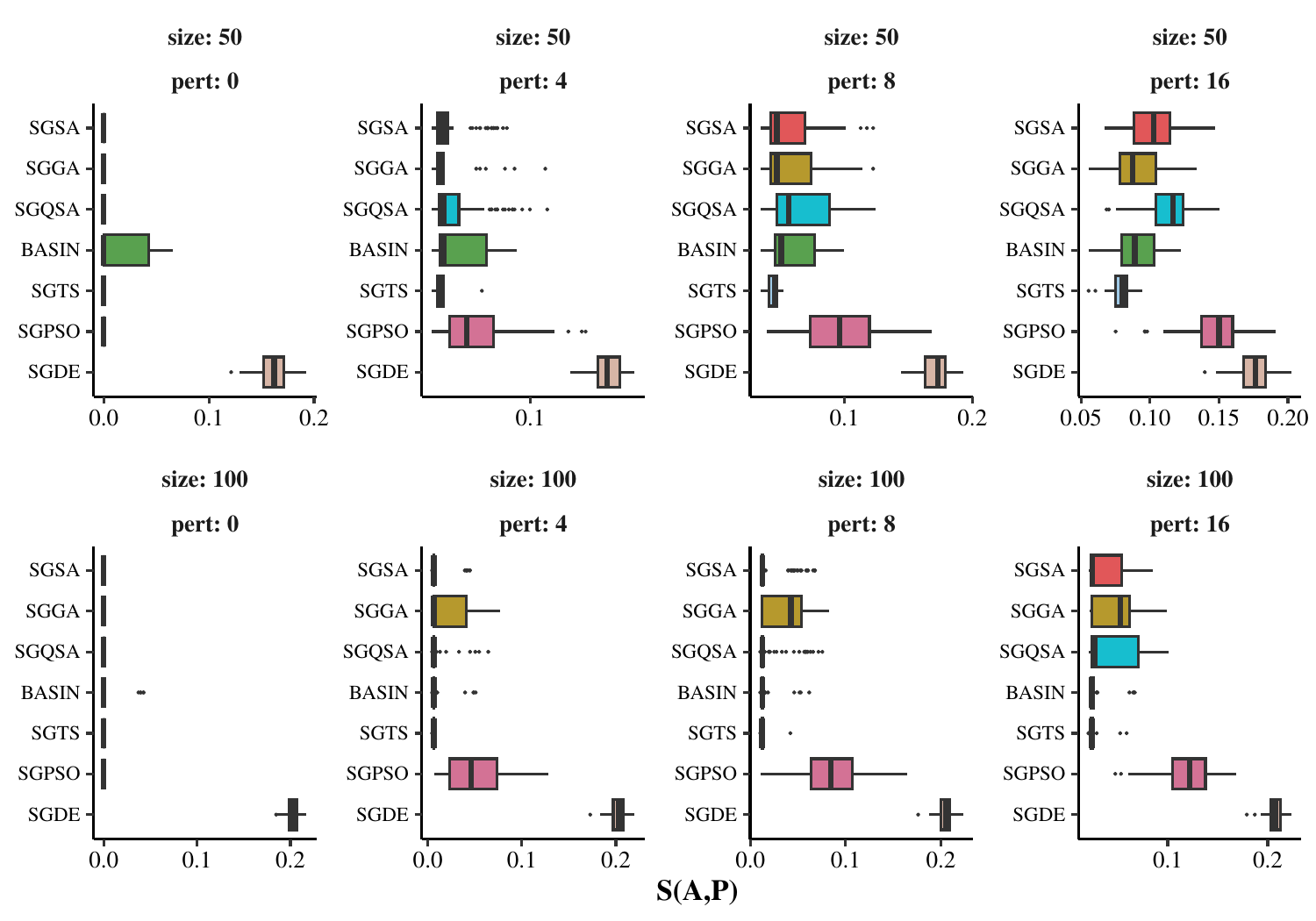}
 \caption{Performance on \ac{MMM} graphs at edge density $30~\%$, under $\mathcal{S}(A,P)$, one variant per algorithm. Boxes show the size-normalized score $\mathcal{S}(A,P)$, which does not correct for fixed points over the 100 test graphs of each condition, one solution per graph; one panel per graph size $n$ and number $\mathit{pert}$ of perturbing edge swaps, each with its own horizontal axis.}
    \label{fig:box_MMM30_S}
\end{figure}
\clearpage

\subsection{Median and interquartile range at every condition}\label{app:median-iqr}

\begingroup\fontsize{8}{10}\selectfont\setlength{\tabcolsep}{3pt}

\begin{longtable}[t]{rrrrrrrrrrr}
\caption{\label{tab:comparison}Median $\mathcal{S}'(A,P)$, with the interquartile range below it in parentheses, over the 100 graphs of each condition; one variant per algorithm. The lowest median of each condition at full precision is in bold; $\mathit{rew}$ applies to the LRM and $\mathit{pert}$ to the MMM.}\\
\toprule
$n$ & $d$ & $\mathit{rew}$/$\mathit{pert}$ & Random & SGSA & SGGA & SGQSA & BASIN & SGTS & SGPSO & SGDE\\
\midrule
\endfirsthead
\caption[]{Median $\mathcal{S}'(A,P)$, with the interquartile range below it in parentheses, over the 100 graphs of each condition; one variant per algorithm. The lowest median of each condition at full precision is in bold; $\mathit{rew}$ applies to the LRM and $\mathit{pert}$ to the MMM. \textit{(continued)}}\\
\toprule
$n$ & $d$ & $\mathit{rew}$/$\mathit{pert}$ & Random & SGSA & SGGA & SGQSA & BASIN & SGTS & SGPSO & SGDE\\
\midrule
\endhead

\endfoot
\bottomrule
\endlastfoot
\addlinespace[0.3em]
\multicolumn{11}{l}{\textbf{BA}}\\
20 & 15\% & -- & 0.1957 & 0.0761 & 0.0667 & 0.0947 & 0.0947 & \textbf{0.0652} & 0.0898 & 0.0869\\
 &  &  & (0.0137) & (0.0207) & (0.0222) & (0.0211) & (0.0211) & (0.0197) & (0.0241) & (0.0172)\\
\addlinespace[0.2em]
20 & 40\% & -- & 0.3175 & 0.1539 & 0.1444 & 0.1895 & 0.1789 & \textbf{0.1413} & 0.1895 & 0.1796\\
 &  &  & (0.0194) & (0.0271) & (0.0222) & (0.0211) & (0.0211) & (0.0199) & (0.0251) & (0.0231)\\
\addlinespace[0.2em]
50 & 15\% & -- & 0.2173 & 0.1143 & 0.1053 & 0.1143 & 0.1094 & \textbf{0.0932} & 0.1437 & 0.1573\\
 &  &  & (0.0036) & (0.0084) & (0.0096) & (0.0065) & (0.0053) & (0.0052) & (0.0080) & (0.0073)\\
\addlinespace[0.2em]
50 & 40\% & -- & 0.4034 & 0.2380 & 0.2226 & 0.2416 & 0.2327 & \textbf{0.2045} & 0.2808 & 0.3059\\
 &  &  & (0.0070) & (0.0091) & (0.0121) & (0.0098) & (0.0098) & (0.0073) & (0.0128) & (0.0130)\\
\addlinespace[0.2em]
100 & 15\% & -- & 0.2331 & 0.1473 & 0.1442 & 0.1416 & 0.1418 & \textbf{0.1334} & 0.1766 & 0.1998\\
 &  &  & (0.0018) & (0.0036) & (0.0039) & (0.0026) & (0.0036) & (0.0055) & (0.0043) & (0.0041)\\
\addlinespace[0.2em]
100 & 40\% & -- & 0.4343 & 0.2720 & 0.2735 & 0.2739 & 0.2760 & \textbf{0.2517} & 0.3271 & 0.3683\\
 &  &  & (0.0041) & (0.0054) & (0.0046) & (0.0049) & (0.0054) & (0.0071) & (0.0072) & (0.0076)\\
\addlinespace[0.2em]
\addlinespace[0.3em]
\multicolumn{11}{l}{\textbf{ER}}\\
20 & 15\% & -- & 0.1789 & 0.0737 & 0.0652 & 0.0947 & 0.0842 & \textbf{0.0632} & 0.0947 & 0.0869\\
 &  &  & (0.0316) & (0.0290) & (0.0217) & (0.0237) & (0.0313) & (0.0123) & (0.0211) & (0.0241)\\
\addlinespace[0.2em]
20 & 40\% & -- & 0.3280 & 0.1784 & 0.1630 & 0.2000 & 0.1895 & \textbf{0.1522} & 0.2105 & 0.2020\\
 &  &  & (0.0211) & (0.0307) & (0.0265) & (0.0211) & (0.0211) & (0.0217) & (0.0271) & (0.0246)\\
\addlinespace[0.2em]
50 & 15\% & -- & 0.2253 & 0.1322 & 0.1259 & 0.1306 & 0.1241 & \textbf{0.1087} & 0.1665 & 0.1725\\
 &  &  & (0.0200) & (0.0147) & (0.0129) & (0.0151) & (0.0163) & (0.0125) & (0.0180) & (0.0170)\\
\addlinespace[0.2em]
50 & 40\% & -- & 0.4212 & 0.2808 & 0.2687 & 0.2792 & 0.2661 & \textbf{0.2451} & 0.3331 & 0.3396\\
 &  &  & (0.0082) & (0.0082) & (0.0075) & (0.0082) & (0.0098) & (0.0105) & (0.0082) & (0.0084)\\
\addlinespace[0.2em]
100 & 15\% & -- & 0.2392 & 0.1665 & 0.1609 & 0.1592 & 0.1582 & \textbf{0.1511} & 0.2028 & 0.2134\\
 &  &  & (0.0099) & (0.0077) & (0.0085) & (0.0084) & (0.0090) & (0.0077) & (0.0092) & (0.0090)\\
\addlinespace[0.2em]
100 & 40\% & -- & 0.4513 & 0.3289 & 0.3255 & 0.3248 & 0.3265 & \textbf{0.3147} & 0.3927 & 0.4070\\
 &  &  & (0.0044) & (0.0039) & (0.0035) & (0.0045) & (0.0051) & (0.0035) & (0.0058) & (0.0057)\\
\addlinespace[0.2em]
\addlinespace[0.3em]
\multicolumn{11}{l}{\textbf{LRM}}\\
50 & 15\% & 0 & 0.1274 & \textbf{0.0000} & \textbf{0.0000} & \textbf{0.0000} & \textbf{0.0000} & \textbf{0.0000} & \textbf{0.0000} & 0.0882\\
 &  &  & (0.0184) & (0.0000) & (0.0000) & (0.0000) & (0.0000) & (0.0000) & (0.0000) & (0.0151)\\
\addlinespace[0.2em]
50 & 15\% & 10 & 0.1257 & 0.0359 & 0.0457 & 0.0522 & 0.0294 & \textbf{0.0278} & 0.0816 & 0.0880\\
 &  &  & (0.0183) & (0.0180) & (0.0226) & (0.0147) & (0.0118) & (0.0033) & (0.0164) & (0.0143)\\
\addlinespace[0.2em]
50 & 15\% & 20 & 0.1290 & 0.0556 & 0.0580 & 0.0620 & 0.0531 & \textbf{0.0431} & 0.0833 & 0.0914\\
 &  &  & (0.0152) & (0.0129) & (0.0124) & (0.0098) & (0.0098) & (0.0092) & (0.0131) & (0.0142)\\
\addlinespace[0.2em]
50 & 15\% & 30 & 0.1306 & 0.0588 & 0.0584 & 0.0620 & 0.0571 & \textbf{0.0449} & 0.0849 & 0.0931\\
 &  &  & (0.0195) & (0.0137) & (0.0140) & (0.0147) & (0.0114) & (0.0112) & (0.0163) & (0.0167)\\
\addlinespace[0.2em]
50 & 40\% & 0 & 0.2754 & \textbf{0.0000} & \textbf{0.0000} & \textbf{0.0000} & \textbf{0.0000} & \textbf{0.0000} & \textbf{0.0000} & 0.1693\\
 &  &  & (0.0212) & (0.0000) & (0.0000) & (0.0000) & (0.0000) & (0.0000) & (0.0000) & (0.0127)\\
\addlinespace[0.2em]
50 & 40\% & 10 & 0.2759 & 0.0996 & 0.1029 & 0.1257 & 0.0718 & \textbf{0.0702} & 0.1755 & 0.1859\\
 &  &  & (0.0201) & (0.0461) & (0.0491) & (0.0310) & (0.0118) & (0.0069) & (0.0135) & (0.0114)\\
\addlinespace[0.2em]
50 & 40\% & 20 & 0.2777 & 0.1451 & 0.1334 & 0.1469 & 0.1371 & \textbf{0.1173} & 0.1935 & 0.2018\\
 &  &  & (0.0215) & (0.0132) & (0.0117) & (0.0114) & (0.0114) & (0.0113) & (0.0180) & (0.0141)\\
\addlinespace[0.2em]
50 & 40\% & 30 & 0.2824 & 0.1569 & 0.1467 & 0.1600 & 0.1486 & \textbf{0.1274} & 0.2057 & 0.2124\\
 &  &  & (0.0199) & (0.0131) & (0.0155) & (0.0131) & (0.0131) & (0.0133) & (0.0149) & (0.0140)\\
\addlinespace[0.2em]
100 & 15\% & 0 & 0.1307 & \textbf{0.0000} & \textbf{0.0000} & \textbf{0.0000} & \textbf{0.0000} & \textbf{0.0000} & \textbf{0.0000} & 0.1140\\
 &  &  & (0.0108) & (0.0000) & (0.0000) & (0.0000) & (0.0000) & (0.0000) & (0.0000) & (0.0105)\\
\addlinespace[0.2em]
100 & 15\% & 10 & 0.1307 & 0.0271 & 0.0730 & 0.0677 & 0.0275 & \textbf{0.0263} & 0.1044 & 0.1133\\
 &  &  & (0.0109) & (0.0032) & (0.0085) & (0.0174) & (0.0053) & (0.0028) & (0.0102) & (0.0107)\\
\addlinespace[0.2em]
100 & 15\% & 20 & 0.1311 & 0.0889 & 0.0770 & 0.0764 & 0.0679 & \textbf{0.0562} & 0.1046 & 0.1146\\
 &  &  & (0.0106) & (0.0058) & (0.0073) & (0.0069) & (0.0152) & (0.0124) & (0.0098) & (0.0092)\\
\addlinespace[0.2em]
100 & 15\% & 30 & 0.1331 & 0.0937 & 0.0808 & 0.0784 & 0.0762 & \textbf{0.0695} & 0.1087 & 0.1162\\
 &  &  & (0.0097) & (0.0066) & (0.0064) & (0.0063) & (0.0065) & (0.0061) & (0.0089) & (0.0085)\\
\addlinespace[0.2em]
100 & 40\% & 0 & 0.2980 & \textbf{0.0000} & \textbf{0.0000} & \textbf{0.0000} & \textbf{0.0000} & \textbf{0.0000} & \textbf{0.0000} & 0.2213\\
 &  &  & (0.0107) & (0.0000) & (0.0000) & (0.0000) & (0.0000) & (0.0000) & (0.0000) & (0.0088)\\
\addlinespace[0.2em]
100 & 40\% & 10 & 0.2994 & 0.1125 & 0.1581 & 0.1588 & \textbf{0.0699} & \textbf{0.0699} & 0.2147 & 0.2390\\
 &  &  & (0.0117) & (0.0516) & (0.0055) & (0.0065) & (0.0049) & (0.0044) & (0.0114) & (0.0087)\\
\addlinespace[0.2em]
100 & 40\% & 20 & 0.2990 & 0.1736 & 0.1725 & 0.1731 & 0.1745 & \textbf{0.1463} & 0.2313 & 0.2535\\
 &  &  & (0.0114) & (0.0085) & (0.0075) & (0.0066) & (0.0113) & (0.0242) & (0.0150) & (0.0110)\\
\addlinespace[0.2em]
100 & 40\% & 30 & 0.2994 & 0.1857 & 0.1834 & 0.1842 & 0.1871 & \textbf{0.1713} & 0.2442 & 0.2620\\
 &  &  & (0.0081) & (0.0078) & (0.0070) & (0.0062) & (0.0061) & (0.0085) & (0.0122) & (0.0093)\\
\addlinespace[0.2em]
\addlinespace[0.3em]
\multicolumn{11}{l}{\textbf{MMM}}\\
50 & 15\% & 0 & 0.1453 & \textbf{0.0000} & \textbf{0.0000} & \textbf{0.0000} & \textbf{0.0000} & \textbf{0.0000} & \textbf{0.0000} & 0.0950\\
 &  &  & (0.0161) & (0.0000) & (0.0000) & (0.0000) & (0.0000) & (0.0000) & (0.0000) & (0.0143)\\
\addlinespace[0.2em]
50 & 15\% & 4 & 0.1420 & 0.0229 & 0.0212 & 0.0294 & 0.0229 & \textbf{0.0212} & 0.0335 & 0.0958\\
 &  &  & (0.0137) & (0.0065) & (0.0049) & (0.0147) & (0.0073) & (0.0050) & (0.0212) & (0.0124)\\
\addlinespace[0.2em]
50 & 15\% & 8 & 0.1420 & 0.0405 & 0.0380 & 0.0539 & 0.0392 & \textbf{0.0361} & 0.0612 & 0.0983\\
 &  &  & (0.0132) & (0.0111) & (0.0101) & (0.0167) & (0.0102) & (0.0079) & (0.0185) & (0.0111)\\
\addlinespace[0.2em]
50 & 15\% & 16 & 0.1420 & 0.0588 & 0.0541 & 0.0678 & 0.0571 & \textbf{0.0466} & 0.0816 & 0.1013\\
 &  &  & (0.0167) & (0.0132) & (0.0121) & (0.0114) & (0.0114) & (0.0095) & (0.0180) & (0.0141)\\
\addlinespace[0.2em]
50 & 30\% & 0 & 0.2336 & \textbf{0.0000} & \textbf{0.0000} & \textbf{0.0000} & \textbf{0.0000} & \textbf{0.0000} & \textbf{0.0000} & 0.1640\\
 &  &  & (0.0161) & (0.0000) & (0.0000) & (0.0000) & (0.0424) & (0.0000) & (0.0000) & (0.0194)\\
\addlinespace[0.2em]
50 & 30\% & 4 & 0.2367 & 0.0253 & \textbf{0.0245} & 0.0261 & 0.0261 & \textbf{0.0245} & 0.0457 & 0.1659\\
 &  &  & (0.0182) & (0.0082) & (0.0049) & (0.0167) & (0.0396) & (0.0049) & (0.0376) & (0.0207)\\
\addlinespace[0.2em]
50 & 30\% & 8 & 0.2385 & 0.0473 & 0.0473 & 0.0563 & 0.0506 & \textbf{0.0457} & 0.0963 & 0.1739\\
 &  &  & (0.0140) & (0.0262) & (0.0314) & (0.0408) & (0.0310) & (0.0065) & (0.0461) & (0.0146)\\
\addlinespace[0.2em]
50 & 30\% & 16 & 0.2392 & 0.1037 & 0.0887 & 0.1167 & 0.0890 & \textbf{0.0800} & 0.1502 & 0.1788\\
 &  &  & (0.0167) & (0.0262) & (0.0288) & (0.0200) & (0.0233) & (0.0082) & (0.0229) & (0.0155)\\
\addlinespace[0.2em]
100 & 15\% & 0 & 0.1420 & \textbf{0.0000} & \textbf{0.0000} & \textbf{0.0000} & \textbf{0.0000} & \textbf{0.0000} & \textbf{0.0000} & 0.1182\\
 &  &  & (0.0086) & (0.0000) & (0.0000) & (0.0000) & (0.0000) & (0.0000) & (0.0000) & (0.0083)\\
\addlinespace[0.2em]
100 & 15\% & 4 & 0.1418 & \textbf{0.0061} & \textbf{0.0061} & \textbf{0.0061} & \textbf{0.0061} & \textbf{0.0061} & 0.0125 & 0.1186\\
 &  &  & (0.0071) & (0.0008) & (0.0008) & (0.0008) & (0.0008) & (0.0008) & (0.0097) & (0.0065)\\
\addlinespace[0.2em]
100 & 15\% & 8 & 0.1440 & \textbf{0.0117} & \textbf{0.0117} & 0.0119 & \textbf{0.0117} & \textbf{0.0117} & 0.0358 & 0.1211\\
 &  &  & (0.0096) & (0.0008) & (0.0008) & (0.0012) & (0.0008) & (0.0008) & (0.0215) & (0.0082)\\
\addlinespace[0.2em]
100 & 15\% & 16 & 0.1418 & \textbf{0.0226} & 0.0226 & 0.0232 & 0.0230 & \textbf{0.0226} & 0.0584 & 0.1201\\
 &  &  & (0.0078) & (0.0016) & (0.0017) & (0.0020) & (0.0017) & (0.0016) & (0.0188) & (0.0090)\\
\addlinespace[0.2em]
100 & 30\% & 0 & 0.2503 & \textbf{0.0000} & \textbf{0.0000} & \textbf{0.0000} & \textbf{0.0000} & \textbf{0.0000} & \textbf{0.0000} & 0.2030\\
 &  &  & (0.0065) & (0.0000) & (0.0000) & (0.0000) & (0.0000) & (0.0000) & (0.0000) & (0.0093)\\
\addlinespace[0.2em]
100 & 30\% & 4 & 0.2505 & \textbf{0.0065} & \textbf{0.0065} & \textbf{0.0065} & \textbf{0.0065} & \textbf{0.0065} & 0.0461 & 0.2037\\
 &  &  & (0.0089) & (0.0004) & (0.0347) & (0.0004) & (0.0004) & (0.0004) & (0.0509) & (0.0102)\\
\addlinespace[0.2em]
100 & 30\% & 8 & 0.2505 & 0.0125 & 0.0426 & 0.0125 & 0.0125 & \textbf{0.0123} & 0.0844 & 0.2058\\
 &  &  & (0.0093) & (0.0012) & (0.0409) & (0.0008) & (0.0008) & (0.0009) & (0.0437) & (0.0097)\\
\addlinespace[0.2em]
100 & 30\% & 16 & 0.2506 & 0.0246 & 0.0525 & 0.0269 & \textbf{0.0242} & \textbf{0.0242} & 0.1220 & 0.2081\\
 &  &  & (0.0097) & (0.0300) & (0.0378) & (0.0462) & (0.0020) & (0.0012) & (0.0326) & (0.0097)\\*
\end{longtable}
\endgroup{}

\section{Scalability}\label{app:scalability}
\begin{figure}[H]
    \centering
    \includegraphics[width=0.95\linewidth]{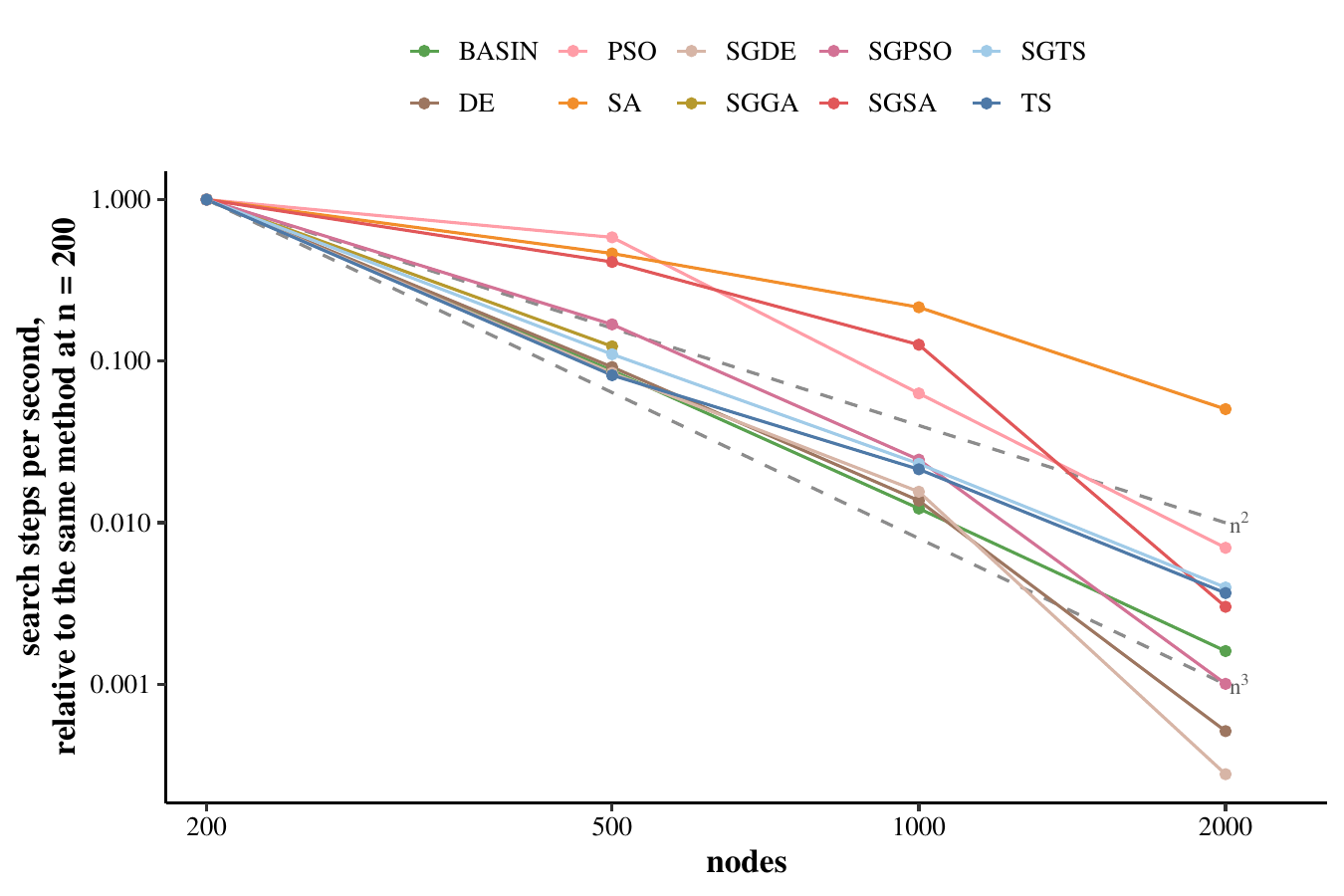}
 \caption{Throughput against graph size in the scalability probe, $20$ graphs per size: the median over graphs of budget units completed per second, relative to each algorithm's own value at $n=200$, since budget units differ between algorithms. Dashed lines are the $O(n^{2})$ and $O(n^{3})$ slopes from the same start. Logarithmic axes.}
    \label{fig:scal_cost}
\end{figure}

\begin{figure}[H]
    \centering
    \includegraphics[width=0.95\linewidth]{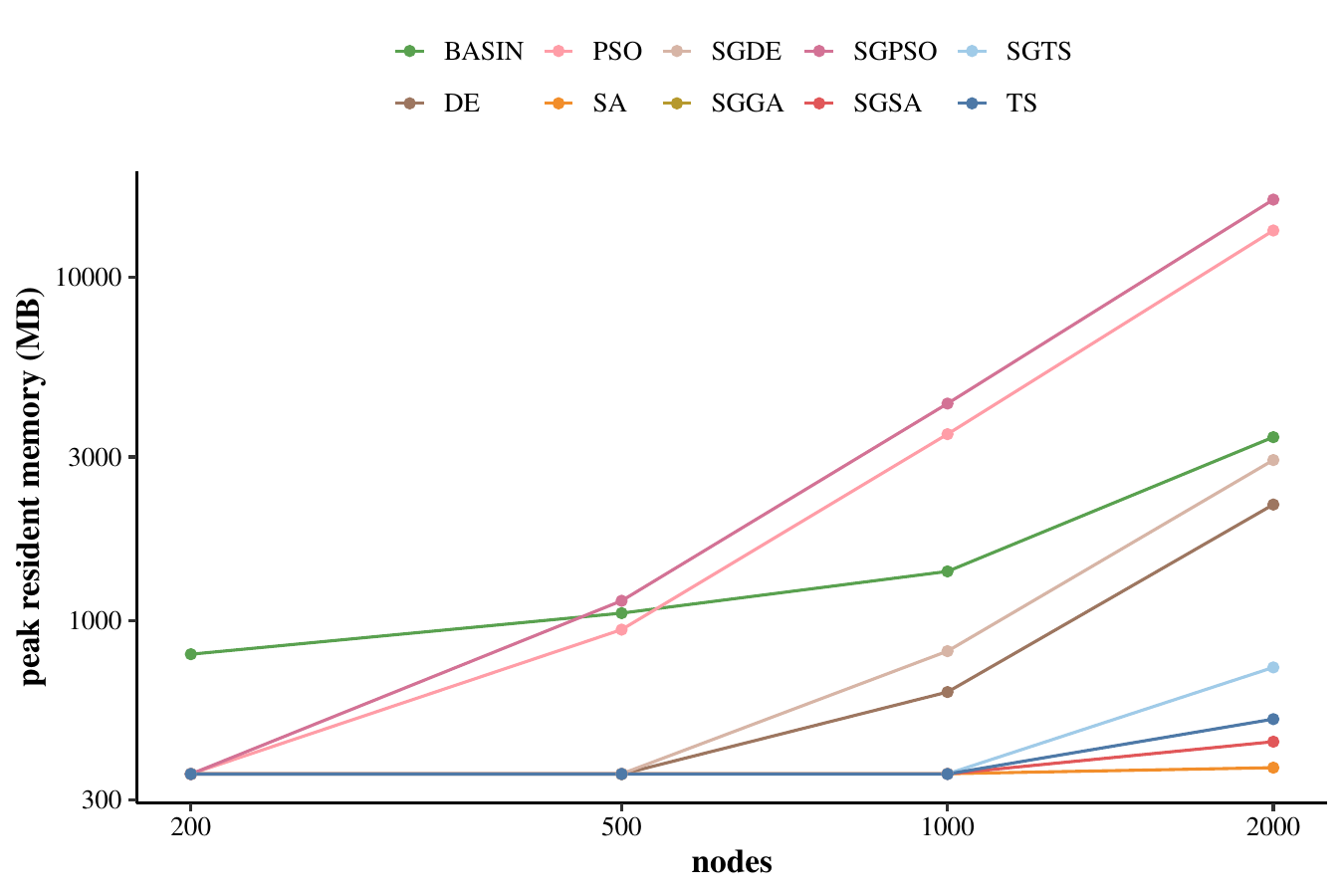}
 \caption{Peak resident memory against graph size in the scalability probe, $20$ graphs per size, median over graphs. Logarithmic axes.}
    \label{fig:scal_mem}
\end{figure}

\section{Stability of strategies}
\begin{figure}[H]
    \centering
    \includegraphics[width=0.95\linewidth]{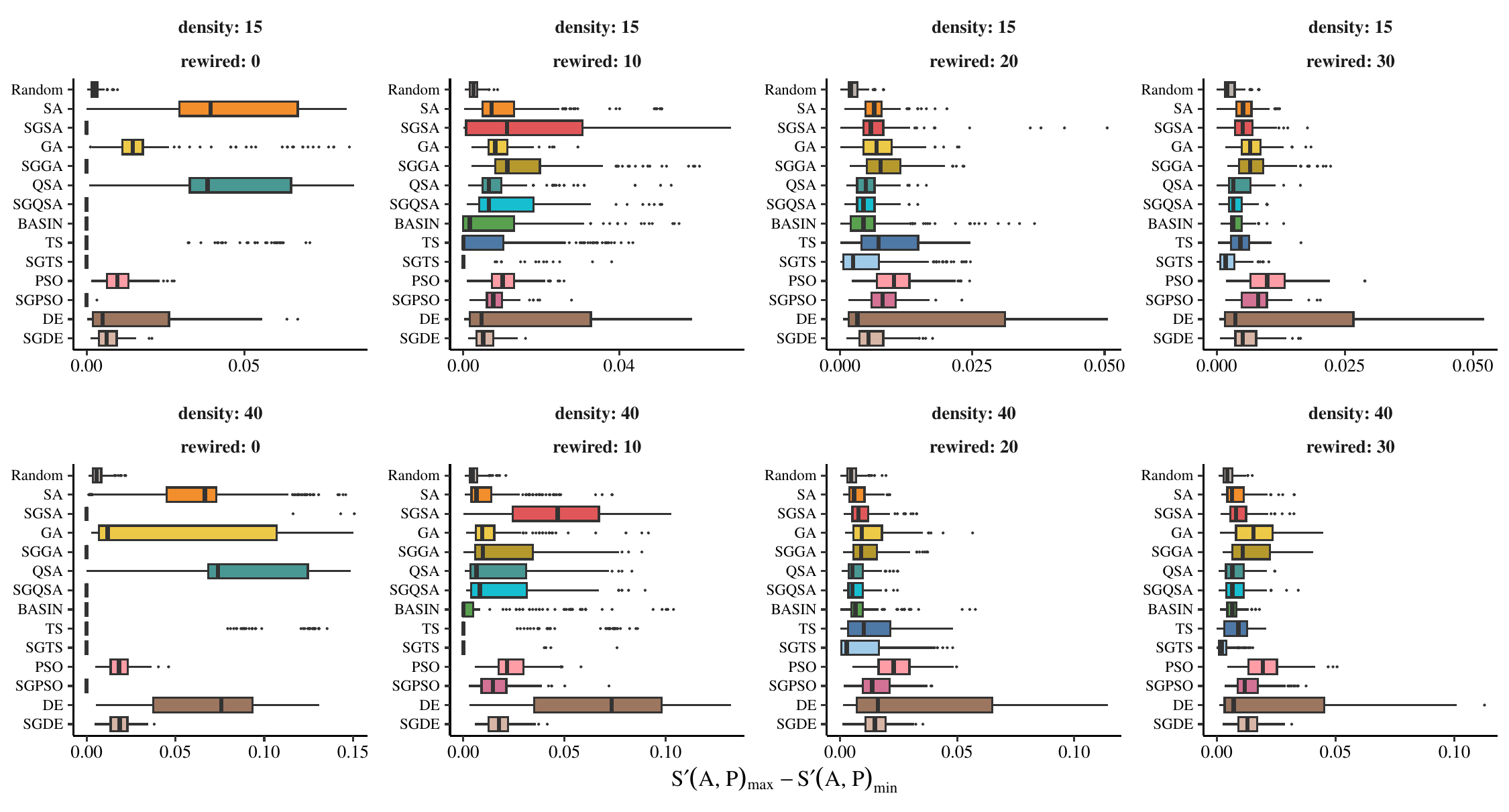}
 \caption{Restart-to-restart spread $\mathcal{S}'(A,P)_{\max}-\mathcal{S}'(A,P)_{\min}$ on \ac{LRM} graphs, every variant of every algorithm and the random-search baseline. Boxes span the test graphs of both sizes, one spread per graph; one panel per edge density and share of rewired edges, each with its own horizontal axis.}
    \label{fig:stability}
\end{figure}

\section{Adopted hyperparameter configurations}\label{app:hyper}

The configuration of every strategy as loaded at evaluation is listed below, and the search that produced it in Table~\ref{tab:hyperparam-searches}. Symbols are those of Section~\ref{s:methods}; \emph{restarts} counts the internal restarts within one reported run, and $w_{\mathrm{deg}}, w_{\mathrm{eig}}, w_{\mathrm{btw}}, w_{\mathrm{cl}}$ are the blend weights of the similarity matrix $M$.

\begin{description}
\item[SA] temperature scale $0.001825$; $10^{6}$ steps; patience $40000$.
\item[SGSA] temperature scale $0.002681$; $\beta = 0.2776$; $\phi = 0.003486$; $w_{\mathrm{deg}}, w_{\mathrm{eig}}, w_{\mathrm{btw}}, w_{\mathrm{cl}} = 0.707, 0.017, 0.180, 0.425$; $10^{6}$ steps; patience $40000$; 4 restarts.
\item[GA] $p_{\mathrm{co}} = 0.102$; initial $p_{\mathrm{mt}} = 0.398$; $k_{\mathrm{t}} = 2$; $N_{\mathrm{e}} = 34$; population $90$ on each of $10$ islands.
\item[SGGA] $w = 8.324$; $p_{\mathrm{co}} = 0.050$; initial $p_{\mathrm{mt}} = 0.314$; $p_{\mathrm{r}} = 0.344$; $k_{\mathrm{t}} = 2$; $N_{\mathrm{e}} = 35$; population $90$ on each of $10$ islands.
\item[TS] $30000$ iterations; $t_{\mathrm{lo}} = t_{\mathrm{hi}} = 1.247$.
\item[SGTS] $7500$ iterations per pull, $16$ pulls; $t_{\mathrm{lo}} = 0.888$, $t_{\mathrm{hi}} = 1.458$; $\kappa = 0.213$; $\beta = 0.005342$; $w_{\mathrm{deg}}, w_{\mathrm{eig}}, w_{\mathrm{btw}}, w_{\mathrm{cl}} = 0.831, 0.640, 0.585, 0.189$; arms: unguided, $\lambda = 0.052$, $\lambda = 1$, $q = 0.088$.
\item[QSA] fixed-point penalty $c = 1.107$; tolerance $1.0\times10^{-6}$; at most $10^{5}$ iterations; 4 restarts.
\item[SGQSA] $c = 13.53$; $\lambda_{\mathrm{c}} = 0.0145$; $\beta = 0.00119$; one centrality per restart, in turn, instead of a blend; tolerance $7.1\times10^{-7}$; at most $10^{5}$ iterations; 4 restarts.
\item[BASIN] $\eta = 0.001077$; $\tau = 4.766$; $m = 13.25n$; $\lambda_{\mathrm{fp}} = 1.225$; $\lambda_{\mathrm{ds}} = 0.068$; $\lambda_{\mathrm{ortho}} = 0$; $4000$ iterations; 4 restarts.
\item[PSO] $55$ particles; $w_0 = 0.643$, $w_1 = 0.619$; $c_1 = 2.598$, $c_2 = 1.081$; $v_{\max} = 0.901$; $40000$ iterations; patience $3200$; 4 restarts.
\item[SGPSO] $60$ particles; $w_0 = 0.939$, $w_1 = 0.323$; $c_1 = 2.998$, $c_2 = 1.264$; $v_{\max} = 0.133$; $c_g = 0.146$; $\beta = 0.00167$; $w_{\mathrm{deg}}, w_{\mathrm{eig}}, w_{\mathrm{btw}}, w_{\mathrm{cl}} = 0.880, 0.722, 0.450, 0.352$; $40000$ iterations; patience $3200$; 4 restarts.
\item[DE] population $59$; $\zeta = 0.329$; $\mathit{CR} = 0.857$; $40000$ iterations; patience $1600$; 4 restarts.
\item[SGDE] population $80$; $\zeta = 1.498$; $\mathit{CR} = 0.068$; $c_g = 3.616$; $\beta = 0.367$; $w_{\mathrm{deg}}, w_{\mathrm{eig}}, w_{\mathrm{btw}}, w_{\mathrm{cl}} = 0.629, 0.248, 0.875, 0.944$; $40000$ iterations; patience $1600$; 4 restarts.
\end{description}

\begin{table}

\caption{\label{tab:hyperparam-searches}Hyperparameter search per strategy. Every search minimizes $\mathcal{S}'$, one vote per graph, under the feasibility constraint. \emph{Complete} counts trials that ran to the end; \emph{Best} $\mathcal{S}'$ is training, not testing.}
\centering
\begin{tabular}[t]{lrrl}
\toprule
Method & Trials & Complete & Best $\mathcal{S}'$\\
\midrule
BASIN & 500 & 225 & 0.12372\\
DE & 264 & 73 & 0.17721\\
GA & 202 & 52 & 0.14597\\
PSO & 201 & 164 & 0.18487\\
QSA & 500 & 144 & 0.14123\\
SA & 500 & 88 & 0.14163\\
SGDE & 202 & 101 & 0.17565\\
SGGA & 200 & 41 & 0.13265\\
SGPSO & 203 & 155 & 0.17478\\
SGQSA & 500 & 232 & 0.13565\\
SGSA & 500 & 111 & 0.13151\\
SGTS & 500 & 125 & 0.10874\\
TS & 500 & 104 & 0.11670\\
\bottomrule
\end{tabular}
\end{table}

\end{document}